\documentclass[10pt,a4paper,oneside,titlepage,openright,fleqn,reqno,english,table]{report} 

\newcommand{\docname}{Stochastic High Fidelity Autonomous Fixed Wing Aircraft Flight Simulator}

\newcommand{\docdate}{May 2023}

\usepackage[hidelinks]{hyperref} 
\usepackage{cmap} 				 

\usepackage{multicol} 

\usepackage{longtable} 
\usepackage{multirow} 
\usepackage{rotating} 
\usepackage[cp1252]{inputenc} 
\usepackage{lscape} 
\usepackage{url}    
\usepackage[section]{placeins}  
\usepackage{afterpage} 
\usepackage{capt-of}        
\usepackage{dirtytalk} 

\usepackage[T1]{fontenc}
\usepackage{tabularx, booktabs}   
\usepackage{makecell} 
\usepackage{array} 
\usepackage{xcolor} 
\usepackage{xfrac} 
\usepackage{tocloft} 
\usepackage{setspace} 
\usepackage{chngcntr} 

\graphicspath{{figs/}} 					
\usepackage{fancyhdr} 					
\fancypagestyle{plain}{ 
	\fancyhead{}																
	\fancyhead[LO,RE]{\raisebox{2mm}{\footnotesize{Eduardo Gallo}}}				
	\fancyhead[C]{}																
	\fancyhead[RO,LE]{\raisebox{2mm}{\footnotesize{\docname}}}	
											
	\fancyfoot{}																
	\fancyfoot[LO,RE]{}															
	\fancyfoot[C]{\rule{0pt}{12pt}}				 												
	\fancyfoot[RO,LE]{\thepage}													
}
  
\fancypagestyle{nopagenumber}{
	\fancyhead{}																
	\fancyhead[LO,RE]{\raisebox{2mm}{\footnotesize{Eduardo Gallo}}}				
	\fancyhead[C]{}																
	\fancyhead[RO,LE]{\raisebox{2mm}{\footnotesize{\docname}}}	
											
	\fancyfoot{}																
	\fancyfoot[LO,RE]{}															
	\fancyfoot[C]{\rule{0pt}{12pt}}				 												
} 
 
\usepackage[parfill]{parskip} 					

\newcounter{level_1}[subsection]  		

\usepackage{amsmath} 			
\usepackage{amsfonts}			
\usepackage{mathtools}			
\everymath{\displaystyle}		
\usepackage{amssymb}  			
\usepackage{nicefrac}  			
\usepackage{accents}			

\usepackage{selinput}
\SelectInputMappings{
  aacute={á},
  eacute={é},
  iacute={í},
  oacute={ó},
  uacute={ú},
  ntilde={ñ}
}

\usepackage{tikz}  
\usetikzlibrary{shapes,arrows,arrows.meta,calc,shadows,fit,intersections,datavisualization.formats.functions,external}
\tikzstyle{block}        = [draw, fill=blue!30, rectangle, text centered, minimum height=3em, minimum width=6em] 
\tikzstyle{blockred}     = [draw, fill=red!30, rectangle, text centered, minimum height=3em, minimum width=6em] 
\tikzstyle{blockyellow}  = [draw, fill=yellow!30, rectangle, text centered, minimum height=3em, minimum width=6em] 
\tikzstyle{blockyellow1} = [draw, fill=yellow!15, rectangle, text centered, minimum height=3em, minimum width=6em] 
\tikzstyle{blockyellow2} = [draw, fill=yellow!30, rectangle, text centered, minimum height=3em, minimum width=6em] 
\tikzstyle{blockyellow3} = [draw, fill=yellow!45, rectangle, text centered, minimum height=3em, minimum width=6em] 
\tikzstyle{blockyellow4} = [draw, fill=yellow!60, rectangle, text centered, minimum height=3em, minimum width=6em] 
\tikzstyle{blockgreen}   = [draw, fill=green!30, rectangle, text centered, minimum height=3em, minimum width=6em] 
\tikzstyle{blockgreen1}  = [draw, fill=green!15, rectangle, text centered, minimum height=3em, minimum width=6em] 
\tikzstyle{blockgreen2}  = [draw, fill=green!30, rectangle, text centered, minimum height=3em, minimum width=6em] 
\tikzstyle{blockgreen3}  = [draw, fill=green!45, rectangle, text centered, minimum height=3em, minimum width=6em] 
\tikzstyle{blockgreen4}  = [draw, fill=green!60, rectangle, text centered, minimum height=3em, minimum width=6em] 
\tikzstyle{blockgrey1}   = [draw, fill=black!8,  rectangle, text centered, minimum height=3em, minimum width=6em] 
\tikzstyle{blockgrey2}   = [draw, fill=black!16, rectangle, text centered, minimum height=3em, minimum width=6em] 
\tikzstyle{blockgrey3}   = [draw, fill=black!24, rectangle, text centered, minimum height=3em, minimum width=6em] 
\tikzstyle{blockgrey4}   = [draw, fill=black!32, rectangle, text centered, minimum height=3em, minimum width=6em] 
\tikzstyle{blockgrey5}   = [draw, fill=black!40, rectangle, text centered, minimum height=3em, minimum width=6em] 
\tikzstyle{blocknofill}  = [draw=black!50, line width=1.5pt, rectangle, rounded corners, text centered, minimum height=3em, minimum width=6em]
\tikzstyle{blockhigh}    = [draw, fill=blue!20, rectangle, text centered, minimum height=6em, minimum width=6em]
\tikzstyle{noblock}      = [draw=black!50, line width=1.5pt, rectangle, rounded corners, text centered, minimum height=3em, minimum width=6em]
\tikzstyle{sum}          = [draw, fill=blue!20, circle, node distance=1cm]
\tikzstyle{sumrest}      = [draw, fill=black, circle, radius=0.5cm]
\tikzstyle{pinstyle}     = [pin edge={to-,thin,black}]
\tikzstyle{title}        = [text centered]
\pgfdeclarelayer{background}
\pgfdeclarelayer{foreground}
\pgfsetlayers{background,main,foreground}

\usepackage{pgfplots}
\pgfplotsset{compat=1.13}
\pgfplotsset{colormap/bluered}
\usepackage{pgfplotstable}
\usepgfplotslibrary{polar}
\usepgfplotslibrary{colormaps}
\usepgfplotslibrary{external} 
 
\usepackage{graphicx, color} 	
\usepackage{epstopdf}  			

\usepackage{emptypage} 

\usepackage{changepage} 

\allowdisplaybreaks 

\newcommand{\ds}    {\displaystyle} 		
\newcommand{\st}	{\scriptsize}           
\newcommand{\sss}   {\scriptscriptstyle}    

\newcommand{\CC}		{{C\nolinebreak[4]\hspace{-.05em}\raisebox{.4ex}{\tiny\bf ++}}}

\newcommand{\hypertt}	[1] {\hyperlink{#1}{\texttt{#1}}}
\newcommand{\hyperbf}	[1] {\hyperlink{#1}{\textbf{#1}}}

\newcommand{\nm}   		[1] {\ensuremath{\mathrm{#1}}} 									
\newcommand{\neweq}     [2] {\begin{equation} \mathrm{#1}\label{#2} \end{equation}} 	
\newcommand{\neweqnn}   [1] {\begin{equation} \mathrm{#1}\nonumber \end{equation}}		

\renewcommand{\vec}		[1] {\mbox{\boldmath{\ensuremath{\mathrm{#1}}}}}	
\newcommand{\deriv}     [1] {\dfrac{d{#1}}{dt}}             				
\newcommand{\derpar}    [2] {\dfrac{d{#1}}{d{#2}}}      					
\newcommand{\pderpar}   [2] {\dfrac{\partial{#1}}{\partial{#2}}}    		
\newcommand{\lrp}       [1] {\left(#1\right)}								
\newcommand{\lrsb}      [1] {\left[#1\right]}								
\newcommand{\lrb}       [1] {\left\{#1\right\}}								

\newcommand{\circled}            [1] {\accentset{\circ}{#1}}

\newcommand{\Deltat}     	{\Delta t}
\newcommand{\DeltatTRUTH}	{\Delta t_{\sss TRUTH}}
\newcommand{\DeltatSENSED}  {\Delta t_{\sss SENSED}}
\newcommand{\DeltatEST}  	{\Delta t_{\sss EST}}
\newcommand{\DeltatCNTR}  	{\Delta t_{\sss CNTR}}
\newcommand{\DeltatGNSS}  	{\Delta t_{\sss GNSS}}
\newcommand{\DeltatION}  	{\Delta t_{\sss ION}}
\newcommand{\DeltatIMG}  	{\Delta t_{\sss IMG}}

\newcommand{\first}         {\nm{1^{st}}}	 
\newcommand{\second}        {\nm{2^{nd}}}   
\newcommand{\third}       	{\nm{3^{rd}}}   
\newcommand{\fourth}        {\nm{4^{th}}}   
\newcommand{\fifth}         {\nm{5^{th}}}   
\newcommand{\sixth}			{\nm{6^{th}}}   

\newcommand{\deltaCNTR}  	{\vec{\delta}_{\sss CNTR}}		
\newcommand{\deltaTARGET}  	{\vec{\delta}_{\sss TARGET}}	    
\newcommand{\deltaTARGETone}  	{\vec{\delta}_{{\sss TARGET},1}}
\newcommand{\deltaTARGETtwo}  	{\vec{\delta}_{{\sss TARGET},2}}

\newcommand{\deltaTARGETn} 			{\vec{\delta}_{{\sss TARGET},n}}

\newcommand{\deltaT}  		{\delta_{\sss T}}            	
\newcommand{\deltaE}		{\delta_{\sss E}}            	
\newcommand{\deltaA}		{\delta_{\sss A}}				
\newcommand{\deltaR}		{\delta_{\sss R}}				
\newcommand{\deltaTARGETT}	{\delta_{\sss{TARGET,T}}}		    
\newcommand{\deltaTARGETE}	{\delta_{\sss{TARGET,E}}}		    
\newcommand{\deltaTARGETA}	{\delta_{\sss{TARGET,A}}}		    
\newcommand{\deltaTARGETR}	{\delta_{\sss{TARGET,R}}}		    
\newcommand{\deltaTRG}  	{\delta_{\sss{TRG}}}			

\newcommand{\deltaTARGETTone}  {\delta_{{\sss TARGET,T},1}}	
\newcommand{\deltaTARGETTtwo}  {\delta_{{\sss TARGET,T},2}}	
\newcommand{\deltaTARGETTn}    {\delta_{{\sss TARGET,T},n}}	
\newcommand{\deltaTARGETEone}  {\delta_{{\sss TARGET,E},1}}	
\newcommand{\deltaTARGETEtwo}  {\delta_{{\sss TARGET,E},2}}	
\newcommand{\deltaTARGETEn}    {\delta_{{\sss TARGET,E},n}}	
\newcommand{\deltaTARGETAone}  {\delta_{{\sss TARGET,A},1}}	
\newcommand{\deltaTARGETAtwo}  {\delta_{{\sss TARGET,A},2}}	
\newcommand{\deltaTARGETAn}    {\delta_{{\sss TARGET,A},n}}	
\newcommand{\deltaTARGETRone}  {\delta_{{\sss TARGET,R},1}}	
\newcommand{\deltaTARGETRtwo}  {\delta_{{\sss TARGET,R},2}}	
\newcommand{\deltaTARGETRn}    {\delta_{{\sss TARGET,R},n}}	
\newcommand{\deltaTRGone}	   {\delta_{{\sss TRG},1}}
\newcommand{\deltaTRGtwo}	   {\delta_{{\sss TRG},2}}
\newcommand{\deltaTRGn}	       {\delta_{{\sss TRG},n}}

\newcommand{\xvec}   			{\vec x}
\newcommand{\xveczero}	 		{\vec x_0}
\newcommand{\xvecdot}			{\vec {\dot x}}
\newcommand{\xvecest}	  		{\hat{\vec x}}
\newcommand{\xvecestzero}		{\hat{\vec x}_0}
\newcommand{\xvectilde} 		{\widetilde{\vec x}}

\newcommand{\xvecvis}			{\circled{\vec x}}

\newcommand{\xTRUTH}			{\vec x_{\sss TRUTH}}
\newcommand{\xSENSED}  			{\vec x_{\sss SENSED}}
\newcommand{\xEST}  			{\vec x_{\sss EST}}
\newcommand{\xIMG}  			{\vec x_{\sss IMG}}
\newcommand{\xREF}	  			{\vec x_{\sss REF}}
\newcommand{\xTRUTHzero}		{\vec x_{0,\sss TRUTH}}
\newcommand{\xTRUTHdot}			{\vec {\dot x}_{\sss TRUTH}}

\newcommand{\FE}		 {F_{\sss E}}
\newcommand{\OECEF}    	 {O_{\sss E}}                           
\newcommand{\TEi}        {T_{1}^{\sss E}}
\newcommand{\TEii}       {T_{2}^{\sss E}}
\newcommand{\TEiii}      {T_{3}^{\sss E}}
\newcommand{\iEi}        {\vec i_{1}^{\sss E}}
\newcommand{\iEii}       {\vec i_{2}^{\sss E}}
\newcommand{\iEiii}      {\vec i_{3}^{\sss E}}

\newcommand{\TEcar}      {\vec T^{\sss E,CAR}}

\newcommand{\TEcarest}   {\hat{\vec T}^{\sss E,CAR}}
\newcommand{\TEgct}      {\vec T^{\sss E,GCT}}
\newcommand{\TEgdt}      {\vec T^{\sss E,GDT}}
\newcommand{\TEgdttilde} {\widetilde{\vec T}^{\sss E,GDT}}
\newcommand{\TEgdtdot}   {\vec{\dot T}^{\sss E,GDT}}

\newcommand{\TEgdtest}   {\hat{\vec T}^{\sss E,GDT}}

\newcommand{\FS}		  {F_{\sss S}}
\newcommand{\OS}          {O_{\sss S}}                            

\newcommand{\iStheta}     {\vec i_\theta^{\sss S}}
\newcommand{\iSlambda}    {\vec i_\lambda^{\sss S}}
\newcommand{\iSr}         {\vec i_r^{\sss S}}

\newcommand{\FN} 		  {F_{\sss N}}
\newcommand{\ON}          {O_{\sss N}}                            

\newcommand{\iNi}         {\vec i_{1}^{\sss N}}
\newcommand{\iNii}        {\vec i_{2}^{\sss N}}
\newcommand{\iNiii}       {\vec i_{3}^{\sss N}}

\newcommand{\FB}		  {F_{\sss B}}
\newcommand{\OB}          {O_{\sss B}}                            

\newcommand{\iBi}         {\vec i_{1}^{\sss B}}
\newcommand{\iBii}        {\vec i_{2}^{\sss B}}
\newcommand{\iBiii}       {\vec i_{3}^{\sss B}}

\newcommand{\FP}		  {F_{\sss P}}
\newcommand{\OP}          {O_{\sss P}}                            

\newcommand{\iPi}         {\vec i_{1}^{\sss P}}
\newcommand{\iPii}        {\vec i_{2}^{\sss P}}
\newcommand{\iPiii}       {\vec i_{3}^{\sss P}}

\newcommand{\FA}		  {F_{\sss A}}							
\newcommand{\OA}          {O_{\sss A}}

\newcommand{\iAi}         {\vec i_{1}^{\sss A}}
\newcommand{\iAii}        {\vec i_{2}^{\sss A}}
\newcommand{\iAiii}       {\vec i_{3}^{\sss A}}

\newcommand{\FY}		  {F_{\sss Y}}							
\newcommand{\OY}          {O_{\sss Y}}

\newcommand{\iYi}         {\vec i_{1}^{\sss Y}}
\newcommand{\iYii}        {\vec i_{2}^{\sss Y}}
\newcommand{\iYiii}       {\vec i_{3}^{\sss Y}}

\newcommand{\FW}		  {F_{\sss W}}
\newcommand{\OW}          {O_{\sss W}}                            

\newcommand{\iWi}         {\vec i_{1}^{\sss W}}
\newcommand{\iWii}        {\vec i_{2}^{\sss W}}
\newcommand{\iWiii}       {\vec i_{3}^{\sss W}}

\newcommand{\FR}		  {F_{\sss R}}
\newcommand{\OR}          {O_{\sss R}}                            

\newcommand{\iRi}         {\vec i_{1}^{\sss R}}
\newcommand{\iRii}        {\vec i_{2}^{\sss R}}
\newcommand{\iRiii}       {\vec i_{3}^{\sss R}}

\newcommand{\FG}		  {F_{\sss G}}
\newcommand{\OG}          {O_{\sss G}}                            

\newcommand{\iGi}         {\vec i_{1}^{\sss G}}
\newcommand{\iGii}        {\vec i_{2}^{\sss G}}
\newcommand{\iGiii}       {\vec i_{3}^{\sss G}}

\newcommand{\FC}		  {F_{\sss C}}
\newcommand{\OC}          {O_{\sss C}}                            
\newcommand{\pCi}         {p_{1}^{\sss C}}
\newcommand{\pCii}        {p_{2}^{\sss C}}
\newcommand{\pCiii}       {p_{3}^{\sss C}}
\newcommand{\iCi}         {\vec i_{1}^{\sss C}}
\newcommand{\iCii}        {\vec i_{2}^{\sss C}}
\newcommand{\iCiii}       {\vec i_{3}^{\sss C}}
\newcommand{\pC}          {\vec p^{\sss C}}
\newcommand{\pCbar} 	  {\vec {\bar p}^{\sss C}}

\newcommand{\pCunit}      {\vec p^{1 \sss C}}

\newcommand{\FT}		  {F_{\sss T}}
\newcommand{\OT}          {O_{\sss T}}                           
\newcommand{\iTi}         {\vec i_{1}^{\sss T}}
\newcommand{\iTii}        {\vec i_{2}^{\sss T}}
\newcommand{\iTiii}       {\vec i_{3}^{\sss T}}

\newcommand{\FI}		 {F_{\sss I}}							

\newcommand{\gS}          {\vec g^{\sss S}}
\newcommand{\gN}          {\vec g^{\sss N}}

\newcommand{\ac}          {\vec a_c}
\newcommand{\acE}         {\vec a_c^{\sss E}}
\newcommand{\gc}     	  {\vec g_c}

\newcommand{\gcNMODEL}    {\vec g_{c,\sss{MOD}}^{\sss N}}

\newcommand{\gcNREAL}     {\vec g_{c,\sss{REAL}}^{\sss N}}

\newcommand{\gcNMODELiii} {g_{c,{{\sss {MOD}},3}}^{\sss N}}
\newcommand{\gcDEV}       {g_{c\sss{DEV}}}
\newcommand{\gcMSL}       {g_{c\sss MSL}}
\newcommand{\gcMSLE}      {g_{c\sss{MSL,E}}}
\newcommand{\gcMSLP}      {g_{c\sss{MSL,P}}}
\newcommand{\sigmagDEV}			{\nm{\sigma_{g\sss{DEV}}}} 
\newcommand{\sigmagammaDEV}		{\nm{\sigma_{\gamma\sss{DEV}}}}
\newcommand{\gammaDEV}			{\nm{\gamma_{\sss DEV}}}
\newcommand{\phiDEV}			{\nm{\phi_{\sss DEV}}}

\newcommand{\vvec} 				{\vec v}
\newcommand{\vEN}  				{\vec v_{\sss EN}}
\newcommand{\vEB}  				{\vec v_{\sss EB}}
\newcommand{\vE}  				{\vec v^{\sss E}}
\newcommand{\vEi}        		{v_{1}^{\sss E}}
\newcommand{\vEii}      	  	{v_{2}^{\sss E}}

\newcommand{\vB} 	   			{\vec v^{\sss B}}
\newcommand{\vBest}    			{\hat{\vec v}^{\sss B}}

\newcommand{\vN}	    		{\vec v^{\sss N}} 
\newcommand{\vNi}        		{v_{1}^{\sss N}}
\newcommand{\vNii}      	  	{v_{2}^{\sss N}}
\newcommand{\vNiii} 	      	{v_{3}^{\sss N}}

\newcommand{\vG}        	  	{\vec v^{\sss G}}
\newcommand{\vIN}				{\vec v_{\sss IN}}
\newcommand{\vIE}				{\vec v_{\sss IE}}
\newcommand{\vIB}           	{\vec v_{\sss IB}}
\newcommand{\vIP}           	{\vec v_{\sss IP}}
\newcommand{\vBP}           	{\vec v_{\sss BP}}
\newcommand{\vBC}           	{\vec v_{\sss BC}}

\newcommand{\vEBI}				{\vec v_{\sss EB}^{\sss I}}

\newcommand{\vEBB}				{\vec v_{\sss EB}^{\sss B}}
\newcommand{\vEBE}				{\vec v_{\sss EB}^{\sss E}}
\newcommand{\vEBN}				{\vec v_{\sss EB}^{\sss N}}
\newcommand{\vENN}				{\vec v_{\sss EN}^{\sss N}}

\newcommand{\vIBIdot}			{\vec{\dot v}_{\sss IB}^{\sss I}}
\newcommand{\vEBEdot}			{\vec{\dot v}_{\sss EB}^{\sss E}}
\newcommand{\vEBBdot}			{\vec{\dot v}_{\sss EB}^{\sss B}}
\newcommand{\vENEdot}			{\vec{\dot v}_{\sss EN}^{\sss E}}
\newcommand{\vENNdot}			{\vec{\dot v}_{\sss EN}^{\sss N}}
\newcommand{\vBdot}				{\vec{\dot v}^{\sss B}}
\newcommand{\vEdot}	    		{\vec{\dot v}^{\sss E}}

\newcommand{\vtilde}			{\widetilde{\vec v}}
\newcommand{\vNtilde}			{\widetilde{\vec v}^{\sss N}}
\newcommand{\vNdot}	    		{\vec{\dot v}^{\sss N}}
\newcommand{\vNest}	    		{\hat{\vec v}^{\sss N}}

\newcommand{\vNskew}			{\widehat{\vec v}^{\sss N}}

\newcommand{\vTAS}          {\vec v_{\sss TAS}}    
\newcommand{\vtas}          {v_{\sss TAS}}

\newcommand{\vtastilde}     {\widetilde{v}_{\sss TAS}}
\newcommand{\vtasdot}       {\dot{v}_{\sss TAS}}

\newcommand{\vtasest}       {\hat{v}_{\sss TAS}}

\newcommand{\vTASB}    		{\vec v_{\sss TAS}^{\sss B}} 
\newcommand{\vTASW}    		{\vec v_{\sss TAS}^{\sss W}}

\newcommand{\vTASBi}        {v_{{\sss TAS},1}^{\sss B}}
\newcommand{\vTASBii}       {v_{{\sss TAS},2}^{\sss B}}
\newcommand{\vTASBiii}      {v_{{\sss TAS},3}^{\sss B}}
\newcommand{\vTASBtilde}	{\widetilde{\vec v}_{\sss TAS}^{\sss B}}

\newcommand{\vWIND}       	{\vec v_{\sss WIND}}

\newcommand{\vwind}      	{v_{\sss WIND}}

\newcommand{\vWINDN}      	{\vec v_{\sss WIND}^{\sss N}}

\newcommand{\vWINDNi}     	{v_{{\sss WIND},1}^{\sss N}}
\newcommand{\vWINDNii}    	{v_{{\sss WIND},2}^{\sss N}}
\newcommand{\vWINDNiii}   	{v_{{\sss WIND}.3}^{\sss N}}
\newcommand{\vWINDNest}    	{\hat{\vec v}_{\sss WIND}^{\sss N}}

\newcommand{\vTURB}       {\vec v_{\sss TURB}}

\newcommand{\vTURBN}      {\vec v_{\sss TURB}^{\sss N}}

\newcommand{\vTURBT}      {\vec v_{\sss TURB}^{\sss T}}

\newcommand{\omegaE}		{\omega_{\sss E}}
\newcommand{\vecomegaE}     {\vec{\omega}_{\sss E}}          	
\newcommand{\vecomegaEE}    {\vec{\omega}_{\sss E}^{\sss E}}     
 
\newcommand{\omegaEEskew}   {\widehat{\vec{\omega}}_{\sss E}^{\sss E}} 

\newcommand{\wIE}			{\vec \omega_{\sss IE}}

\newcommand{\wIEE}			{\vec \omega_{\sss IE}^{\sss E}}
\newcommand{\wIEN}			{\vec \omega_{\sss IE}^{\sss N}}
\newcommand{\wIEB}			{\vec \omega_{\sss IE}^{\sss B}}
\newcommand{\wIEskew}		{\widehat{\vec \omega}_{\sss IE}}
\newcommand{\wIEIskew}		{\widehat{\vec \omega}_{\sss IE}^{\sss I}}
\newcommand{\wIENskew}		{\widehat{\vec \omega}_{\sss IE}^{\sss N}}
\newcommand{\wIEEskew}		{\widehat{\vec \omega}_{\sss IE}^{\sss E}}
\newcommand{\alphaIEIskew} 	{\widehat{\vec \alpha}_{\sss IE}^{\sss I}}
 
\newcommand{\wEN}			{\vec \omega_{\sss EN}}

\newcommand{\wENN}			{\vec \omega_{\sss EN}^{\sss N}}
\newcommand{\wENNskew}		{\widehat{\vec \omega}_{\sss EN}^{\sss N}}

\newcommand{\wENB}			{\vec \omega_{\sss EN}^{\sss B}}
\newcommand{\wENBest}		{\hat{\vec \omega}_{\sss EN}^{\sss B}}

\newcommand{\wEB}			{\vec \omega_{\sss EB}}
\newcommand{\wEBB}			{\vec \omega_{\sss EB}^{\sss B}}
\newcommand{\wEBBest}		{\hat{\vec \omega}_{\sss EB}^{\sss B}}
\newcommand{\wEBBskew}		{\widehat{\vec \omega}_{\sss EB}^{\sss B}}
\newcommand{\wEBBdot}		{\vec{\dot \omega}_{\sss EB}^{\sss B}}
\newcommand{\wIEBdot}		{\vec{\dot \omega}_{\sss IE}^{\sss B}}

\newcommand{\wNB}			{\vec \omega_{\sss NB}}

\newcommand{\wNBB}			{\vec \omega_{\sss NB}^{\sss B}}
\newcommand{\wNBBdot}		{\vec{\dot \omega}_{\sss NB}^{\sss B}}

\newcommand{\wNBBskew}		{\widehat{\vec \omega}_{\sss {NB}}^{\sss B}}
\newcommand{\wNBBest}		{\hat{\vec \omega}_{\sss NB}^{\sss B}}

\newcommand{\wIB}				{\vec \omega_{\sss IB}}
\newcommand{\wIBI}				{\vec \omega_{\sss IB}^{\sss I}}

\newcommand{\wIBN}				{\vec \omega_{\sss IB}^{\sss N}}
\newcommand{\wIBB}				{\vec \omega_{\sss IB}^{\sss B}}
\newcommand{\wBI}				{\vec \omega_{\sss BI}}

\newcommand{\wIBBdot}			{\vec{\dot \omega}_{\sss IB}^{\sss B}}
\newcommand{\wBIBskew}			{\widehat{\vec \omega}_{\sss BI}^{\sss B}}
\newcommand{\wIBBskew}			{\widehat{\vec \omega}_{\sss IB}^{\sss B}}
\newcommand{\wIBBest}			{\hat{\vec \omega}_{\sss IB}^{\sss B}}
\newcommand{\wIBBestskew}		{\widehat{\hat{\vec \omega}}_{\sss IB}^{\sss B}}
\newcommand{\wIBBtildeskew}		{\widehat{\widetilde{\vec \omega}}_{\sss IB}^{\sss B}}
\newcommand{\wIBBtilde}			{\widetilde{\vec \omega}_{\sss IB}^{\sss B}}
\newcommand{\wIBBtildebar}		{\bar{\widetilde{\vec \omega}}_{\sss IB}^{\sss B}}
\newcommand{\wIBBtildebari}		{\bar{\widetilde{\omega}}_{{\sss IB},1}^{\sss B}}
\newcommand{\wIBBtildebarii}	{\bar{\widetilde{\omega}}_{{\sss IB},2}^{\sss B}}
\newcommand{\wIBBtildebariii}	{\bar{\widetilde{\omega}}_{{\sss IB},3}^{\sss B}}
\newcommand{\wIBskew}			{\widehat{\vec \omega}_{\sss IB}}
\newcommand{\alphaIBskew}		{\widehat{\vec \alpha}_{\sss IB}}
\newcommand{\alphaIBBskew}		{\widehat{\vec \alpha}_{\sss IB}^{\sss B}}
\newcommand{\alphaIBBest}		{\hat{\vec \alpha}_{\sss IB}^{\sss B}}
\newcommand{\alphaIBBestskew}	{\widehat{\hat{\vec \alpha}}_{\sss IB}^{\sss B}}
\newcommand{\alphaIBBtildeskew}	{\widehat{\widetilde{\vec \alpha}}_{\sss IB}^{\sss B}}
\newcommand{\alphaIBBtilde}  	{\widetilde{\vec \alpha}_{\sss IB}^{\sss B}}

\newcommand{\wIP}				{\vec \omega_{\sss IP}}

\newcommand{\wIPP}				{\vec \omega_{\sss IP}^{\sss P}}

\newcommand{\wIPPtilde}			{\widetilde{\vec \omega}_{\sss IP}^{\sss P}}
\newcommand{\wIPPtildetilde}	{\widetilde{\widetilde{\vec \omega}}_{\sss IP}^{\sss P}}

\newcommand{\wBP}			{\vec \omega_{\sss BP}}

\newcommand{\wBC}			{\vec \omega_{\sss BC}}

\newcommand{\xiEBB}			{\vec \xi_{\sss EB}^{\sss B}}
\newcommand{\xiEBBest}		{\hat{\vec \xi}_{\sss EB}^{\sss B}}

\newcommand{\nuEBB}			{\vec \nu_{\sss EB}^{\sss B}}
\newcommand{\nuEBBest}		{\hat{\vec \nu}_{\sss EB}^{\sss B}}

\newcommand{\atrs}			{\vec a_{trs}}
\newcommand{\atrsI}			{\vec a_{trs}^{\sss I}} 
\newcommand{\atrsN}			{\vec a_{trs}^{\sss N}} 
\newcommand{\acor}			{\vec a_{cor}}
\newcommand{\acorN}			{\vec a_{cor}^{\sss N}}

\newcommand{\acorE}			{\vec a_{cor}^{\sss E}}

\newcommand{\RNB}			{\vec R_{\sss NB}}
\newcommand{\RNBest}		{\hat{\vec R}_{\sss NB}}

\newcommand{\RBPest}		{\hat{\vec R}_{\sss BP}}
\newcommand{\RBC}			{\vec R_{\sss BC}}

\newcommand{\REN}			{\vec R_{\sss EN}}

\newcommand{\RNW}			{\vec R_{\sss NW}}
\newcommand{\RIE}			{\vec R_{\sss IE}}

\newcommand{\RIB}			{\vec R_{\sss IB}}
\newcommand{\RBI}			{\vec R_{\sss BI}}
\newcommand{\RIN}			{\vec R_{\sss IN}}

\newcommand{\RNG}			{\vec R_{\sss NG}}

\newcommand{\RWB}			{\vec R_{\sss WB}}
\newcommand{\RPA}			{\vec R_{\sss PA}^{\star}}
\newcommand{\RAP}			{\vec R_{\sss AP}^{\star}}
\newcommand{\RPY}			{\vec R_{\sss PY}^{\star}}
\newcommand{\RYP}			{\vec R_{\sss YP}^{\star}}

\newcommand{\RNBdot}		{\vec{\dot R}_{\sss NB}}

\newcommand{\REBdot}		{\vec{\dot R}_{\sss EB}}
\newcommand{\RENdot}		{\vec{\dot R}_{\sss EN}}
\newcommand{\RBIdot}		{\vec{\dot R}_{\sss BI}}

\newcommand{\REB}			{\vec R_{\sss EB}}

\newcommand{\MEB}			{\vec M_{\sss EB}}

\newcommand{\TRBR}			{\vec T_{\sss RB}^{\sss R}}
\newcommand{\TRBRfull}		{\vec T_{{\sss RB},full}^{\sss R}}
\newcommand{\TRBRempty}		{\vec T_{{\sss RB},empty}^{\sss R}}
\newcommand{\TRBB}			{\vec T_{\sss RB}^{\sss B}}
\newcommand{\TBP}			{\vec T_{\sss BP}}

\newcommand{\TBPB}			{\vec T_{\sss BP}^{\sss B}}
\newcommand{\TBPBest}		{\hat{\vec T}_{\sss BP}^{\sss B}}
\newcommand{\TBPBfull}		{\vec T_{{\sss BP},full}^{\sss B}}
\newcommand{\TBPBempty}		{\vec T_{{\sss BP},empty}^{\sss B}}

\newcommand{\TNBN}			{\vec T_{\sss NB}^{\sss N}}
\newcommand{\TNGN}			{\vec T_{\sss NG}^{\sss N}}

\newcommand{\TEBE}			{\vec T_{\sss EB}^{\sss E}}
\newcommand{\TEBEest}		{\hat{\vec T}_{\sss EB}^{\sss E}}

\newcommand{\TESE}			{\vec T_{\sss ES}^{\sss E}}
\newcommand{\TEN}			{\vec T_{\sss EN}}
\newcommand{\TENE}			{\vec T_{\sss EN}^{\sss E}}

\newcommand{\TBC}			{\vec T_{\sss BC}}
\newcommand{\TBCB}			{\vec T_{\sss BC}^{\sss B}}
\newcommand{\TBCBest}		{\hat{\vec T}_{\sss BC}^{\sss B}}
\newcommand{\TBCBfull}		{\vec T_{{\sss BC},full}^{\sss B}}
\newcommand{\TBCBempty}		{\vec T_{{\sss BC},empty}^{\sss B}}
\newcommand{\TECE}			{\vec T_{\sss EC}^{\sss E}}
\newcommand{\TRPR}			{\vec T_{\sss RP}^{\sss R}}
\newcommand{\TRPB}			{\vec T_{\sss RP}^{\sss B}}

\newcommand{\qNB}			{\vec q_{\sss NB}}
\newcommand{\qNG}			{\vec q_{\sss NG}}

\newcommand{\qNBdot}		{\vec{\dot q}_{\sss NB}}

\newcommand{\qNBinit}       {\vec q_{\sss NB,INIT}}
\newcommand{\qNBest}		{\hat{\vec q}_{\sss NB}}

\newcommand{\qNBzero}	    {\vec q_{{\sss NB},0}}
\newcommand{\sigmaqNBinit}  {\sigma_{\sss qNB,INIT}}

\newcommand{\DeltarNBB}        	{\Delta \vec r_{\sss NB}^{\sss B}}
\newcommand{\DeltarNBBdot}      {\Delta \vec {\dot r}_{\sss NB}^{\sss B}}

\newcommand{\qBP}			{\vec q_{\sss BP}}
\newcommand{\qBPest}		{\hat{\vec q}_{\sss BP}}
\newcommand{\zetaEN}		{\vec \zeta_{\sss EN}}
\newcommand{\zetaEG}		{\vec \zeta_{\sss EG}}
\newcommand{\zetaNB}		{\vec \zeta_{\sss NB}}
\newcommand{\zetaNG}		{\vec \zeta_{\sss NG}}

\newcommand{\qNE}			{\vec q_{\sss NE}}

\newcommand{\qEB}			{\vec q_{\sss EB}}

\newcommand{\qEBest}		{\hat{\vec q}_{\sss EB}}

\newcommand{\qEBast}		{\vec q_{\sss EB}^{\ast}}
\newcommand{\qEBdot}		{\vec{\dot q}_{\sss EB}}
\newcommand{\qEBastdot}		{\vec{\dot q}_{\sss EB}^{\ast}}

\newcommand{\qEC}			{\vec q_{\sss EC}}

\newcommand{\qEN}			{\vec q_{\sss EN}}
\newcommand{\qENest}		{\hat{\vec q}_{\sss EN}}

\newcommand{\qBC}			{\vec q_{\sss BC}}

\newcommand{\zetaEB}		{\vec \zeta_{\sss EB}}
\newcommand{\zetaEBest}		{\hat{\vec \zeta}_{\sss EB}}

\newcommand{\zetaEBdot}		{\vec{\dot \zeta}_{\sss EB}}

\newcommand{\zetaEC}		{\vec \zeta_{\sss EC}}

\newcommand{\zetaBC}		{\vec \zeta_{\sss BC}}

\newcommand{\phiNB}			  	{\vec \phi_{\sss NB}} 

\newcommand{\sigmaphiNBinit}    {\sigma_{\sss \phi,NB,INIT}} 
\newcommand{\phiNW}			  	{\vec \phi_{\sss NW}}
\newcommand{\phiNG}		  		{\vec \phi_{\sss NG}}
\newcommand{\phiBP}			  	{\vec \phi_{\sss BP}}
\newcommand{\phiBPest}		  	{\hat{\vec \phi}^{\sss BP}}
\newcommand{\phiBCest}		  	{\hat{\vec \phi}^{\sss BC}}
\newcommand{\phiBC}			  	{\vec \phi_{\sss BC}}
\newcommand{\phiWB}			  	{\vec \phi_{\sss WB}}
\newcommand{\chiTAS}	      	{\chi_{\sss TAS}}                  
\newcommand{\gammaTAS}  	  	{\gamma_{\sss TAS}}
\newcommand{\muTAS}       		{\mu_{\sss TAS}}
\newcommand{\gammaWIND}  		{\gamma_{\sss WIND}}
\newcommand{\chiWIND}       	{\chi_{\sss WIND}}

\newcommand{\psiP}				{\psi^{\sss P}}
\newcommand{\thetaP}			{\theta^{\sss P}}
\newcommand{\xiP}				{\xi^{\sss P}}

\newcommand{\psiC}				{\psi^{\sss C}}
\newcommand{\thetaC}			{\theta^{\sss C}}
\newcommand{\xiC}				{\xi^{\sss C}}
\newcommand{\psiCest}			{\hat{\psi}^{\sss C}}
\newcommand{\thetaCest}	 		{\hat{\theta}^{\sss C}}
\newcommand{\xiCest}			{\hat{\xi}^{\sss C}}
\newcommand{\phiNBest}		  	{\hat{\vec \phi}_{\sss NB}}

\newcommand{\psiest}			{\hat \psi}

\newcommand{\thetaest}			{\hat \theta}

\newcommand{\xiest}				{\hat \xi}

\newcommand{\alphaest}          {\hat{\alpha}}
\newcommand{\alphadot}          {\dot \alpha}

\newcommand{\betaest}			{\hat \beta}

\newcommand{\betadot}           {\dot \beta}

\newcommand{\chiMAG}			{\chi_{\sss MAG}}
\newcommand{\DeltaMAG}			{\Delta_{\sss MAG}}
\newcommand{\chiMAGest}			{\hat{\chi}_{\sss MAG}}
\newcommand{\alphatilde}        {\widetilde{\alpha}}
\newcommand{\betatilde}         {\widetilde{\beta}}

\newcommand{\DeltatOP}          {\Deltat_{\sss OP}}

\newcommand{\sigmapsiP}         {\nm{\sigma_{\psi^{\sss P}}}}
\newcommand{\sigmathetaP}  		{\nm{\sigma_{\theta^{\sss P}}}}
\newcommand{\sigmaxiP}     		{\nm{\sigma_{\xi^{\sss P}}}}
\newcommand{\sigmaphiBPest}		{\nm{\sigma_{\hat{\phi}^{\sss BP}}}}
\newcommand{\sigmaTBPBest}      {\nm{\sigma_{\hat{T}_{\sss BP}^{\sss B}}}}
\newcommand{\sigmapsiC}         {\nm{\sigma_{\psi^{\sss C}}}}
\newcommand{\sigmathetaC}  		{\nm{\sigma_{\theta^{\sss C}}}}
\newcommand{\sigmaxiC}     		{\nm{\sigma_{\xi^{\sss C}}}}
\newcommand{\sigmaphiBCest}		{\nm{\sigma_{\hat{\phi}^{\sss BC}}}}
\newcommand{\sigmaTBCBest}      {\nm{\sigma_{\hat{T}_{\sss BC}^{\sss B}}}}

\newcommand{\Hp}            {H_{\sss P}}

\newcommand{\Hpdot}         {\dot{H}_{\sss P}}
\newcommand{\Hpest}         {\hat{H}_{\sss P}}

\newcommand{\DeltaT}        {\Delta T}
\newcommand{\Deltap}        {\Delta p}
\newcommand{\Ttilde}    	{\widetilde{T}}
\newcommand{\ptilde} 	    {\widetilde{p}}
\newcommand{\Test}			{\hat{T}}

\newcommand{\mest}          {\hat{m}}

\newcommand{\DeltaTest}		{\Delta\hat{T}}
\newcommand{\Deltapest}		{\Delta\hat{p}}

\newcommand{\Tzero}         {T_{\sss 0}}                 
\newcommand{\pzero}         {p_{\sss 0}}                 
\newcommand{\rhozero}       {\rho_{\sss 0}}              
\newcommand{\THpzero}       {T_{\sss \Hp = 0}}           
\newcommand{\TISAHpzero}    {T_{\sss ISA,\Hp = 0}}       
\newcommand{\pHpzero}       {p_{\sss \Hp = 0}}           
\newcommand{\HpHpzero}      {H_{\sss P,\Hp = 0}}     	 
\newcommand{\HHpzero}       {H_{\sss \Hp = 0}}           

\newcommand{\TMSL}          {T_{\sss MSL}}               
\newcommand{\TISAMSL}       {T_{\sss ISA,MSL}}           
\newcommand{\pMSL}          {p_{\sss MSL}}               
\newcommand{\HpMSL}         {H_{\sss P,MSL}}             
\newcommand{\HMSL}          {H_{\sss MSL}}               
\newcommand{\hMSL}          {h_{\sss MSL}}               

\newcommand{\TISA}          {T_{\sss ISA}}              
\newcommand{\Hptrop}        {H_{\sss P,trop}}           
\newcommand{\betaT}         {\beta_{\sss T}}         	
\newcommand{\gzero}     	{g_{\sss 0}} 				
\newcommand{\RE}     	    {R_{\sss E}}   				
\newcommand{\gBR}           {-\dfrac{\gzero}{\betaT R}}	
\newcommand{\BRg}           {-\dfrac{\betaT R}{\gzero}}	

\newcommand{\FAER}  		{\vec F_{\sss AER}}
\newcommand{\FAERB}  		{\vec F_{\sss AER}^{\sss B}}
\newcommand{\FAERBfullskew}	{\widehat{\vec F}_{{\sss AER},full}^{\sss B}}
\newcommand{\FAERW}  		{\vec F_{\sss AER}^{\sss W}}
\newcommand{\MAERB}  		{\vec M_{\sss AER}^{\sss B}}

\newcommand{\FPRO}  		{\vec F_{\sss PRO}}
\newcommand{\FPROB}  		{\vec F_{\sss PRO}^{\sss B}}
\newcommand{\MPROB}  		{\vec M_{\sss PRO}^{\sss B}}

\newcommand{\fIB}				{\vec f_{\sss {IB}}}
\newcommand{\fIBB}				{\vec f_{\sss IB}^{\sss B}}
\newcommand{\fIBN}				{\vec f_{\sss IB}^{\sss N}}
\newcommand{\fIBBtilde}			{\widetilde{\vec f}_{\sss IB}^{\sss B}}
\newcommand{\fIBBtildebar}		{\bar{\widetilde{\vec f}}_{\sss IB}^{\sss B}}
\newcommand{\fIBBest}			{\hat{\vec f}_{\sss IB}^{\sss B}}
\newcommand{\fIBBdot}			{\vec{\dot f}_{\sss IB}^{\sss B}}

\newcommand{\fIBBtildebari}		{\bar{\widetilde{f}}_{{\sss IB},1}^{\sss B}}
\newcommand{\fIBBtildebarii}	{\bar{\widetilde{f}}_{{\sss IB},2}^{\sss B}}
\newcommand{\fIBBtildebariii}	{\bar{\widetilde{f}}_{{\sss IB},3}^{\sss B}}
\newcommand{\fIBBtildebarsquareii}	{\bar{\widetilde{f}}_{{\sss IB},2}^{{\sss B} \ 2}}
\newcommand{\fIBBtildebarsquareiii}	{\bar{\widetilde{f}}_{{\sss IB},3}^{{\sss B} \ 2}}

\newcommand{\fIPP}				{\vec f_{\sss IP}^{\sss P}}
\newcommand{\fIPPtilde}			{\widetilde{\vec f}_{\sss IP}^{\sss P}}
\newcommand{\fIPPtildetilde}	{\widetilde{\widetilde{\vec f}}_{\sss IP}^{\sss P}}

\newcommand{\fIAA}				{\vec f_{\sss IA}^{\sss A}}
\newcommand{\fIAAtilde}			{\widetilde{\vec f}_{\sss IA}^{\sss A}}

\newcommand{\Nu}				{N_u}
\newcommand{\NuACC}				{\vec N_{u,\sss ACC}}

\newcommand{\Nuzero}			{N_{u0}}
\newcommand{\NuzeroACC}			{\vec N_{u0,\sss ACC}}

\newcommand{\NuzeroACCiest}		{\hat N_{u0,\sss{ACC,i}}}
\newcommand{\NuzeroGYR}			{\vec N_{u0,\sss GYR}}

\newcommand{\NuzeroGYRiest}		{\hat N_{u0,\sss{GYR,i}}}
\newcommand{\NuzeroMAG}			{\vec N_{u0,\sss MAG}}

\newcommand{\NuzeroMAGiest}		{\hat N_{u0,\sss{MAG,i}}}
\newcommand{\NhiMAG}			{\vec N_{\sss{HI},\sss MAG}}
\newcommand{\NzeroAOA}			{N_{0,\sss AOA}}
\newcommand{\NzeroAOS}			{N_{0,\sss AOS}}
\newcommand{\NzeroOSP}			{N_{0,\sss OSP}}
\newcommand{\NzeroOAT}			{N_{0,\sss OAT}}
\newcommand{\NzeroTAS}			{N_{0,\sss TAS}}
\newcommand{\Nui}				{N_{ui}}
\newcommand{\NuiACC}			{\vec N_{ui,\sss ACC}}
\newcommand{\NuiGYR}			{\vec N_{ui,\sss GYR}}
\newcommand{\Nv}				{N_v}
\newcommand{\Nvs}				{N_{vs}}
\newcommand{\Nvi}				{N_{vi}}
\newcommand{\NvACC}				{\vec N_{v,\sss ACC}}
\newcommand{\NvsACC}			{\vec N_{vs,\sss ACC}}
\newcommand{\NvGYR}				{\vec N_{v,\sss GYR}}
\newcommand{\NvsGYR}			{\vec N_{vs,\sss GYR}}
\newcommand{\NvsMAG}			{\vec N_{vs,\sss MAG}}
\newcommand{\NsAOA}				{N_{s,\sss AOA}}
\newcommand{\NsAOS}				{N_{s,\sss AOS}}
\newcommand{\NsTAS}				{N_{s,\sss TAS}}
\newcommand{\NsOSP}				{N_{s,\sss OSP}}
\newcommand{\NsOAT}				{N_{s,\sss OAT}}
\newcommand{\NgGNSSPOS}   		{\vec N_{g,\sss GNSS,POS}}
\newcommand{\NgGNSSVEL}			{\vec N_{g,\sss GNSS,VEL}}
\newcommand{\NzeroGNSSION}		{\vec N_{u0,\sss GNSS,ION}}
\newcommand{\NjGNSSION}			{\vec N_{j,\sss GNSS,ION}}
\newcommand{\NpsiP}				{N_{\psi^{\sss P}}}
\newcommand{\NthetaP}			{N_{\theta^{\sss P}}}
\newcommand{\NxiP}				{N_{\xi^{\sss P}}}
\newcommand{\NpsiPest}			{N_{\hat{\psi}^{\sss P}}}
\newcommand{\NthetaPest}		{N_{\hat{\theta}^{\sss P}}}
\newcommand{\NxiPest}			{N_{\hat{\xi}^{\sss P}}}
\newcommand{\NTBPBiest}			{N_{\hat{T}_{{\sss BP},1}^{\sss B}}}
\newcommand{\NTBPBiiest}		{N_{\hat{T}_{{\sss BP},2}^{\sss B}}}
\newcommand{\NTBPBiiiest}		{N_{\hat{T}_{{\sss BP},3}^{\sss B}}}
\newcommand{\NpsiC}				{N_{\psi^{\sss C}}}
\newcommand{\NthetaC}			{N_{\theta^{\sss C}}}
\newcommand{\NxiC}				{N_{\xi^{\sss C}}}
\newcommand{\NpsiCest}			{N_{\hat{\psi}^{\sss C}}}
\newcommand{\NthetaCest}		{N_{\hat{\theta}^{\sss C}}}
\newcommand{\NxiCest}			{N_{\hat{\xi}^{\sss C}}}
\newcommand{\NTBCBiest}			{N_{\hat{T}_{{\sss BC},1}^{\sss B}}}
\newcommand{\NTBCBiiest}		{N_{\hat{T}_{{\sss BC},2}^{\sss B}}}
\newcommand{\NTBCBiiiest}		{N_{\hat{T}_{{\sss BC},3}^{\sss B}}}

\newcommand{\EACC}		    {\vec E_{\sss ACC}}
\newcommand{\EACCdot}	    {\vec{\dot E}_{\sss ACC}}
\newcommand{\EACCinit}      {\vec E_{\sss ACC,INIT}}
\newcommand{\EACCest}	    {\hat{\vec E}_{\sss ACC}}
\newcommand{\EACCzeroest}	{\hat{\vec E}_{0,{\sss ACC}}}
\newcommand{\EACCzero}	    {\vec E_{0,\sss ACC}}

\newcommand{\sigmaEACCinit}	{\sigma_{\sss E,ACC,INIT}}
\newcommand{\BzeroACC}		{B_{0,\sss{ACC}}}
\newcommand{\BzeroACCest}	{\hat{\vec B}_{0,\sss{ACC}}}

\newcommand{\sigmauACC}		{\sigma_{u\sss{ACC}}}  
\newcommand{\sigmavACC}		{\sigma_{v\sss{ACC}}}
\newcommand{\EGYR}			{\vec E_{\sss GYR}}

\newcommand{\EGYRdot}		{\vec{\dot E}_{\sss GYR}}
\newcommand{\EGYRinit}  	{\vec E_{\sss GYR,INIT}}
\newcommand{\EGYRzero}	    {\vec E_{0, \sss GYR}}
\newcommand{\sigmaEGYRinit}	{\sigma_{\sss E,GYR,INIT}}
\newcommand{\EGYRest}		{\hat{\vec E}_{\sss GYR}}

\newcommand{\EGYRzeroest}	{\hat{\vec E}_{0,{\sss GYR}}}

\newcommand{\BzeroGYR}		{B_{0,\sss{GYR}}}
\newcommand{\BzeroGYRest}	{\hat{\vec B}_{0,\sss{GYR}}}

\newcommand{\sigmauGYR}		{\sigma_{u\sss{GYR}}}
\newcommand{\sigmavGYR}		{\sigma_{v\sss{GYR}}}

\newcommand{\SACC}			{\vec S_{\sss ACC}}
\newcommand{\sACC}			{s_{\sss ACC}}
\newcommand{\sACCXi}		{s_{\sss {ACC,i}}}
\newcommand{\sACCXiest}		{\hat{s}_{\sss {ACC,i}}}
\newcommand{\sACCi}			{s_{\sss {ACC},1}}
\newcommand{\sACCii}		{s_{\sss {ACC},2}}
\newcommand{\sACCiii}		{s_{\sss {ACC},3}}
\newcommand{\SGYR}			{\vec S_{\sss GYR}}
\newcommand{\sGYR}			{s_{\sss GYR}}

\newcommand{\sGYRXiest}		{\hat{s}_{\sss {GYR,i}}}
\newcommand{\sGYRi}			{s_{\sss {GYR},1}}
\newcommand{\sGYRii}		{s_{\sss {GYR},2}}
\newcommand{\sGYRiii}		{s_{\sss {GYR},3}}

\newcommand{\alphaACC}			{\alpha_{\sss ACC}}
\newcommand{\alphaACCXi}		{\alpha_{{\sss ACC},i}}

\newcommand{\alphaACCXj}		{\alpha_{{\sss ACC},j}}
\newcommand{\alphaACCi}			{\alpha_{{\sss ACC},1}}
\newcommand{\alphaACCii}		{\alpha_{{\sss ACC},2}}
\newcommand{\alphaACCiii}		{\alpha_{{\sss ACC},3}}

\newcommand{\alphaGYR}			{\alpha_{\sss GYR}}

\newcommand{\alphaGYRXij}		{\alpha_{{\sss GYR},ij}}

\newcommand{\alphaGYRiXii}		{\alpha_{{\sss GYR},12}}
\newcommand{\alphaGYRiiXi}		{\alpha_{{\sss GYR},21}}
\newcommand{\alphaGYRiXiii}		{\alpha_{{\sss GYR},13}}
\newcommand{\alphaGYRiiiXi}		{\alpha_{{\sss GYR},31}}
\newcommand{\alphaGYRiiXiii}	{\alpha_{{\sss GYR},23}}
\newcommand{\alphaGYRiiiXii}	{\alpha_{{\sss GYR},32}}

\newcommand{\MACC}			{\vec M_{\sss ACC}}
\newcommand{\mACC}			{m_{\sss ACC}}
\newcommand{\MACCest}		{\hat{\vec M}_{\sss ACC}}
\newcommand{\mACCXij}		{m_{\sss {ACC,ij}}}
\newcommand{\mACCXijest}	{\hat{m}_{\sss {ACC,ij}}}
\newcommand{\NACC}			{\vec N_{\sss ACC}}
\newcommand{\nACCXij}		{n_{\sss {ACC,ij}}}
\newcommand{\MGYR}			{\vec M_{\sss GYR}}
\newcommand{\mGYR}			{m_{\sss GYR}}
\newcommand{\MGYRest}		{\hat{\vec M}_{\sss GYR}}
\newcommand{\mGYRXij}		{m_{\sss {GYR,ij}}}
\newcommand{\mGYRXijest}	{\hat{m}_{\sss {GYR,ij}}}

\newcommand{\sigmaAOA}		{\sigma_{\sss AOA}}
\newcommand{\sigmaAOS}		{\sigma_{\sss AOS}}
\newcommand{\sigmaOSP}		{\sigma_{\sss OSP}}
\newcommand{\sigmaOAT}		{\sigma_{\sss OAT}}
\newcommand{\sigmaTAS}		{\sigma_{\sss TAS}}
\newcommand{\BzeroAOA}		{B_{0\sss{AOA}}}
\newcommand{\BzeroAOS}		{B_{0\sss{AOS}}}
\newcommand{\BzeroOSP}		{B_{0\sss{OSP}}}
\newcommand{\BzeroOAT}		{B_{0\sss{OAT}}}
\newcommand{\BzeroTAS}		{B_{0\sss{TAS}}}

\newcommand{\sigmaPOSvec}		{\vec {\sigma}_{\sss POS}}
\newcommand{\sigmaGNSSPOS}		{\sigma_{\sss GNSS,POS}}
\newcommand{\sigmaGNSSPOSHOR}	{\sigma_{\sss GNSS,POS,HOR}}
\newcommand{\sigmaGNSSPOSVER}	{\sigma_{\sss GNSS,POS,VER}}

\newcommand{\sigmaVELvec}		{\vec {\sigma}_{\sss VEL}}
\newcommand{\sigmaGNSSVEL}		{\sigma_{\sss GNSS,VEL}}

\newcommand{\sigmaGNSSION}		{\sigma_{\sss GNSS,ION}}
\newcommand{\BzeroGNSSION}		{B_{0,\sss GNSS,ION}}

\newcommand{\BNi}	    	{B_{1}^{\sss N}}
\newcommand{\BNii}	    	{B_{2}^{\sss N}}
\newcommand{\BNiii}	    	{B_{3}^{\sss N}}
\newcommand{\BNhor}	    	{B_{\sss {HOR}}^{\sss N}}
\newcommand{\BNMODEL}    	{\vec B_{\sss MOD}^{\sss N}}
\newcommand{\BNMODELi}	   	{B_{{\sss MOD},1}^{\sss N}}
\newcommand{\BNMODELii}	   	{B_{{\sss MOD},2}^{\sss N}}

\newcommand{\BNREAL}    	{\vec B_{\sss REAL}^{\sss N}}

\newcommand{\BNDEV}			{\vec B_{\sss DEV}^{\sss N}}
\newcommand{\BNDEVi}	   	{B_{{\sss DEV},1}^{\sss N}}
\newcommand{\BNDEVii}	   	{B_{{\sss DEV},2}^{\sss N}}
\newcommand{\BNDEViii}   	{B_{{\sss DEV},3}^{\sss N}}
\newcommand{\BNDEVdot}		{\vec{\dot B}_{\sss DEV}^{\sss N}}
\newcommand{\BNDEVinit}		{\vec B_{\sss DEV,INIT}^{\sss N}}
\newcommand{\BNDEVest}		{\hat{\vec B}_{\sss DEV}^{\sss N}}
\newcommand{\BNDEVzeroest}	{\hat{\vec B}_{0, \sss DEV}^{\sss N}}
\newcommand{\BNDEVzero}		{\vec B_{0, \sss DEV}^{\sss N}}
\newcommand{\sigmaBNDEVinit}{\sigma_{\sss BN,DEV,INIT}}
\newcommand{\BBREAL}    	{\vec B_{\sss REAL}^{\sss B}}
\newcommand{\BBtilde}		{\widetilde{\vec B}^{\sss B}}
\newcommand{\BBtildetilde}	{\widetilde{\widetilde{\vec B}}^{\sss B}}

\newcommand{\EMAG}	    		{\vec E_{\sss MAG}}
\newcommand{\EMAGdot}	    	{\vec{\dot E}_{\sss MAG}}
\newcommand{\EMAGinit}		    {\vec E_{\sss MAG,INIT}}
\newcommand{\EMAGest}	    	{\hat{\vec E}_{\sss MAG}}
\newcommand{\EMAGzeroest}		{\hat{\vec E}_{0,{\sss MAG}}}
\newcommand{\EMAGzero}	    	{\vec E_{0, \sss MAG}}
\newcommand{\sigmaEMAGinit}		{\sigma_{\sss E,MAG,INIT}}

\newcommand{\BzeroMAG}			{B_{0,\sss MAG}}
\newcommand{\BzeroMAGvec}		{\vec B_{0,\sss MAG}}
\newcommand{\BzeroMAGvecest}	{\hat{\vec B}_{0,\sss MAG}}

\newcommand{\BhiMAG}			{B_{\sss{HI,MAG}}}
\newcommand{\BhiMAGvec}		 	{\vec B_{\sss{HI,MAG}}}
\newcommand{\BhiMAGvecest}		{\hat{\vec B}_{\sss{HI,MAG}}}

\newcommand{\BhiMAGXiest}		{\hat{B}_{\sss {HI,MAG,i}}}

\newcommand{\MMAG}			{\vec M_{\sss MAG}}
\newcommand{\MMAGest}		{\hat{\vec M}_{\sss MAG}}
\newcommand{\BBtildebar}	{\bar{\widetilde{\vec B}}^{\sss B}}
\newcommand{\BBtildebari}	{\bar{\widetilde{B}}_{1}^{\sss B}}
\newcommand{\BBtildebarii}	{\bar{\widetilde{B}}_{2}^{\sss B}}
\newcommand{\BBtildebariii}	{\bar{\widetilde{B}}_{3}^{\sss B}}
\newcommand{\sigmavMAG}		{\sigma_{v,\sss MAG}}
\newcommand{\sMAG}			{s_{\sss MAG}}
\newcommand{\mMAG}			{m_{\sss MAG}}

\newcommand{\sMAGXiest}		{\hat{s}_{\sss {MAG,i}}}

\newcommand{\mMAGXijest}	{\hat{m}_{\sss {MAG,ij}}}

\newcommand{\seedA} 		{\Upsilon_{i,\sss A}}
\newcommand{\seedAACC} 		{\upsilon_{i,\sss A,\sss ACC}}
\newcommand{\seedAGYR}		{\upsilon_{i,\sss A,\sss GYR}}
\newcommand{\seedAMAG}		{\upsilon_{i,\sss A,\sss MAG}}
\newcommand{\seedAPLAT}		{\upsilon_{i,\sss A,\sss PLAT}}
\newcommand{\seedACAM}		{\upsilon_{i,\sss A,\sss CAM}}

\newcommand{\seedR} 		{\Upsilon_{j,\sss F}}
\newcommand{\seedRACC} 		{\upsilon_{j,\sss F,\sss ACC}}
\newcommand{\seedRGYR}		{\upsilon_{j,\sss F,\sss GYR}}
\newcommand{\seedRMAG}		{\upsilon_{j,\sss F,\sss MAG}}
\newcommand{\seedROSP}		{\upsilon_{j,\sss F,\sss OSP}}
\newcommand{\seedROAT}		{\upsilon_{j,\sss F,\sss OAT}}
\newcommand{\seedRTAS}		{\upsilon_{j,\sss F,\sss TAS}}
\newcommand{\seedRAOA}		{\upsilon_{j,\sss F,\sss AOA}}
\newcommand{\seedRAOS}		{\upsilon_{j,\sss F,\sss AOS}}
\newcommand{\seedRGNSS}		{\upsilon_{j,\sss F,\sss GNSS}}

\newcommand{\seedRphiNB}	{\upsilon_{j,\sss F,\sss \phi NB}}
\newcommand{\seedREGYR}		{\upsilon_{j,\sss F,\sss EGYR}}
\newcommand{\seedREACC}		{\upsilon_{j,\sss F,\sss EACC}}
\newcommand{\seedREMAG}		{\upsilon_{j,\sss F,\sss EMAG}}
\newcommand{\seedRBNDEV}	{\upsilon_{j,\sss F,\sss BNDEV}}

\newcommand{\seedRWEATHER}	{\upsilon_{j,\sss F,\sss WEATHER}}
\newcommand{\seedRWIND}		{\upsilon_{j,\sss F,\sss WIND}}
\newcommand{\seedRTURB}		{\upsilon_{j,\sss F,\sss TURB}}
\newcommand{\seedRMODELS}	{\upsilon_{j,\sss F,\sss MODELS}}

\newcommand{\seedRGUID}		{\upsilon_{j,\sss F,\sss GUID}}

\newcommand{\uvec}						{\vec u}
\newcommand{\utilde}					{\widetilde{\vec u}}
\newcommand{\wvec}						{\vec w}
\newcommand{\wtilde}					{\widetilde{\vec w}}
\newcommand{\Avec}						{\vec A}
\newcommand{\Bvec}						{\vec B}

\newcommand{\Lvec}						{\vec L}

\newcommand{\Qtilde}					{\widetilde{\vec Q}}
\newcommand{\zvec}						{\vec z}

\newcommand{\Pvec}						{\vec P}

\newcommand{\yvec}						{\vec y}
\newcommand{\Hvec}						{\vec H}
\newcommand{\Mvec}						{\vec M}
\newcommand{\Rtilde}					{\widetilde{\vec R}}

\newcommand{\pIMG}          {\vec p^{\sss IMG}}
\newcommand{\cIMG}          {\vec c^{\sss IMG}}
\newcommand{\FIMG}		    {F_{\sss IMG}}
\newcommand{\OIMG}          {O_{\sss IMG}}                            
\newcommand{\pIMGi}         {p_{1}^{\sss IMG}}
\newcommand{\pIMGii}        {p_{2}^{\sss IMG}}

\newcommand{\iIMGi}         {\vec i_{1}^{\sss IMG}}
\newcommand{\iIMGii}        {\vec i_{2}^{\sss IMG}}
\newcommand{\cIMGi}         {c_{1}^{\sss IMG}}
\newcommand{\cIMGii}        {c_{2}^{\sss IMG}}
\newcommand{\sPX}           {s_{\sss PX}}

\newcommand{\fovh}          {\Theta_{\sss H}}
\newcommand{\fovv}          {\Theta_{\sss V}}

\newcommand{\Sh}            {S_{\sss H}}
\newcommand{\Sv}            {S_{\sss V}}

\begin{document}

\begin{titlepage}
	\begin{center}
		\begin{Huge} \hspace*{0cm} \bigskip \bigskip \bigskip \bigskip \bigskip \end{Huge}
		\\[2.5cm]
		
		\begin{Huge}
			\begin{spacing}{1.5}
				\textcolor{black!80!gray}{\textbf{\docname}}
			\end{spacing}
		\end{Huge}
		
		\vspace{4.0cm}
		
		\begin{Large}
			\begin{spacing}{1.4}
				\textcolor{black!80!gray}{\textbf{Universidad Politécnica de Madrid \\ Centro de Automática y Robótica}}
			\end{spacing}
		\end{Large}
		
		\vspace{1.5cm}
		
		\begin{Large}
			\begin{spacing}{1.4}
				\textcolor{black!80!gray}{\textbf{\docdate}}
			\end{spacing}
		\end{Large}
		
		\vspace{2.5cm}
		
		\begin{large}
			\begin{tabular}{p{1.7cm}p{4.0cm}}
			Author: & \rule[-12pt]{0pt}{12pt}\textbf{Eduardo Gallo} \\
			\end{tabular}
		\end{large}
		
	\end{center}
\end{titlepage}

 		\thispagestyle{empty} \cleardoublepage	

\addtocontents{toc}{\cftpagenumbersoff{chapter}}
\chapter*{Abstract} \label{cha:Abstract} 
\addcontentsline{toc}{chapter}{Abstract} 

This document describes the architecture and algorithms of a high fidelity fixed wing flight simulator intended to test and validate novel guidance, navigation, and control (\hypertt{GNC}) algorithms for autonomous aircraft. The flight simulator aims to replicate the influence of as many factors as possible on the aircraft performances, the Earth model on which the flight takes place, the physics of flight and the associated equations of motion, and in particular the behavior of the onboard sensors, limiting the assumptions or simplifications to the bare minimum, and including multiple relatively minor effects not usually considered in simulation that may in occasions play a role in the \hypertt{GNC} algorithms not performing as intended. The author releases the flight simulator \nm{\CC} implementation as open-source software. 

The simulator modular design enables the user to replace the standard \hypertt{GNC} algorithms with new ones, with the objective of evaluating their performances when subject to specific missions and meteorological conditions (atmospheric properties, wind field, air turbulence). The testing and evaluation is performed by means of Monte Carlo simulations, as most simulation modules (such as the aircraft mission, the meteorological conditions, the errors introduced by the sensors, and the initial conditions) are defined stochastically and hence vary in a pseudo-random way from one execution to the next according to certain user-defined input parameters, ensuring that the results are valid for a wide range of conditions.

In addition to modeling the outputs of all sensors usually present onboard a fixed wing platform, such as accelerometers, gyroscopes, magnetometers, Pitot tube, air vanes, and a Global Navigation Satellite System (\hypertt{GNCC}) receiver, the simulator is also capable of generating realistic images of the Earth surface that resemble what an onboard camera would record if following the resulting trajectory, enabling the use and evaluation of  visual and visual inertial navigation systems.

\textbf{\emph{Keywords}}: flight simulator, \hypertt{GNC}, sensors, stochastic, fixed wing

   		\thispagestyle{empty} \cleardoublepage
\addtocontents{toc}{\cftpagenumbersoff{chapter}}
\chapter*{Notation} \label{cha:Notation} 
\addcontentsline{toc}{chapter}{Notation} 
\addtocontents{toc}{\cftpagenumberson{chapter}}

The meaning of a given variable or symbol is always clarified on its first appearance, or otherwise a cross reference is included to the location where it is properly explained. This document is organized into chapters and appendices (e.g., chapter \ref{cha:Sensors} or appendix \ref{cha:RefSystems}), sections (e.g., section \ref{sec:Sensors_Inertial} or \ref{sec:RefSystems_Acft}), subsections (e.g., section \ref{subsec:Sensors_Accelerometer_Triad_ErrorModel} or \ref{subsec:RefSystems_W}), and so on. Note that the prefix ``sub'' is not employed when referring to subsections. Referencing tables (e.g., table \ref{tab:Summary_trajectory_errors}) and figures (e.g., figure \ref{tab:Intro_trajectory_frequencies}) is also straightforward. When a reference appears in between parenthesis, it refers to an equation or mathematical expression, as in (\ref{eq:EquationsMotion_state_vector}); when in between square brackets, to a bibliographic reference, as in \cite{LIE}; and when preceded by an \say{A} and in between square brackets, it refers to an assumption listed in appendix \ref{cha:Assumptions}, e.g. [A\ref{as:GNC_n_synchronize}]. Acronyms are displayed in uppercase teletype font, and reference to the proper location within the acronym list, as in \hypertt{GNSS}.

With respect to the mathematical notation, vectors and matrices appear in bold (e.g., \nm{\vec x} or \nm{\vec R}), and in general the former are lowercase and the latter uppercase, although not always. A variable or vector with a hat over it \nm{< \hat{\cdot} >} refers to its (inertial) estimated value, with a circular accent \nm{<\circled{\cdot}>} to its (visual) estimated value, with a tilde \nm{< \widetilde{\cdot} >} to its measured value, and with a dot \nm{< \dot{\cdot} >} to its time derivative. Other commonly employed symbols for vectors are the wide hat \nm{< \widehat{\cdot} >}, which refers to its skew-symmetric form, the double vertical bars \nm{< \| \cdot \| >}, which refer to its norm, and the wedge \nm{<\cdot^\wedge>}, which implies a tangent space representation. In the case of a scalar, the vertical bars \nm{< | \cdot | >} refer to its absolute value.

Appendix \ref{cha:RefSystems} defines the various reference frames employed throughout this document, which are listed in table \ref{tab:RefSystems_RefSystems}. Each is represented by an uppercase F followed by a subindex identifying the frame, such as \nm{\FB} for the body frame. The reference frames are employed so often that it is not practical to use cross references to the proper sections within appendix \ref{cha:RefSystems} every time a frame appears in the text; most have thus been omitted.

Superindexes are employed over vectors to specify the reference system in which they are viewed or evaluated (e.g., \nm{\vN} refers to ground velocity viewed in \hypertt{NED} or \nm{\FN}, while \nm{\vB} is the same vector but viewed in the body frame \nm{\FB}). Subindexes may be employed to clarify the meaning of the variable or vector, such as in \nm{\vTAS} for airspeed vector instead of the ground velocity \nm{\vec v}, or to refer to a given component of a vector (e.g. \nm{\vNii} refers to the second component of \nm{\vN}). In addition, where two reference systems appear as subindexes to a vector, it means that the vector goes from the first system to the second. For example, \nm{\wNBB} refers to the angular velocity from \nm{\FN} to \nm{\FB} viewed in \nm{\FB}\footnote{This is analogous to the angular velocity of \nm{\FB} with respect to \nm{\FN} viewed in \nm{\FB}.}.

This document heavily relies on the linear and Lie algebra notions discussed in \cite{LIE}, which include:
\begin{itemize}
\item Random variables, distributions, stochastic processes, white noise, and robust statistics.
\item Euclidean spaces and Lie groups, including the various parameterizations of the Special Orthogonal group of \nm{\mathbb{R}^3} or \nm{\mathbb{SO}(3)}, which represents the rigid body rotations (attitudes), and the Special Euclidean group of \nm{\mathbb{R}^3} or \nm{\mathbb{SE}(3)} representing full rigid body motions (poses).
\item Notions of Lie algebra, such as manifolds, tangent spaces, perturbations, plus and minus operators, time derivatives and velocities, adjoints, covariances, and Jacobians.
\item Common mathematical techniques such as discrete integration, gradient descent optimization, and state estimation for Euclidean spaces and their adaptations to Lie groups.
\item Composition of the linear and angular velocities and accelerations of multiple rigid bodies.
\end{itemize}

As in the case of the reference frames, it is not practical to include an external reference to \cite{LIE} every time a concept discussed in it is employed. References to \cite{LIE} are present in some occasions, but the reader should automatically refer to it every time that additional information about one of the above topics is required.

Many different types of Jacobians or function partial derivative matrices can be found in this document (a formal definition can be found in \cite{LIE}), and a special notation is hence required. Jacobians are represented by a \nm{\vec J} combined with a subindex and a superindex:
\begin{itemize}
\item The subindex provides information about the function domain, and is composed by a symbol representing how increments are added to the domain, followed by the domain itself. A \nm{+} is employed when the domain belongs to an Euclidean space\footnote{The subindex symbol may sometimes be omitted when it is clear from the context that the domain is Euclidean.} but not when it is a manifold or Lie group; in these cases, \nm{\oplus} means that increments are added in the local tangent space and \nm{\boxplus} that they are added in the global tangent space. 
\item The superindex is composed by a second symbol representing how differences are computed in the function image or codomain, followed by the codomain itself. A \nm{-} is employed when the codomain is Euclidean\footnote{The superindex symbol may sometimes be omitted when it is clear from the context that the codomain is Euclidean.}, \nm{\ominus} when the differences are evaluated in the local tangent space to the codomain manifold, and \nm{\boxminus} when evaluated in the global tangent space.
\end{itemize}

Several \nm{\mathbb{SO}\lrp{3}} and \nm{\mathbb{SE}\lrp{3}} representations are possible \cite{LIE}, and while they are fully equivalent, each application has its preferred parameterization. As such, rotation matrices (e.g., \nm{\RNB}), rotation vectors (e.g., \nm{\vec r_{\sss NB}}), unit quaternions (e.g., \nm{\qNB}), and Euler angles (e.g., \nm{\phiNB}) are all employed as required, without indicating that they refer to the same concept, e.g., the \nm{\mathcal{R}_{\sss NB}} rotation between the spatial \nm{\FN} and local \nm{\FB} frames. Similarly, affine representations (e.g., \nm{\REB, \, \TEBE}), homogeneous matrices (e.g., \nm{\MEB}), transform vectors (e.g., \nm{\vec \tau_{\sss EB}}), and unit dual quaternions (e.g., \nm{\zetaEB}), are also employed as required referring to the \nm{\mathcal{M}_{\sss EB}} rigid body motion between a spatial frame \nm{\FE} and a local one \nm{\FB}. The most widely used parameterizations are however the unit quaternion \nm{\vec q} for rotations and the unit dual quaternion \nm{\vec \zeta} for rigid body motions.

The most common symbols related with the use of \nm{\mathbb{SO}(3)} and \nm{\mathbb{SE}(3)} Lie groups are \nm{\circ} for concatenation, \nm{\otimes} for both the unit quaternion and the unit dual quaternion products, \nm{< \cdot^{-1} >} for the inverse, \nm{\oplus} and \nm{\ominus} for local perturbations, \nm{\boxplus} and \nm{\boxminus} for global perturbations, \nm{\vec{Ad}} for the adjoint matrix, \nm{\vec g()} for the transformation of points, and \nm{\vec g_*()} for that of vectors.

Last, note that there are instances in this document in which it is necessary to integrate, optimize, or estimate states that include \nm{\mathbb{SO}(3)} and \nm{\mathbb{SE}(3)} instances. As these are not Euclidean\footnote{Although Euclidean spaces are formally defined in \cite{LIE}, they can be informally understood as those that comply with the five axioms of Euclidean geometry (\nm{\first} things that are equal to the same thing are also equal to one another; \nm{\second} if equals be added to equals, the wholes are equal; \nm{\third} if equals be subtracted from equals, the remainders are equal; \nm{\fourth} things that coincide with one another are equal to one another; \nm{\fifth} the whole is greater than the part).}, the author has relied on Lie theory to adapt common calculus techniques such as Runge-Kutta integration, Gauss-Newton minimization, and Extended Kalman Filtering or \hypertt{EKF} state estimation so they operate within the \nm{\mathbb{SO}(3)} or \nm{\mathbb{SE}(3)} manifolds, as explained in \cite{LIE}. This approach, which has recently experienced increased usage in the field of robotics and avoids the simplifications usually employed in aircraft navigation and trajectory prediction, is not only more rigorous and elegant, but also avoids the accumulation of errors caused by continuously reprojecting the resulting poses back into the manifold.

   		\thispagestyle{empty} \cleardoublepage
\addtocontents{toc}{\cftpagenumbersoff{chapter}}
\chapter*{Acronyms} \label{cha:Acronyms} 
\addcontentsline{toc}{chapter}{Acronyms} 
\addtocontents{toc}{\cftpagenumberson{chapter}}

\renewcommand{\arraystretch}{1.25}
\begin{table}[ht]
\begin{tabular}{lp{6.0cm}p{0.1cm}lp{6.0cm}}
	\hypertarget{ACC}{\texttt{ACC}}			& ACCelerometers 								& & \hypertarget{ISO}{\texttt{ISO}}			& International Organization for			\\
	\hypertarget{ADS}{\texttt{ADS}}			& Air Data System								& &											& Standardization							\\
	\hypertarget{AOA}{\texttt{AOA}}			& Angle Of Attack								& & \hypertarget{MAG}{\texttt{MAG}}			& MAGnetometers								\\
	\hypertarget{AOS}{\texttt{AOS}}			& Angle Of Sideslip								& & \hypertarget{MEMS}{\texttt{MEMS}}		& Micro Machined Electromechanical			\\
	\hypertarget{AT}{\texttt{AT}} 			& Actual Trajectory								& &											& System									\\ 
	\hypertarget{AVL}{\texttt{AVL}}			& Athena Vortex Lattice							& & \hypertarget{MSL}{\texttt{MSL}}			& Mean Sea Level							\\
	\hypertarget{CAM}{\texttt{CAM}}			& CAMera										& & \hypertarget{NED}{\texttt{NED}}			& North East Down							\\
	\hypertarget{CEP}{\texttt{CEP}}			& Circular Error Probability					& & \hypertarget{NSE}{\texttt{NSE}}			& Navigation System Error					\\
	\hypertarget{D}{\texttt{D}}				& Derivative									& & \hypertarget{OAT}{\texttt{OAT}}			& Outside Air Temperature					\\
	\hypertarget{ECEF}{\texttt{ECEF}}		& Earth Centered Earth Fixed					& & \hypertarget{OSP}{\texttt{OSP}}			& Outside Static Pressure					\\
	\hypertarget{EGM96}{\texttt{EGM96}}		& Earth Gravitational Model 1996				& & \hypertarget{OT}{\texttt{OT}}			& Observed Trajectory						\\
	\hypertarget{EKF}{\texttt{EKF}}			& Extended Kalman Filter						& & \hypertarget{P}{\texttt{P}}				& Proportional								\\
	\hypertarget{EUROCONTROL}{\texttt{EUROCONTROL}} & European Organization for the Safety	& & \hypertarget{PID}{\texttt{PID}}			& Proportional Integral Derivative			\\
											& of Air Navigation								& & \hypertarget{PSD}{\texttt{PSD}}			& Power Spectral Density					\\
	\hypertarget{FTE}{\texttt{FTE}}			& Flight Technical Error						& & \hypertarget{PSFC}{\texttt{PSFC}} 		& Power Specific Fuel Consumption			\\
	\hypertarget{GLONASS}{\texttt{GLONASS}}	& GLObal NAvigation Satellite System			& & \hypertarget{ROC}{\texttt{ROC}}			& Rate Of Climb								\\
	\hypertarget{GNC}{\texttt{GNC}}			& Guidance, Navigation, and Control				& & \hypertarget{RT}{\texttt{RT}}			& Reference Trajectory						\\
	\hypertarget{GNSS}{\texttt{GNSS}}		& Global Navigation Satellite System			& & \hypertarget{ST}{\texttt{ST}}			& Sensed Trajectory							\\
	\hypertarget{GPS}{\texttt{GPS}}			& Global Positioning System						& & \hypertarget{SWaP}{\texttt{SWaP}}		& Size, Weight, and Power					\\
	\hypertarget{GYR}{\texttt{GYR}}			& GYRoscopes									& & \hypertarget{TAS}{\texttt{TAS}}			& True Air Speed							\\
	\hypertarget{I}{\texttt{I}}				& Integral										& & \hypertarget{TSE}{\texttt{TSE}}			& Total System Error						\\
	\hypertarget{ICAO}{\texttt{ICAO}}		& International Civil Aviation					& & \hypertarget{UAS}{\texttt{UAS}}			& Unmanned Aircraft System					\\
											& Organization									& & \hypertarget{UAV}{\texttt{UAV}}			& Unmanned Aerial Vehicle					\\
	\hypertarget{IMU}{\texttt{IMU}}			& Inertial Measurement Unit				        & & \hypertarget{VLM}{\texttt{VLM}}			& Vortex Lattice Method						\\
	\hypertarget{INS}{\texttt{INS}}			& Inertial Navigation System					& &	\hypertarget{VT}{\texttt{VT}}			& Visual Trajectory							\\
	\hypertarget{INSA}{\texttt{INSA}}		& \hypertt{ICAO} Non Standard Atmosphere		& & \hypertarget{WGS84}{\texttt{WGS84}}		& World Geodetic System 1984				\\
	\hypertarget{ISA}{\texttt{ISA}}			& \hypertt{ICAO} Standard Atmosphere			& &	\hypertarget{WMM}{\texttt{WMM}}			& World Magnetic Model						\\
											&												& & 										&											\\
											&												& & 										&											\\
											&												& & 										& 											\\
											&												& & 										& 											\\
											&												& &  										&											\\
											&												& &  										&											\\
											& 												& &  										&											\\
											& 												& & 										& 											\\	
											& 												& & 										& 											\\
											&												& &  										&											\\
\end{tabular}
\end{table}
\renewcommand{\arraystretch}{1.0}

\thispagestyle{empty}

 		\thispagestyle{empty} \cleardoublepage

\pagestyle{plain}
\pagenumbering{roman}
\tableofcontents \cleardoublepage
\pagenumbering{arabic}
 
\chapter{Introduction}\label{cha:Intro}

An \emph{unmanned aerial vehicle} (\hypertt{UAV}) or drone is an aircraft without a human pilot onboard. It is just one component of an \emph{unmanned aircraft system} (\hypertt{UAS}), which does not only include the \hypertt{UAV} platform, but also a controller based on the ground, and a communications system between the two \cite{OACI2011}. In addition to their size, weight, onboard equipment, and performances, \hypertt{UAV}s can be classified according to their power source (electric, piston engine, turbine), the lift generation method (fixed wing, single rotor, multi rotor, combined), and the degree of autonomy. In this last category, \hypertt{UAS} are divided into \emph{remotely controlled} aircraft, in which the ground human operator continuously provides control commands to the aircraft, and \emph{autonomous} aircraft, which rely on onboard computers to execute the previously uploaded mission objectives.

In the case of remotely piloted platforms, the \hypertt{UAV} also employs the communications channel to provide the ground with its current status (position, velocity, attitude, etc.) and in some cases photos or video, on which the ground operator relies to continuously generate the control instructions. Autonomous aircraft can also employ the communications channel to provide information to the ground, which may decide to update the mission objectives based on it, but as its own name implies, they can operate without any kind of communication with the ground; they just continue executing their mission until the flight concludes or communications are restored.

The number, variety, and applications of \hypertt{UAS} have grown exponentially over the last few years, and this trend is expected to continue in the future \cite{Hassanalian2017, Shakhatreh2019}. This is particularly true in the case of low \hypertt{SWaP} (\emph{size, weight, and power}) vehicles because their reduced cost makes them suitable for a wide range of users and applications, both civil and military. A significant percentage of these vehicles are capable of operating autonomously. With small variations, these platforms rely on a suite of sensors that continuously provides noisy data about the airframe state, a navigation algorithm to estimate the aircraft pose (position plus attitude), and a control system that, based on the navigation output, adjusts the aircraft control mechanisms to successfully execute the preloaded mission.

This document focuses on fixed wing autonomous platforms, which are usually equipped with a Global Navigation Satellite System (\hypertt{GNSS}) receiver, accelerometers, gyroscopes, magnetometers, an air data system, a Pitot tube, air vanes, and one or multiple cameras\footnote{A single camera is employed in this document.}. The combination of accelerometers and gyroscopes is known as the \hypertt{IMU} or Inertial Measurement Unit. The errors introduced by all sensors grow significantly as their \hypertt{SWaP} decreases, in particular in the case of the \hypertt{IMU}. The recent introduction of solid state accelerometers and gyroscopes has dramatically improved the performance of low \hypertt{SWaP} \hypertt{IMU}s, with new models showing significant improvements when compared to those fabricated only a few years ago. The problem of noisy measurement readings is compounded in the case of low \hypertt{SWaP} vehicles by the low mass of the platforms, which results in less inertia and hence more high frequency accelerations and rotations caused by the atmospheric turbulence.

Aircraft navigation has traditionally relied on the measurements provided by accelerometers, gyroscopes, and magnetometers, incurring in an slow but unbounded position drift that could only be stopped by triangulation with the use of external navigation (radio) aids. More recently, the introduction of satellite navigation has completely removed the position drift and enabled autonomous \emph{inertial navigation} in low \hypertt{SWaP} platforms \cite{Farrell2008, Groves2008, Chatfield1997}. A comprehensive review of low \hypertt{SWaP} \hypertt{UAV} systems and the problems they face, including the degradation or absence of \hypertt{GNSS} signals, is provided by \cite{Bijjahalli2020}.

This chapter discusses the document objectives (section \ref{sec:Intro_objective}), lists the main blocks of the associated high fidelity flight simulator, which match the outline of the document (section \ref{sec:Intro_sim_outline}), and then introduces key concepts required to understand the simulation, such as the various types of trajectories, frequencies, and errors (section \ref{sec:Intro_traj}). 


\section{Objective}\label{sec:Intro_objective}

The main objective of this document is to describe the capabilities and underlying algorithms of the associated autonomous fixed wing aircraft high fidelity flight simulator. Equipped with standard guidance, navigation, and control (\hypertt{GNC}) systems, the simulation can be employed not only to test and validate novel \hypertt{GNC} algorithms, but also to evaluate their behavior when faced with sensors of different grades, severe and varying wind and atmospheric conditions, challenging missions, or even the absence of \hypertt{GNSS} signals. The source code of the simulator has been implemented in \nm{\CC} and is available as open source code in \cite{CODE}.

Although simulation can never constitute a full replacement for field testing novel algorithms, the simulator described in this document aims to replicate the influence of as many factors as possible on the performances of the \hypertt{GNC} system, limiting the assumptions or simplifications to the bare minimum, and including multiple relatively minor effects not usually considered in simulation that may in occasions play a role in the algorithms not performing as intended. 

Note that although the sensor suite described in chapter \ref{cha:Sensors} includes a simulated camera capable of generating realistic images of the Earth surface that resemble what a real onboard camera would record if it followed the resulting trajectory, the images are not employed by the \hypertt{GNC} system. By including the camera, the simulation is however ready to accept visual navigation algorithms that may complement or even replace the proposed inertial navigation system.

The simulator has been designed so most of its inputs are defined stochastically instead of in a deterministic way. This design facilitates the use of Monte Carlo simulations, in which the specific parameters defining the weather, the wind field, the turbulence, the initial conditions, the errors introduced by the onboard sensors, and the Earth gravity and magnetic fields, vary from one execution or run to the next in a pseudo-random way. In addition, the simulator is prepared so the guidance objectives that represent the aircraft mission can also be defined stochastically. The use of Monte Carlo simulations facilitates the evaluation of a given algorithm over a wide range of conditions, ensuring that virtually all parameters with any influence over the results are different for every run, enhancing their significance.


\section{Flight Simulation and Outline}\label{sec:Intro_sim_outline}

As the primary objective of the high fidelity simulator is to act as a test bench in which to implement, test, and evaluate novel \hypertt{GNC} algorithms, it must be capable of accurately modeling the flight of a low \hypertt{SWaP} fixed wing autonomous aircraft considering the influence on the resulting aircraft trajectory of many different factors, such as the atmospheric conditions, the wind, the air turbulence, the aircraft aerodynamic and propulsive performances, its onboard sensors, the guidance objectives that make up the mission, the control system that moves the throttle and the aerodynamic controls so the trajectory conforms to the guidance objectives, and the navigation system that processes the data obtained by the sensors and feeds the control system with the appropriate targets. 

This section provides an overall view of the different blocks that work together in the flight simulator, together with the interactions among them, so that the following chapters can be understood in the context of the broader system of which they are part and how they contribute to the overall simulation objectives, and not only as stand-alone entities. 

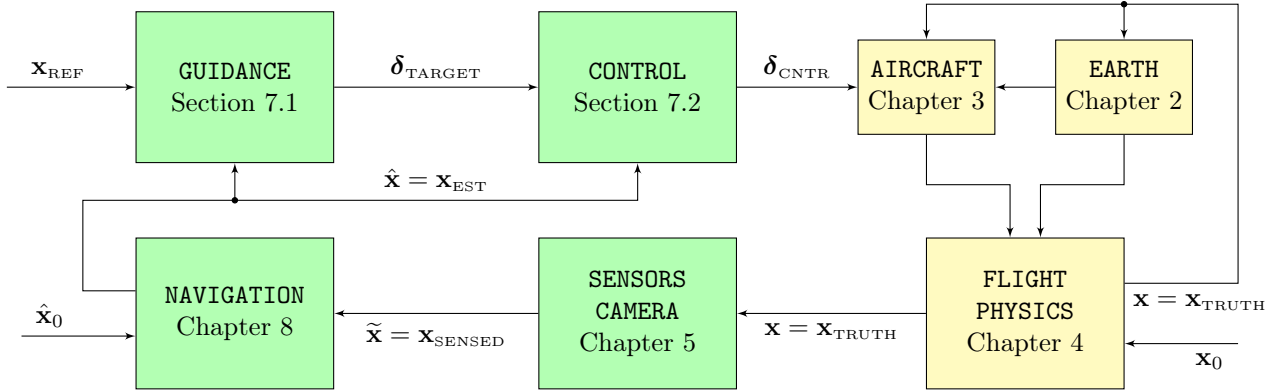
\begin{figure}[h]
\centering
\begin{tikzpicture}[auto, node distance=2cm,>=latex']

	\node [coordinate](xrefinput) {};
	\node [blockgreen, right of=xrefinput, minimum width=2.6cm, node distance=3.0cm, align=center, minimum height=2.0cm] (GUIDANCE) {\texttt{GUIDANCE} \\ Section \ref{sec:GNC_Guidance}};
	\draw [->] (xrefinput) -- node[pos=0.4] {\nm{\xREF}} (GUIDANCE.west);

	\node [blockgreen, right of=GUIDANCE, minimum width=2.6cm, node distance=5.3cm, align=center, minimum height=2.0cm] (CONTROL) {\texttt{CONTROL} \\ Section \ref{sec:GNC_Control}};
	\draw [->] (GUIDANCE.east) -- node[pos=0.5] {\nm{\deltaTARGET}} (CONTROL.west);

	\node [blockyellow, right of=CONTROL, minimum width=1.8cm, node distance=3.8cm, align=center, minimum height=1.25cm] (AIRCRAFT) {\texttt{AIRCRAFT} \\ Chapter \ref{cha:AircraftModel}};
	\draw [->] (CONTROL.east) --  node[pos=0.5] {\nm{\deltaCNTR}} (AIRCRAFT.west);

	\node [blockyellow, right of=AIRCRAFT, minimum width=1.8cm, node distance=2.6cm, align=center, minimum height=1.25cm] (EARTH) {\texttt{EARTH} \\ Chapter \ref{cha:EarthModel}};	
	\draw [->] (EARTH.west) -- (AIRCRAFT.east);

	\node [coordinate, right of=AIRCRAFT, node distance=1.3cm] (midpoint){};
	\node [blockyellow, below of=midpoint, minimum width=2.6cm, node distance=3.0cm, align=center, minimum height=2.0cm] (FLIGHT) {\texttt{FLIGHT} \\ \texttt{PHYSICS} \\ Chapter \ref{cha:FlightPhysics}};
	\draw [->] ($(EARTH.south)$) |- ($(FLIGHT.north)+(0.20cm,0.70cm)$) -- ($(FLIGHT.north)+(0.20cm,0cm)$);
	\draw [->] ($(AIRCRAFT.south)$) |- ($(FLIGHT.north)+(-0.20cm,0.70cm)$) -- ($(FLIGHT.north)-(0.20cm,0cm)$);

	\node [blockgreen, below of=GUIDANCE, minimum width=2.6cm, node distance=3.0cm, align=center, minimum height=2.0cm] (NAVIGATION) {\texttt{NAVIGATION} \\ Chapter \ref{cha:nav}};
	\node [coordinate, left of=NAVIGATION, node distance=2.0cm] (pointnav1){};
	\node [coordinate, above of=pointnav1, node distance=1.50cm] (pointnav2){};
	\node [coordinate, below of=GUIDANCE, node distance=1.50cm] (pointnav3){};
	\filldraw [black] (pointnav3) circle [radius=1pt];
	\draw [->] ($(NAVIGATION.west)+(0cm,0.3cm)$) -| (pointnav2) -- (pointnav3) -- (GUIDANCE.south);
	\draw [->] (pointnav3) -| node[pos=0.25] {\nm{\xvecest = \xEST}} (CONTROL.south);
	\draw [->] ($(NAVIGATION.west)+(-1.5cm,-0.3cm)$) -- node[pos=0.25] {\nm{\xvecestzero}} ($(NAVIGATION.west)+(0cm,-0.3cm)$);

	\node [blockgreen, below of=CONTROL, minimum width=2.6cm, node distance=3.0cm, align=center, minimum height=2.0cm] (SENSORS) {\texttt{SENSORS} \\ \texttt{CAMERA} \\ Chapter \ref{cha:Sensors}};
	\draw [->] (SENSORS.west) -- node[pos=0.5] {\nm{\xvectilde = \xSENSED}} (NAVIGATION.east);

	\node [coordinate, right of=EARTH, node distance=1.5cm] (pointflight1){};
	\node [coordinate, above of=EARTH, node distance=1.1cm] (pointflight2){};
	\filldraw [black] (pointflight2) circle [radius=1pt];
	\draw [->] (FLIGHT.west) -- node[pos=0.5] {\nm{\xvec = \xTRUTH}} (SENSORS.east);
	\draw [->] ($(FLIGHT.east)+(0cm,+0.4cm)$) -| (pointflight1) |- (pointflight2) -| (AIRCRAFT.north);
	\node [coordinate, right of=FLIGHT, node distance=2.3cm] (pointname){};	
	\node [above of=pointname, node distance=0.15cm] (nametruth) {\nm{\xvec = \xTRUTH}};	
	\draw [->] (pointflight2) -- (EARTH.north);

	\node [coordinate, right of=FLIGHT, node distance=2.8cm] (point1){};
	\node [coordinate, below of=point1, node distance=0.4cm] (point2){};
	\draw [->] (point2) -- node[pos=0.25] {\nm{\xveczero}} ($(FLIGHT.east)+(0cm,-0.4cm)$);
\end{tikzpicture}
\caption{Components of the high fidelity simulation}
\label{fig:Trajectory_flow_diagram}
\end{figure}

Figure \ref{fig:Trajectory_flow_diagram} represents the two distinct processes that take place during the flight of an autonomous air vehicle. The first one, represented by the yellow blocks on the right, focuses on the physics of flight and the interaction between the aircraft and its surroundings that results in the actual aircraft trajectory, while the second, represented by the green blocks on the left, contains the aircraft systems in charge of ensuring that the resulting trajectory is as close to the original objectives as possible. As shown in the figure, the two parts of the simulation are not independent. The aircraft flight depends on the variation with time of the control parameters \nm{\deltaCNTR} (position of the throttle and aerodynamic control surfaces) provided by the aircraft systems, while the systems behavior depends exclusively on the real aircraft state \nm{\xvec = \xTRUTH}.

The first part of the simulation (yellow blocks), which models the aircraft flight and its simulation leading to the actual or real trajectory, is the focus of chapters \ref{cha:EarthModel}, \ref{cha:AircraftModel}, and \ref{cha:FlightPhysics}. They discuss the aircraft platform and its interaction with the atmosphere while considering the time variation of the aircraft controls as an input. The objective is to develop high accuracy models of most physical phenomena involved in the aircraft flight, as well as to identify and justify those cases in which simplifications are introduced. In this way, the simulated aircraft response to variations in the control inputs (generated by algorithms discussed later in the document) closely resembles what would occur in the flight of a real aircraft, hence enabling the validation or rejection of the \hypertt{GNC} algorithms. Chapter \ref{cha:EarthModel} describes the environment in which the aircraft motion takes place and its influences on the aircraft trajectory, such as the gravitational field, the atmosphere, the wind, and the air turbulence. It is followed by chapter \ref{cha:AircraftModel}, which focuses on the aircraft aerodynamic and propulsive performances, and their dependencies on the atmospheric conditions as well as on the position of the throttle lever and aerodynamic control surfaces. The first part concludes with chapter \ref{cha:FlightPhysics}, which introduces the equations of motion that describe the physics of flight, and the actual or real aircraft trajectory obtained with their integration.

The second part of the simulation (green blocks) focuses on the systems installed onboard the autonomous aircraft, whose primary mission is to adjust the throttle lever and the aerodynamic control surfaces in such a way that the resulting trajectory complies with the previously loaded mission objectives. To do so, the aircraft is equipped with a suite of sensors that measure the aircraft state, a navigation system in charge of estimating the vehicle pose, and guidance and control systems that together compare that estimation with the mission objectives and adjust the throttle lever and control surfaces accordingly so the aircraft deviation from its objectives is minimal. Chapter \ref{cha:Sensors} presents the different sensors installed onboard the aircraft, analyzes their error sources, and introduces models describing their behavior. Chapter \ref{cha:PreFlight} discusses the calibration and alignment procedures that need to be executed before the flight to reduce the errors introduced by the different sensors. In the first of the two chapters allocated to the \hypertt{GNC} system, chapter \ref{cha:gc} describes both the guidance system, which is in charge of converting the mission objectives into control targets, and the control system, which compares these targets with the estimations provided by the navigation system to obtain the appropriate adjustments to the throttle and control surfaces. Finally, chapter \ref{cha:nav} describes the inertial navigation system, whose mission is to obtain the best possible estimation of the aircraft state based on the sensor measurements. 

The document also includes three appendices. Appendix \ref{cha:Assumptions} groups together the different assumptions employed in the simulator, appendix \ref{cha:PhysicalConstants} provides a list of the different constants referenced throughout the document, and appendix \ref{cha:RefSystems} groups together the various reference frames employed in all previous chapters, describes the variables required to define them, and explains their relationships with one another.


\section{The Concept of Aircraft Trajectory}\label{sec:Intro_traj}

As graphically depicted in figure \ref{fig:Trajectory_flow_diagram}, the various types of aircraft trajectories or aircraft states constitute the main interfaces among the different simulation blocks. The \emph{aircraft trajectory} is a time stamped series of a group of variables representing the \emph{aircraft state} at a given moment of time. If available, the aircraft state should be composed by variables describing the aircraft pose (position plus attitude), its linear and angular kinematics, as well as its dynamics. Depending on their accuracy and their completeness (the identity and number of the aircraft state variables), it is possible to define various different aircraft trajectory representations; they are listed in table \ref{tab:Into_trajectory_types} and described below:
\begin{center}
\begin{tabular}{lccclc}
	\hline
	Trajectory & Acronym	 & Other Names		& Math							  & Description						  & Section \\
	\hline
	Actual    & \hypertt{AT} & real, true       & \nm{\xvec = \xTRUTH}            & Real aircraft states              & \ref{sec:EquationsMotion_AT} \\
	Sensed    & \hypertt{ST} & -                & \nm{\xvectilde = \xSENSED}      & Measured by onboard sensors       & \ref{sec:Sensors_ST} \\ 
	Observed  & \hypertt{OT} & estimated        & \nm{\xvecest = \xEST}           & Inertial navigation output        & \ref{sec:GNC_OT} \\ 
	Visual    & \hypertt{VT} & image            & \nm{\xvecvis = \xIMG}           & Visual navigation output          & N/A \\
	Reference & \hypertt{RT} & guidance, script & \nm{\xREF}                      & Taken from mission objectives     & \ref{subsec:GNC_RT} \\
	\hline
\end{tabular}
\end{center}
\captionof{table}{Types of aircraft trajectories} \label{tab:Into_trajectory_types}

\begin{itemize}
\item The \emph{actual trajectory} (\hypertt{AT}), \emph{real trajectory}, or \emph{true trajectory} is composed by a time stamped series of \emph{actual aircraft states} represented by \nm{\xvec\lrp{t} = \xTRUTH\lrp{t}}, and is the real trajectory flown by the aircraft. It is the outcome of the interaction of an aircraft whose throttle and control surfaces positions are set with the environment that surrounds it. This process is depicted by the yellow blocks of figure \ref{fig:Trajectory_flow_diagram}, and starts with the dependency of the environmental variables (temperature, pressure, wind, turbulence, gravitation) provided by the Earth model (chapter \ref{cha:EarthModel}) on the aircraft true position. The aircraft aerodynamic and propulsive forces and moments (chapter \ref{cha:AircraftModel}) depend on these environmental variables, together with the positions of the throttle and control surfaces \nm{\deltaCNTR\lrp{t_c}}, where \nm{t_c = c \cdot \DeltatCNTR}, which are set by the control system (in the simulation the control system works at a rate of \nm{50 \ Hz}). The aircraft motion then depends exclusively on the environmental conditions and the aircraft performances, a process modeled in chapter \ref{cha:FlightPhysics} by the integration of the equations of motion, which requires the initial conditions \nm{\xveczero}.

The actual or real trajectory is defined in section \ref{sec:EquationsMotion_AT}. It is an easy to understand concept, but the actual aircraft state is unknown to all actors involved, as there do not exist sensors accurate enough to measure its components without errors. In the real world all its components vary in a continuous way, but in simulation it is computed at a discrete rate of \nm{500 \ Hz}, resulting in  \nm{\xvec\lrp{t_t} = \xTRUTH\lrp{t_t}}, where \nm{t_t = t \cdot \DeltatTRUTH}. The real trajectory is however only employed for evaluation and as the input to the aircraft sensors, as the aircraft \hypertt{GNC} system operates based on other less accurate representations of the trajectory.

\item The \emph{sensed trajectory} (\hypertt{ST}), represented by \nm{\xvectilde\lrp{t_s} = \xSENSED\lrp{t_s}}, where \nm{t_s = s \cdot \DeltatSENSED}, is a discrete representation of the aircraft trajectory that puts together the outputs of the different onboard sensors, known as \emph{sensed aircraft states}, at a series of discrete times. The sensed trajectory is defined in section \ref{sec:Sensors_ST}, and in the simulation it is obtained at a rate of \nm{100 \ Hz}; some of its components are however generated at a lower rate, such as the images at \nm{10 \ Hz}, and the \hypertt{GNSS} receiver measurements at \nm{1 \ Hz}. As such, the sensed states are not only incomplete, in the sense that they do not contain all the variables that make up the actual state, but they are also partially incorrect, as they include the errors introduced by the sensors themselves. However, the sensed trajectory is the only view of the actual trajectory available to the \hypertt{GNC} system. The errors introduced by the onboard sensors when obtaining the sensed trajectory are described in chapter \ref{cha:Sensors}. The sensed state is then taken by the navigation system, which processes it with the objective of reducing the measurement errors.

\item The \emph{observed trajectory} (\hypertt{OT}) or \emph{estimated trajectory}, represented by \nm{\xvecest\lrp{t_n} = \xEST\lrp{t_n}}, where \nm{t_n = n \cdot \DeltatEST}, is the output of the inertial navigation system, and contains the \emph{observed aircraft states}, this is, the most accurate knowledge that the aircraft systems posses about the aircraft actual states. The observed trajectory is defined in section \ref{sec:GNC_OT}, and in the simulation it is computed at a rate of \nm{100 \ Hz}. It can be considered as an improved version of the sensed trajectory, as it is complete (it not only contains the sensed variables but also other components of the actual trajectory that are more relevant to the guidance and control systems) and also less incorrect (it is a better representation of the actual trajectory). Its obtainment by the inertial navigation system is described in chapter \ref{cha:nav}.

\item The \emph{visual trajectory} (\hypertt{VT}) or \emph{image trajectory}, represented by \nm{\xvecvis\lrp{t_i} = \xIMG\lrp{t_i}}, where \nm{t_i = i \cdot \DeltatIMG}, is the output of the visual navigation system. Considered as an alternative or a complement to the observed trajectory, it is composed by a series of \emph{visual aircraft states}, and represents an estimation of the actual trajectory obtained by visual algorithms at the same rate as the images are generated, this is, \nm{10 \ Hz}. Note that as previously mentioned, although the simulator is capable of generating realistic images of the Earth surface so they can potentially be employed by a visual navigation system, such a system has not been implemented and as a consequence the visual trajectory does not appear in figure \ref{fig:Trajectory_flow_diagram}.

\item The \emph{reference trajectory} (\hypertt{RT}), \emph{guidance trajectory}, or \emph{trajectory script}, represented by \nm{\xREF}, is defined in section \ref{subsec:GNC_RT}, and contains the target trajectory that should be achieved according to the mission objectives. As such, it is quite incomplete because for each flight segment it only contains those components of the trajectory to which the aircraft shall adhere. Note that the main objective of the \hypertt{GNC} system is for the real or actual trajectory flown by the aircraft to be as close as possible to the reference trajectory, but in fact what it does is to diminish the differences between the observed and reference trajectories.
\end{itemize}
\begin{center}
\begin{tabular}{lrrcl}
	\hline
	\multicolumn{1}{c}{Discrete Time}	& Frequency		& Period		 & Variables						& Systems \\
	\hline
	\nm{t_t = t \cdot \DeltatTRUTH}		& \nm{500 \ Hz} & \nm{0.002 \ s} & \nm{\xvec = \xTRUTH}             & Flight physics \\
	\nm{t_s = s \cdot \DeltatSENSED}    & \nm{100 \ Hz} & \nm{0.01 \ s}  & \nm{\xvectilde = \xSENSED}       & Sensors \\ 
	\nm{t_n = n \cdot \DeltatEST}       & \nm{100 \ Hz} & \nm{0.01 \ s}  & \nm{\xvecest = \xEST}            & Inertial navigation \\ 
	\nm{t_c = c \cdot \DeltatCNTR}      & \nm{ 50 \ Hz} & \nm{0.02 \ s}  & \nm{\deltaTARGET, \, \deltaCNTR} & Guidance \& control \\
	\nm{t_i = i \cdot \DeltatIMG}       & \nm{ 10 \ Hz} & \nm{0.1 \ s}   & \nm{\xvecvis = \xIMG}            & Visual navigation \& camera (sensor) \\
	\nm{t_g = g \cdot \DeltatGNSS}      & \nm{  1 \ Hz} & \nm{1 \ s}     &                                  & \hypertt{GNSS} receiver (sensor) \\	
	\hline
\end{tabular}
\end{center}
\captionof{table}{Working frequencies of the different systems and trajectory representations} \label{tab:Intro_trajectory_frequencies}

As graphically depicted in figure \ref{fig:Trajectory_flow_diagram}, the simulator also requires the real initial conditions \nm{\xveczero} to start the integration that leads to the actual trajectory, and the navigation system needs its own estimated initial conditions \nm{\xvecestzero}, which as explained in chapter \ref{cha:nav} are mostly taken directly from the sensors themselves.

Table \ref{tab:Intro_trajectory_frequencies} shows the various operation frequencies that coexist in the simulation, together with the variables that are sampled or evaluated at those frequencies, and also the systems that operate at such frequencies.
\begin{center}
\begin{tabular}{lccl}
	\hline
	\multicolumn{1}{c}{Error} & Acronym & Compares & \multicolumn{1}{c}{Notes} \\
	\hline
	Flight Technical Error  & \hypertt{FTE} & \hypertt{OT} \& \hypertt{RT} & Evaluates guidance \& control systems performance \\
	Navigation System Error & \hypertt{NSE} & \hypertt{AT} \& \hypertt{OT} & Evaluates navigation system performance \\ 
	Total System Error      & \hypertt{TSE} & \hypertt{AT} \& \hypertt{RT} & Combination of \hypertt{FTE} plus \hypertt{NSE} \\ 
	\hline
\end{tabular}
\end{center}
\captionof{table}{Errors between different trajectory types} \label{tab:Summary_trajectory_errors}

\begin{figure}[h]
\centering
\begin{tikzpicture}[auto, node distance=2cm,>=latex']
	\node [block, minimum width=4.0cm, align=center, minimum height=2.0cm] (ACTUAL TRAJECTORY) {\texttt{ACTUAL TRAJECTORY} \\ \nm{\xvec\lrp{t} = \xTRUTH\lrp{t}}}; 
	\node [coordinate, below of=ACTUAL TRAJECTORY, node distance=4.0cm] (middlepoint){};
	\node [block, right of=middlepoint, minimum width=4.0cm, node distance=4.5cm, align=center, minimum height=2.0cm] (REFERENCE TRAJECTORY) {\texttt{REFERENCE TRAJECTORY} \\ \nm{\xREF\lrp{t}}}; 
	\node [block, left of=middlepoint, minimum width=4.0cm, node distance=4.5cm, align=center, minimum height=2.0cm] (OBSERVED TRAJECTORY) {\texttt{OBSERVED TRAJECTORY} \\ \nm{\xvecest\lrp{t} = \xEST\lrp{t}}};
		
	\draw [<->] (ACTUAL TRAJECTORY.east) -- node[pos=0.5] {Total System Error} (REFERENCE TRAJECTORY.north);	
	\draw [<->] (OBSERVED TRAJECTORY.east) -- node[pos=0.5] {Flight Technical Error} (REFERENCE TRAJECTORY.west);
	\draw [<->] (OBSERVED TRAJECTORY.north) -- node[pos=0.5] {Navigation System Error} (ACTUAL TRAJECTORY.west);
\end{tikzpicture}
\caption{Trajectory errors}
\label{fig:Trajectory_errors}
\end{figure}
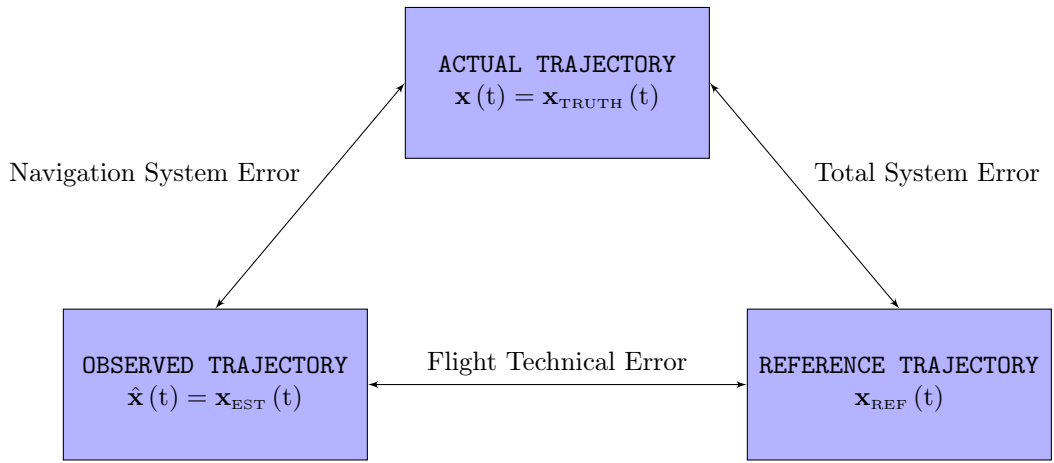

Table \ref{tab:Summary_trajectory_errors} and figure \ref{fig:Trajectory_errors} show the differences between the different types of trajectories \cite{Corraro2009}, which enable the evaluation of the performances of the different aircraft systems. The final objective of every \hypertt{GNC} system is to reduce the \emph{total system error} (\hypertt{TSE}) or difference between the actual and reference trajectories. This objective can however be divided into two complimentary objectives that are quite independent from each other. The first is to reduce the \emph{navigation system error} (\hypertt{NSE}) or difference between the actual and observed trajectories, and depends on the performance of the onboard sensors and navigation system, while the second is the reduction of the \emph{flight technical error} (\hypertt{FTE}) or difference between the observed and reference trajectories, which is the responsibility of the guidance and control systems. 

 \cleardoublepage
\chapter{Earth Model}\label{cha:EarthModel}
	
The Earth model encompasses a series of concepts, hypotheses, and models that together conform the underlying frame in which the aircraft motion and its interaction with the environment takes place. The accuracy of the trajectory computation process relies heavily on its proper definition, as both the aircraft performances and its motion depend on the atmospheric properties and a proper modeling of the Earth physics.

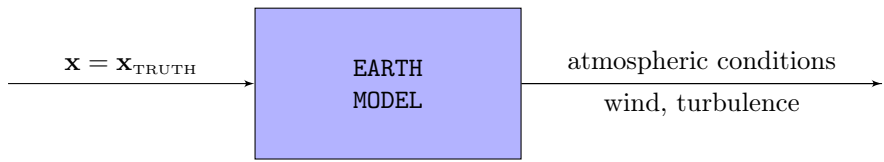
\begin{figure}[h]
\centering
\begin{tikzpicture}[auto, node distance=2cm,>=latex']
	\node [coordinate](xtruthinput) {};
	\node [block, right of=xtruthinput, minimum width=3.5cm, node distance=5.0cm, align=center, minimum height=2.0cm] (EARTH MODEL) {\texttt{EARTH} \\ \texttt{MODEL}};	
	\node [coordinate, right of=EARTH MODEL, node distance=6.5cm] (output){};
	
	\draw [->] (xtruthinput) -- node[pos=0.5] {\nm{\xvec = \xTRUTH}} (EARTH MODEL.west);
	\draw [->] (EARTH MODEL.east) -- node[pos=0.5] {atmospheric conditions} (output);
	\draw [-] (output) -- node[pos=0.5] {wind, turbulence} (EARTH MODEL.east);
\end{tikzpicture}
\caption{Earth model flow diagram}
\label{fig:EarthModel_flow_diagram_generic}
\end{figure}

The behavior of the Earth model within the simulation is graphically depicted in figure \ref{fig:EarthModel_flow_diagram_generic}, in which all outputs (atmospheric conditions, wind, air turbulence, etc.) are based on the true aircraft state, and in particular on the time and geodetic position.

The chapter starts with the introduction of the reference ellipsoid to model the Earth surface (section \ref{sec:EarthModel_WGS84}), which is the basis for the \hypertt{ECEF} reference frame \nm{\FE} and the Cartesian, geocentric, and geodetic coordinates employed to position the aircraft with respect to the Earth. It then defines an ellipsoidal gravity model representing the combination of the gravitational and centrifugal potentials, which establishes the vertical equilibrium of the atmosphere and conforms the basis for the definition of the mean sea level and the geopotential altitude (section \ref{sec:EarthModel_GEOP}). The weather and atmospheric models are introduced in section \ref{sec:EarthModel_ISA}, while section \ref{sec:EarthModel_WIND} describes both the low frequency wind component that is an important factor in the resulting aircraft trajectory, as well as the turbulence or high frequency wind component that heavily influences the aircraft dynamics. The Earth magnetic field, indispensable for navigation, is described in section \ref{sec:EarthModel_WMM}. Section \ref{sec:EarthModel_accuracy} analyzes the differences between the gravity and magnetic fields provided by the previously described models, which are employed by the aircraft systems, and those that exist in reality, which are measured by the aircraft sensors; the design of the navigation system shall consider the inexactitude of the models to lower their influence on the resulting aircraft trajectory. The chapter concludes with a discussion on the Earth model implementation and validation (section \ref{sec:EarthModel_implementation}).


\section{World Geodetic System 1984 and Associated Ellipsoid}\label{sec:EarthModel_WGS84}

Geodesy is the scientific discipline that deals with the measurement and representation of the Earth, including its gravitational field, in a three dimensional time varying space. A good geodetic reference system of the Earth is necessary for aircraft geopositioning and navigation. The World Geodetic System 1984 (\texttt{WGS84}) \cite{WGS84} is the de facto standard for these purposes \cite{Groves2008,Chatfield1997,Rogers2007,Kayton1997,Stevens2003}; it defines the appropriate reference system, introduces the different sets of coordinates, and provides the fundamental constants.

The \hypertt{WGS84} system is based on the following assumptions:
\begin{itemize}
\item The surface of the Earth is a revolution ellipsoid [A\ref{as:WGS84_ellipsoid}].
\item The Earth ellipsoid is centered at the Earth center of mass [A\ref{as:WGS84_center_mass}].
\item The Earth rotates at constant speed around the ellipsoid symmetry axis [A\ref{as:WGS84_rotation}].
\end{itemize}

The \hypertt{WGS84} \emph{ellipsoid} is defined by its semimajor \emph{a} and its flattening \emph{f} \cite{WGS84} (appendix \ref{cha:PhysicalConstants}). Other ellipsoid parameters, such as the semiminor \emph{b}, the \nm{\first} eccentricity \emph{e}, and the \nm{\second} eccentricity \emph{g}, can be directly obtained:
\begin{eqnarray}
	\nm{b} & = & \nm{a \lrp{1 -f}} \label{eq:EarthModel_WGS84_ELL_b} \\
	\nm{e} & = & \nm{\sqrt{\frac{a^2 - b^2}{a^2}} = \sqrt{f \lrp{2 - f}}} \label{eq:EarthModel_WGS84_ELL_e}  \\
    \nm{g} & = & \nm{\sqrt{\frac{a^2 - b^2}{b^2}}} \label{eq:EarthModel_WGS84_ELL_g} 
\end{eqnarray}
	
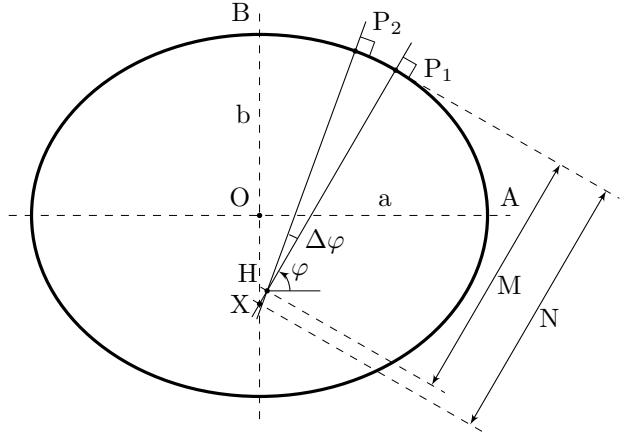
\begin{figure}[h]
\centering
\begin{tikzpicture}[auto,>=latex']
	\coordinate (origin)     at (+0.0,+0.0);
	\coordinate (pointH)     at (+0.1,-1.0);

	\fill (origin) circle [radius=1pt] node [above left] {\nm{O}};
	\fill (pointH) circle [radius=1pt] node [above left] {\nm{H}};
	\draw [name path = ellipse, very thick] (origin) ellipse [x radius=3, y radius=2.4];
	\draw [name path = OB, dashed, ultra thin] ($(origin) - (0,2.7)$) -- ($(origin) + (0,2.7)$) node [pos=0.75, left] {\nm{b}} node [left] {\nm{B}};
	\draw [dashed, ultra thin] ($(origin) - (3.3,0)$) -- ($(origin) + (3.3,0)$) node [pos=0.75, above] {\nm{a}} node [above] {\nm{A}};
	
	\draw [name path = HP1]  ($(pointH) + (-120:0.4)$) -- ($(pointH) + (60:3.8)$);
	\fill [name intersections={of=ellipse and HP1, by={pointP1}}] (pointP1) circle [radius=1pt] node [right=7pt] {\nm{P_1}};
	\fill [name intersections={of=OB and HP1, by={pointX}}] (pointX) circle [radius=1pt] node [left] {\nm{X}};
	\draw [name path = HP2] ($(pointH) + (-110:0.4)$) -- ($(pointH) + (70:3.8)$);
	\fill [name intersections={of=ellipse and HP2, by={pointP2}}] (pointP2) circle [radius=1pt]  node [above right=3pt] {\nm{P_2}};	
	
	\draw [dashed, ultra thin] ($(pointH) + (150:0.1)$) -- ($(pointH) + (-30:2.7)$);
	\draw [<->] ($(pointH) + (-30:2.5)$) -- ($(pointP1) + (-30:2.5)$) node [pos=0.5,below right=-3pt] {\nm{M}};
	\draw [dashed, ultra thin] ($(pointX) + (150:0.1)$) -- ($(pointX) + (-30:3.4)$);
	\draw [dashed, ultra thin] (pointP1) -- ($(pointP1) + (-30:3.4)$);	
	\draw [<->] ($(pointX) + (-30:3.2)$) -- ($(pointP1) + (-30:3.2)$) node [pos=0.5,below right=-3pt] {\nm{N}};	
	\draw [ultra thin] ($(pointP1) + (60:0.2)$) -- ++(-30:0.2) -- ++(-120:0.2) -- ++(+150:0.2);
	\draw [ultra thin] ($(pointP2) + (70:0.2)$) -- ++(-20:0.2) -- ++(-110:0.2) -- ++(+160:0.2);

	\draw [ultra thin] (pointH) -- ($(pointH) + (0:0.7)$);
	\draw [->] ($(pointH) + (0:0.3)$) arc [radius=0.3, start angle=0, end angle=60];
	\path node at ($(pointH) + (30:0.5)$) {\nm{\varphi}};
	\draw [-] ($(pointH) + (60:0.8)$) arc [radius=0.7, start angle=60, end angle=70];
	\path node at ($(pointH) + (55:0.8) + (0.3,0)$) {\nm{\Delta\varphi}};
	
\end{tikzpicture}
\caption{Ellipse resulting from ellipsoid cut along a meridian}
\label{fig:EarthModel_ellipsoid}
\end{figure}

The ellipsoid radii of curvature are of high interest for navigation \cite{Groves2008,Chatfield1997,Rogers2007,Kayton1997,Stevens2003}. Given any point \nm{P_1} at the ellipsoid surface, the prime vertical rotation center X is the intersection between the rotation axis and the line that passes through \nm{P_1} and is normal to the ellipsoid surface (figure \ref{fig:EarthModel_ellipsoid}). The \emph{radius of curvature of prime vertical} N is the radius of curvature of the ellipsoid section produced by a plane normal to the meridian and that passes through \nm{P_1}X\footnote{Equation (\ref{eq:EarthModel_WGS84_ELL_N}) introduces the latitude \nm{\varphi}, defined in section \ref{subsubsec:RefSystems_E_GeodeticCoord} as the angle between the Equator plane and the line \nm{P_1}X.}.
\neweq{N = \frac{a}{\sqrt{1-e^2\sin^2{\varphi}}}}{eq:EarthModel_WGS84_ELL_N}

The meridian rotation center H is the intersection between the normals to the ellipsoid surface of two differentially spaced points \nm{P_1} and \nm{P_2} placed along the same meridian. The \emph{radius of curvature of the meridian} M is the distance from H to the point \nm{P_1}.
\neweq{M = \frac{N}{1+g^2\cos^2{\varphi}}}{eq:EarthModel_WGS84_ELL_M}

Based on the ellipsoid representing the surface of the Earth, the \hypertt{WGS84} also introduces the \hypertt{ECEF} reference frame and associated coordinates (Cartesian, geocentric, geodetic) defined in section \ref{subsec:RefSystems_E}, required to represent the position of any point with respect to the Earth.


\section{Comprehensive Gravity Model}\label{sec:EarthModel_GEOP}

The objective of this section is to obtain models for the gravitation and centrifugal accelerations that act on all aircraft, and to define the concepts of mean sea level and geopotential altitude H that rely on them. Mean sea level acts as a reference to measure altitudes, while the geopotential altitude H is key to linking the geometric or geodetic altitude \emph{h} (section \ref{subsubsec:RefSystems_E_GeodeticCoord}) employed to position an aircraft with respect to the Earth, with the pressure altitude \nm{\Hp} (section \ref{sec:EarthModel_ISA}) on which the atmospheric properties (and hence the aircraft performances) are based. 


\subsection{Earth Gravitational Model 1996}\label{subsec:EarthModel_EGM96}

Gravitation has a significant influence in the motion of all aircraft and as such must be modeled with sufficient accuracy. All bodies in presence of a gravitational field \vec g are subject to a potential acceleration that derives from the gravitational potential function \nm{U_g}, such that:
\neweq{\vec{g}=\vec \nabla U_g\,}{eq:EarthModel_EGM96_g}

The Earth Gravitational Model 1996 (\hypertt{EGM96}) \cite{EGM96}, associated to \hypertt{WGS84} \cite{WGS84}, contains a very accurate model of the Earth gravitational field. It models the gravitational potential function \nm{U_g} as a development in series of Legendre polynomials up to degree and order 360. It expands (\ref{eq:EarthModel_EGM96_g}) based on geocentric coordinates (section \ref{subsubsec:RefSystems_E_GeocentricCoord}) and the spherical frame \nm{\FS}:
\neweq{\gS(\theta,\,\lambda,\,r) = \Big[\frac 1r \ \pderpar{U_g}{\theta} \ \ \ \ \frac 1{r\ \sin\theta} \ \pderpar{U_g}{\lambda} \ \ \ \ \pderpar{U_g}{r} \Big]^T}{eq:EarthModel_EGM96_g2}

Aircraft navigation and trajectory prediction do not require such accuracy, so the \hypertt{EGM96} model can be simplified into an ellipsoidal gravitational model \cite{Rogers2007,Stevens2003}. Assuming that the gravitational field varies only with the geocentric distance and colatitude [A\ref{as:EGM96_gravity}], only the first two terms of the \hypertt{EGM96} Legendre polynomials (spherical and ellipsoidal variations) are required, resulting in:
\neweq{U_g(\theta,r) = \frac{GM}{r} \lrsb{1 + \lrp{\frac{a}{r}}^2 \, \bar{C}_{20} \, \frac{\sqrt{5}}2 \, \lrp{3 \, \cos^2\theta-1}}}{eq:EarthModel_EGM96_Ug}

where GM is the Earth gravitational constant (atmosphere mass included) \cite{WGS84}, and \nm{\bar{C}_{20}} is the \hypertt{EGM96} second degree zonal harmonical \cite{EGM96} (appendix \ref{cha:PhysicalConstants}). The application of (\ref{eq:EarthModel_EGM96_g2}) results in an \emph{ellipsoidal gravitational field}:
\neweq{\gS(\theta,r) = - \frac{GM}{r^2} \lrsb{ \lrp{\frac{a}{r}}^2 \, \bar{C}_{20} \, 3 \sqrt{5} \, \sin\theta \cos\theta \ \ \ \ 0 \ \ \ \ 1 + 3 \lrp{\frac{a}{r} }^2 \, \bar{C}_{20} \, \frac{\sqrt{5}}{2} \, \lrp{3 \, \cos^2\theta - 1} }^T}{eq:EarthModel_EGM96_g3}


\subsection{Centrifugal and Gravity Accelerations}\label{subsec:EarthModel_GEOP_acc}

The \emph{centrifugal acceleration} \nm{\ac} caused by the Earth rotation around its poles axis does not only influence the aircraft trajectory, but also the atmosphere vertical equilibrium, and as such plays a key role in the definition of the geopotential altitude and in establishing the mean sea level \cite{Rogers2007,Stevens2003}. According to \cite{WGS84}, the angular velocity of the Earth \nm{\omega_{\sss E}} around the ellipsoid symmetry axis can be considered constant (appendix \ref{cha:PhysicalConstants}).
\neweq{\vecomegaEE = \omega_{\sss E} \, \iEiii}{eq:EarthModel_GEOP_omegaE}

Any object whose position with respect to the Earth center of mass \nm{\OECEF} is given by its Cartesian coordinates \nm{\TENE = \TEcar} experiences a centrifugal acceleration \nm{\vec{a_c}} as a consequence of the Earth rotation that is parallel to the Equatorial plane and derives from the centrifugal potential \nm{U_c}.
\begin{eqnarray}
\nm{\ac} & = & \nm{\vec \nabla U_c} \label{eq:EarthModel_GEOP_ac_Uc} \\
\nm{\acE} & = & \nm{- \ \omegaEEskew \ \omegaEEskew \ \TENE} \label{eq:EarthModel_GEOP_ac} 
\end{eqnarray}

Using geocentric coordinates and the spherical frame \nm{\FS} results in:
\neweq{\vec a_{c}^{\sss S}(\theta,\,\lambda,\,r) = \Big[\frac 1r \, \pderpar{U_c}{\theta} \ \ \ \ \frac 1{r \, \sin\theta} \ \pderpar{U_c}{\lambda} \ \ \ \ \pderpar{U_c}{r} \Big]^T} {eq:EarthModel_GEOP_ac3} 

The final expressions for the centrifugal potential and its acceleration are the following:
\begin{eqnarray}
\nm{U_c(\theta,r)} & = & \nm{\frac{\vecomegaE^T \ \vecomegaE}{2} = \frac{(\omegaE \ r \ \sin\theta)^2}{2}} \label{eq:EarthModel_GEOP_Uc} \\
\nm{\vec a_{c}^{\sss S}(\theta,r)} & = & \nm{\Big[\omegaE^2 \, r \, \sin\theta \, \cos\theta \ \ \ \ 0 \ \ \ \ \omegaE^2 \, r \, \sin^2\theta \Big]^T} \label{eq:EarthModel_EGM96_ac2}
\end{eqnarray}

The \emph{gravity acceleration} \nm{\gc} is defined as the sum of the gravitational and centrifugal accelerations. It is required to define the concept of mean sea level and to establish the relationship between geodetic and geopotential altitudes, which influences the atmospheric properties described in section \ref{sec:EarthModel_ISA}. The gravity acceleration can be defined as deriving from a potential function U, such that:
\begin{eqnarray}
	\nm{\vec{f}}& = & \nm{\vec \nabla U} \label{eq:EarthModel_GEOP_f} \\
	\nm{U} 		& = & \nm{U_g + U_c} \label{eq:EarthModel_GEOP_U} \\
	\nm{\gc}	& = & \nm{\vec g + \ac} \label{eq:EarthModel_GEOP_f2}
\end{eqnarray}
  		

\subsection{Mean Sea Level and WGS84 Ellipsoidal Gravity Formula}\label{subsec:EarthModel_GEOP_msl_and_ellip_gravity}

\emph{Mean Sea Level} (\hypertt{MSL}) is defined as the equigeopotential surface that better represents the average sea level\footnote{The surface of the ocean is not a fixed level. Even when the effects of tides and waves are removed, the oceanic streams and the saline gradients are responsible of oscillations of about 1\,m around the mean sea level.}. Because of the way the \hypertt{WGS84} and \hypertt{EGM96} models are constructed, the Earth surface (defined by the \hypertt{WGS84} ellipsoid), composed by those points whose geodetic altitude \emph{h} is zero, is a geopotential surface. The \hypertt{MSL} surface is hence characterized as overlapping the \hypertt{WGS84} ellipsoid \cite{Stevens2003,Eshelby2000}.
\neweq{\hMSL = 0}{eq:EarthModel_GEOP_msl}

However, the gravity acceleration \nm{\gc} computed by (\ref{eq:EarthModel_GEOP_f2}) based on the simplified gravitational model provided by (\ref{eq:EarthModel_EGM96_g3}) possesses a non-negligible component along \nm{\iNi} and hence is not orthogonal to the geopotential surface represented by the \hypertt{WGS84} ellipsoid. It would be necessary to include many more terms of the \hypertt{EGM96} polynomials for a gravity vector \nm{\gc} computed in such a way to be vertical. Instead of computing gravity as the sum of gravitation plus centrifugal accelerations, the \hypertt{WGS84} directly provides expression (\ref{eq:EarthModel_Gravity})\footnote{This expression is known as the Somigliana ellipsoidal gravity formula.}, which depends on geodetic altitude \nm{\varphi} and altitude \emph{h} (section \ref{subsubsec:RefSystems_E_GeodeticCoord}), and is based on constants already introduced in sections \ref{sec:EarthModel_WGS84} and \ref{subsec:EarthModel_EGM96}, together with theoretical \hypertt{MSL} gravity values at the Equator and the poles (\nm{\gcMSLE, \ \gcMSLP}) \cite{WGS84} (appendix \ref{cha:PhysicalConstants}):
\begin{eqnarray}
\nm{\gcNMODEL}    & = & \nm{\lrsb{0 \ \ \ \ 0 \ \ \ \ \gcNMODELiii}^T} \label{eq:EarthModel_Gravity} \\
\nm{\gcNMODELiii} & = & \nm{\gcMSL \ \lrsb{1 - \frac{2}{a} \, \lrp{1 + f + m - 2 \ f \ \sin^2{\varphi}} \, h + \frac{3}{a^2} \ h^2}}\label{eq:EarthModel_Gravity1}\\
\nm{\gcMSL}       & = & \nm{\gcMSLE \ \frac{1 + k \ \sin^2{\varphi}}{\sqrt{1-e^2\sin^2{\varphi}}}}\label{eq:EarthModel_Gravity2} \\
\nm{k}            & = & \nm{\frac{b \ \gcMSLP}{a \ \gcMSLE} - 1}\label{eq:EarthModel_Gravity3} \\ 
\nm{m}            & = & \nm{\frac{\omegaE^2 \ a^2 \ b}{GM}}\label{EarthModel_Gravity4}
\end{eqnarray}

The simulator does not employ (\ref{eq:EarthModel_EGM96_g3}) and directly computes gravity based on (\ref{eq:EarthModel_Gravity}). If gravitation is required, it obtains it by subtracting the centrifugal acceleration provided by (\ref{eq:EarthModel_EGM96_ac2}). Expression (\ref{eq:EarthModel_Gravity}) employs the subindex \nm{\sss{MOD}} (model) to distinguish it from (\ref{eq:EarthModel_accuracy_gravity}); refer to section \ref{sec:EarthModel_accuracy} for an explanation.


\subsection{Geopotential Altitude}\label{subsec:EarthModel_GEOP_geopotential_altitude}

\emph{Geopotential altitude} H is defined so that the same differential work is performed by the standard acceleration of free fall \nm{\gzero}\footnote{The standard acceleration of free fall \nm{g_0} is obtained using the equation of Lambert for the acceleration of free fall \cite{ISA}
\neweqnn{g = 9.80616 \lrp{1 - 0.0026373\cos(2\varphi) + 0.0000059 \cos^2 (2\varphi)} \ m/s^2}
at latitude \nm{\varphi=45^o\,32'\, 33''}.} \cite{ISA} (appendix \ref{cha:PhysicalConstants}) when displacing the unit of mass a distance dH, as that performed by the gravity acceleration \nm{g_c} when displacing the unit of mass a geodetic distance dh \cite{Stevens2003,Eshelby2000}. The relationship between geopotential and geodetic altitudes is hence the following:
\neweq{- g_c \ dh = - g_0 \ dH}{eq:EarthModel_GEOP_h_H}

\hypertt{MSL} can also be defined as the surface of zero geopotential altitude:
\neweq{\HMSL = 0}{eq:EarthModel_GEOP_msl2}

The relationship between the geodetic and geopotential altitudes is obtained by integrating (\ref{eq:EarthModel_GEOP_h_H}) between \hypertt{MSL} (\nm{\hMSL = \HMSL = 0}) and a generic point. 
\neweq{\int_{\hMSL=0}^{h} g_c \ dh = g_0 \int_{\HMSL=0}^{H} dH}{eq:EarthModel_GEOP_h_H2}

No explicit expression can be obtained, but the solution can be easily tabulated:
\begin{eqnarray}
	\nm{H} & = & \nm{f_H\lrp{h,\varphi}} \label{eq:EarthModel_GEOP_h2H_table} \\
	\nm{h} & = & \nm{f_h\lrp{H,\varphi}} \label{eq:EarthModel_GEOP_H2h_table} 
\end{eqnarray}         		
\begin{center}
\begin{tabular}{rrrrrrrrr}
	\hline
	\nm{h \lrsb{m}} & \nm{0^{\circ}} & \nm{15^{\circ}} & \nm{30^{\circ}} & \nm{45^{\circ}} & \nm{60^{\circ}} & \nm{75^{\circ}} & \nm{90^{\circ}} & (\ref{eq:EarthModel_GEOP_h2H}) \\
	\hline
       0  &     0.00  &     0.00  &     0.00  &     0.00  &     0.00  &     0.00  &     0.00  &     0.00 \\
    1000  &   997.15  &   997.51  &   998.48  &   999.80  &  1001.12  &  1002.08  &  1002.43  &   999.84 \\
    2000  &  1993.99  &  1994.70  &  1996.64  &  1999.29  &  2001.93  &  2003.85  &  2004.55  &  1999.37 \\
    3000  &  2990.52  &  2991.58  &  2994.49  &  2998.46  &  3002.42  &  3005.31  &  3006.36  &  2998.58 \\
    4000  &  3986.73  &  3988.15  &  3992.03  &  3997.32  &  4002.59  &  4006.45  &  4007.85  &  3997.48 \\
    5000  &  4982.62  &  4984.40  &  4989.25  &  4995.86  &  5002.46  &  5007.27  &  5009.03  &  4996.07 \\
	\hline
\end{tabular}
\end{center}
\captionof{table}{Geopotential altitude versus geodetic altitude for various latitudes}\label{tab:H_vs_h}

As shown in table \ref{tab:H_vs_h}, the influence of latitude is very small, accounting for approximately \nm{5 \ m} of difference in geopotential altitude between the Equator and the poles for every \nm{1000 \ m} of geometric altitude. This fact, together with the penalties involved by employing a tabulated solution instead of an explicit one, explains why it is common (\cite{ISA} for example) to employ simplified expressions obtained by integrating (\ref{eq:EarthModel_GEOP_h_H}) based on an spherical Earth surface of radius \nm{\RE} \ \cite{SMITHSONIAN} (appendix \ref{cha:PhysicalConstants}), an spherical gravitation obtained by removing the \second\ term from (\ref{eq:EarthModel_EGM96_Ug}), and an average centrifugal effect. 
\begin{eqnarray}
	\nm{H} & = & \nm{\dfrac{\RE \cdot h}{\RE + h}} \label{eq:EarthModel_GEOP_h2H} \\
	\nm{h} & = & \nm{\dfrac{\RE \cdot H}{\RE - H}} \label{eq:EarthModel_GEOP_H2h} 
\end{eqnarray}         		

This simplification is adopted in the simulation [A\ref{as:GEOP_geopotential}] for convenience purposes, and results in minimum errors at temperate latitudes, as shown in table \ref{tab:H_vs_h}, which compares the results of applying (\ref{eq:EarthModel_GEOP_h2H_table}) at different latitudes with those of (\ref{eq:EarthModel_GEOP_h2H}):


\section{Weather and ICAO Non Standard Atmosphere}\label{sec:EarthModel_ISA}

The atmosphere is the medium in which aircraft motion and its interaction with the environment takes place. The aircraft performances are directly dependent on the atmospheric properties (pressure, temperature, density) at the aircraft locations, and it is hence necessary to define a sufficiently accurate and realistic atmospheric model that properly captures the dependencies of the atmospheric properties.

The International Civil Aviation Organization (\hypertt{ICAO}) Standard Atmosphere (\hypertt{ISA}) \cite{ISA} is an international standard employed in aviation for computing and evaluating aircraft performances, testing, and calibration of instruments \cite{Eshelby2000}. The standard atmosphere properly captures the physics and interrelationships among the different variables, but it is by construction a static model intended for standardization and not capable of capturing the weather induced atmospheric variations with both horizontal position and time. For this reason the simulator employs a generalization of \hypertt{ISA} created by the author that introduces a quasi static atmospheric model based on two parameters (pressure and temperature offsets) that defaults to \hypertt{ISA} when both parameters are zero. The author has named this model the \hypertt{ICAO} Non Standard Atmosphere (\hypertt{INSA}) and published it in \cite{INSA}.
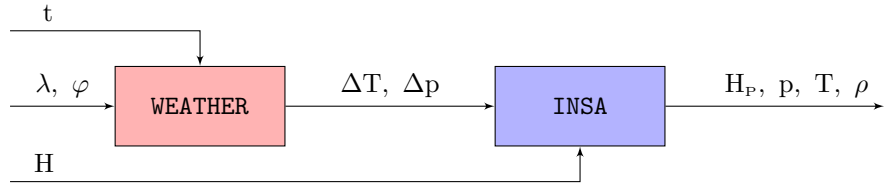
\begin{figure}[h]
\centering
\begin{tikzpicture}[auto, node distance=2cm,>=latex']
	\node [coordinate](lambdaphiinput) {};
    \node [blockred, right of=lambdaphiinput, text width=2cm, node distance=2.5cm] (WEATHER) {\texttt{WEATHER}};	
    \node [block, right of=WEATHER, node distance=5.0cm, text width=2cm] (ISA) {\hypertt{INSA}};
    \node [coordinate, above of=lambdaphiinput, node distance=1cm] (tinput) {};
	\node [coordinate, below of=lambdaphiinput, node distance=1cm] (Hinput) {};
	\node [coordinate, right of=ISA, node distance=4cm](ISAoutput) {};
    \draw [->] (lambdaphiinput) -- node[pos=0.5] {\nm{\lambda, \ \varphi}} (WEATHER);
	\draw [->] (tinput) -| node[pos=0.1] {\nm{t}} (WEATHER);
	\draw [->] (Hinput) -| node[pos=0.03] {\nm{H}} (ISA);	
    \draw [->] (WEATHER) -- node {\nm{\DeltaT, \ \Deltap}} (ISA);
	\draw [->] (ISA) -- node[pos=0.6] {\nm{\Hp, \ p, \ T, \ \rho}} (ISAoutput);
\end{tikzpicture}
\caption{Weather and INSA flow diagram}
\label{fig:EarthModel_ISA_flow_diagram}
\end{figure}

The \hypertt{INSA} model was developed by the author many years ago under contract for the European Organization for the Safety of Air Navigation (\hypertt{EUROCONTROL}), who possesses all relevant intellectual proprietary rights (\copyright 2023 All rights reserved). \hypertt{EUROCONTROL} has awarded the author permission to use the \hypertt{INSA} model in this document.

As shown in figure \ref{fig:EarthModel_ISA_flow_diagram}, the weather model is created ad hoc to recreate the expected meteorological conditions for a given day. Provided with the time and the horizontal position (given by the longitude \nm{\lambda}\ and latitude \nm{\varphi}), it returns the temperature and pressure offsets (\nm{\DeltaT} and \nm{\Deltap}). These two parameters realize an instance of the \hypertt{INSA} model, which relies on the same hypotheses as \hypertt{ISA} and provides the variations of all the atmospheric variables with geopotential altitude at the column of air corresponding to that specific location and time. Although the \hypertt{INSA} model presented in \cite{INSA} encompasses both the troposphere and stratosphere, the simulation is restricted to the lowest layer (the troposphere). 


\subsection{Definitions}\label{subsec:EarthModel_ISA_Definitions}

The following definitions are necessary to properly comprehend and derive an atmospheric model:

\begin{itemize}

\item \emph{Column atmospheric model} is a set of relationships providing the atmospheric pressure, temperature, and density as a function of the geodetic altitude \emph{h} along an infinitesimally narrow column of air orthogonal to the geopotential surfaces of the Earth\footnote{In this way, the geopotential acceleration is always contained within the air column, which is in vertical equilibrium, with no lateral accelerations.}:
\neweq{\lrsb{p \ \ T \ \ \rho}^T = \vec f_{CAM,h}\lrp{h}}{eq:EarthModel_ISA_pTrho}

The insertion of (\ref{eq:EarthModel_GEOP_H2h}) results in:
\neweq{\lrsb{p \ \ T \ \ \rho}^T = \vec f_{CAM,H}\lrp{H}}{eq:EarthModel_ISA_pTrho2}

\item \emph{Standard atmosphere}, also referred to as the \hypertt{ICAO} Standard Atmosphere or \hypertt{ISA}, is the column model defined by \cite{ISA}. It provides expressions for the standard atmospheric pressure, temperature, and density as functions of the pressure altitude \nm{\Hp}:
\neweq{\lrsb{p \ \ \TISA \ \ \rho}^T = \vec f_{ISA}\lrp{\Hp}}{eq:EarthModel_ISA_standard}

Note that the \emph{standard temperature} \nm{\TISA} is defined as the atmospheric temperature that occurs in standard conditions, and the \emph{pressure altitude} \nm{\Hp} is defined as the geopotential altitude H (section \ref{subsec:EarthModel_GEOP_geopotential_altitude}) in those specific conditions. As the \hypertt{ISA} is a standard, these two variables are also widely employed in nonstandard conditions, but it is important to remark that in general \nm{\TISA \neq T} and \nm{\Hp \neq H}.

Standard mean sea level conditions are those that occur in the standard atmosphere at the points where the pressure altitude \nm{\Hp} is zero. They are \nm{\Tzero}, \nm{\pzero}, and \nm{\rhozero} \cite{ISA} (appendix \ref{cha:PhysicalConstants}).

\item \emph{Nonstandard atmospheres} or \hypertt{INSA} are the column models defined by \cite{INSA}, which follow the same hypotheses as the standard atmosphere \hypertt{ISA} but differ from it in that one or both of the following parameters is not zero\footnote{The standard atmosphere can be considered a particularization of a nonstandard atmosphere, in which both the temperature \nm{\DeltaT} and the pressure \nm{\Deltap} offsets are zero.}:

\begin{itemize}
\item The temperature offset \nm{\DeltaT} is the difference in atmospheric temperature between a given \hypertt{INSA} and \hypertt{ISA} at mean sea level.
\item The pressure offset \nm{\Deltap} is the difference in atmospheric pressure at mean sea level between a given \hypertt{INSA} and \hypertt{ISA}.
\end{itemize}

Mean sea level conditions, identified by the subindex \hypertt{MSL}, are those that occur in a nonstandard atmosphere at the points where the geopotential altitude H is zero (section \ref{subsec:EarthModel_GEOP_geopotential_altitude}), and differ from \nm{\lrp{\Tzero, \, \pzero, \, \rhozero}} in nonstandard conditions.
            
The values of these two parameters uniquely identify any nonstandard atmosphere\footnote{Both offsets, \nm{\DeltaT} and \nm{\Deltap}, are constant for any given nonstandard atmosphere.}. Thus, a nonstandard atmosphere provides expressions for the atmospheric pressure, temperature, and density as functions of the geopotential altitude H and its two offset parameters.
\neweq{\lrsb{p \ \ T \ \ \rho}^T = \vec f_{INSA}\lrp{H,\DeltaT,\Deltap}}{eq:EarthModel_ISA_non_standard}

\item \emph{Atmospheric ratios} are dimensionless variables containing the ratios between the atmospheric properties at a given point and those found in standard mean sea level conditions.
\begin{eqnarray}
	\nm{\delta} & = \nm{\frac{p}{\pzero}} \label{eq:EarthModel_ISA_delta} \\
	\nm{\theta} & = \nm{\frac{T}{\Tzero}} \label{eq:EarthModel_ISA_theta} \\
	\nm{\sigma} & = \nm{\frac{\rho}{\rhozero}} \label{eq:EarthModel_ISA_sigma}
\end {eqnarray}

\item \emph{Weather model} is a set of relationships providing the temperature and pressure offsets at mean sea level based on horizontal position and time.
\neweq{\lrsb{\DeltaT \ \ \Deltap}^T = \vec f_{WM}\lrp{\lambda, \varphi, t}}{eq:EarthModel_ISA_weather}

\end{itemize}

The obtainment of the atmospheric properties experienced by the aircraft is a two step process. It is first necessary to build a nonstandard column model based on the temperature and pressure offsets at mean sea level provided by the weather model based on the aircraft longitude, latitude, and time. Once the column model is defined, it is possible to correlate geopotential and pressure altitudes, and then obtain the atmospheric pressure, temperature, and density.


\subsection{Hypotheses}\label{subsec:EarthModel_ISA_Hypotheses}

The \hypertt{INSA} nonstandard column atmospheric models are based in the following hypotheses \cite{INSA}: 

\begin{itemize}

\item The atmosphere is composed by pure air, which is a perfect gas [A\ref{as:ISA_perfect_gas}]. Its thermodynamic properties (pressure, temperature, and density) at any point are thus related by the law of perfect gases:
\neweq{p = \rho \ R \ T}{eq:EarthModel_ISA_ideal_gas}

where R is the specific air constant \cite{ISA} (appendix \ref{cha:PhysicalConstants}).
 
\item The atmosphere is static in relation to the Earth surface [A\ref{as:ISA_static_equilibrium}], so the fluid static equilibrium of forces\footnote{The gravity acceleration \nm{\gc} encompasses the gravitational and centrifugal acceleration fields to which the atmosphere is subjected  to, as explained in section \ref{subsec:EarthModel_GEOP_msl_and_ellip_gravity}.} in the vertical direction (normal to the Earth surface) results in:
\neweq{dp = - \rho \ g_c \ dh}{eq:EarthModel_ISA_static_equilibrium}

Using (\ref{eq:EarthModel_GEOP_h_H}) leads to the pressure derivative with respect to geopotential altitude H:
\neweq{dp = - \rho \ g_0 \ dH}{eq:EarthModel_ISA_static_equilibrium2}

\item The temperature decreases linearly with altitude according to a given gradient \nm{\betaT} (appendix \ref{cha:PhysicalConstants}) when expressed in terms of pressure altitude \nm{\Hp} [A\ref{as:ISA_T_gradient}]. The value of \nm{\betaT} depends on the atmospheric layer:
\neweq{dT = \betaT \ d\Hp}{eq:EarthModel_ISA_T_gradient}

\item The tropopause is the upper limit of the troposphere, and hence of the validity of the expressions in this document. Its altitude is constant when expressed in terms of pressure altitude (\nm{\Hptrop}) [A\ref{as:ISA_tropopause}] (appendix \ref{cha:PhysicalConstants}).

\item The air humidity does not have any influence on the value of any other atmospheric property [A\ref{as:ISA_humidity}], and hence is not taken into account.

\end{itemize}


\subsection{Relationship between Geopotential and Pressure Altitudes}\label{subsec:EarthModel_ISA_H_Hp}

Although geopotential altitude H is defined in section \ref{subsec:EarthModel_GEOP_geopotential_altitude} and pressure altitude \nm{\Hp} in section \ref{subsec:EarthModel_ISA_Definitions}, the obtainment of the mathematical expression linking them is of particular importance, as the performances of an aircraft are usually provided in terms of pressure altitude \nm{\Hp}\footnote{Performance data are provided as functions of \nm{\Hp} to make the data as general as possible. As there is a one to one relationship between pressure altitude and atmospheric pressure (\ref{eq:EarthModel_ISA_p_Hp}), many performance parameters that do not depend on temperature cannot be expressed as functions of geopotential altitude H in a simple manner, as the relationship between H and \nm{\Hp} (\ref{eq:EarthModel_ISA_H_Hp}) does depend on the temperature offset \nm{\DeltaT}.}, while its movement must be expressed in geodetic altitude h, which is directly linked with the geopotential altitude H per (\ref{eq:EarthModel_GEOP_h_H2}).

The combination of the law of perfect gases and fluid static vertical equilibrium (\ref{eq:EarthModel_ISA_ideal_gas}, \ref{eq:EarthModel_ISA_static_equilibrium2}) results in:
\neweq{dp = - \frac{p}{R \ T} \, \gzero \ dH}{eq:EarthModel_ISA_static_equilibrium3}

In the case of the standard atmosphere, and according to the definitions of section \ref{subsec:EarthModel_ISA_Definitions}, the atmospheric temperature becomes the standard temperature (\nm{T = \TISA}), and the geopotential altitude turns into the pressure altitude (\nm{H = \Hp}):
\neweq{dp = - \frac{p}{R \ \TISA} \, \gzero \ d\Hp}{eq:EarthModel_ISA_static_equilibrium4}

Dividing the last two expressions results in the ratio between the derivatives of both types of altitudes:
\neweq{\frac{dH}{d\Hp} = \frac{T}{\TISA}}{eq:EarthModel_ISA_H_Hp_ratio}

The ratio between incremental changes of geopotential altitude and pressure altitude is thus the same as that between the atmospheric temperature at that point and the temperature that would occur in standard atmospheric conditions at the same point. It is represented in figure \ref{fig:earth_dHdHp_vs_Hp}.


\subsection{Expressions}\label{subsec:EarthModel_ISA_Expressions}

According to the flow diagram shown in figure \ref{fig:EarthModel_ISA_flow_diagram}, once the temperature and pressure offsets (\nm{\DeltaT}, \nm{\Deltap}) have been obtained from the weather model for a given horizontal position and time, the corresponding \hypertt{INSA} model is responsible for providing the pressure, temperature, and pressure altitude \nm{\Hp} corresponding to the input geopotential altitude H. The atmospheric density \nm{\rho} can then be easily obtained by means of (\ref{eq:EarthModel_ISA_ideal_gas}), while the atmospheric ratios (pressure \nm{\delta}, temperature \nm{\theta}, and density \nm{\sigma}) can be determined by means of (\ref{eq:EarthModel_ISA_delta}, \ref{eq:EarthModel_ISA_theta}, \ref{eq:EarthModel_ISA_sigma}).

The atmospheric relationships are the result of integrating the differential equations introduced in section \ref{subsec:EarthModel_ISA_Hypotheses}, so their values at certain reference altitudes are required to obtain the complete expressions. These reference altitudes are the following:

\begin{itemize}

\item \emph{Standard mean sea level conditions}. Represented by subindex \nm{\Hp = 0}, they are defined in section \ref{subsec:EarthModel_ISA_Definitions} as those occurring in standard atmospheric conditions where the pressure altitude \nm{\Hp} is zero. The corresponding pressure \nm{\pHpzero} and standard temperature \nm{\TISAHpzero} are \nm{\pzero} and \nm{\Tzero} respectively. The temperature offset \nm{\DeltaT} sets the value of the real temperature \nm{\THpzero} for nonstandard atmospheres, while the value of the geopotential altitude \nm{\HHpzero} is provided by applying (\ref{eq:EarthModel_ISA_H_Hp}) to standard mean sea level conditions.
\begin{eqnarray}
\nm{\HpHpzero}   & = & 0 \label{eq:EarthModel_ISA_smsl_Hp} \\
\nm{\TISAHpzero} & = & \nm{\Tzero} \label{eq:EarthModel_ISA_smsl_Tisa} \\
\nm{\pHpzero}    & = & \nm{\pzero} \label{eq:EarthModel_ISA_smsl_p} \\
\nm{\THpzero}    & = & \nm{\Tzero + \DeltaT} \label{eq:EarthModel_ISA_smsl_T} \\
\nm{\HHpzero}    & = & \nm{\frac{1}{\betaT}\lrp{\Tzero - \TISAMSL + \DeltaT \ \text{Ln} \frac{\Tzero}{\TISAMSL}}} \label{eq:EarthModel_ISA_smsl_H}
\end{eqnarray}

\item \emph{Mean sea level conditions}. Represented by subindex \hypertt{MSL}, the geopotential altitude \nm{\HMSL} is always zero per the definition of section \ref{subsec:EarthModel_ISA_Definitions}, while the pressure offset \nm{\Deltap} sets the value of the atmospheric pressure \nm{\pMSL}. The values of the pressure altitude \nm{\HpMSL}, standard temperature \nm{\TISAMSL}, and temperature \nm{\TMSL} are provided by (\ref{eq:EarthModel_ISA_msl_Hp2}, \ref{eq:EarthModel_ISA_msl_Tisa2}, \ref{eq:EarthModel_ISA_msl_T2}).
\begin{eqnarray}
\nm{\HMSL}    & = & \nm{0} \label{eq:EarthModel_ISA_msl_H} \\
\nm{\pMSL}    & = & \nm{\pzero + \Deltap} \label{eq:EarthModel_ISA_msl_p} \\
\nm{\HpMSL}   & = & \nm{\frac{\Tzero}{\betaT} \lrsb{\lrp{\frac{\pMSL}{\pzero}}^{\sss \BRg} - 1}} \label{eq:EarthModel_ISA_msl_Hp} \\
\nm{\TISAMSL} & = & \nm{\Tzero + \betaT \ \HpMSL} \label{eq:EarthModel_ISA_msl_Tisa} \\
\nm{\TMSL}    & = & \nm{\Tzero + \DeltaT + \betaT \ \HpMSL} \label{eq:EarthModel_ISA_msl_T}
\end{eqnarray}

\end{itemize}


\subsubsection{Relationship between standard temperature and pressure altitude}\label{subsubsec:EarthModel_ISA_Relationship_TisaHp}

The application of the temperature gradient (\ref{eq:EarthModel_ISA_T_gradient}) to the standard atmosphere results in:
\neweq{d\TISA = \betaT \ d\Hp} {eq:EarthModel_ISA_T_gradient_iSA}

This equation can be integrated between standard mean sea level (\nm{\TISAHpzero = \Tzero}, \nm{\HpHpzero = 0}) and a generic point, resulting in:
\neweq{\TISA = \TISA \lrp{\Hp} =  \Tzero + \betaT \ \Hp} {eq:EarthModel_ISA_Tisa_Hp}

The relationship between \nm{\TISA} and \nm{\Hp} is hence unique for all \hypertt{INSA} atmospheres and does not depend on \nm{\DeltaT} nor \nm{\Deltap}.


\subsubsection{Relationship between temperature and pressure altitude}\label{subsubsec:EarthModel_ISA_Relationship_THp}

The integration of the temperature gradient (\ref{eq:EarthModel_ISA_T_gradient}) between standard mean sea level (\nm{\THpzero = \Tzero + \DeltaT}, \nm{\HpHpzero = 0}) and a generic point results in:
\neweq{T = T\lrp{\Hp, \DeltaT} = \Tzero + \DeltaT + \betaT \ \Hp}{eq:EarthModel_ISA_T_Hp}

The above expression, represented in figure \ref{fig:earth_Tp_vs_Hp}, describes how the atmospheric temperature T varies linearly with both the pressure altitude \nm{\Hp} and the temperature offset \nm{\DeltaT}, but does not depend on the pressure offset \nm{\Deltap}.


\subsubsection{Relationship between pressure and pressure altitude}\label{subsubsec:EarthModel_ISA_Relationship_pHp}

The combination of the hydrostatic equilibrium (\ref{eq:EarthModel_ISA_static_equilibrium2}) and the law of perfect gases (\ref{eq:EarthModel_ISA_ideal_gas}) for standard atmospheres (\nm{\Hp = H} and \nm{\TISA = T}) results in:
\neweq{\frac{dp}{p} = -\frac{\gzero}{R \ \TISA} \ d\Hp}{eq:EarthModel_ISA_p_Hp1}

Expression (\ref{eq:EarthModel_ISA_Tisa_Hp}) can be inserted into the above relationship:
\neweq{\frac{dp}{p} = -\frac{\gzero}{R} \ \frac{d\Hp}{\Tzero + \betaT \ \Hp}} {eq:EarthModel_ISA_p_Hp2}

Its integration between standard mean sea level conditions (\nm{\pHpzero = \pzero}, \nm{\HpHpzero = 0}) and a generic point turns into:
\begin{eqnarray}
\nm{p = p\lrp{\Hp}} & = & \nm{\pzero \ \lrp{1 + \frac{\betaT}{\Tzero} \ \Hp}^{\sss \gBR}}\label{eq:EarthModel_ISA_p_Hp} \\
\nm{\Hp = \Hp\lrp{p}} & = & \nm{\frac{\Tzero}{\betaT} \, \lrsb{\lrp{\frac{p}{\pzero}}^{\sss \BRg} - 1}}\label{eq:EarthModel_ISA_Hp_p}
\end{eqnarray}

This expression, shown in figure \ref{fig:earth_Tp_vs_Hp}, describes the one to one relationship that exists between p and \nm{\Hp} as their relationship does not depend on \nm{\DeltaT} nor \nm{\Deltap}. As it easier to work with length units than pressure ones, aircraft performances are usually given in terms of pressure altitude, although obviously the aerodynamic and propulsive responses of an aircraft are based on atmospheric pressure.

The pressure altitude at mean sea level \nm{\HpMSL} previously shown at (\ref{eq:EarthModel_ISA_msl_Hp}) can be obtained from \ref{eq:EarthModel_ISA_Hp_p}:
\neweq{\HpMSL = \frac{\Tzero}{\betaT} \ \lrsb{\lrp{\frac{\pMSL}{\pzero}}^{\sss \BRg} - 1}}{eq:EarthModel_ISA_msl_Hp2}

Inserting this result into (\ref{eq:EarthModel_ISA_Tisa_Hp}) and (\ref{eq:EarthModel_ISA_T_Hp}) results in the standard and nonstandard temperatures at mean sea level (\nm{\TISAMSL} and \nm{\TMSL}) previously shown at (\ref{eq:EarthModel_ISA_msl_Tisa}) and (\ref{eq:EarthModel_ISA_msl_T}):
\begin{eqnarray}
\nm{\TISAMSL} & = & \nm{\Tzero + \betaT \ \HpMSL} \label{eq:EarthModel_ISA_msl_Tisa2} \\
\nm{\TMSL}    & = & \nm{\Tzero + \DeltaT + \betaT \ \HpMSL} \label{eq:EarthModel_ISA_msl_T2}
\end{eqnarray}
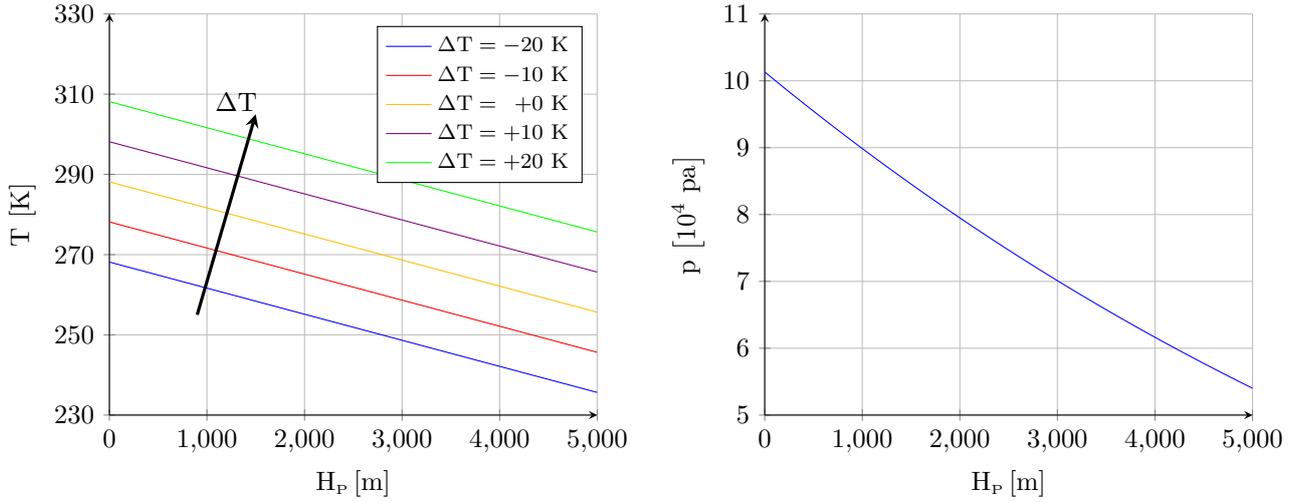
\begin{figure}[h]
\centering
\begin{tikzpicture}
\begin{axis}[
cycle list={{blue,no markers},{red,no markers},{orange!50!yellow, no markers},{violet, no markers},{green, no markers}},
width=8.0cm,
xmin=0, xmax=5000, xtick={0,1000,...,5000},
xlabel={\nm{\Hp \left[m\right]}},
xmajorgrids,
ymin=230, ymax=330, ytick={230,250,...,330},
ylabel={\nm{T \, \left[K\right]}},
ymajorgrids,
axis lines=left,
axis line style={-stealth},
legend entries={\nm{\DeltaT = - 20 \ K},\nm{\DeltaT = - 10 \ K}, \nm{\DeltaT = \ \, + 0 \ K}, \nm{\DeltaT = + 10 \ K}, \nm{\DeltaT = + 20 \ K}},
legend pos=north east,
legend style={font=\footnotesize},
legend cell align=left,
]
\pgfplotstableread{figs/ch02_earth/T_vs_Hp__DeltaT.txt}\mytable
\addplot table [header=false, x index=0,y index=1] {\mytable};
\addplot table [header=false, x index=0,y index=2] {\mytable};
\addplot table [header=false, x index=0,y index=3] {\mytable};
\addplot table [header=false, x index=0,y index=4] {\mytable};
\addplot table [header=false, x index=0,y index=5] {\mytable};
\draw [-stealth, very thick] (900,255) -- (1500,305);
\path node at (1300,308) {\nm{\DeltaT}};
\end{axis}	
\end{tikzpicture}%
\hskip 10pt
\begin{tikzpicture}
\begin{axis}[
cycle list={{blue,no markers},{red,no markers},{orange!50!yellow, no markers},{violet, no markers},{green, no markers}},
width=8.0cm,
xmin=0, xmax=5000, xtick={0,1000,...,5000},
xlabel={\nm{\Hp \left[m\right]}},
xmajorgrids,
ymin=5, ymax=11,
ytick={5,6,...,11},
ylabel={\nm{p \, \left[10^4 \ pa\right]}},
ymajorgrids,
axis lines=left,
axis line style={-stealth},
]
\pgfplotstableread{figs/ch02_earth/p_vs_Hp.txt}\mytable
\addplot table [header=false, x index=0,y index=1] {\mytable};
\end{axis}		
\end{tikzpicture}%
\caption{Temperature and pressure versus pressure altitude for various \nm{\DeltaT}}
\label{fig:earth_Tp_vs_Hp}
\end{figure}

 
\subsubsection{Relationship between geopotential altitude and pressure altitude}\label{subsubsec:EarthModel_ISA_HHP}

The starting point is the relationship between the differential increments of both altitudes provided by (\ref{eq:EarthModel_ISA_H_Hp_ratio}):
\neweq{\frac{dH}{d\Hp} = \frac{T}{\TISA}}{eq:EarthModel_ISA_H_Hp_ratio2}

This expression shows that geopotential altitude H grows quicker than pressure altitude \nm{\Hp} on warm atmospheres (\nm{\DeltaT > 0}), and slower in cold ones. For warm atmospheres, the ratio grows with altitude until reaching a maximum in the tropopause; the opposite is true for cold atmospheres.

Introducing (\ref{eq:EarthModel_ISA_Tisa_Hp}) and (\ref{eq:EarthModel_ISA_T_Hp}) results in:
\neweq{\frac{dH}{d\Hp} = \frac{\Tzero + \DeltaT + \betaT \ \Hp}{\Tzero + \betaT \ \Hp} = 1 + \frac{\DeltaT}{\Tzero + \betaT \ \Hp}}{eq:EarthModel_ISA_HHp1}

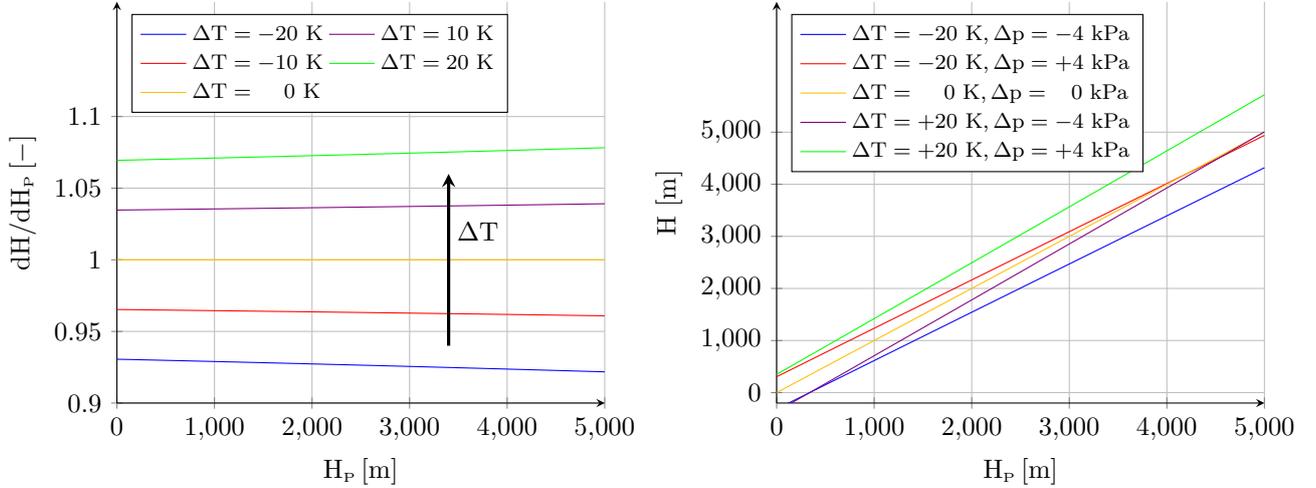
\begin{figure}[h]
\centering
\begin{tikzpicture}
\begin{axis}[
cycle list={{blue,no markers},{red,no markers},{orange!50!yellow, no markers},{violet, no markers},{green, no markers}},
width=8.0cm,
xmin=0, xmax=5000, xtick={0,1000,...,5000},
xlabel={\nm{\Hp \left[m\right]}},
xmajorgrids,
ymin=0.90, ymax=1.18, ytick={0.90,0.95,1.00,1.05,1.10},
ylabel={\nm{dH/d\Hp \left[-\right]}},
ymajorgrids,
axis lines=left,
axis line style={-stealth},
legend entries={\nm{\DeltaT = -20 \ K},\nm{\DeltaT = - 10 \ K}, \nm{\DeltaT = \ \ \ \, 0 \ K}, \nm{\DeltaT = 10 \ K}, \nm{\DeltaT = 20 \ K}},
legend pos=north west,
transpose legend,
legend columns=3,
legend style={font=\footnotesize},
legend cell align=left,
]
\pgfplotstableread{figs/ch02_earth/dHdHp_vs_Hp__DeltaT.txt}\mytable
\addplot table [header=false, x index=0,y index=1] {\mytable};
\addplot table [header=false, x index=0,y index=2] {\mytable};
\addplot table [header=false, x index=0,y index=3] {\mytable};
\addplot table [header=false, x index=0,y index=4] {\mytable};
\addplot table [header=false, x index=0,y index=5] {\mytable};
\draw [-stealth, very thick] (3400,0.94) -- (3400,1.06);
\path node at (3700,1.02) {\nm{\DeltaT}};
\end{axis}	
\end{tikzpicture}%
\hskip 1pt
\begin{tikzpicture}
\begin{axis}[
cycle list={{blue,no markers},{red,no markers},{orange!50!yellow, no markers},{violet, no markers},{green, no markers}},
width=8.0cm,
xmin=0, xmax=5000, xtick={0,1000,...,5000},
xlabel={\nm{\Hp \left[m\right]}},
xmajorgrids,
ymin=-200, ymax=7500, ytick={0,1000,2000,3000,4000,5000},
ylabel={\nm{H \, \left[m\right]}},
ymajorgrids,
axis lines=left,
axis line style={-stealth},
legend entries={\nm{\DeltaT = - 20 \ K, \Deltap = - 4 \ kPa},\nm{\DeltaT = - 20 \ K, \Deltap = + 4 \ kPa}, \nm{\DeltaT = \ \, \ \ 0 \ K, \Deltap = \ \ \, 0 \ kPa}, \nm{\DeltaT = + 20 \ K, \Deltap = - 4 \ kPa}, \nm{\DeltaT = + 20 \ K, \Deltap = + 4 \ kPa}},
legend pos=north west,
legend style={font=\footnotesize},
legend cell align=left,
]
\pgfplotstableread{figs/ch02_earth/H_vs_Hp__DeltaT.txt}\mytable
\addplot table [header=false, x index=0,y index=1] {\mytable};
\addplot table [header=false, x index=0,y index=2] {\mytable};
\addplot table [header=false, x index=0,y index=3] {\mytable};
\addplot table [header=false, x index=0,y index=4] {\mytable};
\addplot table [header=false, x index=0,y index=5] {\mytable};
\end{axis}	
\end{tikzpicture}%
\caption{Geopotential altitude versus pressure altitude for various \nm{\DeltaT}}
\label{fig:earth_dHdHp_vs_Hp}
\end{figure}

Its integration between mean sea level conditions (\nm{\HpMSL} provided by (\ref{eq:EarthModel_ISA_msl_Hp2}), \nm{\HMSL = 0}) and any point below the tropopause results in:
\neweq{H = H\lrp{\Hp, \DeltaT, \Deltap} = \Hp - \HpMSL + \frac{\DeltaT}{\betaT} \ Ln \frac{\Tzero + \betaT \ \Hp}{\TISAMSL}} {eq:EarthModel_ISA_H_Hp}

where \nm{\TISAMSL} is given by (\ref{eq:EarthModel_ISA_msl_Tisa2}). Expressions (\ref{eq:EarthModel_ISA_H_Hp_ratio2}) and (\ref{eq:EarthModel_ISA_H_Hp}) are graphically represented in figure \ref{fig:earth_dHdHp_vs_Hp} and show the influences of both the temperature and pressure offsets: while \nm{\DeltaT} sets the ratio between increments of both types of altitudes, the influence of \nm{\Deltap} is by means of the pressure at mean sea level \nm{\pMSL} (\ref{eq:EarthModel_ISA_msl_Hp}). In other words, when representing H versus \nm{\Hp}, the temperature offset \nm{\DeltaT} sets the slope and the pressure offset \nm{\Deltap} marks the zero point.

It is impossible to reverse [\ref{eq:EarthModel_ISA_H_Hp}] to obtain an explicit expression of the pressure altitude \nm{\Hp} as a function of the geopotential altitude H, but their one to one relationship converges quickly by iteration.


\section{Wind and Turbulence}\label{sec:EarthModel_WIND}

An aircraft flight is heavily influenced by the wind encountered during its execution, which can be divided into low and high frequency components [A\ref{as:WIND_composition}]. Each component influences the flight in different ways.
 
The low frequency wind, referred to as \emph{wind field} in this document, is the quasi stationary airspeed caused by large scale pressure and temperature variations in the atmosphere. Its variations with position (longitude, latitude, altitude) and time are very small when compared with the accelerations experienced by the aircraft, so its influence on the flight dynamics is quite reduced. On the other hand, the sustained effect of the wind field over time has a major influence on the aircraft kinematics and its resulting trajectory, as the wind field can be considered to be continuously displacing the aircraft. The wind field can be obtained from a meteorological service \cite{GFS} and is usually provided in the \hypertt{NED} reference frame (section \ref{subsec:RefSystems_N}), with the \third\ component being generally discarded:
\neweq{\vWINDN = \lrsb{\vWINDNi \ \ \vWINDNii \ \ \vWINDNiii}^T = \vec f_{WIND} \lrp{\TEgdt, t} = \vec f_{WIND}\lrp{\lambda, \ \varphi, \ h, \ t}}{eq:EarthModel_WIND_low_freq}

The \emph{wind bearing} \nm{\chiWIND} is defined as the angle that exists between geodetic North (\nm{\iNi}) and the projection of the wind velocity (\nm{\vWIND}) onto the horizontal plane (defined by \nm{\iNi} and \nm{\iNii}), while the \emph{wind path angle} \nm{\gammaWIND} represents the angle existing between \nm{\vWIND} and its projection onto the horizontal plane.
\neweq{\vWINDN = \vwind \ \lrsb{\cos\gammaWIND \ \cos\chiWIND \ \ \ \ \cos\gammaWIND \ \sin\chiWIND \ \ \ \ - \sin\gammaWIND}^T}{eq:EarthModel_WIND_low_freq2}

\emph{Turbulence} encompasses the high frequency wind variations\footnote{Other infrequent wind phenomena such as wind shear and wind gusts are not considered in this document as both are short lived variations in the wind intensity and/or direction that, although important from a design point of view, do not have significant effects on the overall aircraft trajectory and are not encountered often enough as to influence the aircraft dynamics.} that balance themselves out over time and hence have a relatively small effect on the resulting aircraft trajectory. The turbulence however produces strong but high frequency accelerations that have a significant influence on both the aircraft dynamics and its control system, which needs to continuously adjust the control surfaces to reduce the turbulence effect on the aircraft guidance objectives.

The simulation relies on the Dryden turbulence model described in \cite{Turbulence1980, Gage2003}, where turbulence is modeled as a stochastic process defined by its velocity spectra \cite{Etkin1972}. The air turbulence is assumed to be frozen in time and space, implying that the turbulence experienced by the aircraft results exclusively from its motion relative to the frozen turbulence, not from the variations or motion of the turbulence itself [A\ref{as:WIND_turbulence}].

The model distinguishes between two distinct turbulence regions based on the aircraft \emph{height} (altitude over the terrain): below \nm{1000 \ ft} and above \nm{2000 \ ft}, with linear interpolation in between. The Dryden model is based on expressions that provide the turbulence scale lengths and turbulence intensities based on the aircraft height\footnote{The turbulence intensity expressions also rely on the intensity level (light, moderate, severe).} for each of the two height regions and each of the three \nm{\FT} turbulence frame directions (section \ref{subsec:RefSystems_T}). 

The velocity spectrum for each of the three \nm{\FT} axes depends on the previously obtained turbulence scale lengths and intensities, as well as on the aircraft height, airspeed, and wingspan. Turbulence is then generated by filtering band limited white noise based on the spectral representation. The result is a time varying sequence of turbulence intensities \nm{\vTURBT = \vec f\lrp{t}} than can be converted to its \hypertt{NED} form by (\ref{eq:RefSystems_qNT}):
\neweq{\vTURBN \lrp{t} = \vec g_{\ds{\vec \zeta_{\sss NT}\lrp{t}_{*}}} \big(\vTURBT \lrp{t}\big) = \vec g_{\ds{\vec q_{\sss NT}\lrp{t}_{*}}} \big(\vTURBT \lrp{t}\big) = \vec g_{\ds{\vec R_3\lrp{\chiWIND(t)}_{*}  }} \big(\vTURBT \lrp{t}\big)} {eq:EarthModel_WIND_turbulence}


\section{World Magnetic Model}\label{sec:EarthModel_WMM}	 
  
The Earth magnetic field is a valuable tool for navigation since its measurement by magnetometers (section \ref{subsec:Sensors_NonInertial_Magnetometers}) provides the aircraft with an absolute reference in the form of a vector that always points towards the magnetic North, and which can be used to control the aircraft attitude, specially its heading (section \ref{subsec:RefSystems_B}). The Earth magnetic field is in fact a composite of several others that superimpose and interact with each other \cite{WMM}. The most important of these are: 
\begin{itemize}
\item The Earth's main magnetic field generated in the conducting, fluid outer core.
\item The crustal field generated in Earth's crust and upper mantle.
\item The combined disturbance field from electrical currents flowing in the upper atmosphere and magnetosphere, which induce electrical currents in the sea and ground.
\end{itemize}  	 
  	 
The World Magnetic Model (\hypertt{WMM}) \cite{WMM} consists of a degree and order 12 spherical harmonic main (core generated) field model comprised of 168 spherical harmonic Gauss coefficients and degree and order 12 spherical harmonic secular variation (core generated, slow temporal variation) field model. The observed magnetic field forming the basis for the model is a sum of contributions of the main field (varying in both time and space), the crustal field (varies spatially, but considered constant in time for the time scales of the \hypertt{WMM}), and the disturbance fields (varying in space and rapidly in time). Earth's main magnetic field dominates, accounting for over 95\% of the field strength at the Earth's surface. Secular variation is the slow change in time of the main magnetic field. 	
	
The \hypertt{WMM} provides the magnetic field intensity \nm{\Bvec} viewed in the \hypertt{NED} frame \nm{\FN} as a function of the geodetic coordinates \nm{\TEgdt} and time. As in the case of gravity, the subindex \nm{\sss MOD} (model) distinguishes this expression from that developed in section \ref{sec:EarthModel_accuracy}.
\neweq{\BNMODEL = \vec f_B\lrp{\TEgdt, \, t} = \vec f_B\lrp{\lambda, \, \varphi, \, h, \, t}}{eq:EarthModel_WMM}

The \emph{magnetic deviation} \nm{\chiMAG} is the angle between geodetic North (\nm{\iNi}) and the projection of the magnetic field onto the horizontal plane (made by \nm{\iNi} and \nm{\iNii}), while the \emph{magnetic inclination} \nm{\DeltaMAG} is the angle between the magnetic field and its horizontal projection \cite{WMM}. While \nm{\chiMAG} follows the usual criterion for yaw angles of being positive when East of North, \nm{\DeltaMAG} is considered positive when heading down.
\neweq{\chiMAG = \arctan \dfrac{\BNii}{\BNi} \hspace{80pt} \DeltaMAG = \arctan \dfrac{\BNiii}{\sqrt{{\BNi}^2 + {\BNii}^2}} = \arctan \dfrac{\BNiii}{\BNhor}}{eq:EarthModel_WMM_deviation_inclination}

Instead of implementing the \hypertt{WMM}, the simulator relies on preloaded values taken from the magnetic field calculator found in \cite{WMMCalculator}. Assuming that the Earth magnetic field can be locally approximated by a bilinear function on longitude and latitude while neglecting the influence of altitude and time [A\ref{as:WMM_linear}], a fact numerically checked by the author, it is possible to use the above calculator to  obtain the magnetic field at four points forming a square encompassing the expected aircraft trajectory, employing the current time and the expected flight altitude, and then linearly interpolate based on longitude and latitude to obtain the expected magnetic field at every location. This approach not only saves the significant effort required to implement the \hypertt{WMM}, but also employs less computing resources at execution time.


\section{Accuracy of Gravity and Magnetic Models}\label{sec:EarthModel_accuracy}

The navigation algorithms described in chapter \ref{cha:nav} rely on a precise knowledge of the gravity acceleration and the magnetic field to which the aircraft is exposed, as the former is measured by the accelerometers and is the main factor in the estimation of the aircraft pitch and bank angles, while the latter is measured by the magnetometers and employed to estimate the aircraft heading. 

Hence, any differences between the real gravity and magnetic field experienced by the aircraft and those computed on board have a negative influence on the accuracy of the navigation algorithms. For this reason it is important to distinguish between the gravity and magnetic field models employed by the navigation system to estimate the value of \nm{\vec g_c^{\sss N}} and \nm{\vec B^{\sss N}} based on the estimated aircraft position and time, and the real values employed in chapter \ref{cha:FlightPhysics} for the obtainment of the aircraft trajectory, and in chapter \ref{cha:Sensors} as inputs to the accelerometers and magnetometers.

The real gravity and magnetic field experienced by the aircraft are thus not the same as those that the aircraft system estimates. To account for the differences, the simulator employs the models derived in sections \ref{sec:EarthModel_GEOP} and \ref{sec:EarthModel_WMM}, this is, expressions (\ref{eq:EarthModel_Gravity}) and (\ref{eq:EarthModel_WMM}), when determining the navigation system estimations (chapter \ref{cha:nav}), but corrupts them with stochastic perturbations when computing the real gravity and magnetic field experienced by the aircraft, this is, the values determining its motion and being measured by the sensors. 

Note that although it is possible to employ more accurate models, the objective of this section is to introduce realism into the simulation by acknowledging that there are differences between reality and the onboard models. It is much more important to introduce realistic differences between the real and modeled gravity and magnetic fields, than to ignore those differences (that always exist) and in turn dedicate scarce resources to implement more accurate models.
\begin{center}
\begin{tabular}{lcccl}
	\hline
	Parameter					  & Values & Unit &   & Expression \\
	\hline
    \nm{\sigmagDEV}               & 0.0001 & \nm{m/\,s^{\sss2}}	& \hspace{30pt} & \nm{\gcDEV \sim N\lrp{0, \ \sigmagDEV^2}} \\
    \nm{\sigmagammaDEV}           & 0.0028 & \nm{^{\circ}}      & \hspace{30pt} & \nm{\gammaDEV \hspace{4pt} \sim N\lrp{0, \ \sigmagammaDEV^2}} \\
	\nm{\phiDEV}				  & 	   & \nm{^{\circ}}      & \hspace{30pt} & \nm{\phiDEV \hspace{3pt} \sim U\lrp{-179, +180}} \\
	\hline
\end{tabular}
\end{center}
\captionof{table}{Gravity deviation model parameters}\label{tab:EarthModel_gravity_accuracy}

While the intensity of the gravity acceleration is accurately captured by the \hypertt{WGS84} models, deflections of the vertical at the surface of the Earth are at most of the order of \nm{60^{\prime\prime} = 0.0167^{\circ}} in mountain regions and seldom larger than \nm{20^{\prime\prime} = 0.0056^\circ} in lowland areas \cite{Chatfield1997}, understood as \nm{3 \sigma}. At altitudes above the Earth, differences are even smaller. An analysis of the differences between various gravity models is provided by \cite{Madden2006}. The author has developed a model based on three stochastic variables that modifies the gravity vector intensity and direction from the \nm{\gcNMODEL} value provided by the (\ref{eq:EarthModel_Gravity}) model. Each flight results in a different realization of this gravity deviation, but the difference between the real and modeled gravity vectors does not change during a given flight [A\ref{as:MODELS_gc}].

The model is described by table \ref{tab:EarthModel_gravity_accuracy} and expressions (\ref{eq:EarthModel_accuracy_gravity}) and (\ref{eq:EarthModel_accuracy_gravity_aux}), and makes use of the normal and uniform distributions described in \cite{LIE}. The gravity field intensity is first modified by \nm{\gcDEV}, and then it gets rotated away from the vertical by rotating an angle \nm{\gammaDEV} around a horizontal direction provided by \nm{\phiDEV}. Note the use of the \nm{\mathbb{SO}\lrp{3}} exponential map:
\begin{eqnarray}
\nm{\gcNREAL} & = & \nm{\vec g_{Exp\ds{\lrp{{\vec n}_{\sss DEV}^{\sss N} \, \gammaDEV}_{*}}} \Big(\iNiii \ \lrp{\gcNMODELiii - \gcDEV}\Big)}\label{eq:EarthModel_accuracy_gravity} \\
\nm{\vec n_{\sss DEV}^{\sss N}} & = & \nm{\lrsb{\cos \phi_{\sss DEV} \ \ \ \ \sin \phi_{\sss DEV} \ \ \ \ 0}^T}\label{eq:EarthModel_accuracy_gravity_aux}
\end{eqnarray}

Local variations are more significant in the case of the Earth magnetic field. Fortunately, \hypertt{WMM} \cite{WMM} provides values for the standard deviations of the differences between the magnetic field intensities provided by the model and those measured throughout the Earth. These are listed in table \ref{tab:EarthModel_magnetism_accuracy}:
\begin{center}
\begin{tabular}{lrccl}
	\hline
	Parameter & Values & Unit &   & Expression \\
	\hline
    \nm{\sigma_{B \sss{DEV,1}}} & 138 & nT & \hspace{30pt} & \nm{\BNDEVi   \sim N\lrp{0, \ {\sigma_{B \sss{DEV,1}}}^2}} \\
    \nm{\sigma_{B \sss{DEV,2}}} &  89 & nT & \hspace{30pt} & \nm{\BNDEVii  \sim N\lrp{0, \ {\sigma_{B \sss{DEV,2}}}^2}} \\
	\nm{\sigma_{B \sss{DEV,3}}} & 165 & nT & \hspace{30pt} & \nm{\BNDEViii \sim N\lrp{0, \ {\sigma_{B \sss{DEV,3}}}^2}} \\
	\hline
\end{tabular}
\end{center}
\captionof{table}{Magnetism deviation model parameters}\label{tab:EarthModel_magnetism_accuracy}

The author has implemented a simple model that modifies the magnetic field provided by (\ref{eq:EarthModel_WMM}) based on the three standard distributions on table \ref{tab:EarthModel_magnetism_accuracy}. Each flight results in a different realization of this magnetic field deviation, but the difference between the real and modeled magnetic field vectors does not change during a given flight [A\ref{as:MODELS_B}]. 
\begin{eqnarray}
\nm{\BNREAL} & = & \nm{\BNMODEL - \BNDEV}\label{eq:EarthModel_accuracy_magnetism}\\
\nm{\BNDEV}  & = & \nm{\lrsb{\BNDEVi \ \ \BNDEVii \ \ \BNDEViii}^T}\label{eq:EarthModel_accuracy_magnetism_aux}
\end{eqnarray}

As indicated above, the real gravity and magnetic fields represented by (\ref{eq:EarthModel_accuracy_gravity}) and (\ref{eq:EarthModel_accuracy_magnetism}) are employed when determining the aircraft motion (chapter \ref{cha:FlightPhysics}) and the inputs to the aircraft sensors (chapter \ref{cha:Sensors}), while models represented by (\ref{eq:EarthModel_Gravity}) and (\ref{eq:EarthModel_WMM}) are used by the navigation system in chapter \ref{cha:nav} when estimating those vectors.


\section{Implementation, Validation, and Execution} \label{sec:EarthModel_implementation}

The Earth model as described in this chapter has been implemented as an object oriented \texttt{C++} library based on the flow diagram shown in figure \ref{fig:EarthModel_flow_diagram}, which contains four modules that are either user configurable or stochastic (red blocks in the figure) and four deterministic ones (blue). Given the time and the aircraft position represented by its geodetic coordinates \nm{\TEgdt}, the Earth model computes the atmospheric properties (\nm{\Hp, \; p, \; T, \; \rho}) that determine the aircraft performances, external influences over the aircraft dynamics such as the gravity acceleration \nm{\gcNMODEL}, the wind velocity \nm{\vWINDN}, and the turbulence field \nm{\vTURBN}, as well as the Earth magnetic field \nm{\BNMODEL} and the local radii of curvature N and M. As indicated in section \ref{sec:EarthModel_accuracy}, it is important to distinguish between the gravity and magnetic fields measured by the aircraft sensors (\nm{\gcNREAL, \, \BNREAL}) and those estimated by the navigation system (\nm{\gcNMODEL, \, \BNMODEL}).
\begin{itemize}
\item The \hypertt{WGS84} block contains the ellipsoid shape described in section \ref{sec:EarthModel_WGS84}, the Earth focused reference systems of section \ref{sec:RefSystems_Earth}, and the gravity model of section \ref{subsec:EarthModel_GEOP_msl_and_ellip_gravity}. Its validation is simple given the low complexity of the equations contained in those sections.

\item The \texttt{GEOP} block represents the relationships between geometric and geopotential altitudes described in section \ref{sec:EarthModel_GEOP}. The explicit expressions are also simple to verify.

\item The \texttt{WEATHER} block allows the user to introduce different functions (from constant to three dimensional tables) to represent the weather experienced by the aircraft that provides the temperature and pressure offsets at mean sea level as functions of longitude, latitude, and time (\ref{eq:EarthModel_ISA_weather}).

\item The \hypertt{INSA} block contains an implementation of the standard and nonstandard atmospheric models described in section \ref{sec:EarthModel_ISA}. The official standard atmosphere literature \cite{ISA} provides detailed tables of the pressure, temperature, and geopotential altitude variations with pressure altitude, against which the implementation has been validated. The nonstandard atmospheres \cite{INSA} have however been developed by the author to provide a framework capable of simulating different atmospheric conditions while preserving all the hypotheses on which the standard atmosphere is based. Validation of the nonstandard atmospheres has involved computing the numerical derivatives of the resulting expressions (section \ref {subsec:EarthModel_ISA_Expressions}) to verify that they comply with the hypotheses of section \ref{subsec:EarthModel_ISA_Hypotheses}.
\end{itemize}
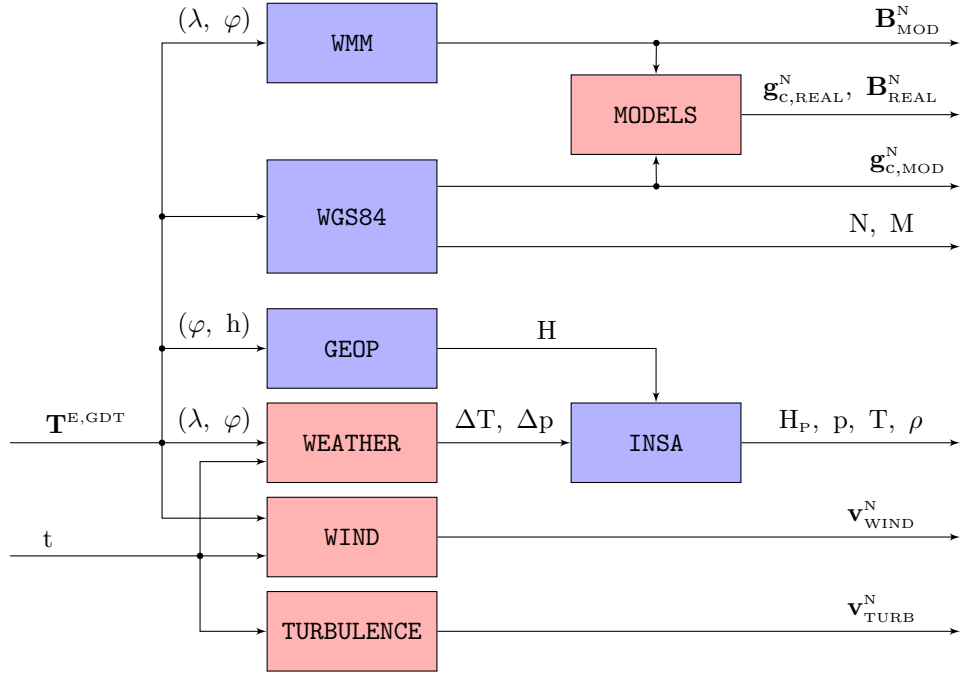
\begin{figure}[h]
\centering
\begin{tikzpicture}[auto, node distance=2cm,>=latex']
	\node [coordinate](xEgdtinput) {};
    \node [coordinate, right of=xEgdtinput, node distance=2cm] (tmp1){};
	\filldraw [black] (tmp1) circle [radius=1pt];
    \node [blockred, right of=tmp1, text width=2cm, node distance=2.5cm] (WEATHER) {\texttt{WEATHER}};	
    \node [block, right of=WEATHER, node distance=4.0cm, text width=2cm] (ISA) {\hypertt{INSA}};
	\node [coordinate, right of=ISA, node distance=4cm](ISAoutput) {};
    \draw [->] (xEgdtinput) -- node[pos=0.5] {\nm{\TEgdt}} (tmp1) -- node {\nm{\lrp{\lambda, \ \varphi}}} (WEATHER);
    \draw [->] (WEATHER) -- node {\nm{\DeltaT, \ \Deltap}} (ISA);
	\draw [->] (ISA) -- node[pos=0.5] {\nm{\Hp, \ p, \ T, \ \rho}} (ISAoutput);
	
    \node [block, above of=WEATHER, node distance=3.0cm, text width=2cm, minimum height=1.5cm] (WGS84) {\hypertt{WGS84}};
	\node [coordinate, above of=tmp1, node distance=3.0cm] (tmp3){};
	\filldraw [black] (tmp3) circle [radius=1pt];
    \draw [->] (tmp1) |- node {} (WGS84);
	
	\node [block, above of=WEATHER, node distance=5.30cm, text width=2cm] (WMM) {\hypertt{WMM}};
	\draw [->] (tmp3) |- node [pos=0.75] {\nm{\lrp{\lambda, \ \varphi}}} (WMM);
		
	\node [block, above of=WEATHER, node distance=1.25cm, text width=2cm] (GEOP) {\texttt{GEOP}};
	\node [coordinate, above of=tmp1, node distance=1.25cm](tmp2) {};
	\filldraw [black] (tmp2) circle [radius=1pt];		
	\draw [->] (tmp2) -- node {\nm{\lrp{\varphi, \ h}}} (GEOP);
	\draw [->] (GEOP.east) -| node[near start] {\nm{H}} (ISA);	
	
	\node [coordinate, right of=WGS84, node distance=8cm](WGS84output) {};
	\node [coordinate, above of=WGS84output, node distance=0.4cm](gNoutput) {};
	\node [coordinate, below of=WGS84output, node distance=0.4cm](NMoutput) {};
	\draw [->] ($(WGS84.east)+(0cm,0.4cm)$) -- node[pos=0.9] {\nm{\gcNMODEL}} (gNoutput);
	\draw [->] ($(WGS84.east)-(0cm,0.4cm)$) -- node[pos=0.85] {\nm{N, \ M}} (NMoutput);
	
    \node [blockred, below of=WEATHER, node distance=1.25cm, text width=2cm] (WIND) {\texttt{WIND}};
    \draw [->] (tmp1) |- node {} ($(WIND.west)+(0cm,0.25cm)$);
	\node [coordinate, right of=WIND, node distance=8cm](vWINDNoutput) {};
	\draw [->] (WIND) -- node[pos=0.85] {\nm{\vWINDN}} (vWINDNoutput);		
    \node [blockred, below of=WIND, node distance=1.25cm, text width=2cm] (TURBULENCE) {\texttt{TURBULENCE}};				
	\node [coordinate, right of=TURBULENCE, node distance=8cm](vTURBNoutput) {};
	\draw [->] (TURBULENCE) -- node[pos=0.85] {\nm{\vTURBN}} (vTURBNoutput);		
	\node [coordinate, below of=xEgdtinput, node distance=1.5cm](tinput) {};				
    \node [coordinate, right of=tinput, node distance=2.5cm] (tmp2){};
	\filldraw [black] (tmp2) circle [radius=1pt];
    \draw [->] (tinput) -- node[pos=0.2] {\nm{t}} (tmp2) |- (TURBULENCE);
    \draw [->] (tmp2) |- node {} ($(WIND.west)-(0cm,0.25cm)$);
	\draw [->] (tmp2) |- node {} ($(WEATHER.west)-(0cm,0.25cm)$);
	
	\node [blockred, above of=ISA, node distance=4.35cm, text width=2cm] (MODELS) {\texttt{MODELS}};
	\coordinate (tmpA) at ($(WGS84.east)+(2.88cm,0.4cm)$) {};
	\filldraw [black] (tmpA) circle [radius=1pt];
	\draw [->] (tmpA) -- (MODELS.south);
	\coordinate (tmpB) at ($(WMM.east)+(2.88cm,0cm)$) {};
	\filldraw [black] (tmpB) circle [radius=1pt];
	\draw [->] (tmpB) -- (MODELS.north);
	\node [coordinate, right of=MODELS, node distance=4cm](MODELSoutput) {};
	\draw [->] (MODELS.east) -- node[pos=0.5] {\nm{\gcNREAL, \ \BNREAL}} (MODELSoutput);	
	
	\node [coordinate, right of=WMM, node distance=8cm](WMMoutput) {};
	\draw [->] (WMM.east) -- node[pos=0.9] {\nm{\BNMODEL}} (WMMoutput);
		
\end{tikzpicture}
\caption{Earth Model flow diagram}
\label{fig:EarthModel_flow_diagram}
\end{figure}
\begin{itemize}
\item The \texttt{WIND} block is similar to the \texttt{WEATHER} one in the sense that it allows the user to use different functions (up to four dimensional tables) to represent the low frequency wind field (\ref{eq:EarthModel_WIND_low_freq}) experienced by the aircraft.

\item The \texttt{TURBULENCE} block implementation relies on a two step process. The first, containing the turbulence generating methodology described in section \ref{sec:EarthModel_WIND} that results in the \nm{\FT} frame vector \nm{\vTURBT}, has been implemented in \texttt{MatLab}\textsuperscript{\textregistered} for a different number of random seeds, tabulated at a very high sample rate, and then saved in text files. The second part, implemented in \texttt{C++}, simply evaluates these tables at the required times, and converts the results to \hypertt{NED}. The implementation has been statistically validated\footnote{An exact validation is not possible as it relies on randomly generated distributions.} against the Dryden model implementation available in Simulink\textsuperscript{\textregistered}.

\item The \hypertt{WMM} block provides the magnetic field models described in section \ref{sec:EarthModel_WMM}. As it is based on preloaded values, its validation is simple.

\item The \texttt{MODELS} block implements the stochastic distortion models described in section \ref{sec:EarthModel_accuracy}. Validation is straightforward as the standard deviation of the outputs should approach the input value as the number of executions increases.
\end{itemize}

The execution of the deterministic blocks (blue in figure \ref{fig:EarthModel_flow_diagram}) needs little explanation, as they are called when required by other modules of the simulation, and always return the same outputs given the same inputs. The behavior of the remaining blocks is explained below on a one by one basis, but in all cases their output varies for each individual simulation run or execution \emph{j}, and hence depend on the flight seed \nm{\seedR}. Refer to section \ref{sec:Sensors_implementation} for a detailed explanation on the generation and meaning of \nm{\seedR}.
\begin{center}
\begin{tabular}{lcc}
	\hline
	Type & Group & Seeds \\
	\hline
	Flight & \nm{\seedR} & \nm{\seedRWEATHER, \, \seedRWIND, \, \seedRMODELS, \, \seedRTURB} \\
	\hline
\end{tabular}
\end{center}
\captionof{table}{Earth model seeds} \label{tab:EarthModel_seeds}

Each execution is initialized with a flight seed \nm{\seedR}, which is the only input to all the stochastic variables and hence it ensures that the results are repeatable if the same seed is employed again. The flight seed \nm{\seedR} is then employed to initialize a discrete uniform distribution, which in addition to the executions described in sections \ref{sec:Sensors_implementation}, \ref{sec:PreFlight_implementation}, and \ref{sec:gc_implementation}, is executed three times to provide the weather seed \nm{\seedRWEATHER}, the wind seed \nm{\seedRWIND}, and the models accuracy seed \nm{\seedRMODELS}. In addition, the turbulence seed \nm{\seedRTURB} is set equal to the trajectory execution number (\nm{\seedRTURB = j}). These seeds hence become the initialization seeds for each of the stochastic Earth model modules described below:

\begin{itemize}
\item The \texttt{WEATHER} block is customizable and enables the use of any two functions that provide the temperature and pressure offsets at mean sea level based on longitude, latitude, and time (\ref{eq:EarthModel_ISA_weather}). These two functions may depend on stochastic parameters so their values vary among the various Monte Carlo simulation runs. If this is the case, standard normal and discrete uniform distributions are initialized with the weather seed \nm{\seedRWEATHER} and realized as necessary to obtain the required parameters.

\item The \texttt{WIND} block is similar to \texttt{WEATHER} as it enables the use of any function that provides the \hypertt{NED} wind as function of longitude, latitude, altitude, and time (\ref{eq:EarthModel_WIND_low_freq}). Similarly to the previous case, this function may be defined stochastically so it varies from execution to execution; in this case, standard normal and discrete uniform distributions are initialized with the wind seed \nm{\seedRWIND} and realized as necessary to obtain the required parameters

\item The \texttt{TURBULENCE} block execution relies on the use of a previously stored text file with precomputed values, as explained above. The turbulence seed (\nm{\seedRTURB = j}) is employed to select which text file to use, and hence ensures that the results vary from execution to execution.

\item The \texttt{MODELS} block provides the differences between the real gravity and magnetic fields, which influence the aircraft motion and the sensor readings, and those estimated by the models loaded on the aircraft systems. As explained in section \ref{sec:EarthModel_accuracy}, these differences do not change during flight. The model accuracy seed \nm{\seedRMODELS} is employed to initialize both a standard normal distribution \nm{N\lrp{0, \, 1}} and a discrete uniform distribution \nm{U\lrp{-179, 180}}. The former is realized five times to obtain the three magnetic deviations of table \ref{tab:EarthModel_magnetism_accuracy} plus the two gravity deviations of table \ref{tab:EarthModel_gravity_accuracy}, while the latter is realized only once to obtain the gravity deviation direction shown in the last row of table \ref{tab:EarthModel_gravity_accuracy}.

\end{itemize}

 \cleardoublepage 
\chapter{Aircraft Model} \label{cha:AircraftModel}

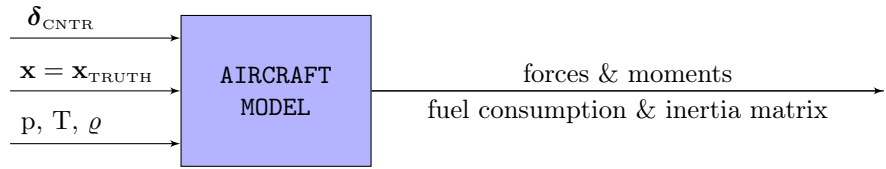
\begin{figure}[h]
\centering
\begin{tikzpicture}[auto, node distance=2cm,>=latex']
	\node [coordinate](xtruthinput) {};
	\node [coordinate, above of=xtruthinput, node distance=0.7cm](deltacntrinput) {};
	\node [coordinate, below of=xtruthinput, node distance=0.7cm](atminput) {};
	\node [block, right of=xtruthinput, minimum width=2.5cm, node distance=3.5cm, align=center, minimum height=2.0cm] (AIRCRAFT MODEL) {\texttt{AIRCRAFT} \\ \texttt{MODEL}};	
	\node [coordinate, right of=AIRCRAFT MODEL, node distance=8.0cm] (output){};
	
	\draw [->] (xtruthinput) -- node[pos=0.45] {\nm{\xvec = \xTRUTH}} (AIRCRAFT MODEL.west);
	\draw [->] (deltacntrinput) -- node[pos=0.3] {\nm{\deltaCNTR}} ($(AIRCRAFT MODEL.west)+(0cm,0.7cm)$);
	\draw [->] (atminput) -- node[pos=0.3] {\nm{p, \, T, \, \varrho}} ($(AIRCRAFT MODEL.west)-(0cm,0.7cm)$);
	\draw [->] (AIRCRAFT MODEL.east) -- node[pos=0.5] {forces \& moments} (output);
	\draw [-] (output) -- node[pos=0.5] {fuel consumption \& inertia matrix} (AIRCRAFT MODEL.east);
\end{tikzpicture}
\caption{Aircraft model flow diagram}
\label{fig:AircraftModel_flow_diagram_generic}
\end{figure}

The aircraft model is a key component of the trajectory computation process as it provides the aerodynamic and propulsive actions (forces and moments) generated by the aircraft while in flight. The role of the aircraft model within the simulation is graphically depicted in figure \ref{fig:AircraftModel_flow_diagram_generic}; its mission is to evaluate the forces and moments generated by the platform (both aerodynamic and propulsive), together with the fuel consumption, the center of mass position, and the inertia matrix\footnote{As the aircraft mass is reduced because of the burnt fuel, the aircraft center of mass and its inertia properties vary slightly, as described in section \ref{sec:AircraftModel_MassInertia}}. The inputs are the position of the throttle and aerodynamic control surfaces \nm{\deltaCNTR} provided by the control system (section \ref{sec:GNC_Control}), various components of the actual aircraft state \nm{\xvec = \xTRUTH} (section \ref{sec:EquationsMotion_AT}), such as the airspeed, angles of attack and sideslip, and aircraft angular velocity, and finally the air pressure, temperature, and density computed by the Earth model in chapter \ref{cha:EarthModel}.

The aircraft is a key component of the simulation and as such it must be modeled with care, as it has a heavy influence in the resulting trajectories. However, as the primary purpose of the simulation is to evaluate different guidance, navigation, and control algorithms, the aircraft is a means and not an end in itself, and there is no real preference for one aircraft or group of aircraft versus another, as long as the selected one is properly modeled and behaves adequately from a dynamic point of view. In other words, it is not necessary for the model to accurately represent an existing aircraft, but its flight dynamics together with its stability and control properties must be as realistic as possible. From the point of view of the simulation it is better to have a detailed performance model of an aircraft that does not exist, than a less accurate one of a real aircraft. For that reason, the author has chosen to virtually design an aircraft with publicly available aerodynamic and propulsion software and qualitatively verify the results, instead of taking an existing aircraft and trying to obtain models that capture its performances.

Each section of this chapter describes a different characteristic of the aircraft: its geometry and mass properties (section \ref{sec:AircraftModel_MassInertia}), the location and orientation of its sensors (section \ref{sec:AircraftModel_Sensor}), the controls with which the guidance and control system influences the aircraft dynamics (section \ref{sec:AircraftModel_Control_Parameters}), and the aerodynamic (section \ref{sec:AircraftModel_aerodynamics}) and propulsion (section \ref{sec:AircraftModel_propulsion}) forces and moments generated by the aircraft. It concludes with a short discussion about the aircraft model implementation and validation (section \ref{sec:AircraftModel_implementation}).
\begin{figure}[!h]
	\centering \includegraphics[width=10cm]{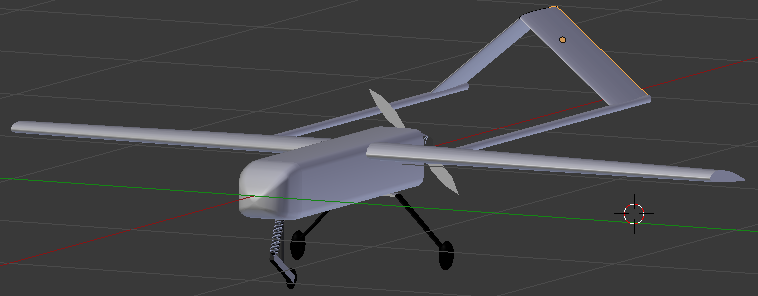}
    \caption{Aircraft overall view}\label{fig:AircraftModel_overall_view}
\end{figure}

\begin{figure}[!h]
	\centering \includegraphics[width=10cm]{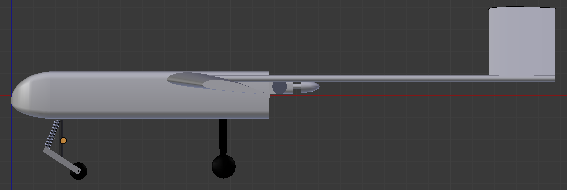}
    \caption{Aircraft left view}\label{fig:AircraftModel_left_view}
\end{figure}

\begin{figure}[!h]
	\centering \includegraphics[width=10cm]{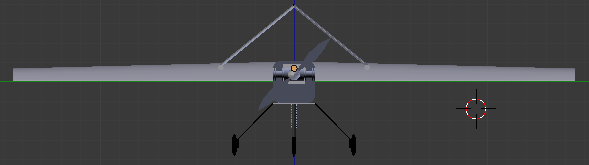}
    \caption{Aircraft front view}\label{fig:AircraftModel_front_view}
\end{figure}

\begin{figure}[!h]
	\centering \includegraphics[width=8cm]{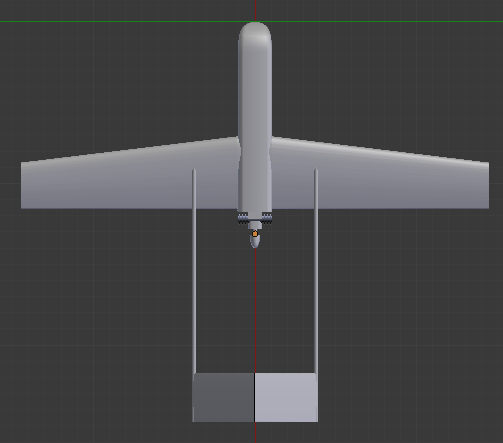}
    \caption{Aircraft top view}\label{fig:AircraftModel_top_view}
\end{figure}

Figure \ref{fig:AircraftModel_overall_view} shows an overall view of the resulting aircraft, equipped with a high wing, pusher propeller, and \nm{\wedge} tail. Additional views are presented in figures \ref{fig:AircraftModel_left_view}, \ref{fig:AircraftModel_front_view}, and \ref{fig:AircraftModel_top_view}. Its main characteristics are shown in table \ref{tab:Aircraft_characteristics}.

\begin{center}
\begin{tabular}{lcrc}
	\hline
	Variable & Symbol & Value & Unit \\
	\hline 
	Total mass (with fuel) 		& \nm{m_{full}}  & 19.715 	& kg \\
	Total mass (without fuel) 	& \nm{m_{empty}} & 17.835 	& kg \\
	Wing area   				& S              & 0.878 	& \nm{m^2} \\
	Wing chord	 				& c              & 0.359 	& m \\
	Wing span 					& b              & 2.680 	& m \\
	Aircraft length 			& l              & 2.320 	& m \\
	\hline
\end{tabular}
\end{center}
\captionof{table}{Aircraft characteristics}\label{tab:Aircraft_characteristics}	


\section{Mass and Inertia Analysis}\label{sec:AircraftModel_MassInertia}

The aircraft is assumed to be a rigid body with lateral symmetry in both mass and geometry [A\ref{as:APM_rigid_body},A\ref{as:APM_symmetric}]. Each of the aircraft main components (wings, tail fins, tail booms, fuselage, fuel tank, fuel, engine, ballast, forward landing gear, main landing gear, and avionics) has been assigned a mass and a position for its center of mass. The total mass of the aircraft shown in table \ref{tab:Aircraft_characteristics} is obtained by adding together those of its components, with the aircraft center of mass \nm{\OB}\footnote{The aircraft center of mass \nm{\OB} is defined in section \ref{subsec:RefSystems_B} as the origin of the body frame \nm{\FB}.} located at the aircraft plane of symmetry \nm{207 \ mm} forward and \nm{5 \ mm} above the reference point \nm{\OR}\footnote{The reference point \nm{\OR} is defined in section \ref{subsec:RefSystems_R} as the origin of the reference frame \nm{\FR}.} when the fuel tank is full, and \nm{219 \ mm} forward \nm{6 \ mm} above when it is empty, as depicted in figure \ref{fig:Sensor_symmetry_plane}. This results in:
\begin{eqnarray}
\nm{\TRBRfull}   & = & \nm{\begin{bmatrix} 0.207 & 0 & -0.005 \end{bmatrix}^T \ m} \label{eq:AircraftModel_Trbb_full} \\
\nm{\TRBRempty}  & = & \nm{\begin{bmatrix} 0.219 & 0 & -0.006 \end{bmatrix}^T \ m} \label{eq:AircraftModel_Trbb_empty}
\end{eqnarray}

For the inertia analysis, each of the aircraft components has been assigned a simple shape to estimate its moments of inertia (\nm{I_{x}, I_{y}, I_{z}}) with respect to its own center of mass \cite{Goldstein2002}. In addition, the products of inertia (\nm{J_{xz}}) of the wings and tail fins have also been estimated\footnote{The products of inertia of the remaining components are all zero as they are modeled possessing lateral and vertical symmetry, while \nm{J_{xy}} and \nm{J_{yz}} for the wings and tail fins are also zero because of their lateral symmetry.}.

The inertia properties of all components are added together to obtain the aircraft inertia matrix with respect to the aircraft center of mass:
\begin{eqnarray}
\nm{\vec I_{full}^{\sss B}} & = & \nm{\begin{bmatrix} \nm{I_{x}} & 0 & \nm{- J_{xz}} \\ 0 & \nm{I_{y}} & 0 \\ \nm{- J_{xz}} & 0 & \nm{I_{z}} \end{bmatrix}_{full} \ \ \  
			  = \begin{bmatrix} 2.202 & 0 & -0.191 \\ 0 & 3.462 & 0 \\ -0.191 & 0 & 5.490 \end{bmatrix} \ kg \ m^2} \label{eq:AircraftModel_Ifull} \\
\nm{\vec I_{empty}^{\sss B}} & = & \nm{\begin{bmatrix} \nm{I_{x}} & 0 & \nm{- J_{xz}} \\ 0 & \nm{I_{y}} & 0 \\ \nm{- J_{xz}} & 0 & \nm{I_{z}} \end{bmatrix}_{empty} 
			  = \begin{bmatrix} 2.198 & 0 & -0.192 \\ 0 & 3.430 & 0 \\ -0.192 & 0 & 5.458 \end{bmatrix} \ kg \ m^2} \label{eq:AircraftModel_Iempty}
\end{eqnarray}

Because of the little difference in the center of mass position and inertia matrix between the empty and full tank configurations, linear interpolation can be employed to estimate their value based on the mass [A\ref{as:APM_center_mass}, A\ref{as:APM_inertia}]:
\begin{eqnarray}
\nm{\TRBR} & = & \nm{\TRBB = \vec f_T\lrp{m} = \TRBRfull + \dfrac{m_{full} - m}{m_{full} - m_{empty}} \; \lrp{\TRBRempty - \TRBRfull}} \label{eq:AircraftModel_Trbb} \\
\nm{\vec I^{\sss B}} & = & \nm{\vec f_I\lrp{m} = \vec I_{full}^{\sss B} + \dfrac{m_{full} - m}{m_{full} - m_{empty}} \; \lrp{\vec I_{empty}^{\sss B} - \vec I_{full}^{\sss B}}} \label{eq:AircraftModel_I}
\end{eqnarray}


\section{Sensor Location and Orientation}\label{sec:AircraftModel_Sensor}

Chapter \ref{cha:Sensors} describes the different sensors with which the aircraft is equipped to assist with its navigation and control. Each has its own requirements related to how they should be mounted in or outside the aircraft:
\begin{itemize}
\item The magnetometer triad, as described in section \ref{subsec:Sensors_NonInertial_Magnetometers}, shall be located within the aircraft as far as possible from the aircraft magnets, electrical currents, and ferrous materials; once properly mounted, and assisted by their calibration or swinging process (section \ref{sec:PreFlight_swinging}), the magnetometers provide measurements of the Earth magnetic field viewed in \nm{\FB} (\nm{\BBtilde}) that are independent of their pose with respect to \nm{\FB}.

\item The air data sensors, described in section \ref{subsec:Sensors_NonInertial_ADS}, can be located anywhere outside the aircraft as long as they have a clear view of the undisturbed air stream, and their readings are not influenced by their pose with respect to \nm{\FB}.

\item For simplifications purposes, the \hypertt{GNSS} receiver modeled in section \ref{subsec:Sensors_NonInertial_GNSS} is assumed to directly provide an estimation of the aircraft center of mass position instead of that of the receiver itself [A\ref{as:APM_GNSS}].

\item The accelerometers and gyroscopes are described in section \ref{sec:Sensors_Inertial}. They are intended to measure the specific force and angular velocity (sections \ref{subsubsec:EquationsMotion_specific_force} and \ref{sec:EquationsMotion_velocity}) of the body frame \nm{\FB} with respect to the inertial frame \nm{\FI} when viewed in \nm{\FB} (\nm{\fIBBtilde} and \nm{\wIBBtilde}), but in fact they measure those of the platform frame \nm{\FP} and also view them in \nm{\FP} (\nm{\fIPPtilde} and \nm{\wIPPtilde}). Given their importance for inertial navigation, it is indispensable to determine the relative pose between \nm{\FB} and \nm{\FP}, which is described in detail in section \ref{subsec:Sensors_Inertial_Mounting}.

\item The digital camera described in section \ref{sec:Sensors_camera} is located on the lower part of the fuselage facing down, and its field of view can not be blocked by the landing gear or any other aircraft elements. The resulting images are hence viewed in the camera frame \nm{\FC}, so the relative pose between \nm{\FB} and \nm{\FC} is indispensable to visual navigation (section \ref{sec:PreFlight_camera_frame}).
\end{itemize}


\section{Control Parameters}\label{sec:AircraftModel_Control_Parameters}

The control parameters are the different means at the disposal of the guidance and control system to influence the aircraft forces and moments, and hence control the aircraft attitude and kinematics. Its number (4) coincides with the degrees of freedom in the equations of motion, and are the following:
\begin{itemize}
\item \emph{Throttle parameter} \nm{\deltaT} [-] models the position of the throttle lever, with 1 meaning maximum power or lever fully forward and 0 meaning idle.
\item \emph{Elevator parameter} \nm{\deltaE \lrsb{^{\circ}}} represents the elevator\footnote{In a conventional aircraft, the elevator is composed by two surfaces located at the trailing edge of each horizontal stabilizer that move in the same direction to introduce pitching moments.} deflection, with positive meaning horns forward, elevator down, and negative moment\footnote{The elevator behaves differently from the other controls in the sense that a negative deflection creates a positive moment.} along \nm{\iBii}.
\item \emph{Ailerons parameter} \nm{\deltaA \lrsb{^{\circ}}} represents the ailerons\footnote{In a conventional aircraft, the ailerons are two surfaces located at the trailing edge of both wings that move in opposite directions to introduce rolling moments.} deflection, with positive meaning horns to the right, right aileron up, left aileron down, and positive moment along \nm{\iBi}.
\item \emph{Rudder parameter} \nm{\deltaR \lrsb{^{\circ}}} represents the rudder\footnote{In a conventional aircraft, the rudder is a moving surface located at the trailing edge of the vertical stabilizer that moves to introduce a yaw moment.} deflection, with positive meaning right pedal down, rudder to the right, and positive moment along \nm{\iBiii}.
\end{itemize}

They can be grouped into the control parameter vector:
\neweq{\deltaCNTR = \lrsb{\deltaT \ \ \deltaE \ \ \deltaA \ \ \deltaR}^T} {eq:AircraftModel_deltaCNTR}

Because of its \nm{\wedge} configuration, the modeled aircraft lacks a conventional elevator or rudder, but instead possesses a moving control surface at the trailing edge of each of the two tail fins. If they are called \nm{\delta_{\sss LF}} and \nm{\delta_{\sss RF} \lrsb{^{\circ}}} for left and right fins respectively, with positive meaning surface deflection down, the elevator effect is achieved when both tail surfaces move in the same direction, while the rudder effect relies on both tail surfaces moving in opposite directions. 
\begin{eqnarray}
\nm{\deltaE} & = & \nm{0.5 \ \lrp{\delta_{\sss LF} + \delta_{\sss RF}}} \label{eq:AircraftModel_deltaE} \\
\nm{\deltaR} & = & \nm{0.5 \ \lrp{\delta_{\sss LF} - \delta_{\sss RF}}} \label{eq:AircraftModel_deltaR} 
\end{eqnarray}

Based on the above expressions, any combination of elevator and rudder can be achieved with proper movements of the left and right tail fins control parameters. From this point on, the tail fin surfaces are not mentioned again in this document, which relies on the conventional parameters. Wherever it is necessary, the elevator and rudder positions are always internally converted into left and right fin control positions to obtain their aerodynamic effects.


\section{Aerodynamics}\label{sec:AircraftModel_aerodynamics}

The objective of this section is to obtain realistic expressions for the aerodynamic actions (forces and moments) generated by the aircraft during flight, which are a key part of the equations of motion required to compute the aircraft trajectory. When expressed in the body frame \nm{F_{\sss B}}, they are called the \emph{longitudinal} (\nm{F_x}), \emph{lateral} (\nm{F_y}), and \emph{vertical} (\nm{F_z}) aerodynamic \emph{forces}, and the \emph{rolling} (\nm{M_l}), \emph{pitching} (\nm{M_m}), and \emph{yawing} (\nm{M_n}) \emph{moments}, respectively \cite{Stevens2003,Etkin1972,Ashley1974,Miele1962,Katz2001,Anderson2001,Drela2014}.
\begin{eqnarray}
\nm{\FAERB} & = & \nm{\lrsb{F_x \ \ F_y \ \ F_z}^T} \label{eq:AircraftModel_FaerB} \\
\nm{\MAERB} & = & \nm{\lrsb{M_l \ \ M_m \ \ M_n}^T} \label{eq:AircraftModel_MaerB} 
\end{eqnarray}

The aerodynamic forces are however historically viewed in the wind frame \nm{F_{\sss W}} (section \ref{subsec:RefSystems_W}), where (with the sign inverted) they are called \emph{drag} (\nm{D}), \emph{side force} (\nm{Y}), and \emph{lift} (\nm{L}), respectively.
\neweq{\FAERW = \vec g_{\ds{\vec \zeta_{{\sss WB}*}}} \big(\FAERB\big) = \lrsb{-D \ \ -Y \ \ -L}^T} {eq:AircraftModel_FaerW}
\input{scripts/ch03_acft/AircraftModel_forces}

Figure \ref{fig:AircraftModel_forces} shows an schematic aircraft with the body and wind frames depicted in figure \ref{fig:RefSystems_W2B}, highlighting the aerodynamic control surfaces and their associated control parameters in different colors, together with the induced aerodynamic forces and moments.

The process employed to obtain the aerodynamic model is the following:
\begin{enumerate}
\item Use dimensional analysis to obtain the true physical dependencies based on dimensionless parameters.
\item Use fluid dynamics analysis software to obtain the aerodynamic actions for different aerodynamic configurations (combinations of the dimensionless parameters).
\item Construct equispaced tables based on the results from the previous step.
\item Evaluate the table when requested by the trajectory computation process.
\item Correct for the aircraft center of mass variation caused by the fuel burn.
\end{enumerate}


\subsection{Dimensional Analysis}\label{subsec:AircraftModel_aerodynamics_dimensional}

The Buckingham \nm{\pi} theorem states that any physically meaningful equation involving a certain number of n physical variables (with dimensions) can be rewritten without loss of information in terms of a reduced set of \emph{p} dimensionless parameters built from the original variables such that \nm{p = n - k}, where \emph{k} is the number of physical dimensions involved \cite{Szirtes1997}. The dimensionless form is always preferred when modeling a physical process as it contains the true physical dependencies with a minimum of complexity. 

The aerodynamic actions are generated because of the interaction between the aircraft exterior structure and the air that flows around it. The physical variables involved are hence the following:
\begin{itemize}
\item The aerodynamic forces (\nm{F_x, \, F_y, \, F_z}) and moments (\nm{M_l, \, M_m, \, M_n}) being modeled.
\item The characteristics of the air flow, such as the pressure \emph{p}, temperature T, density \nm{\rho}, adiabatic index \nm{\kappa}, specific air constant R, viscosity \nm{\mu_v}, and relative humidity \nm{R_H}.
\item The aircraft dimensions, defined by its wing surface S, wing chord \emph{c}, wing span \emph{b}, and aircraft length \emph{l}.
\item The aircraft geometry, defined by a long array of dimensionless parameters \nm{p_k} describing the aircraft exterior shape, together with the control parameters (\nm{\deltaE, \, \deltaA, \, \deltaR}) and their variation with time (\nm{\dot{\delta}_{\sss E}, \, \dot{\delta}_{\sss A}, \, \dot{\delta}_{\sss R}}).
\item The interaction between the aircraft and the air flow, defined by the airspeed \nm{\vtas}, the angle of attack \nm{\alpha}, the angle of sideslip \nm{\beta}, their variations with time \nm{\dot{v}_{\sss {TAS}}, \, \dot{\alpha}, \, and \, \dot{\beta}}, plus the three components (\nm{p, \, q, \, r}) of the aircraft angular speed \nm{\wNBB} (section \ref{sec:EquationsMotion_velocity}).
\end{itemize}

As the number of physical dimensions is \nm{k = 4} (length, mass, time, and temperature), four physical variables (wing surface S, air pressure \emph{p}, air temperature T, and specific air constant R) can be eliminated by employing them to convert all others into dimensionless parameters \cite{Gallo2012}. This results in:
\neweq{\lrsb{\dfrac{\FAERB}{p \; S} \ \ \dfrac{\MAERB}{p \; S^{\frac{3}{2}}}}^T = \vec f_{A,1} \lrp{
\begin{array}{l}
\nm{\dfrac{\rho \, R \, T}{p}, \kappa, \dfrac{\mu_v \, \sqrt{R \, T}}{p \, \sqrt{S}}, R_H, \dfrac{c}{\sqrt{S}}, \dfrac{b}{\sqrt{S}}, \dfrac{l}{\sqrt{S}}, p_k, \deltaE, \deltaA, \deltaR, \dfrac{\dot{\delta}_{\sss E} \, \sqrt{S}}{\sqrt{R\,T}}, \dfrac{\dot{\delta}_{\sss A} \, \sqrt{S}}{\sqrt{R\,T}},} \\ 
\nm{\dfrac{\dot{\delta}_{\sss R} \, \sqrt{S}}{\sqrt{R\,T}}, \dfrac{\vtas}{\sqrt{R\,T}}, \alpha, \beta, \dfrac{\dot{v}_{\sss TAS}\,\sqrt{S}}{R\,T}, \dfrac{\dot{\alpha} \, \sqrt{S}}{\sqrt{R\,T}}, \dfrac{\dot{\beta} \, \sqrt{S}}{\sqrt{R\,T}}, \dfrac{p \, \sqrt{S}}{\sqrt{R\,T}}, \dfrac{q \, \sqrt{S}}{\sqrt{R\,T}}, \dfrac{r \, \sqrt{S}}{\sqrt{R\,T}}}
\end{array}
}}{eq:AircraftModel_dimensional_analysis}

The assumptions of perfect gas [A\ref{as:ISA_perfect_gas}], no air humidity [A\ref{as:ISA_humidity}], quasi-stationary flow\footnote{Quasi stationary means that the forces and moments depend on certain variables but not on their time derivatives. A minor exception is made by not discarding the influence of the angular velocity.} [A\ref{as:APM_quasi_stationary}], and no viscosity [A\ref{as:APM_viscosity}], together with the known aircraft geometry, enable the elimination of many parameters:
\neweq{\lrsb{\dfrac{\FAERB}{p \; S} \ \ \dfrac{\MAERB}{p \; S^{\frac{3}{2}}}}^T = \vec f_{A,2} \lrp{\kappa, \, \deltaE, \, \deltaA, \, \deltaR, \, \dfrac{\vtas}{\sqrt{R\,T}}, \, \alpha, \, \beta, \, \dfrac{p \, \sqrt{S}}{\sqrt{R\,T}}, \, \dfrac{q \, \sqrt{S}}{\sqrt{R\,T}}, \, \dfrac{r \, \sqrt{S}}{\sqrt{R\,T}}}}{eq:AircraftModel_dimensional_analysis2}

Several factors enable a further simplification to (\ref{eq:AircraftModel_dimensional_analysis2}). First is that any dimensionless parameter can be multiplied or divided by any combination of other dimensionless parameters that results in no change of dimensions \cite{Szirtes1997}. Second is the assumption of incompressible air flow [A\ref{as:APM_incompressible}], which means that the Mach number \nm{M = \vtas / \sqrt{\kappa \, R \, T}} can be considered zero. And third is classic aerodynamic theory \cite{Etkin1972,Ashley1974,Miele1962}, that states that the aerodynamic actions are proportional to the air adiabatic index \nm{\kappa}, to the air pressure \emph{p}, and to the square of the Mach number\footnote{Note that \nm{\kappa \, p \, M^2} is equal to \nm{\rho \, v_{tas}^2}.}. This results in the definition of the aerodynamic force and moment coefficients \cite{Stevens2003,Etkin1972,Ashley1974,Miele1962,Katz2001,Anderson2001,Drela2014}:
\begin{eqnarray}
\nm{\FAERB} & = & \nm{\lrsb{F_x \ \ F_y \ \ F_z}^T \ \ \ = \frac{1}{2} \; \rho \; S \; v_{\sss TAS}^2 \; \lrsb{C_{Fx} \ \ C_{Fy} \ \ C_{Fz}}^T}\label{eq:AircraftModel_CFaerB} \\
\nm{\MAERB} & = & \nm{\lrsb{M_l \ \ M_m \ \ M_n}^T = \frac{1}{2} \; \rho \; S \; v_{\sss TAS}^2 \; \lrsb{b \; C_{Ml} \ \ c \; C_{Mm} \ \ b \; C_{Mn}}^T} \label{eq:AircraftModel_CMaerB} 
\end{eqnarray}

where
\neweq{\lrsb{C_{Fx} \ \ C_{Fy} \ \ C_{Fz} \ \ C_{Ml} \ \ C_{Mm} \ \ C_{Mn}}^T = \vec f_{A,3} \lrp{\deltaE, \, \deltaA, \, \deltaR, \, \alpha, \, \beta, \, \dfrac{p \, b}{2 \, \vtas}, \, \dfrac{q \, c}{2 \ \vtas}, \, \dfrac{r \, b}{2 \ \vtas}}}{eq:AircraftModel_coefficientsB}

The final expression results from the use of dimensionless angular speed parameters (p', q', r'):
\neweq{\lrsb{C_{Fx} \ \, C_{Fy} \ \ C_{Fz} \ \ C_{Ml} \ \ C_{Mm} \ \ C_{Mn}}^T = \vec f_{A,4} \lrp{\deltaE, \, \deltaA, \, \deltaR, \, \alpha, \, \beta, \, p', \, q', \, r'}}{eq:AircraftModel_coefficientsC}

Equation (\ref{eq:AircraftModel_coefficientsC}) contains the simplest form of the physical dependencies involved in the generation of the aerodynamic forces and moments. The drag, side, and lift coefficients are also defined in the same way:
\neweq{\FAERW = - \lrsb{D \ \, Y \ \, L}^T = - \frac{1}{2} \; \rho \; S \; v_{\sss TAS}^2 \; \lrsb{C_D \ \ C_Y \ \ C_L}^T}{eq:AircraftModel_CFaerW} 


\subsection{Fluid Dynamics Analysis}\label{subsec:AircraftModel_aerodynamics_fluid_dynamics}

The Vortex Lattice Method (\hypertt{VLM}) is a numerical method used in computational fluid dynamics at the early stages of aircraft design. It models the aircraft lifting surfaces as infinitely thin sheets of discrete vortices to compute lift and induced drag for each vortex, discarding the influence of compressibility and viscosity. The forces and moments are then obtained by adding together the lift and induced drag of each vortex \cite{Katz2001,Anderson2001,Drela2014}. Athena Vortex Lattice (\hypertt{AVL}) is a software for the aerodynamic analysis of rigid aircraft of arbitrary configuration \cite{AVL}. It employs an extended vortex lattice model for the lifting surfaces combined with a slender body model for fuselages. It is best suited for aerodynamic configurations that consist mainly of thin lifting surfaces at small angles of attack and sideslip. These surfaces and their trailing wakes are represented as single layer vortex sheets, discretized into horseshoe vortex filaments whose trailing legs are assumed to be parallel to the \nm{\iBi} axis \cite{Ashley1985}.

The aircraft exterior shape has been modeled in detail to facilitate the use of \hypertt{AVL}. With the exception of the fuselage, which has been modeled as a slender body, the remaining surfaces (wings, tail fins, ailerons, tail control surfaces) have been modeled by means of sixty-four strips or aerodynamic profiles\footnote{The airfoil employed for the wings is \texttt{NACA4415}, while that of the fins is \texttt{NACA0005}.}, which translate into a total of 620 vortices. Each execution of \hypertt{AVL} requires as input the value of the eight independent parameters shown at the right hand side of (\ref{eq:AircraftModel_coefficientsC}) and returns as output not only the force and moment coefficients but also their derivatives with respect to the input parameters.

The aerodynamic model relies on multi dimensional tables constructed by repeated executions of \hypertt{AVL}. The approach taken is slightly different for the longitudinal actions than for the lateral ones, as explained in the following sections.


\subsection{Longitudinal Forces and Moments}\label{subsec:AircraftModel_aerodynamics_long}

This section describes the aerodynamic forces and moments whose effect resides mostly within the aircraft longitudinal plane: the longitudinal  (\nm{F_x}) and vertical (\nm{F_z}) forces and the pitch moment (\nm{M_m}). Given that there are eight independent parameters in (\ref{eq:AircraftModel_coefficientsC}), it is imperative to select the most important ones in order to tabulate them while linearizing the influence of the remaining ones. In the case of the longitudinal actions, the three parameters with a bigger influence are the angles of attack \nm{\alpha} and sideslip \nm{\beta}, together with the elevator deflection \nm{\deltaE}. As shown in (\ref{eq:AircraftModel_longitudinal}), each longitudinal coefficient (\nm{C_{lon}} stands for \nm{C_{Fx}}, \nm{C_{Fz}}, or \nm{C_{Mm}}) can be modeled by the sum of six tridimensional tables (the independent parameters are \nm{\alpha}, \nm{\beta}, and \nm{\deltaE}) evaluated considering all the remaining parameters as zero (\nm{\vec\delta_0 \rightarrow \deltaA = \deltaR = p' = q' = r' = 0}).
\begin{eqnarray}
\nm{C_{lon}} & = & \nm{C_{lon} \big|_{\sss {\vec\delta}_0} \lrp{\alpha, \, \beta, \, \deltaE} + 
\pderpar{C_{lon}}{\deltaA} \bigg|_{\sss {\vec\delta}_0} \! \lrp{\alpha, \, \beta, \, \deltaE} \cdot \deltaA +
\pderpar{C_{lon}}{\deltaR} \bigg|_{\sss {\vec\delta}_0} \! \lrp{\alpha, \, \beta, \, \deltaE} \cdot \deltaR} \nonumber \\
& & \nm{+ \pderpar{C_{lon}}{p'} \bigg|_{\sss {\vec\delta}_0} \! \lrp{\alpha, \, \beta, \, \deltaE} \cdot p' +
\pderpar{C_{lon}}{q'} \bigg|_{\sss {\vec\delta}_0} \! \lrp{\alpha, \, \beta, \, \deltaE} \cdot q' +
\pderpar{C_{lon}}{r'} \bigg|_{\sss {\vec\delta}_0} \! \lrp{\alpha, \, \beta, \, \deltaE} \cdot r'} \label{eq:AircraftModel_longitudinal}
\end{eqnarray}

The tables that make up the longitudinal actions have been constructed by means of a total of 360 executions of \hypertt{AVL} with the inputs shown in table \ref{tab:Aircraft_AVL_executions_longitudinal}:
\begin{center}
\begin{tabular}{lcccc}
	\hline
	Input & & Unit & Range & \# \\
	\hline
	Angle of attack 	& \nm{\alpha}	& \nm{^{\circ}} & \nm{-5.0, -2.5, \dots, 10.0, 12.5}	& 8 \\
	Angle of sideslip	& \nm{\beta}	& \nm{^{\circ}} & \nm{-10.0, -7.5, \dots, 7.5, 10.0}	& 9 \\
	Elevator			& \nm{\deltaE}	& \nm{^{\circ}} & \nm{-8.0, -4.0, 0, 4.0, 8.0}			& 5 \\
	\hline
\end{tabular}
\end{center}
\captionof{table}{Inputs to \texttt{AVL} for longitudinal model generation}\label{tab:Aircraft_AVL_executions_longitudinal}	

Once constructed, the integration of the equations of motion to obtain the aircraft trajectory requires the continuous evaluation of these tables to obtain the aerodynamic forces and moments. Although the tables have been constructed with equispaced independent variables to avoid the costly binary searches required otherwise, it is imperative to select an evaluation algorithm that provides a good balance between accuracy and speed because of the high frequency at which the tables are evaluated. Another important factor to take into consideration is the continuity and smoothness characteristics of the solution as well as those of the first derivatives, as discontinuities may cause convergence problems with both the trajectory integrator and the navigation filter.

The selected method is \emph{biparabolic interpolation}, which provides continuous and smooth results with a first derivative that is also continuous, although not smooth at the nodes. Although described next for one dimension, the required tridimensional process is obtained by a series of one dimensional interpolations, each based on the results of the previous one. Biparabolic interpolation relies on four table points bracketing the input (two above and two below) and executes two different parabolic interpolations (each based on three points), one discarding the first point and another discarding the last, and then linearly combines the results.
\begin{figure}[h]
\centering
\begin{tikzpicture}
\begin{axis}[
cycle list={{blue,no markers},{red,no markers},{orange!50!yellow, no markers},{violet, no markers},{green, no markers}},
width=8.0cm,
xmin=-5, xmax=12.5, xtick={-5,-2.5,...,12.5},
xlabel={\nm{\alpha \left[^{\circ}\right]}},
xmajorgrids,
ymin=0, ymax=1.5, ytick={0,0.25,...,1.5},
ylabel={\nm{C_L \left[-\right]}},
ymajorgrids,
axis lines=left,
axis line style={-stealth},
legend entries={\nm{\deltaE = - 8^{\circ}},\nm{\deltaE = - 4^{\circ}}, \nm{\deltaE = + 0^{\circ}}, \nm{\deltaE = + 4^{\circ}}, \nm{\deltaE = + 8^{\circ}}},
legend pos=south east,
legend style={font=\footnotesize},
legend cell align=left,
]
\pgfplotstableread{figs/ch03_acft/test1_cl_vs_alpha__deltaE.txt}\mytable
\addplot table [header=false, x index=0,y index=1] {\mytable};
\addplot table [header=false, x index=0,y index=2] {\mytable};
\addplot table [header=false, x index=0,y index=3] {\mytable};
\addplot table [header=false, x index=0,y index=4] {\mytable};
\addplot table [header=false, x index=0,y index=5] {\mytable};
\draw [-stealth, very thick] (9.2,0.98) -- (6.7,1.22);
\path node at (6.2,1.18) {\nm{\deltaE}};
\path node [draw, shape=rectangle, fill=white] at (+1.2,+1.35) {\footnotesize \nm{\beta = \deltaA = \deltaR = p' = q' = r' = 0}};
\end{axis}	
\end{tikzpicture}%
\hskip 10pt
\begin{tikzpicture}
\begin{axis}[
cycle list={{blue,no markers},{red,no markers},{orange!50!yellow, no markers},{violet, no markers},{green, no markers}},
width=8.0cm,
xmin=-5, xmax=12.5, xtick={-5,-2.5,...,12.5},
xlabel={\nm{\alpha \left[^{\circ}\right]}},
xmajorgrids,
ymin=0, ymax=1.5, ytick={0,0.25,...,1.5},
ylabel={\nm{C_L \left[-\right]}},
ymajorgrids,
axis lines=left,
axis line style={-stealth},
legend entries={\nm{|\beta| = \ \, 0.0^{\circ}},\nm{|\beta| = \ \, 2.5^{\circ}}, \nm{|\beta| = \ \, 5.0^{\circ}}, \nm{|\beta| = \ \, 7.5^{\circ}}, \nm{|\beta| = 10.0^{\circ}}},
legend pos=south east,
legend style={font=\footnotesize},
legend cell align=left,
]
\pgfplotstableread{figs/ch03_acft/test2_cl_vs_alpha__beta.txt}\mytable
\addplot table [header=false, x index=0,y index=1] {\mytable};
\addplot table [header=false, x index=0,y index=2] {\mytable};
\addplot table [header=false, x index=0,y index=3] {\mytable};
\addplot table [header=false, x index=0,y index=4] {\mytable};
\addplot table [header=false, x index=0,y index=5] {\mytable};
\draw [-stealth, very thick] (8.0,1.27) -- (9.9,1.1);
\path node at (8.8,1.05) {\nm{|\beta|}};
\path node [draw, shape=rectangle, fill=white] at (+1.3,+1.35) {\footnotesize \nm{\deltaE = \deltaA = \deltaR = p' = q' = r' = 0}};
\end{axis}		
\end{tikzpicture}%
\caption{\nm{C_L} versus \nm{\alpha} for various \nm{\deltaE} and \nm{\beta}}
\label{fig:aircraft_cl_vs_alpha}
\end{figure}

The last step is to verify that the results obtained when evaluating the tables are realistic and representative of the way a properly designed aircraft behaves. The first validity check of any aerodynamic model is the linearity of the lift coefficient \nm{C_L} with the angle of attack \nm{\alpha} \cite{Stevens2003,Etkin1972,Miele1962}, as shown in figure \ref{fig:aircraft_cl_vs_alpha}. As the lift force L is proportional to the square of the airspeed \nm{\vtas} (\ref{eq:AircraftModel_CFaerW}), it is possible to generate the same lift (for example to fly at the same pressure altitude \nm{\Hp}) by flying quicker at a smaller angle of attack or slower at a bigger one. Figure \ref{fig:aircraft_cl_vs_alpha} also shows that \nm{C_L} grows with the elevator deflection \nm{\deltaE} because of the added lift from the horizontal stabilizer, and decreases slightly with the sideslip \nm{\beta} because of the diminished flow symmetry \cite{Stevens2003,Etkin1972,Miele1962}. 
\begin{figure}[h]
\centering
\begin{tikzpicture}
\begin{axis}[
cycle list={{blue,no markers},{red,no markers},{orange!50!yellow, no markers},{violet, no markers},{green, no markers}},
width=8.0cm, 
xmin=0, xmax=1.5, xtick={0,0.25,...,1.5},
xlabel={\nm{C_L \left[-\right]}},
xmajorgrids,
ymin=0.12, ymax=0.26, ytick={0.12,0.14,0.16,0.18,0.20,0.22,0.24,0.26},
ylabel={\nm{C_D \left[-\right]}},
ymajorgrids,
axis lines=left,
axis line style={-stealth},
legend entries={\nm{\deltaE = - 8^{\circ}},\nm{\deltaE = - 4^{\circ}}, \nm{\deltaE = + 0^{\circ}}, \nm{\deltaE = + 4^{\circ}}, \nm{\deltaE = + 8^{\circ}}},
legend pos=north west,
legend style={font=\footnotesize},
legend cell align=left,
]
\pgfplotstableread{figs/ch03_acft/test1_cd_vs_cl__deltaE.txt}\mytable
\addplot table [header=false, x index=0,y index=1] {\mytable};
\addplot table [header=false, x index=2,y index=3] {\mytable};
\addplot table [header=false, x index=4,y index=5] {\mytable};
\addplot table [header=false, x index=6,y index=7] {\mytable};
\addplot table [header=false, x index=8,y index=9] {\mytable};
\draw [-stealth, very thick] (0.47,0.16) -- (0.74,0.13);
\path node at (0.65,0.13) {\nm{\deltaE}};
\path node [draw, shape=rectangle, fill=white, align=center] at (+1.20,+0.135) {\footnotesize \nm{\beta = \deltaA = \deltaR = 0}\\[-3pt] \footnotesize \nm{p' = q' = r' = 0}};
\end{axis}	
\end{tikzpicture}%
\hskip 10pt
\begin{tikzpicture}
\begin{axis}[
cycle list={{blue,no markers},{red,no markers},{orange!50!yellow, no markers},{violet, no markers},{green, no markers}},
width=8.0cm,
xmin=0, xmax=1.5, xtick={0,0.25,...,1.5},
xlabel={\nm{C_L \left[-\right]}},
xmajorgrids,
ymin=0.12, ymax=0.26, ytick={0.12,0.14,0.16,0.18,0.20,0.22,0.24,0.26},
ylabel={\nm{C_D \left[-\right]}},
ymajorgrids,
axis lines=left,
axis line style={-stealth},
legend entries={\nm{|\beta| = \ \, 0.0^{\circ}},\nm{|\beta| = \ \, 2.5^{\circ}}, \nm{|\beta| = \ \, 5.0^{\circ}}, \nm{|\beta| = \ \, 7.5^{\circ}}, \nm{|\beta| = 10.0^{\circ}}},
legend pos=north west,
legend style={font=\footnotesize},
legend cell align=left,
]
\pgfplotstableread{figs/ch03_acft/test2_cd_vs_cl__beta.txt}\mytable
\addplot table [header=false, x index=0,y index=1] {\mytable};
\addplot table [header=false, x index=2,y index=3] {\mytable};
\addplot table [header=false, x index=4,y index=5] {\mytable};
\addplot table [header=false, x index=6,y index=7] {\mytable};
\addplot table [header=false, x index=8,y index=9] {\mytable};
\draw [-stealth, very thick] (0.7,0.135) -- (0.55,0.155);
\path node at (0.5,0.148) {\nm{|\beta|}};
\path node [draw, shape=rectangle, fill=white, align=center] at (+1.20,+0.135) {\footnotesize \nm{\deltaE = \deltaA = \deltaR = 0}\\[-3pt] \footnotesize \nm{p' = q' = r' = 0}};
\end{axis}		
\end{tikzpicture}%
\caption{\nm{C_D} versus \nm{C_L} for various \nm{\deltaE} and \nm{\beta}}
\label{fig:aircraft_cd_vs_cl}
\end{figure}
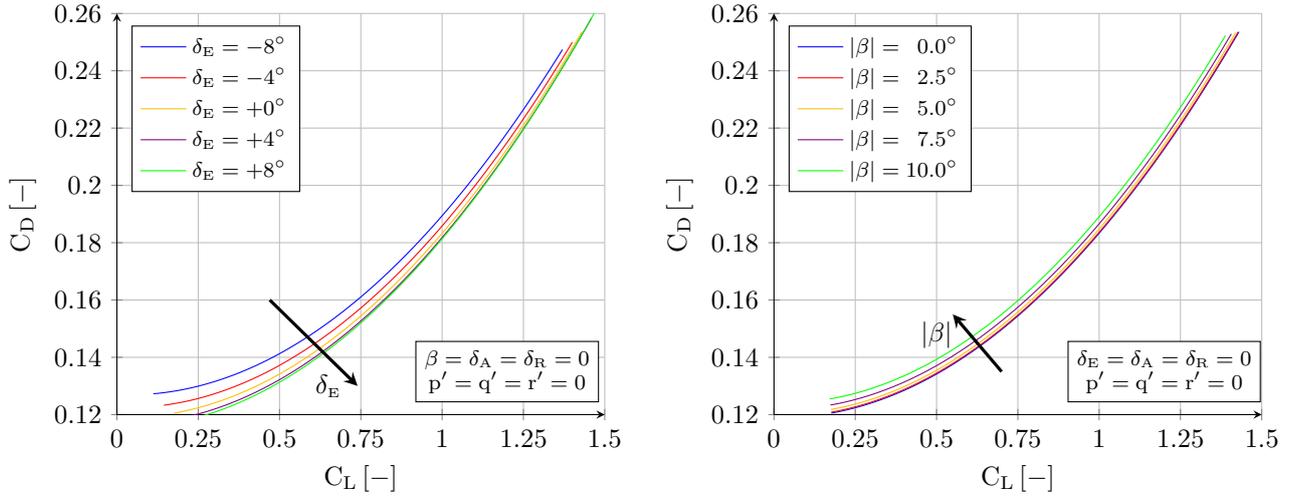

The second check is the quadratic nature of the drag coefficient \nm{C_D} with the angle of attack \nm{\alpha} \cite{Stevens2003,Etkin1972,Miele1962}. When coupled with the linearity between \nm{C_L} and \nm{\alpha} described above, this results in a parabolic drag polar, such as that shown in figure \ref{fig:aircraft_cd_vs_cl}. The drag polar is the relationship between the drag and lift coefficients \nm{C_D} and \nm{C_L}, and represents the main indicator of the aerodynamic efficiency of a given aircraft. For a given configuration, it enables the selection of the optimum \nm{C_L} (and hence optimum \nm{\alpha}) as that resulting in a minimum \nm{C_D / C_L}\footnote{To obtain it, draw a straight line tangent to the drag polar from the origin. The tangency point indicates the optimum position.}. The figure shows that the most efficient aerodynamic configuration (that providing more lift with the same drag) is obtained with no sideslip \nm{\beta} and a high elevator deflection \nm{\deltaE}.
\begin{figure}[h]
\centering
\begin{tikzpicture}
\begin{axis}[
cycle list={{blue,no markers},{red,no markers},{orange!50!yellow, no markers},{violet, no markers},{green, no markers}},
width=8.0cm,
xmin=-5, xmax=12.5, xtick={-5,-2.5,...,12.5},
xlabel={\nm{\alpha \left[^{\circ}\right]}},
xmajorgrids,
ymin=-0.5, ymax=0.55, ytick={-0.5,-0.3,-0.1,0.1,0.3,0.5},
ylabel={\nm{C_{Mm} \left[-\right]}},
ymajorgrids,
axis lines=left,
axis line style={-stealth},
legend entries={\nm{\deltaE = - 8^{\circ}},\nm{\deltaE = - 4^{\circ}}, \nm{\deltaE = + 0^{\circ}}, \nm{\deltaE = + 4^{\circ}}, \nm{\deltaE = + 8^{\circ}}},
legend pos=north east,
legend style={font=\footnotesize},
legend cell align=left,
]
\draw [ultra thin] (-5.0,0.55) -- (12.5,0.55);
\pgfplotstableread{figs/ch03_acft/test1_cmm_vs_alpha__deltaE.txt}\mytable
\addplot table [header=false, x index=0,y index=1] {\mytable};
\addplot table [header=false, x index=0,y index=2] {\mytable};
\addplot table [header=false, x index=0,y index=3] {\mytable};
\addplot table [header=false, x index=0,y index=4] {\mytable};
\addplot table [header=false, x index=0,y index=5] {\mytable};
\draw [-stealth, very thick] (+3.0,0.25) -- (1.2,-0.38);
\path node at (0.8,-0.4) {\nm{\deltaE}};
\path node [draw, shape=rectangle, fill=white, align=center] at (+0.70,+0.42) {\footnotesize \nm{\beta = \deltaA = \deltaR = 0}\\[-3pt] \footnotesize \nm{p' = q' = r' = 0}};
\end{axis}	
\end{tikzpicture}%
\hskip 0pt
\begin{tikzpicture}
\begin{axis}[
cycle list={{blue,no markers},{red,no markers},{orange!50!yellow, no markers},{violet, no markers},{green, no markers}},
width=8.0cm,
xmin=-5, xmax=12.5, xtick={-5,-2.5,...,12.5},
xlabel={\nm{\alpha \left[^{\circ}\right]}},
xmajorgrids,
ymin=-0.5, ymax=0.55, ytick={-0.5,-0.3,-0.1,0.1,0.3,0.5},
ylabel={\nm{C_{Mm} \left[-\right]}},
ymajorgrids,
axis lines=left,
axis line style={-stealth},
legend entries={\nm{|\beta| = \ \, 0.0^{\circ}},\nm{|\beta| = \ \, 2.5^{\circ}}, \nm{|\beta| = \ \, 5.0^{\circ}}, \nm{|\beta| = \ \, 7.5^{\circ}}, \nm{|\beta| = 10.0^{\circ}}},
legend pos=north east,
legend style={font=\footnotesize},
legend cell align=left,
]
\draw [ultra thin] (-5.0,0.55) -- (12.5,0.55);
\pgfplotstableread{figs/ch03_acft/test2_cmm_vs_alpha__beta.txt}\mytable
\addplot table [header=false, x index=0,y index=1] {\mytable};
\addplot table [header=false, x index=0,y index=2] {\mytable};
\addplot table [header=false, x index=0,y index=3] {\mytable};
\addplot table [header=false, x index=0,y index=4] {\mytable};
\addplot table [header=false, x index=0,y index=5] {\mytable};
\draw [-stealth, very thick] (10.5,-0.35) -- (11.0,-0.2);
\path node at (11.7,-0.18) {\nm{|\beta|}};
\draw [-stealth, very thick] (-2.6,0.1) -- (-3.1,-0.02);
\path node at (-3.8,-0.04) {\nm{|\beta|}};
\path node [draw, shape=rectangle, fill=white, align=center] at (-0.20,+0.42) {\footnotesize \nm{\deltaE = \deltaA = \deltaR = 0}\\[-3pt] \footnotesize \nm{p' = q' = r' = 0}};
\end{axis}		
\end{tikzpicture}%
\caption{\nm{C_{Mm}} versus \nm{\alpha} for various \nm{\deltaE} and \nm{\beta}}
\label{fig:aircraft_cmm_vs_alpha}
\end{figure}

Another validity check is the aircraft longitudinal stability, analyzed by means of figure \ref{fig:aircraft_cmm_vs_alpha}. It shows that the pitching moment coefficient \nm{C_{Mm}} decreases with the angle of attack \nm{\alpha}, implying that the aircraft is longitudinally stable \cite{Stevens2003,Etkin1972,Miele1962}, since any disturbance that increases the body pitch angle \nm{\theta} (and hence the angle of attack \nm{\alpha}) causes a negative pitching moment that opposes the disturbance and moves the aircraft back towards equilibrium. Figure \ref{fig:aircraft_cmm_vs_alpha} also shows that the elevator is the main way to control the aircraft pitch \nm{\theta} and its angle of attack \nm{\alpha}; when the horns are moved forward displacing the elevator down, \nm{\deltaE} increases and induces a diminution in the pitching moment coefficient \nm{C_{Mm}}, which in turn causes the aircraft to reduce its pitch and hence the angle of attack.

The throttle \nm{\deltaT} and elevator \nm{\deltaE} are employed by the aircraft control system to adhere to its two longitudinal guidance objectives: one is used to reach and follow a given airspeed, while the second is generally employed to maintain altitude or to climb and descend at a given rate.
  

\subsection{Lateral Forces and Moments}\label{subsec:AircraftModel_aerodynamics_lat}

The lateral actions described in this section are the lateral force (\nm{F_y}) together with the yaw (\nm{M_n}) and roll (\nm{M_l}) moments. The analysis is similar to that described in the previous section for the longitudinal actions, with the exception that in this case the most significant parameters are the sideslip \nm{\beta}, the ailerons deflection \nm{\deltaA}, and the rudder deflection \nm{\deltaR}.
\begin{eqnarray}
\nm{C_{lat}} & = & \nm{C_{lat} \big|_{\sss {\vec\delta}_5} \lrp{\beta, \, \deltaA, \, \deltaR} + 
\pderpar{C_{lat}}{\alpha} \bigg|_{\sss {\vec\delta}_5} \! \lrp{\beta, \, \deltaA, \, \deltaR} \cdot \lrp{\alpha - 5} +
\pderpar{C_{lat}}{\deltaE} \bigg|_{\sss {\vec\delta}_5} \! \lrp{\beta, \, \deltaA, \, \deltaR} \cdot \deltaE} \nonumber \\
& & \nm{+ \pderpar{C_{lat}}{p'} \bigg|_{\sss {\vec\delta}_5} \! \lrp{\beta, \, \deltaA, \, \deltaR} \cdot p' +
\pderpar{C_{lat}}{q'} \bigg|_{\sss {\vec\delta}_5} \! \lrp{\beta, \, \deltaA, \, \deltaR} \cdot q' +
\pderpar{C_{lat}}{r'} \bigg|_{\sss {\vec\delta}_5} \! \lrp{\beta, \, \deltaA, \, \deltaR} \cdot r'} \label{eq:AircraftModel_lateral}
\end{eqnarray}

Although each lateral coefficient (\nm{C_{lat}} stands for \nm{C_{Fy}}, \nm{C_{Ml}}, or \nm{C_{Mn}}) is also modeled by the sum of six tridimensional tables (the independent parameters are \nm{\beta}, \nm{\deltaA}, and \nm{\deltaR}), they are evaluated at a point where the angle of attack is \nm{5^{\circ}}\footnote{A nonzero angle of attack is more characteristic of the normal situation of the aircraft while in flight.} while all other parameters are zero (\nm{\vec\delta_5 \rightarrow \alpha = 5^{\circ}, \, \deltaE = p' = q' = r' = 0}).

In this case, the construction of the lateral tables requires 225 executions of \hypertt{AVL} with the following inputs:
\begin{center}
\begin{tabular}{lcccc}
	\hline
	Input & & Unit & Range & \# \\
	\hline
	Angle of sideslip	& \nm{\beta}	& \nm{^{\circ}} & \nm{-10.0, -7.5, \dots, 7.5, 10.0}	& 9 \\
	Ailerons			& \nm{\deltaA}	& \nm{^{\circ}} & \nm{-8.0, -4.0, 0, 4.0, 8.0}			& 5 \\
	Rudder				& \nm{\deltaR}	& \nm{^{\circ}} & \nm{-8.0, -4.0, 0, 4.0, 8.0}			& 5 \\
	\hline
\end{tabular}
\end{center}
\captionof{table}{Inputs to \texttt{AVL} for lateral model generation}\label{tab:Aircraft_AVL_executions_lateral}	
\begin{figure}[h]
\centering
\begin{tikzpicture}
\begin{axis}[
cycle list={{blue,no markers},{red,no markers},{orange!50!yellow, no markers},{violet, no markers},{green, no markers}},
width=8.0cm,
xmin=-10, xmax=10, xtick={-10,-5,0,5,10},
xlabel={\nm{\beta \left[^{\circ}\right]}},
xmajorgrids,
ymin=-4, ymax=5, ytick={-4,-2,0,2,4},
ylabel={\nm{C_{Mn} \, \left[10^{-2} \ -\right]}},
ymajorgrids,
axis lines=left,
axis line style={-stealth},
legend entries={\nm{\deltaR = - 8^{\circ}},\nm{\deltaR = - 4^{\circ}}, \nm{\deltaR = + 0^{\circ}}, \nm{\deltaR = + 4^{\circ}}, \nm{\deltaR = + 8^{\circ}}},
legend pos=north west,
legend style={font=\footnotesize},
legend cell align=left,
]
\draw [ultra thin] (-10.0,5.0) -- (10.0,5.0);
\pgfplotstableread{figs/ch03_acft/test3_cmn_vs_beta__deltaR.txt}\mytable
\addplot table [header=false, x index=0,y index=1] {\mytable};
\addplot table [header=false, x index=0,y index=2] {\mytable};
\addplot table [header=false, x index=0,y index=3] {\mytable};
\addplot table [header=false, x index=0,y index=4] {\mytable};
\addplot table [header=false, x index=0,y index=5] {\mytable};
\draw [-stealth, very thick] (6.7,-1.8) -- (3.0,3.5);
\path node at (1.9,3.3) {\nm{\deltaR}};
\path node [draw, shape=rectangle, fill=white, align=center] at (+4.0,-3.0) {\footnotesize \nm{\alpha = 4^{\circ}}\\[-3pt]\footnotesize \nm{\deltaE = \deltaA = p' = q' = r' = 0}};
\end{axis}	
\end{tikzpicture}%
\hskip 10pt
\begin{tikzpicture}
\begin{axis}[
cycle list={{blue,no markers},{red,no markers},{orange!50!yellow, no markers},{violet, no markers},{green, no markers}},
width=8.0cm,
xmin=-10, xmax=10, xtick={-10,-5,0,5,10},
xlabel={\nm{\beta \left[^{\circ}\right]}},
xmajorgrids,
ymin=-4, ymax=5, ytick={-4,-2,0,2,4},
ylabel={\nm{C_{Mn} \, \left[10^{-2} \ -\right]}},
ymajorgrids,
axis lines=left,
axis line style={-stealth},
legend entries={\nm{\deltaA = - 8^{\circ}},\nm{\deltaA = - 4^{\circ}}, \nm{\deltaA = + 0^{\circ}}, \nm{\deltaA = + 4^{\circ}}, \nm{\deltaA = + 8^{\circ}}},
legend pos=north west,
legend style={font=\footnotesize},
legend cell align=left,
]
\draw [ultra thin] (-10.0,5.0) -- (10.0,5.0);
\pgfplotstableread{figs/ch03_acft/test4_cmn_vs_beta__deltaA.txt}\mytable
\addplot table [header=false, x index=0,y index=1] {\mytable};
\addplot table [header=false, x index=0,y index=2] {\mytable};
\addplot table [header=false, x index=0,y index=3] {\mytable};
\addplot table [header=false, x index=0,y index=4] {\mytable};
\addplot table [header=false, x index=0,y index=5] {\mytable};
\draw [-stealth, very thick] (1.9,1.1) -- (3.0,-0.5);
\path node at (4.1,-0.5) {\nm{\deltaA}};
\path node [draw, shape=rectangle, fill=white, align=center] at (+4.0,-2.5) {\footnotesize \nm{\alpha = 4^{\circ}}\\[-3pt]\footnotesize \nm{\deltaE = \deltaR = p' = q' = r' = 0}};
\end{axis}		
\end{tikzpicture}%
\caption{\nm{C_{Mn}} versus \nm{\beta} for various \nm{\deltaR} and \nm{\deltaA}}
\label{fig:aircraft_cmn_vs_beta}
\end{figure}

The aircraft yawing moment \nm{M_n} can be analyzed by means of figure \ref{fig:aircraft_cmn_vs_beta}. Note that because of the aircraft symmetry [A\ref{as:APM_symmetric}], if given rudder \nm{\delta_{{\sss{R}}0}} and sideslip \nm{\beta_0} values result in a yawing coefficient \nm{C_{Mn0}}, then their negatives (\nm{- \delta_{{\sss{R}}0}}, \nm{- \beta_0}) result in (\nm{- C_{Mn0}}), and the same occurs with the ailerons \nm{\deltaA} effects. The rudder is the main way to control the aircraft yaw angle \nm{\psi}; when the right pedal is pushed down displacing the rudder to the right, \nm{\deltaR} increases and induces a higher yawing moment coefficient \nm{C_{Mn}}, which in turn causes the aircraft to turn rightwards. Although the ailerons position \nm{\deltaA} also influences the yawing moment, their effect on \nm{C_{Mn}} is much smaller. Figure \ref{fig:aircraft_cmn_vs_beta} also shows that if the airflow hits the aircraft from its right hand side (positive sideslip \nm{\beta}), the lateral force that this creates on the vertical stabilizer causes the aircraft to rotate to its right (positive \nm{C_{Mn}}). The aircraft is thus laterally stable \cite{Stevens2003,Etkin1972,Miele1962}, as any perturbation that increases its sideslip \nm{\beta}, rotating the aircraft to its left, generates an increased yawing moment that opposes the disturbance and rotates the aircraft back towards balance.
\begin{figure}[h]
\centering
\begin{tikzpicture}
\begin{axis}[
cycle list={{blue,no markers},{red,no markers},{orange!50!yellow, no markers},{violet, no markers},{green, no markers}},
width=8.0cm,
xmin=-10, xmax=10, xtick={-10,-5,0,5,10},
xlabel={\nm{\beta \left[^{\circ}\right]}},
xmajorgrids,
ymin=-6, ymax=6, ytick={-6,-4,-2,0,2,4,6},
ylabel={\nm{C_{Ml} \, \left[10^{-2} \ -\right]}},
ymajorgrids,
axis lines=left,
axis line style={-stealth},
legend entries={\nm{\deltaR = - 8^{\circ}},\nm{\deltaR = - 4^{\circ}}, \nm{\deltaR = + 0^{\circ}}, \nm{\deltaR = + 4^{\circ}}, \nm{\deltaR = + 8^{\circ}}},
legend pos=south west,
legend style={font=\footnotesize},
legend cell align=left,
]
\pgfplotstableread{figs/ch03_acft/test3_cml_vs_beta__deltaR.txt}\mytable
\addplot table [header=false, x index=0,y index=1] {\mytable};
\addplot table [header=false, x index=0,y index=2] {\mytable};
\addplot table [header=false, x index=0,y index=3] {\mytable};
\addplot table [header=false, x index=0,y index=4] {\mytable};
\addplot table [header=false, x index=0,y index=5] {\mytable};
\draw [-stealth, very thick] (3.5,-1.2) -- (4.5,0.8);
\path node at (3.1,0.8) {\nm{\deltaR}};
\path node [draw, shape=rectangle, fill=white, align=center] at (+3.2,+4.3) {\footnotesize \nm{\alpha = 4^{\circ}}\\[-3pt]\footnotesize \nm{\deltaE = \deltaA = p' = q' = r' = 0}};
\end{axis}		
\end{tikzpicture}
\hskip 10pt
\begin{tikzpicture}
\begin{axis}[
cycle list={{blue,no markers},{red,no markers},{orange!50!yellow, no markers},{violet, no markers},{green, no markers}},
width=8.0cm,
xmin=-10, xmax=10, xtick={-10,-5,0,5,10},
xlabel={\nm{\beta \left[^{\circ}\right]}},
xmajorgrids,
ymin=-6, ymax=6, ytick={-6,-4,-2,0,2,4,6},
ylabel={\nm{C_{Ml} \, \left[10^{-2} \ -\right]}},
ymajorgrids,
axis lines=left,
axis line style={-stealth},
legend entries={\nm{\deltaA = - 8^{\circ}},\nm{\deltaA = - 4^{\circ}}, \nm{\deltaA = + 0^{\circ}}, \nm{\deltaA = + 4^{\circ}}, \nm{\deltaA = + 8^{\circ}}},
legend pos=north east,
legend style={font=\footnotesize},
legend cell align=left,
]
\pgfplotstableread{figs/ch03_acft/test4_cml_vs_beta__deltaA.txt}\mytable
\addplot table [header=false, x index=0,y index=1] {\mytable};
\addplot table [header=false, x index=0,y index=2] {\mytable};
\addplot table [header=false, x index=0,y index=3] {\mytable};
\addplot table [header=false, x index=0,y index=4] {\mytable};
\addplot table [header=false, x index=0,y index=5] {\mytable};
\draw [-stealth, very thick] (-8.0,-3.0) -- (-3.0,+4.5);
\path node at (-2.5,+5.0) {\nm{\deltaA}};
\path node [draw, shape=rectangle, fill=white, align=center] at (-4.0,-4.7) {\footnotesize \nm{\alpha = 4^{\circ}}\\[-3pt]\footnotesize \nm{\deltaE = \deltaR = p' = q' = r' = 0}};
\end{axis}	
\end{tikzpicture}%
\caption{\nm{C_{Ml}} versus \nm{\beta} for various \nm{\deltaR} and \nm{\deltaA}}
\label{fig:aircraft_cml_vs_beta}
\end{figure}

The rolling moment \nm{M_l} shown in figure \ref{fig:aircraft_cml_vs_beta} presents the same symmetry characteristics with respect to sideslip \nm{\beta}, ailerons \nm{\deltaA}, and rudder \nm{\deltaR}, as those of the yawing moment \nm{M_n} described in the previous paragraph. The ailerons mission is to control the aircraft bank angle \nm{\xi}; when the horns are moved rightwards (\nm{\deltaA} increases), raising the right aileron and lowering the left one, the lift of the right wing decreases while that of the left one grows, creating a positive rolling moment coefficient \nm{C_{Ml}} that causes the aircraft to bank to its right hand side. Although the rudder deflection \nm{\deltaR} also influences the rolling moment, its effect on \nm{C_{Ml}} is much smaller. Figure \ref{fig:aircraft_cml_vs_beta} also shows that if the airflow hits the aircraft from its right hand side (positive sideslip \nm{\beta}), the fuselage partially blocks the airflow reaching the left wing, reducing its lift, and hence causing the aircraft to roll to its left (negative \nm{C_{Ml}}).
\begin{figure}[h]
\centering
\begin{tikzpicture}
\begin{axis}[
cycle list={{blue,no markers},{red,no markers},{orange!50!yellow, no markers},{violet, no markers},{green, no markers}},
width=8.0cm,
xmin=-10, xmax=10, xtick={-10,-5,0,5,10},
xlabel={\nm{\beta \left[^{\circ}\right]}},
xmajorgrids,
ymin=-10, ymax=10, ytick={-10,-5,0,5,10},
ylabel={\nm{C_Y \, \left[10^{-2} \ -\right]}},
ymajorgrids,
axis lines=left,
axis line style={-stealth},
legend entries={\nm{\deltaR = - 8^{\circ}},\nm{\deltaR = - 4^{\circ}}, \nm{\deltaR = + 0^{\circ}}, \nm{\deltaR = + 4^{\circ}}, \nm{\deltaR = + 8^{\circ}}},
legend pos=north west,
legend style={font=\footnotesize},
legend cell align=left,
]
\pgfplotstableread{figs/ch03_acft/test3_cy_vs_beta__deltaR.txt}\mytable
\addplot table [header=false, x index=0,y index=1] {\mytable};
\addplot table [header=false, x index=0,y index=2] {\mytable};
\addplot table [header=false, x index=0,y index=3] {\mytable};
\addplot table [header=false, x index=0,y index=4] {\mytable};
\addplot table [header=false, x index=0,y index=5] {\mytable};
\draw [-stealth, very thick] (6.0,-3.0) -- (4.0,8.0);
\path node at (2.9,8.0) {\nm{\deltaR}};
\path node [draw, shape=rectangle, fill=white, align=center] at (+3.7,-7.5) {\footnotesize \nm{\alpha = 4^{\circ}}\\[-3pt]\footnotesize \nm{\deltaE = \deltaA = p' = q' = r' = 0}};
\end{axis}		
\end{tikzpicture}
\hskip 10pt
\begin{tikzpicture}
\begin{axis}[
cycle list={{blue,no markers},{red,no markers},{orange!50!yellow, no markers},{violet, no markers},{green, no markers}},
width=8.0cm,
xmin=-10, xmax=10, xtick={-10,-5,0,5,10},
xlabel={\nm{\beta \left[^{\circ}\right]}},
xmajorgrids,
ymin=-10, ymax=10, ytick={-10,-5,0,5,10},
ylabel={\nm{C_Y \, \left[10^{-2} \ -\right]}},
ymajorgrids,
axis lines=left,
axis line style={-stealth},
legend entries={\nm{\deltaA = - 8^{\circ}},\nm{\deltaA = - 4^{\circ}}, \nm{\deltaA = + 0^{\circ}}, \nm{\deltaA = + 4^{\circ}}, \nm{\deltaA = + 8^{\circ}}},
legend pos=north west,
legend style={font=\footnotesize},
legend cell align=left,
]
\pgfplotstableread{figs/ch03_acft/test4_cy_vs_beta__deltaA.txt}\mytable
\addplot table [header=false, x index=0,y index=1] {\mytable};
\addplot table [header=false, x index=0,y index=2] {\mytable};
\addplot table [header=false, x index=0,y index=3] {\mytable};
\addplot table [header=false, x index=0,y index=4] {\mytable};
\addplot table [header=false, x index=0,y index=5] {\mytable};
\path node [draw, shape=rectangle, fill=white, align=center] at (+3.7,-7.5) {\footnotesize \nm{\alpha = 4^{\circ}}\\[-3pt]\footnotesize \nm{\deltaE = \deltaR = p' = q' = r' = 0}};
\end{axis}	
\end{tikzpicture}%
\caption{\nm{C_Y} versus \nm{\beta} for various \nm{\deltaR} and \nm{\deltaA}}
\label{fig:aircraft_cy_vs_beta}
\end{figure}

Although very small when compared with the lift and drag forces, figure \ref{fig:aircraft_cy_vs_beta} enables the analysis of the side force Y, which shares the symmetry characteristics of the yawing and rolling moments. The slope of the plots is positive as an airflow reaching the fuselage and vertical stabilizer from their right hand side (positive \nm{\beta}) generates a force towards the left (positive \nm{C_Y}). The ailerons \nm{\deltaA} influence on the side force is negligible, but a positive rudder deflection \nm{\deltaR} increases the lateral force generated by the vertical stabilizer (its lift) and hence that of the aircraft.

The coordinated action of the rudder \nm{\deltaR} and ailerons \nm{\deltaA} is employed by the aircraft control system to adhere to its two lateral directional guidance objectives. One is nearly always the obtainment of symmetric flight (\nm{\beta = 0}), since according to figure \ref{fig:aircraft_cd_vs_cl} this is the most efficient aerodynamic configuration, while the second entails following a given bearing to reach the intended destination or maintaining a bank angle while in turn.


\subsection{Correction for Center of Mass Variation}\label{subsec:AircraftModel_aerodynamics_correction}

The expressions for the aerodynamic forces and moments shown in the previous sections have been obtained with a full tank configuration, so it is necessary to correct them for the variation in the center of mass position caused by the fuel consumption. Note that the variation of the center of mass does not change the orientation of the \nm{\FB} axes, only their origin \nm{\OB}, so the final aerodynamic forces and moments are the following:
\begin{eqnarray}
\nm{\FAERB} & = & \nm{\vec F_{{\sss AER},full}^{\sss B}}\label{eq:eq:AircraftModel_FaerB_final} \\
\nm{\MAERB} & = & \nm{\vec M_{{\sss AER},full}^{\sss B} + \FAERBfullskew \; \lrsb{\vec T_{\sss RB}^{\sss B}\lrp{m} - \vec T_{{\sss RB},full}^{\sss B}}}\label{eq:eq:AircraftModel_MaerB_final} 
\end{eqnarray}


\section{Propulsion}\label{sec:AircraftModel_propulsion}

The propulsion analysis follows the same approach as the aerodynamics one performed in section \ref{sec:AircraftModel_aerodynamics} with the objective of obtaining realistic expressions for the forces and moments generated by the aircraft propulsion system, which are called \emph{thrust} (\nm{T}) and \emph{propeller torque} (\nm{M_T}):
\begin{eqnarray}
\nm{\FPROB} & = & \nm{\lrsb{T_x \ \ T_y \ \ T_z}^T} \label{eq:AircraftModel_FproB_bis} \\
\nm{\MPROB} & = & \nm{\lrsb{M_{Tl} \ \ M_{Tm} \ \ M_{Tn}}^T} \label{eq:AircraftModel_MproB_bis} 
\end{eqnarray}

These equations can be simplified based on the hypotheses of aircraft lateral symmetry [A\ref{as:APM_symmetric}] and alignment between the thrust force, propeller axis, and aircraft fuselage [A\ref{as:APM_propeller_axis},A\ref{as:APM_propeller_thrust}].:
\begin{eqnarray}
\nm{\FPROB} & = & \nm{\lrsb{T_x \ \ 0 \ \ 0}^T = \ \lrsb{T \ \ 0 \ \ 0}^T} \label{eq:AircraftModel_FproB} \\
\nm{\MPROB} & = & \nm{\lrsb{M_{Tl} \ \ 0 \ \ 0}^T = \ \lrsb{M_T \ \ 0 \ \ 0}^T} \label{eq:AircraftModel_MproB} 
\end{eqnarray}


\subsection{Dimensional Analysis}\label{subsec:AircraftModel_propulsion_dimensional}

The generation of the thrust force and propeller torque is the result of two different physical processes. The first is the conversion of the fuel internal energy into mechanical rotating energy by means of the combustion that occurs inside the power plant \cite{Taylor1985,Oates1989,Heywood1988,Hill1992}, and the second is the propeller transfer of that mechanical energy into the air stream, increasing its velocity and hence generating thrust \cite{Hill1992,Hitches2015}. The two processes are not independent, as both the engine and the propeller share the same power shaft. The physical variables involved are the following:
\begin{itemize}
\item The generated propulsive forces (\nm{T_x, \, T_y, \, T_z}) and moments (\nm{M_{Tl}, \, M_{Tm}, \, M_{Tn}}) by means of the power (P) transmitted through the shaft.
\item The fuel flow \nm{F} being injected into the cylinders, as well as its lower heating value \nm{L_{\sss HV}}.
\item The characteristics of the air flow, such as the pressure \emph{p}, temperature T, density \nm{\rho}, adiabatic index \nm{\kappa}, specific air constant R, viscosity \nm{\mu_v}, and relative humidity \nm{R_H}.
\item The state of the power shaft, defined by its rotation speed \emph{n} and torque  \emph{Q}.
\item The mixture strength \nm{f_{\sss E}}, equivalent to the fuel to air weight ratio.
\item The piston engine thermodynamical efficiency \nm{\eta_{\sss {E,t}}} depicting the percentage of the fuel internal energy that gets transformed into power, and the volumetric efficiencies \nm{\eta_{\sss {E,v}}} representing the percentage of the cylinders volume filled up by fresh air at the intake stroke.
\item The piston engine geometry, represented by the total engine displacement \nm{V_{\sss E}} and a series of dimensionless parameters \nm{p_{\sss E}} describing the engine shape.
\item The propeller geometry, represented by its diameter \nm{D_{\sss P}}, its angle of attack at at a section located at 75\% of the blades length \nm{\beta_{75}}, and a series of dimensionless parameters \nm{p_{\sss P}} describing the propeller shape.
\item The interaction of the air stream with both the power plant and the propeller, defined by the airspeed \nm{\vtas}, the angle of attack \nm{\alpha}, the angle of sideslip \nm{\beta}, and their variations with time \nm{\dot{v}_{\sss {TAS}}, \, \dot{\alpha}, \, and \, \dot{\beta}}.
\end{itemize}


\subsection{Piston Engine Power Generation and Fuel Consumption}\label{subsec:AircraftModel_propulsion_power}

At this point it is better to separate the two processes, starting with the power generation at the piston engine. As in the aerodynamics case, the number of physical dimensions is \nm{k = 4} (length, mass, time, and temperature), so the application of Buckingham \nm{\pi} theorem \cite{Szirtes1997} enables the elimination of four physical variables (air temperature T, air density \nm{\rho}, specific air constant R, and total engine displacement \nm{V_{\sss E}}) while removing the dimensions of all others \cite{Gallo2012}:
\neweq{\lrsb{\dfrac{P}{p \; V_{\sss E}^{2/3} \; \sqrt{R \; T} \; \eta_{\sss {E,t}} \; \eta_{\sss {E,v}}} \ \ \dfrac{F \; L_{\sss HV}}{p \; V_{\sss E}^{2/3} \; \sqrt{R \; T}}}^T = \vec f_{P,1} \lrp{
\begin{array}{l}
\nm{\dfrac{L_{\sss HV}}{R \; T}, \, \dfrac{p}{\rho \; R \; T}, \, \kappa, \, R_H, \, \dfrac{n \; V_{\sss E}^{1/3}}{\rho \; R \; T}, \, \dfrac{n \; Q}{P}, \, f_{\sss E}, \, p_{\sss E}} \\ 
\nm{\dfrac{\vtas}{\sqrt{R \; T}}, \, \alpha, \, \beta, \, \dfrac{\dot{v}_{\sss TAS} \; V_{\sss E}^{1/3}}{R \; T}, \, \dfrac{\dot{\alpha} \; V_{\sss E}^{1/3}}{\sqrt{R \; T}}, \, \dfrac{\dot{\beta} \; V_{\sss E}^{1/3}}{\sqrt{R \; T}}}
\end{array}
}}{eq:AircraftModel_power_dimensional_analysis}

This expression can be further simplified by means of the following assumptions: perfect gas [A\ref{as:ISA_perfect_gas}], no air humidity [A\ref{as:ISA_humidity}], quasi stationary flow [A\ref{as:APM_quasi_stationary}], the relationship between shaft power and torque (\nm{P = n \; Q}), the constant engine thermodynamic efficiency [A\ref{as:APM_power_efficiency}], and the particularization for a given engine. In addition, the full recovery of the air flow kinetic energy at the inlet [A\ref{as:APM_flow_piston}] enables the elimination of the airspeed \nm{\vtas}, and the replacement of the pressure and temperature by their total equivalents\footnote{Total magnitudes (pressure \nm{p_t}, temperature \nm{T_t}, and density \nm{\rho_t}) are those obtained if a fluid moving at static atmospheric magnitudes (p, T, \nm{\rho}) completely decelerates through a process that is stationary, has no viscosity or inertial forces, is adiabatic, and presents fixed boundaries for the analyzed volume control. Such a process conserves the fluid entropy as well as its total enthalpy.} \nm{p_t} and \nm{T_t}:
\neweq{\lrsb{\dfrac{P}{p_t \; V_{\sss E}^{2/3} \; \sqrt{R \; T_t} \; \eta_{\sss {E,v}}} \ \ \dfrac{F \; L_{\sss HV}}{p_t \; V_{\sss E}^{2/3} \; \sqrt{R \; T_t}}}^T = \vec f_{P,2} \lrp{\dfrac{L_{\sss HV}}{R \; T_t}, \, \dfrac{n \; V_{\sss E}^{1/3}}{\rho \; R \; T_t}, \, f_{\sss E}}}{eq:AircraftModel_power_dimensional_analysis2}

This indicates that the power generated by a piston engine and its fuel consumption depend on three parameters: the rotation speed, the energy stored in the fuel, and the mixture ratio \cite{Gallo2012}. The power plant model considers that aviation gasoline is always employed, and that the mixture strength is automatically controlled by the engine incorporated control unit, leaving the dimensionless rotating speed as the only independent parameter. 
\neweq{\lrsb{\dfrac{P}{p_t \; V_{\sss E}^{2/3} \; \sqrt{R \; T_t} \; \eta_{\sss {E,v}}} \ \ \dfrac{F \; L_{\sss HV}}{p_t \; V_{\sss E}^{2/3} \; \sqrt{R \; T_t}}}^T = \vec f_{P,3} \lrp{\dfrac{n \; V_{\sss E}^{1/3}}{\rho \; R \; T_t}}}{eq:AircraftModel_power_dimensional_analysis3}

This expression can be analyzed in two ways: first considering that the throttle position \nm{\deltaT} controls the piston engine by means of the rotating speed of its output power shaft, and second by switching terms and considering that the output power and speed depend on the amount of fuel injected into the cylinders. In practical terms the choice does not matter, as the important fact is that the power plant behavior is controlled from the flight deck by a single parameter, \nm{\deltaT}.
\begin{figure}[h]
\centering
\begin{tikzpicture}
\begin{axis}[
cycle list={{blue,no markers},{red,no markers},{orange!50!yellow, no markers},{violet, no markers},{green, no markers}},
width=8.0cm,
xmin=0, xmax=5000, xtick={0,1000,2000,3000,4000,5000},
xlabel={\nm{\Hp \left[m\right]}},
xmajorgrids,
ymin=2.0, ymax=4.5, ytick={2.0,2.5,3.0,3.5,4.0,4.5},
ylabel={\nm{P \left[kW\right]}},
ymajorgrids,
axis lines=left,
axis line style={-stealth},
legend entries={\nm{\deltaT = 1.0},\nm{\deltaT = 0.9}, \nm{\deltaT = 0.8}, \nm{\deltaT = 0.7}, \nm{\deltaT = 0.6}},
legend pos=north east,
legend style={font=\footnotesize},
legend cell align=left,
]
\pgfplotstableread{figs/ch03_acft/test7_P_vs_Hp__deltaT.txt}\mytable
\addplot table [header=false, x index=0,y index=1] {\mytable};
\addplot table [header=false, x index=0,y index=2] {\mytable};
\addplot table [header=false, x index=0,y index=3] {\mytable};
\addplot table [header=false, x index=0,y index=4] {\mytable};
\addplot table [header=false, x index=0,y index=5] {\mytable};
\draw [-stealth, very thick] (300,2.2) -- (800,4.3);
\path node at (1100,4.2) {\nm{\deltaT}};
\path node [draw, shape=rectangle, fill=white] at (4000,3.2) {\footnotesize \nm{\DeltaT = 0 \ K}};
\end{axis}	
\end{tikzpicture}%
\hskip 10pt
\begin{tikzpicture}
\begin{axis}[
cycle list={{blue,no markers},{red,no markers},{orange!50!yellow, no markers},{violet, no markers},{green, no markers}},
width=8.0cm,
xmin=0, xmax=5000, xtick={0,1000,2000,3000,4000,5000},
xlabel={\nm{\Hp \left[m\right]}},
xmajorgrids,
ymin=0.8, ymax=1.6, ytick={0.8,1.0,1.2,1.4,1.6},
ylabel={\nm{F \left[kg/hr\right]}},
ymajorgrids,
axis lines=left,
axis line style={-stealth},
legend entries={\nm{\deltaT = 1.0},\nm{\deltaT = 0.9}, \nm{\deltaT = 0.8}, \nm{\deltaT = 0.7}, \nm{\deltaT = 0.6}},
legend pos=north east,
legend style={font=\footnotesize},
legend cell align=left,
]
\pgfplotstableread{figs/ch03_acft/test7_F_vs_Hp__deltaT.txt}\mytable
\addplot table [header=false, x index=0,y index=1] {\mytable};
\addplot table [header=false, x index=0,y index=2] {\mytable};
\addplot table [header=false, x index=0,y index=3] {\mytable};
\addplot table [header=false, x index=0,y index=4] {\mytable};
\addplot table [header=false, x index=0,y index=5] {\mytable};
\draw [-stealth, very thick] (300,0.83) -- (800,1.5);
\path node at (1100,1.5) {\nm{\deltaT}};
\path node [draw, shape=rectangle, fill=white] at (4000,1.18) {\footnotesize \nm{\DeltaT = 0 \ K}};
\end{axis}		
\end{tikzpicture}%
\caption{Piston engine power and fuel consumption}
\label{fig:aircraft_P_F_vs_Hp}
\end{figure}
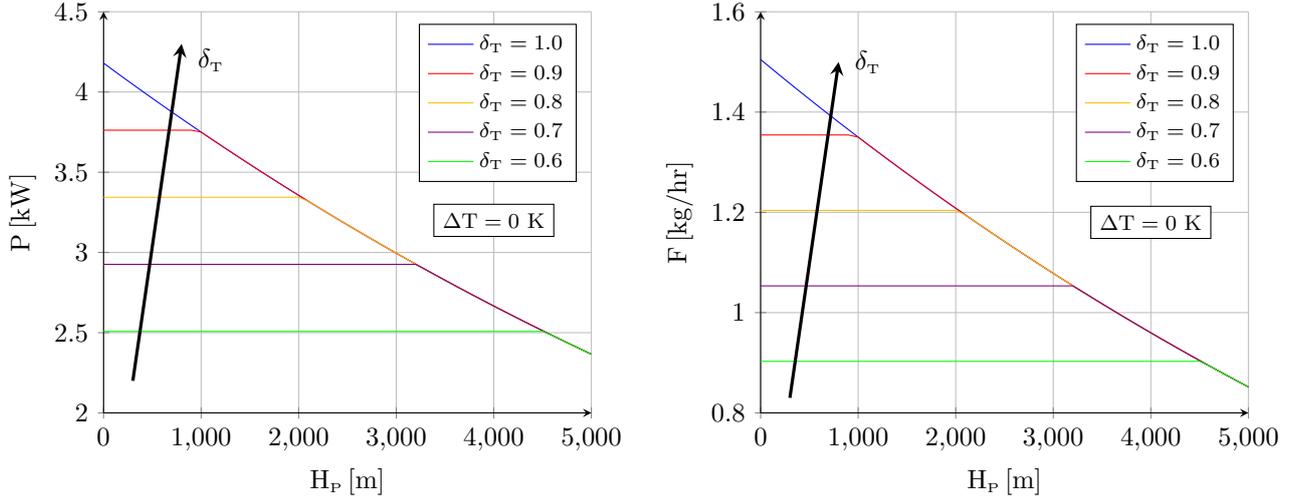

Although most piston engines employed in aviation are equipped with a dynamical inlet and centrifugal compressor (driven by a centrifugal turbine located at the outlet) to improve the cylinders admission pressure and reduce the performance degradation with altitude of atmospheric engines, the reduced size of this aircraft makes the compressor installation impractical. The selected engine is the \texttt{CRRCpro GF55II} two stroke atmospheric gasoline engine \cite{GF55II}, whose characteristics are listed in table \ref{tab:Power_plant_characteristics}.
\begin{center}
\begin{tabular}{lrcclrc}
	\hline
	Variable & Value & Unit & \ \ \ \ \ \ \ \ & Variable & Value & Unit \\
	\hline
	Number of cylinders 			& 2 	& - 				& & Maximum power 		& 4180 		& \nm{W} \\
	Total displacement  			& 55.0 	& \nm{cm^3} 		& & Maximum power speed & 7600 		& \nm{rpm} \\
	Bore 							& 3.4 	& \nm{cm} 			& & Speed range 		& 1600-7800 & \nm{rpm} \\
	Stroke 							& 3.0 	& \nm{cm} 			& & Mass 				& 1780 		& \nm{g} \\
	Power specific fuel consumption & 1.0 	& \nm{s^2 / m^2} 	& & & & \\
	\hline
\end{tabular}
\end{center}
\captionof{table}{Power plant characteristics}\label{tab:Power_plant_characteristics}	

Experimental data shows that the volumetric efficiency \nm{\eta_{\sss {E,v}}} is proportional to the square root of the total temperature \cite{Gallo2012,GalloBADA4}, which enables the establishment of the following model for the piston engine behavior:
\begin{eqnarray}
\nm{P} & = & \nm{min\lrp{P_{\sss max} \cdot \deltaT, \, P_{\sss max} \cdot \dfrac{\delta}{\sqrt{\theta}}}}\label{eq:AircraftModel_power} \\
\nm{F} & = & \nm{PSFC \cdot P}\label{eq:AircraftModel_fuel_consumption}
\end{eqnarray}

where \nm{P_{\sss max}} and \hypertt{PSFC} are the maximum power and power specific fuel consumption shown in table \ref{tab:Power_plant_characteristics}. Figure \ref{fig:aircraft_P_F_vs_Hp} shows the variations of both power and fuel consumption (they are proportional) with pressure altitude \nm{\Hp} at various positions of the throttle lever \nm{\deltaT} for a standard atmosphere\footnote{The relationship between temperature and pressure altitude depends on \nm{\DeltaT} (section \ref{sec:EarthModel_ISA}).} (\nm{\DeltaT = 0 \ K}). The significance of the power parameter \nm{\deltaT} is that the engine tries to generate the indicated power with respect to the maximum \nm{P_{\sss max}} independently of the altitude. At full throttle (\nm{\deltaT = 1}), the generated power quickly decreases with altitude per (\ref{eq:AircraftModel_power}), while at inferior settings (\nm{\deltaT < 1}) the power is constant with altitude until it can no longer be sustained. 


\subsection{Propeller, Thrust, and Torque}\label{subsec:AircraftModel_propulsion_propeller}

The application of Buckingham \nm{\pi} theorem \cite{Szirtes1997} to the propeller also permits the elimination of four physical variables (air temperature T, air density \nm{\rho}, shaft speed \emph{n}, and propeller diameter \nm{D_{\sss P}}) while removing the dimensions of all others \cite{Gallo2012}:
\neweq{\dfrac{\FPROB}{\rho \; n^2 \; D_{\sss P}^4}  = \vec f_{P,4} \lrp{\dfrac{p}{\rho \, R \, T}, \kappa, \dfrac{R \, T}{n^2 \, D_{\sss P}^2}, \dfrac{\mu_v \, \sqrt{R \, T}}{\rho \; n^2 \; D_{\sss P}^3}, R_H, \dfrac{P}{\rho \; n^3 \; D_{\sss P}^5}, \dfrac{n\;Q}{P}, \beta_{75}, p_{\sss P}, \dfrac{\vtas}{n\;D_{\sss P}}, \alpha, \beta, \dfrac{\dot{v}_{\sss TAS}}{n^2\;D_{\sss P}}, \dfrac{\dot{\alpha}}{n}, \dfrac{\dot{\beta}}{n}}}{eq:AircraftModel_propeller_dimensional_analysis}

This expression can be further simplified by means of the following assumptions: perfect gas [A\ref{as:ISA_perfect_gas}], no air humidity [A\ref{as:ISA_humidity}], quasi stationary flow [A\ref{as:APM_quasi_stationary}], no viscosity [A\ref{as:APM_viscosity}], incompressible flow [A\ref{as:APM_incompressible}], the relationship between shaft power and torque (\nm{P = n \; Q}), the alignment of the thrust force with the propeller axis [A\ref{as:APM_propeller_thrust}], the exclusive dependence on the axial airspeed [A\ref{as:APM_flow_propeller}], and the particularization for a given propeller:
\neweq{\dfrac{T}{\rho \; n^2 \; D_{\sss P}^4} = f_{P,5} \lrp{\dfrac{P}{\rho \; n^3 \; D_{\sss P}^5}, \, \dfrac{\vTASBi}{n\;D_{\sss P}} }}{eq:AircraftModel_propeller_dimensional_analysis2}

This indicates that the thrust coefficient \nm{C_T} generated by a fixed pitch propeller depends on two parameters \cite{Gallo2012,Hill1992,Hitches2015}: the power coefficient \nm{C_P} it receives by means of the rotation shaft and the advance ratio J, defined as the ratio between the propeller axial and rotating velocities. The propeller efficiency \nm{\eta_P} is then defined accordingly:
\begin{eqnarray}
\nm{C_T} & = & \nm{\dfrac{T}{\rho \; n^2 \; D_{\sss P}^4}} \label{eq:AircraftModel_propeller_Ct} \\
\nm{C_P} & = & \nm{\dfrac{P}{\rho \; n^3 \; D_{\sss P}^5}} \label{eq:AircraftModel_propeller_Cp} \\
\nm{J} & = & \nm{\dfrac{\vTASBi}{n\;D_{\sss P}}} \label{eq:AircraftModel_propeller_J} \\
\nm{\eta_P} & = & \nm{\dfrac{T \; \vTASBi}{P} = \dfrac{C_T \; J}{C_P}} \label{eq:AircraftModel_propeller_eta}
\end{eqnarray}

\texttt{JavaProp} is a simple software program intended for the design and analysis of propellers and wind turbines \cite{JavaProp}, applicable within its limits to aeronautical as well as marine applications. It is based on the coupling of moments with two dimensional airfoil characteristics, enabling the consideration of different airfoil sections and their impact on propeller performance. The estimation of the geometry of a two blade fixed pitch propeller with a diameter of \nm{0.51 \ m} (as recommended in \cite{GF55II}) followed by the execution of \texttt{JavaProp} results in the propeller map represented by (\ref{eq:AircraftModel_propeller_Ct_map}) and (\ref{eq:AircraftModel_propeller_Cp_map}):
\begin{eqnarray}
\nm{C_T} & = & \nm{f_{P,6}\lrp{J}} \label{eq:AircraftModel_propeller_Ct_map} \\
\nm{C_P} & = & \nm{f_{P,7}\lrp{J}} \label{eq:AircraftModel_propeller_Cp_map}
\end{eqnarray}

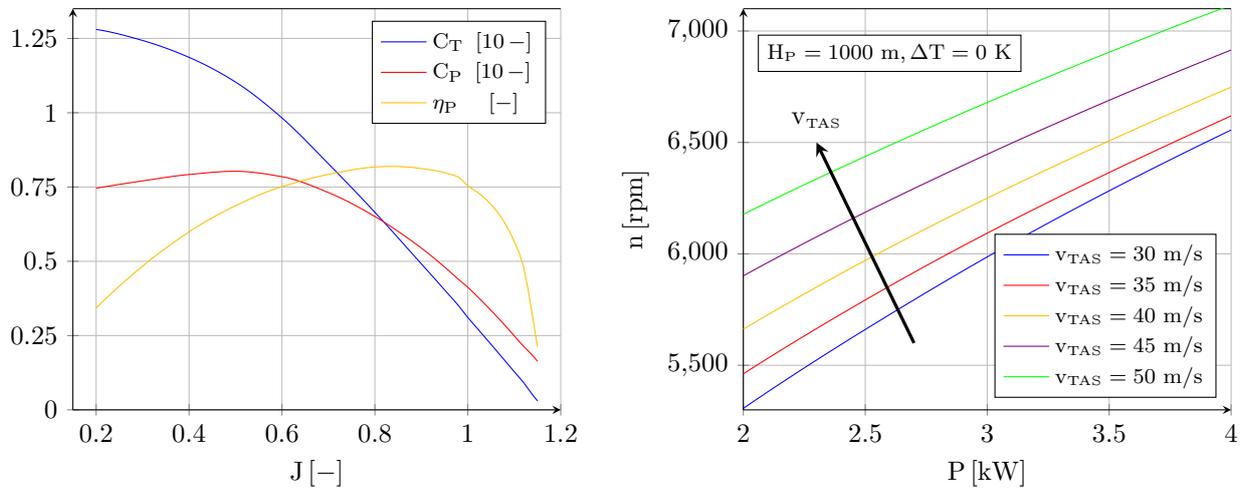
\begin{figure}[h]
\centering
\begin{tikzpicture}
\begin{axis}[
cycle list={{blue,no markers},{red,no markers},{orange!50!yellow, no markers},{violet, no markers},{green, no markers}},
width=8.0cm,
xmin=0.15, xmax=1.2, xtick={0.2,0.4,0.6,0.8,1.0,1.2},
xlabel={\nm{J \left[-\right]}},
xmajorgrids,
ymin=0, ymax=1.35, ytick={0,0.25,0.5,0.75,1,1.25},
ymajorgrids,
axis lines=left,
axis line style={-stealth},
legend entries={\nm{C_T \ \left[10 \, -\right]},\nm{C_P \ \left[10 \, -\right]}, \nm{\eta_P \ \ \ \, \left[-\right]}},
legend pos=north east,
legend style={font=\footnotesize},
legend cell align=left,
]
\draw [ultra thin] (0.15,1.35) -- (1.2,1.35);
\pgfplotstableread{figs/ch03_acft/test8_Ct_Cp_eta_vs_J.txt}\mytable
\addplot table [header=false, x index=0,y index=1] {\mytable};
\addplot table [header=false, x index=0,y index=2] {\mytable};
\addplot table [header=false, x index=0,y index=3] {\mytable};
\end{axis}	
\end{tikzpicture}%
\hskip 10pt
\begin{tikzpicture}
\begin{axis}[
cycle list={{blue,no markers},{red,no markers},{orange!50!yellow, no markers},{violet, no markers},{green, no markers},{magenta, no markers}},
width=8.0cm,
xmin=2, xmax=4, xtick={2.0,2.5,3.0,3.5,4.0},
xlabel={\nm{P \left[kW\right]}},
xmajorgrids,
ymin=5300, ymax=7100, ytick={5500,6000,6500,7000},
ylabel={\nm{n \left[rpm\right]}},
ymajorgrids,
axis lines=left,
axis line style={-stealth},
legend entries={\nm{\vtas = 30 \ m/s}, \nm{\vtas = 35 \ m/s}, \nm{\vtas = 40 \ m/s}, \nm{\vtas = 45 \ m/s}, \nm{\vtas = 50 \ m/s}},
legend pos=south east,
legend style={font=\footnotesize},
legend cell align=left,
]
\draw [ultra thin] (2,7100) -- (4,7100);
\pgfplotstableread{figs/ch03_acft/test9_n_vs_P__vtas.txt}\mytable
\addplot table [header=false, x index=0,y index=1] {\mytable};
\addplot table [header=false, x index=0,y index=2] {\mytable};
\addplot table [header=false, x index=0,y index=3] {\mytable};
\addplot table [header=false, x index=0,y index=4] {\mytable};
\addplot table [header=false, x index=0,y index=5] {\mytable};
\draw [-stealth, very thick] (2.7,5600) -- (2.3,6500);
\path node at (2.3,6600) {\nm{\vtas}};
\path node [draw, shape=rectangle, fill=white] at (2.60,6900) {\footnotesize \nm{\Hp = 1000 \ m, \DeltaT = 0 \ K}};
\end{axis}	
\end{tikzpicture}%
\caption{Propeller efficiency and rotating speed}
\label{fig:aircraft_engine_propeller}
\end{figure}

The propeller map is graphically shown in the left hand side of figure \ref{fig:aircraft_engine_propeller}. The different rates at which \nm{C_T} and \nm{C_P} decrease with J create the classic shape of the relationship between the propeller efficiency \nm{\eta_P} and the advance ratio J, in which there is a relatively wide range of advance ratios that result in elevated efficiencies with sharp drop offs on either side, but particularly at elevated airspeeds. The maximum efficiency is obtained at a specific ratio between the airspeed and the propeller rotating speed \cite{Hill1992,Hitches2015}.

From the aircraft point of view the interesting performances are those of the combined system of piston engine plus propeller, in which the former output power and rotating speed are the inputs to the later. Once the atmospheric properties and the airspeed are known, the shaft power can be obtained through (\ref{eq:AircraftModel_power}), and then there is a single rotating speed \emph{n} for which the power coefficients \nm{C_P} obtained by both its definition (\ref{eq:AircraftModel_propeller_Cp}) and the propeller map (\ref{eq:AircraftModel_propeller_Cp_map}) coincide [A\ref{as:APM_propulsion_shaft}]. The results when flying at different speeds at the same pressure altitude are shown at the right hand side of figure \ref{fig:aircraft_engine_propeller}.
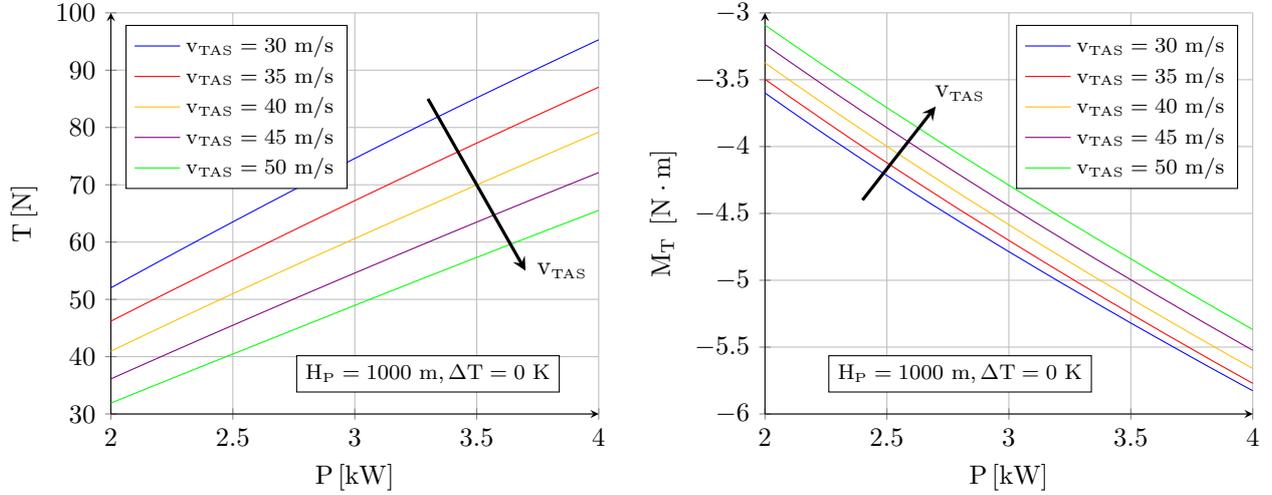
\begin{figure}[h]
\centering
\begin{tikzpicture}
\begin{axis}[
cycle list={{blue,no markers},{red,no markers},{orange!50!yellow, no markers},{violet, no markers},{green, no markers},{magenta, no markers}},
width=8.0cm,
xmin=2, xmax=4, xtick={2.0,2.5,3.0,3.5,4.0},
xlabel={\nm{P \left[kW\right]}},
xmajorgrids,
ymin=30, ymax=100, ytick={30,40,50,60,70,80,90,100},
ylabel={\nm{T \left[N\right]}},
ymajorgrids,
axis lines=left,
axis line style={-stealth},
legend entries={\nm{\vtas = 30 \ m/s}, \nm{\vtas = 35 \ m/s}, \nm{\vtas = 40 \ m/s}, \nm{\vtas = 45 \ m/s}, \nm{\vtas = 50 \ m/s}},
legend pos=north west,
legend style={font=\footnotesize},
legend cell align=left,
]
\pgfplotstableread{figs/ch03_acft/test9_T_vs_P__vtas.txt}\mytable
\addplot table [header=false, x index=0,y index=1] {\mytable};
\addplot table [header=false, x index=0,y index=2] {\mytable};
\addplot table [header=false, x index=0,y index=3] {\mytable};
\addplot table [header=false, x index=0,y index=4] {\mytable};
\addplot table [header=false, x index=0,y index=5] {\mytable};
\draw [-stealth, very thick] (3.3,85) -- (3.7,55);
\path node at (3.85,55) {\nm{\vtas}};
\path node [draw, shape=rectangle, fill=white] at (3.3,37) {\footnotesize \nm{\Hp = 1000 \ m, \DeltaT = 0 \ K}};
\end{axis}	
\end{tikzpicture}%
\hskip 10pt
\begin{tikzpicture}
\begin{axis}[
cycle list={{blue,no markers},{red,no markers},{orange!50!yellow, no markers},{violet, no markers},{green, no markers},{magenta, no markers}},
width=8.0cm,
xmin=2, xmax=4, xtick={2.0,2.5,3.0,3.5,4.0},
xlabel={\nm{P \left[kW\right]}},
xmajorgrids,
ymin=-6, ymax=-3, ytick={-6.0,-5.5,-5.0,-4.5,-4.0,-3.5,-3.0},
ylabel={\nm{M_T \; \left[N \cdot m\right]}},
ymajorgrids,
axis lines=left,
axis line style={-stealth},
legend entries={\nm{\vtas = 30 \ m/s}, \nm{\vtas = 35 \ m/s}, \nm{\vtas = 40 \ m/s}, \nm{\vtas = 45 \ m/s}, \nm{\vtas = 50 \ m/s}},
legend pos=north east,
legend style={font=\footnotesize},
legend cell align=left,
]
\pgfplotstableread{figs/ch03_acft/test9_M_vs_P__vtas.txt}\mytable
\addplot table [header=false, x index=0,y index=1] {\mytable};
\addplot table [header=false, x index=0,y index=2] {\mytable};
\addplot table [header=false, x index=0,y index=3] {\mytable};
\addplot table [header=false, x index=0,y index=4] {\mytable};
\addplot table [header=false, x index=0,y index=5] {\mytable};
\draw [-stealth, very thick] (2.4,-4.4) -- (2.7,-3.7);
\path node at (2.8,-3.6) {\nm{\vtas}};
\path node [draw, shape=rectangle, fill=white] at (2.8,-5.7) {\footnotesize \nm{\Hp = 1000 \ m, \DeltaT = 0 \ K}};
\end{axis}	
\end{tikzpicture}%
\caption{Propeller thrust and torque}
\label{fig:aircraft_engine_propeller2}
\end{figure}

Once the rotating speed has been determined, the advance ratio J is fixed, and hence the thrust coefficient \nm{C_T} can be obtained from (\ref{eq:AircraftModel_propeller_Ct_map}) and the thrust force from (\ref{eq:AircraftModel_propeller_Ct}), with the results shown on the left hand side of figure \ref{fig:aircraft_engine_propeller2}. Note that the shape of the last two figures derives from the definition of J and \nm{\eta_P} in (\ref{eq:AircraftModel_propeller_J}) and (\ref{eq:AircraftModel_propeller_eta}), as they imply inverse relationships between axial and radial speeds as well as between thrust and airspeed.

The propeller torque \nm{M_T} shown at the right hand side of figure \ref{fig:aircraft_engine_propeller2} is the reaction caused by Newton's third law of motion to the mechanical torque transmitted through the shaft. The chosen propeller configuration is inverse pusher\footnote{A pusher propeller is one located behind the fuselage instead of the traditional puller propeller located at its front. Inverse means that it rotates in the same direction that puller propellers (counterclockwise when observed from the front of the aircraft).}, rotating along positive \nm{\iBi}, which means that the propeller torque \nm{M_T} is negative and induces the aircraft to roll to its left.
\neweq{M_T = - \dfrac{P}{2 \; \pi \; n}}{eq:AircraftModel_propeller_torque}

To offset the propeller torque, the pilot shall introduce positive ailerons (\nm{\deltaA > 0}), which creates positive \nm{M_l} but also negative \nm{M_n}, which would initiate a left yaw unless also compensated with positive rudder (\nm{\deltaR > 0}).


\subsection{Correction for Center of Mass Variation}\label{subsec:AircraftModel_propulsion_correction}

As the propulsive force and moment are both aligned with \nm{\iBi} per (\ref{eq:AircraftModel_FproB}) and (\ref{eq:AircraftModel_MproB}), and the position of the aircraft center of mass only varies \nm{1 \ mm} along \nm{\iBiii} and none along \nm{\iBii} from full to empty tank, it can be assumed with very little error that neither \nm{\FPROB} nor \nm{\MPROB} experience any variation with the position of the aircraft center of mass [A\ref{as:APM_propulsion_center_mass}].


\section{Implementation, Validation, and Execution} \label{sec:AircraftModel_implementation}

\begin{figure}[h]
\centering
\begin{tikzpicture}[auto, node distance=2cm,>=latex']
	\node [coordinate](middleinput) {};
	\node [coordinate, above of=middleinput, node distance=2.50cm] (minput){};
		
	\node [block, right of=middleinput, text width=2.5cm, node distance=5.5cm, minimum height=3.0cm] (AERODYNAMICS) {\texttt{AERODYNAMICS}};	
	\node [block, below of=AERODYNAMICS, text width=2.5cm, node distance=3.0cm, minimum height=2.5cm] (PROPULSION) {\texttt{PROPULSION}};	
	\node [block, right of=minput, node distance=3.5cm, minimum height=1.5cm] (INERTIA) {\texttt{INERTIA}};	
	
	\node [coordinate, above of=middleinput, node distance=1.12cm] (wBBNinput){};		
	\node [coordinate, below of=middleinput, node distance=3.75cm] (pTrhoinput){};
	\node [coordinate, below of=middleinput, node distance=3.0cm] (deltaCNTRinput){};	
	\node [coordinate, below of=middleinput, node distance=0.37cm] (vTASBinput){};

	\node [coordinate, right of=pTrhoinput, node distance=3.25cm] (pTrhodot){};
	\filldraw [black] (pTrhodot) circle [radius=1pt];
	\node [coordinate, right of=deltaCNTRinput, node distance=1.75cm] (deltaCNTRdot){};
	\filldraw [black] (deltaCNTRdot) circle [radius=1pt];	
	\node [coordinate, right of=vTASBinput, node distance=2.5cm] (vTASBdot){};
	\filldraw [black] (vTASBdot) circle [radius=1pt];	
	\node [coordinate, above of=AERODYNAMICS, node distance=2.13cm] (xcgdot){};
	\filldraw [black] (xcgdot) circle [radius=1pt];	
		
	\draw [->] (minput) -- node[pos=0.2] {\nm{m}} (INERTIA.west);
	\draw [->] (wBBNinput) -- node[pos=0.15] {\nm{\wNBB}} ($(AERODYNAMICS.west)+(0cm,1.12cm)$);
	\draw [->] (pTrhoinput) -- node[pos=0.25] {\nm{p, \, T, \, \rho}} (pTrhodot) |- ($(PROPULSION.west)-(0cm,0.75cm)$);
	\draw [->] (pTrhodot) |- node [pos=0.7] {\nm{\rho}} ($(AERODYNAMICS.west)-(0cm,1.12cm)$);
    \draw [->] (deltaCNTRinput) -- node[pos=0.4] {\nm{\deltaCNTR}} (deltaCNTRdot) |- node[pos=0.6] {\nm{\deltaT}} (PROPULSION.west);
	\draw [->] (deltaCNTRdot) |- node[pos=0.7] {\nm{\deltaE, \, \deltaA, \, \deltaR}} ($(AERODYNAMICS.west)+(0cm,0.37cm)$);
	\draw [->] (vTASBinput) -- node[pos=0.25] {\nm{\vTASB}} (vTASBdot) |- ($(PROPULSION.west)+(0cm,0.75cm)$);		
	\draw [->] (vTASBdot) |- ($(AERODYNAMICS.west)-(0cm,0.37cm)$);	    
		
	\node [coordinate, right of=PROPULSION, node distance=4.0cm](MPROBoutput) {};
	\node [coordinate, above of=MPROBoutput, node distance=0.75cm](FPROBoutput) {};
	\node [coordinate, below of=MPROBoutput, node distance=0.75cm](Foutput) {};	
	\node [coordinate, right of=INERTIA, node distance=6.0cm](inertiaoutput) {};
	\node [coordinate, above of=inertiaoutput, node distance=0.37cm](IBoutput) {};
	\node [coordinate, below of=inertiaoutput, node distance=0.37cm](xcgoutput) {};
	\node [coordinate, right of=AERODYNAMICS, node distance=4cm](AERODYNAMICSoutput) {};
	\node [coordinate, above of=AERODYNAMICSoutput, node distance=0.37cm](FBAERoutput) {};
	\node [coordinate, below of=AERODYNAMICSoutput, node distance=0.37cm](MBAERoutput) {};
		
	\draw [->] ($(PROPULSION.east)+(0cm,0.75cm)$) -- node[pos=0.7] {\nm{\FPROB}} (FPROBoutput);
	\draw [->] (PROPULSION.east) -- node[pos=0.7] {\nm{\MPROB}} (MPROBoutput);	
	\draw [->] ($(PROPULSION.east)-(0cm,0.75cm)$) -- node[pos=0.6] {\nm{F}} (Foutput);
	\draw [->] ($(INERTIA.east)+(0cm,0.37cm)$) -- node[pos=0.85] {\nm{\vec I^{\sss B}}} (IBoutput);	
	\draw [->] (xcgdot) -- (AERODYNAMICS.north);
	\draw [->] ($(INERTIA.east)-(0cm,0.37cm)$) -- node[pos=0.85] {\nm{\TRBR}} (xcgoutput);		
	\draw [->] ($(AERODYNAMICS.east)+(0cm,0.37cm)$) -- node[pos=0.7] {\nm{\FAERB}} (FBAERoutput);
	\draw [->] ($(AERODYNAMICS.east)-(0cm,0.37cm)$) -- node[pos=0.7] {\nm{\MAERB}} (MBAERoutput);		
		
\end{tikzpicture}
\caption{Aircraft Model flow diagram}
\label{fig:AircraftModel_flow_diagram}
\end{figure}
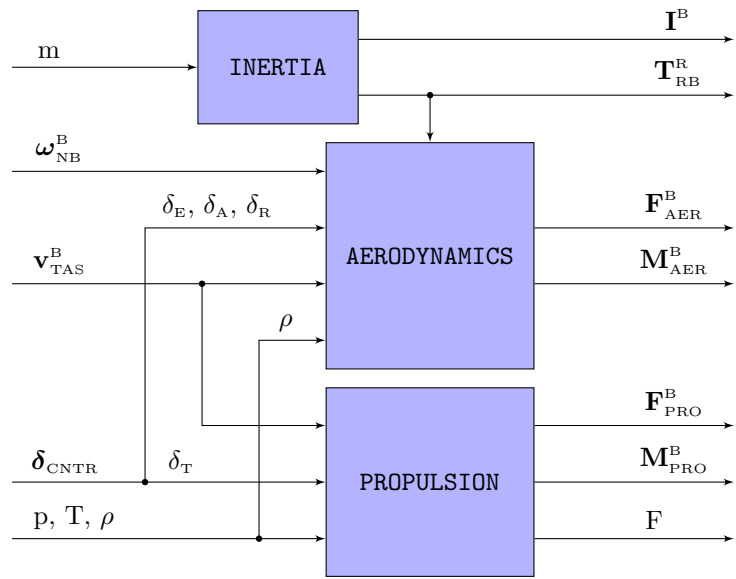

The aircraft model as described in this chapter has been implemented as an object oriented \texttt{C++} library based on the flow diagram shown in figure \ref{fig:AircraftModel_flow_diagram}. Provided with the atmospheric conditions (\nm{p, \, T, \, \varrho}) described in section \ref{sec:EarthModel_ISA}, the control parameters \nm{\deltaCNTR} defined in section \ref{sec:AircraftModel_Control_Parameters}, the air velocity viewed in the body frame \nm{\vTASB} (which is equivalent to the airspeed \nm{\vtas} plus the angles of attack \nm{\alpha} and sideslip \nm{\beta} as explained in section \ref{sec:EquationsMotion_velocity}), the aircraft angular velocity \nm{\wNB} defined in section \ref{sec:EquationsMotion_velocity}, and the aircraft mass \emph{m}, it returns the position of the aircraft center of mass \nm{\TRBR}, its body frame \nm{\FB} moment of inertia \nm{\vec I^{\sss B}}, the \nm{\FB} aerodynamic and propulsive forces (\nm{\FAERB} and \nm{\FPROB}), the fuel consumption F, and the \nm{\FB} aerodynamic and propulsive moments with respect to the aircraft center of mass (\nm{\MAERB} and \nm{\MPROB}). 

\begin{itemize}
\item The \texttt{INERTIA} block contains the position of the aircraft center of mass and its matrix of inertia in both the full and empty fuel tank configurations as described in section \ref{sec:AircraftModel_MassInertia}, and interpolates linearly when provided with an aircraft mass by the integrator. It has been manually validated because of its low complexity.     

\item The \texttt{AERODYNAMICS} block computes the aerodynamic forces and moments based on the process described in section \ref{sec:AircraftModel_aerodynamics}. The different data sets generated by \hypertt{AVL} each containing the aerodynamic forces and moments of an specific aircraft configuration are processed by a \texttt{MatLab}\textsuperscript{\textregistered} script that tabulates the data and saves it as a single text file. This file is then loaded by the corresponding \texttt{C++} class at construction time to save the different tables into memory, and finally these tables are evaluated at execution time to provide the aerodynamic forces and moments required by the trajectory computation process.

The aerodynamics model has been qualitatively validated with the objective of not only achieving longitudinal and lateral stability, but also making sure that the dependencies of all components of the aerodynamic forces and moments with the parameters on which they depend are the right ones, this is, that the model represents a properly designed aircraft. A summary of the validation process is included in section \ref{sec:AircraftModel_aerodynamics}, together with multiple plots describing the results. As the \hypertt{AVL} generated files are based on a detailed geometric description (size, position, and shape) of all the aircraft aerodynamic surfaces, obtaining the model has involved a continuous improvement process of analyzing the results and then tweaking the geometry to correct the observed deficiencies.

\item The \texttt{PROPULSION} block obtains the propulsive forces and moments, together with the fuel consumption, based on the power plant and propeller models described in section \ref{sec:AircraftModel_propulsion}. The propeller map is loaded into memory at construction time, and evaluated when required by the trajectory computation process based on the shaft rotating speed obtained by matching the power generated by the engine and that accepted by the propeller. The validation has been performed verifying that the performances obtained under different conditions conform with the hypotheses listed in section \ref{sec:AircraftModel_propulsion}.
\end{itemize}

In contrast with other modules of the flight simulation, the aircraft model is fully deterministic and always provides the same results if the inputs do not change. The high fidelity simulation always employs the same aircraft and does not consider the influence of its inertial, aerodynamic, and propulsive characteristics on the aircraft control and navigation capabilities. Although this document does not prove it, the author believes that as long as they represent a properly designed and balanced aircraft, the aircraft performances do not have any influence on its navigation capabilities. This does not mean that the control and navigation algorithms are independent from the aircraft performances, but that they are valid, with minor modifications, for other fixed wing aircraft with different performances than those described in this chapter.

 \cleardoublepage
\chapter{Physics of Flight} \label{cha:FlightPhysics}

This chapter introduces the differential equations that describe the physics of flight and whose integration results in the real aircraft trajectory. These equations establish the interrelationships among the variables describing the aircraft state at a given moment of time, such as its position, velocity, and attitude, with their variations with time and the forces and moments that cause those variations. Their integration results in a time series of aircraft states that together comprise the aircraft actual trajectory \nm{\xTRUTH\lrp{t_t}}, first introduced in section \ref{sec:Intro_traj} and defined in detail in section \ref{sec:EquationsMotion_AT} below. Note that although the real or actual trajectory is continuous in the real world, in the simulation it is computed at a discrete rate of \nm{500 \ Hz}, with \nm{t_t = t \cdot \DeltatTRUTH}.

\begin{figure}[h]
\centering
\begin{tikzpicture}[auto, node distance=2cm,>=latex']
	\node [coordinate](x0input) {};
	\node [coordinate, below of=x0input, node distance=2.2cm] (deltaCNTRinput){};
	\node [block, right of=x0input, minimum width=4.5cm, node distance=5.0cm, align=center, minimum height=2.0cm] (FLIGHT PHYSICS) {\texttt{FLIGHT PHYSICS}};
	\node [coordinate, below of=FLIGHT PHYSICS, node distance=2.2cm] (midblocks){};
	\node [block, right of=midblocks, minimum width=2.0cm, node distance=1.25 cm, align=center, minimum height=1.25cm] (EARTH) {\texttt{EARTH}};	
	\node [block, left of=midblocks, minimum width=2.0cm, node distance=1.25 cm, align=center, minimum height=1.25cm] (AIRCRAFT) {\texttt{AIRCRAFT}};	
	\node [coordinate, right of=FLIGHT PHYSICS, node distance=3.5cm] (crosspoint1){};
	\node [coordinate, right of=FLIGHT PHYSICS, node distance=6.0cm] (xTRUTHoutput){};
	\node [coordinate, below of=crosspoint1, node distance=3.4cm] (crosspoint2){};
	\node [coordinate, left of=crosspoint2, node distance=2.25cm] (crosspoint3){};
	
	\draw [->] (EARTH.north) -- ($(FLIGHT PHYSICS.south)+(1.25cm,0cm)$);
	\draw [->] (EARTH.west) -- (AIRCRAFT.east);
	\draw [->] (AIRCRAFT.north) -- ($(FLIGHT PHYSICS.south)-(1.25cm,0cm)$);	
	\draw [->] (x0input) -- node[pos=0.5] {\nm{\xveczero = \xTRUTHzero}} (FLIGHT PHYSICS.west);
	\filldraw [black] (crosspoint1) circle [radius=1pt];
	\filldraw [black] (crosspoint3) circle [radius=1pt];
	\draw [->] (crosspoint1) |- (crosspoint3) -- (EARTH.south);	
	\draw [->] (crosspoint3) -| (AIRCRAFT.south);	
	\draw [->] (deltaCNTRinput) -- node[pos=0.4] {\nm{\deltaCNTR\lrp{t_c}}} (AIRCRAFT.west);
	\draw [->] (FLIGHT PHYSICS.east) -- node[pos=0.5] {\nm{\xvec\lrp{t_t} = \xTRUTH\lrp{t_t}}} (xTRUTHoutput);
\end{tikzpicture}
\caption{Trajectory computation flow diagram}
\label{fig:FlightPhysics_flow_diagram}
\end{figure}
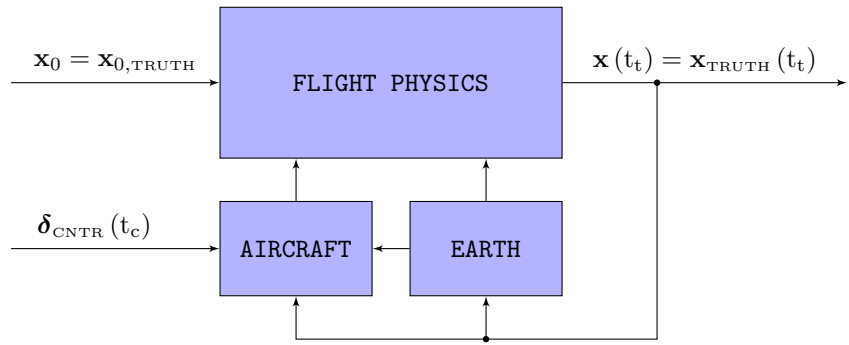

The process is graphically depicted in figure \ref{fig:FlightPhysics_flow_diagram}, in which the variation with time of the actual state vector \nm{\xvec\lrp{t_t} = \xTRUTH\lrp{t_t}} can be obtained from its initial value \nm{\xveczero = \xTRUTHzero} and the time variation of the aircraft control parameters \nm{\deltaCNTR \lrp{t_c}} updated at \nm{50 \ Hz} by the control system described in section \ref{sec:GNC_Control}, where \nm{t_c = c \cdot \DeltatCNTR}. This process requires the atmospheric properties and aircraft performances described in chapters \ref{cha:EarthModel} and \ref{cha:AircraftModel}, respectively.

The selection of the most adequate frame on which to apply the equations of motion is discussed in section \ref{sec:EquationsMotion_inertial_selection}. Next, section \ref{sec:EquationsMotion_velocity} introduces the various required linear and angular velocities. After defining the real or actual aircraft trajectory in section \ref{sec:EquationsMotion_AT}, the equations of motion representing the aircraft flight are established in section \ref{sec:EquationsMotion_equations_motion}, and their integration to obtain the actual trajectory discussed in section \ref{sec:EquationsMotion_integration}. The chapter concludes in section \ref{sec:EquationsMotion_implementation} with a few remarks about implementation and validation.

	
\section{Selection of Inertial Frame}\label{sec:EquationsMotion_inertial_selection}
	
The equations describing the motion of any object can be established either in a non-inertial frame, where it is necessary to consider the inertial forces and moments induced by the accelerations present in the frame, or in an inertial one, which is the approach adopted in this document. An \emph{inertial frame} is that in which bodies whose net force acting upon them is zero do not experience any acceleration and remain either at rest or moving at constant velocity in a straight line. A reference frame centered at the Sun center of mass and with axes fixed with respect to other stars is always considered inertial for the motion of objects in the Earth atmosphere [A\ref{as:MOTION_inertial_system}]. 

Any reference frame can be considered inertial for the analysis of the motion of a given object if the accelerations (linear and angular) of that frame with respect to an accepted inertial system can be discarded when compared with the accelerations characteristic of the movement being studied. The inertial frame \nm{\FI} employed in the simulation is one centered at the Earth center of mass and whose axes do not rotate with respect to any stars other than the Sun \cite{Farrell2008,Groves2008,Chatfield1997,Rogers2007,Etkin1972,Ashley1974,Miele1962,Titterton2004}. Such a frame does not rotate with respect to the accepted inertial frame and its linear acceleration can be approximated by \nm{\lrp{\omega_{\sss E-S}^2 \cdot R_{\sss E-S}}}, where \nm{\omega_{\sss E-S}} is the rotation speed caused by the Earth movement around the Sun, and \nm{R_{\sss E-S}} is the distance between the Earth and the Sun (appendix \ref{cha:PhysicalConstants}). Its value (\nm{5.95 \cdot 10^{-3} \; m/s^2}) is sufficiently small when compared with the accelerations characteristic of flight\footnote{As an example, this value represents only 0.6\textperthousand \ of the gravitational acceleration.} as to validate the use of \nm{\FI} as inertial within this document, meaning that the accelerations caused by the translation motion of the Earth center of mass around the Sun can be discarded [A\ref{as:MOTION_earth_translation}].


\section{Linear and Angular Velocities}\label{sec:EquationsMotion_velocity}

Prior to the introduction of the equations of motion in section \ref{sec:EquationsMotion_equations_motion}, it is helpful to review the different linear and angular velocities that appear on them. This way the equations of motion can be presented with more ease as many of the involved concepts are explained in this section.
\begin{itemize}

\item The aircraft \emph{absolute} or \emph{ground velocity} \nm{\vec v} is that of its center of mass (\nm{\ON \equiv \OB \equiv \OG}) with respect to the \hypertt{ECEF} frame \nm{\FE} (section \ref{subsec:RefSystems_E}), this is, represents how the aircraft moves with respect to the Earth surface. It is the sum of the airspeed \nm{\vTAS}, the wind field \nm{\vWIND}, and the turbulence \nm{\vTURB}\footnote{Wind field is a synonym for low frequency wind component, and turbulence for high frequency wind component (section \ref{sec:EarthModel_WIND}).}:
\neweq{\vvec = \vEN = \vEB = \vTAS + \vWIND + \vTURB}{eq:EquationsMotion_auxiliary_ground_speed1}

As a free vector, the absolute velocity can be viewed in different coordinate frames:
\neweq{\vN = \vec g_{\ds{\vec \zeta_{{\sss NB}*}}} \big(\vB\big)}{eq:EquationsMotion_auxiliary_ground_speed1b}

The ground velocity has a single component (\nm{v = \|\vvec\|}) when viewed in \nm{\FG} (section \ref{subsec:RefSystems_G}), although its \hypertt{NED} or \nm{\FN} form is more useful because its three components have very clear physical meanings (North, East and down speeds, section \ref{subsec:RefSystems_N}). The ground velocity \nm{\FN} components are geometrically related with the time derivatives of the geodetic coordinates \nm{\TEgdt} defined in section \ref{subsubsec:RefSystems_E_GeodeticCoord} because of the alignment between the geodetic coordinates and the \hypertt{NED} frame. They also depend on the ellipsoid radii of curvature defined in section \ref{sec:EarthModel_WGS84}:
\begin{eqnarray}
\nm{\vG} & = & \nm{\lrsb{v \ \ 0 \ \ 0}^T} \label{eq:EquationsMotion_auxiliary_ground_speed2}\\
\nm{\vN} & = & \nm{\vec g_{\ds{\vec \zeta_{{\sss NG}*}}} \big(\vG\big) = \lrsb{\dot \varphi \; \lrsb{M\lrp{\varphi} + h} \ \ \ \ \dot \lambda \; \lrsb{N\lrp{\varphi} + h} \; \cos \varphi\ \ \ \ - \dot h}^T} \label{eq:EquationsMotion_auxiliary_ground_speed3}
\end{eqnarray}

\item The \emph{air velocity} \nm{\vTAS} is the speed of the aircraft center of mass (\nm{\ON \equiv \OB \equiv \OW}) with respect to the mass of air that surrounds it, and represents the intensity and direction with which the air stream interacts with the aircraft. It has a single component (\nm{\vtas = \|\vTAS\|}) when expressed in \nm{\FW} (section \ref{subsec:RefSystems_W}), while its \nm{\FB} form depends on the angles of attack and sideslip defined in section \ref{subsec:RefSystems_W}:
\begin{eqnarray}
\nm{\vTASW} & = & \nm{\lrsb{\vtas \ \ 0 \ \ 0}^T} \label {eq:EquationsMotion_auxiliary_airspeed1} \\
\nm{\vTASB} & = & \nm{\vec g_{\ds{\vec \zeta_{{\sss BW}*}}} \big(\vTASW\big) = \lrsb{\vtas \; \cos \beta \; \cos \alpha \ \ \ \ \vtas \; \sin \beta \ \ \ \ \vtas \; \cos \beta \; \sin \alpha}^T} \label {eq:EquationsMotion_auxiliary_airspeed2} 
\end{eqnarray}

As a matter of fact, it is often more practical to replace \nm{\vTASB} by its norm and the angles of attack and sideslip:
\neweq{\vtas = \|\vTASB\| \ \ \ \ \ \ \ \alpha = \arctan{\dfrac{\vTASBiii}{\vTASBi}} \ \ \ \ \ \ \ \beta = \arcsin{\dfrac{\vTASBii}{\vtas}}} {eq:EquationsMotion_auxiliary_airspeed03}

\item The wind velocity is defined as the velocity of the air mass surrounding the aircraft with respect to the Earth frame \nm{\FE}, and can be divided into its low frequency (\emph{wind field}) and high frequency (\emph{turbulence}) components (section \ref{sec:EarthModel_WIND}). Expressions (\ref{eq:EarthModel_WIND_low_freq}) and (\ref{eq:EarthModel_WIND_turbulence}) provide the value of \nm{\vWINDN} and \nm{\vTURBN}, respectively.

\item The aircraft \emph{inertial velocity} \nm{\vec v_{\sss IN}} is that of its center of mass (\nm{\ON \equiv \OB}) with respect to the inertial frame \nm{\FI}, and is indirectly employed in the obtainment of the equations of motion in section \ref{sec:EquationsMotion_equations_motion}. Based on the composition of velocities discussed in \cite{LIE}, it responds to:
\neweq{\vIN = \vEN + \vIE + \wIEskew \; \TEN}{eq:EquationsMotion_auxiliary_inertial_speed1}

where \nm{\vEN} is the ground velocity defined above, \nm{\wIE} is the Earth angular velocity provided by (\ref{eq:EquationsMotion_auxiliar_angular_velocities_earth1}), \nm{\TEN} is the aircraft position with respect to \nm{\OECEF}, and \nm{\vIE} is zero because the \nm{\FE} and \nm{\FI} frames both have their origin located at the Earth center of mass:
\neweq{\vIN = \vEN + \wIEskew \; \TEN}{eq:EquationsMotion_auxiliary_inertial_speed2}

\item The aircraft \emph{inertial angular velocity} can be decomposed into the sum of three angular velocities: the aircraft angular velocity (that of \nm{\FB} with respect to \nm{\FN}), the motion angular velocity (\nm{\FN} with respect to \nm{\FE}), and the Earth angular velocity (\nm{\FE} with respect to \nm{\FI}):
\begin{eqnarray}
\nm{\wIB} & = & \nm{\wNB + \wEN + \wIE} \label{eq:EquationsMotion_auxiliary_angular_velocities01} \\
\nm{\wIBB} & = & \nm{\wNBB + \wENB + \wIEB = \wNBB + \vec {Ad}_{\ds{\vec q_{\sss NB}}}^{-1} \ \wENN + \vec {Ad}_{\ds{\vec q_{\sss EB}}}^{-1} \ \wIEE} \label{eq:EquationsMotion_auxiliary_angular_velocities02}
\end{eqnarray}

\item The \emph{aircraft angular velocity} \nm{\wNB} represents the variation with time of \nm{\FB} representing the aircraft with respect to its associated \hypertt{NED} frame \nm{\FN} containing the North, East, and Down directions. Its components when expressed in \nm{\FB} are historically known as \nm{\{p, \, q, \, r\}}, and its relationship with the quaternion \nm{\qNB} representing the aircraft attitude can be obtained from \cite{LIE}:
\neweq{\wNBB = \lrsb{p \ \ q \ \ r}^T = 2 \, \qNB^{\ast} \otimes \qNBdot}{eq:EquationsMotion_auxiliary_angular_velocicites_aircraft1}

\item The \emph{motion angular velocity} \nm{\wEN} represents the rotation experienced by any object that moves across the Earth surface without modifying its attitude with respect to the \hypertt{WGS84} ellipsoid because of the curvature of the Earth surface. It can be obtained by employing the rotation vector to represent the two consecutive rotations listed in section \ref{subsec:RefSystems_N}: 
\neweq{\wEN = \dot{\lambda} \; \iEiii - \dot{\varphi} \; \iNii \ \ \rightarrow \ \ \wENN = \dot{\lambda} \; \vec g_{\ds{\vec \zeta_{{\sss EN}*}}}^{-1} \big(\iEiii\big) - \dot{\varphi} \; \iNii = \lrsb{\dot{\lambda} \; \cos \varphi \ \ \ \ - \dot{\varphi} \ \ \ \ - \dot{\lambda} \; \sin \varphi }^T}{eq:EquationsMotion_auxiliary_angular_velocities_motion}

By means of (\ref{eq:EquationsMotion_auxiliary_ground_speed3}), this is equivalent to:
\neweq{\wENN = \lrsb{\dfrac{\vNii}{N\lrp{\varphi} + h} \ \ \ \ \dfrac{- \; \vNi}{M\lrp{\varphi} + h} \ \ \ \ \dfrac{- \; \vNii \; \tan \varphi}{N\lrp{\varphi} + h} }^T}{eq:EquationsMotion_auxiliary_angular_velocities_motion2}

The frequently used derivative or Jacobian of \nm{\wENN} with respect to \nm{\vN} hence responds to:
\neweq{\vec J_{\ds{\; \vN}}^{\ds{\; \wENN}} = \begin{bmatrix} 0 & \nm{\dfrac{1}{N + h}} & 0 \\ \nm{\dfrac{-1}{M + h}} & 0 & 0 \\ 0 & \nm{\dfrac{- \tan \varphi}{N + h}} & 0 \end{bmatrix} }{eq:EquationsMotion_jac_wENN_vN}

\item The \emph{Earth angular velocity} \nm{\wIE} is caused by the rotation of the \hypertt{WGS84} ellipsoid (section \ref{sec:EarthModel_WGS84}) around its symmetry axis at a constant rate \nm{\omegaE} (appendix \ref{cha:PhysicalConstants}) [A\ref{as:WGS84_rotation}], and represents the full turn experienced by any object fixed with respect to the Earth surface (\hypertt{NED}) every twenty four hours:
\begin{eqnarray}
\nm{\wIEE} & = & \nm{\omegaE \; \iEiii = \lrsb{0 \ \ 0 \ \ \omegaE }^T} \label{eq:EquationsMotion_auxiliar_angular_velocities_earth1} \\
\nm{\wIEN} & = & \nm{\vec {Ad}_{\ds{\vec q_{\sss EN}}}^{-1} \ \wIEE = \lrsb{\omegaE \; \cos \varphi \ \ \ \ 0 \ \ \ \ - \; \omegaE \; \sin \varphi }^T} \label{eq:EquationsMotion_auxiliar_angular_velocities_earth3} 
\end{eqnarray}
\end{itemize}


\section{Actual Trajectory}\label{sec:EquationsMotion_AT}

The main objective of this chapter is to obtain a set of ordinary differential equations whose integration from known initial conditions \nm{\xveczero = \xTRUTHzero}, and considering the variation with time of the throttle and aerodynamic control surfaces \nm{\deltaCNTR\lrp{t_c}} as an input, provides the aircraft \emph{actual trajectory} \nm{\xvec\lrp{t_t} = \xTRUTH\lrp{t_t}}, also known as the real or true trajectory (\hypertt{AT}). First introduced in section \ref{sec:Intro_traj}, the actual trajectory contains the variation with time of a group of variables that together describe the aircraft state \cite{Stevens2003,Etkin1972}, in the sense that the values of all other variables can be obtained with the algebraic equations shown in this chapter together with those of chapters \ref{cha:EarthModel} and \ref{cha:AircraftModel}. The system of equations responds to:
\begin{eqnarray}
\nm{\dot {\vec x}_{\sss TRUTH}} & = & \nm{\vec f_{TR}\big(\xTRUTH\lrp{t}, \, \deltaCNTR\lrp{t}, \, t\big)}\label{eq:EquationsMotion_equations_motion01} \\
\nm{\xTRUTHzero} & = & \nm{\xTRUTH\lrp{t = t_0}}\label{eq:EquationsMotion_equations_motion02} 
\end{eqnarray}

where the variation with time of the control parameters \nm{\deltaCNTR\lrp{t}} is considered an input, as it is provided by the control system described in section \ref{sec:GNC_Control}. The actual trajectory is hence a time stamped series of \emph{actual state vectors} \nm{\xvec = \xTRUTH}, defined as:
\neweq{\xvec = \xTRUTH = \lrsb{\TEgdt \ \ \vB \ \ \qNB \ \ \wIBB \ \ m}^T} {eq:EquationsMotion_state_vector}

\begin{itemize}
\item Position of the aircraft center of mass, represented by its geodetic coordinates \nm{\TEgdt} defined in section \ref{subsubsec:RefSystems_E_GeodeticCoord}.
\item Ground velocity \nm{\vB} viewed in \nm{\FB}, defined in section \ref{sec:EquationsMotion_velocity}.
\item Attitude \nm{\qNB} with respect to \nm{\FN}, defined in section \ref{subsec:RefSystems_B}.
\item Inertial angular velocity \nm{\wIBB} viewed in \nm{\FB}, defined in section \ref{sec:EquationsMotion_velocity}.
\item Mass of the aircraft, which diminishes with time because of the fuel consumption.
\end{itemize}


\section{Equations of Motion}\label{sec:EquationsMotion_equations_motion}

The equations of motion comprise the system of first order differential equations (\ref{eq:EquationsMotion_equations_motion01}) that provide the derivatives with time of the different components of the state vector, so they can be integrated to obtain the actual trajectory. The equation corresponding to each state vector component is treated in its own section below, while their integration is described in section \ref{sec:EquationsMotion_integration}.


\subsection{Kinematic Equations}\label{subsec:EquationsMotion_kinematic_equations}

The kinematic equations provide the time derivatives of the geodetic coordinates \nm{\TEgdt} and are obtained by reversing (\ref{eq:EquationsMotion_auxiliary_ground_speed3}):
\neweq{\TEgdtdot = \lrsb{\dfrac{\vNii}{\lrsb{N\lrp{\varphi} + h} \; \cos\varphi} \ \ \ \ \dfrac{\vNi}{M\lrp{\varphi} + h} \ \ \ \ - \vNiii}^T}{eq:EquationsMotion_equations_kinematic}

Its derivative or Jacobian with respect to the velocity is hence the following:
\neweq{\vec J_{\ds{\; \vN}}^{\ds{\; \TEgdtdot}} = \begin{bmatrix} 0 & \nm{\dfrac{1}{\lrp{N + h} \, \cos\varphi}} & 0 \\ \nm{\dfrac{1}{M + h}} & 0 & 0 \\ 0 & 0 & \nm{- 1} \end{bmatrix}}{eq:EquationsMotion_jac_TEgdt_vN}

Making use of the section \ref{sec:EquationsMotion_velocity} relationships, the dependencies of \nm{\TEgdtdot} are the following:
\neweq{\TEgdtdot = \vec f_{T,TR} \lrp{\TEgdt, \, \vB, \, \qNB }}{eq:EquationsMotion_equations_kinematic_depend}


\subsection{Force Equations}\label{subsec:EquationsMotion_force_equations}

The force equations provide the time derivatives of the absolute velocity \nm{\vB} and result from the equilibrium of forces applied to the aircraft center of mass when viewed in the inertial frame \nm{\FI}:
\neweq{\sum {\vec F}^{\sss I} = \deriv{\lrp{m \; \vIB}^{\sss I}} = m \; \vIBIdot = m \; \vec a_{\sss IB}^{\sss I}} {eq:EquationsMotion_equations_force1}

where the mass variation with time is considered slow enough as to be quasi stationary [A\ref{as:MOTION_mass_stationary}]. The application of the composition of accelerations \cite{LIE} considering \nm{\FI} as \nm{F_{\sss0}}, \nm{\FE} as \nm{F_{\sss1}}, and \nm{\FB} as \nm{F_{\sss2}} results in the following expression for the inertial acceleration of the body frame \nm{\vec a_{\sss IB}^{\sss I}}:
\neweq{\vec a_{\sss IB}^{\sss I} = \vec a_{\sss EB}^{\sss I} + \lrp{ \vec a_{\sss IE}^{\sss I} + \alphaIEIskew \; \vec x_{\sss EB}^{\sss I} + \wIEIskew \; \wIEIskew \; \vec x_{\sss EB}^{\sss I} } +  2 \; \wIEIskew \; \vEBI} {eq:EquationsMotion_equations_force2}

The first term at the right hand side of (\ref {eq:EquationsMotion_equations_force2}) is the linear acceleration \nm{\vec a_{\sss EB}^{\sss I}}, which is the time derivative of the aircraft absolute velocity \nm{\vec v_{\sss EB}} viewed in \nm{\FI}. It can be converted into a function of the aircraft and motion angular velocities (\nm{\wEB = \wNB + \wEN}) and the aircraft absolute velocity \nm{\vec v = \vec v_{\sss EB}}:
\begin{eqnarray}
\nm{\vec a_{\sss EB}^{\sss I}} & = & \nm{\vec g_{\ds{\vec \zeta_{{\sss IE}*}}} \big(\vec a_{\sss EB}^{\sss E}\big) = \vec g_{\ds{\vec q_{{\sss IE}*}}} \big(\vec a_{\sss EB}^{\sss E}\big) = \RIE \; \vec a_{\sss EB}^{\sss E} = \RIE \; \vEBEdot = \RIE \; \deriv{\big(\vec g_{\ds{\vec \zeta_{{\sss EB}*}}} \big(\vEBB\big)\big)} = \RIE \; \deriv{\big(\vec g_{\ds{\vec q_{{\sss EB}*}}} \big(\vEBB\big)\big)}} \nonumber \\
& = & \nm{\RIE \; \deriv{\lrp{\REB \; \vEBB}} = \RIE \; \lrp{\REBdot \; \vEBB + \REB \; \vEBBdot} = \RIB \; \lrp{\wEBBskew \; \vEBB + \vEBBdot}} \label{eq:EquationsMotion_equations_force3a} \\
\nm{\vec a_{\sss EB}^{\sss B}} & = & \nm{\wEBBskew \; \vEBB + \vEBBdot = \wEBBskew \; \vB + \vBdot} \label{eq:EquationsMotion_equations_force3b}
\end{eqnarray}

Noting that \nm{\vEB = \vEN = \vvec} per (\ref{eq:EquationsMotion_auxiliary_ground_speed1}), the same process can be repeated replacing \nm{\FB} by \nm{\FN}:
\begin{eqnarray}
\nm{\vec a_{\sss EB}^{\sss I}} & = & \nm{\RIE \; \vEBEdot = \RIE \; \vENEdot = \RIE \; \deriv{\big(\vec g_{\ds{\vec \zeta_{{\sss EN}*}}} \big(\vENN\big)\big)} = \RIE \; \deriv{\big(\vec g_{\ds{\vec q_{{\sss EN}*}}} \big(\vENN\big)\big)}} \nonumber \\
& = & \nm{\RIE \; \deriv{\lrp{\REN \; \vENN}} = \RIE \; \lrp{\RENdot \; \vENN + \REN \; \vENNdot} = \RIN \; \lrp{\wENNskew \; \vENN + \vENNdot}} \label{eq:EquationsMotion_equations_force3_Na} \\
\nm{\vec a_{\sss EB}^{\sss N}} & = & \nm{\wENNskew \; \vENN + \vENNdot = \wENNskew \; \vN + \vNdot} \label{eq:EquationsMotion_equations_force3_Nb}
\end{eqnarray}

The process is significantly more simple if performed in the \nm{\FE} frame:
\begin{eqnarray}
\nm{\vec a_{\sss EB}^{\sss I}} & = & \nm{\RIE \; \vEBEdot} \label{eq:EquationsMotion_equations_force3_Ea} \\
\nm{\vec a_{\sss EB}^{\sss E}} & = & \nm{\vEBEdot = \vENEdot = \vEdot} \label{eq:EquationsMotion_equations_force3_Eb} 
\end{eqnarray}

The second term of (\ref{eq:EquationsMotion_equations_force2}), called the \emph{transport acceleration} \nm{\atrs}, can be simplified because the linear acceleration \nm{\vec a_{\sss IE}} of \nm{\FE} with respect to \nm{\FI} is zero as both have their origins located at the Earth center of mass (\nm{{\vec T}_{\sss IE}^{\sss I} = \vec 0}), while the angular acceleration \nm{\vec \alpha_{\sss IE}} is also zero as \nm{\FE} rotates about \nm{\iEiii} at a constant rate \nm{\omegaE}.
\neweq{\atrsI = \wIEIskew \; \wIEIskew \; \vec T_{\sss EB}^{\sss I}} {eq:EquationsMotion_equations_force4} 

The transport acceleration hence coincides with the \emph{centripetal acceleration} caused by the Earth rotation about its symmetry axis, this is, it is the opposite of the centrifugal acceleration \nm{\ac} provided by (\ref{eq:EarthModel_GEOP_ac}). Its \hypertt{NED} form \nm{\atrsN} can be directly obtained from the Earth angular velocity (\ref{eq:EquationsMotion_auxiliar_angular_velocities_earth3}) and the definition of the Cartesian coordinates \nm{\TEcar} (\ref{eq:RefSystems_E_geod2geoc}):
\neweq{\atrsN = - \ac^{\sss N} = \wIENskew \; \wIENskew \; \vec T_{\sss EB}^{\sss N} = \lrsb{\omegaE^2 \; \lrp{N + h} \; \sin \varphi \; \cos \varphi \ \ \ \ 0 \ \ \ \ \omegaE^2 \; \lrp{N + h} \, \cos^2 \varphi}^T}{eq:EquationsMotion_equations_force5}

The third term of (\ref{eq:EquationsMotion_equations_force2}) is called the \emph{Coriolis acceleration} \nm{\acor}. Its \hypertt{NED} form \nm{\acorN} can also be obtained from the Earth angular velocity \nm{\wIE} and the aircraft absolute velocity \nm{\vEB}:
\neweq{\acorN =  2 \; \wIENskew \; \vEBN = \lrsb{2 \; \omegaE \; \vNii \; \sin \varphi \ \ \ \ 2 \; \omegaE \; \lrp{- \vNi \; \sin \varphi - \vNiii \; \cos \varphi} \ \ \ \ 2 \; \omegaE \; \vNii \; \cos \varphi}^T}{eq:EquationsMotion_equations_force6}

Its derivative or Jacobian with respect to the velocity, employed later in this document, responds to:
\neweq{\vec J_{\ds{\; \vN}}^{\ds{\; \acorN}} = 2 \; \wIENskew = 2 \; \omegaE \; \begin{bmatrix} 0 & \nm{\sin \varphi} & 0 \\ \nm{- \sin \varphi} & 0 & \nm{- \cos \varphi} \\ 0 & \nm{\cos \varphi} & 0 \end{bmatrix}}{eq:EquationsMotion_jac_acorN_vN}

The Coriolis acceleration and Jacobian may also be employed viewed in the \nm{\FE} frame:
\begin{eqnarray}
\nm{\acorE} & = & \nm{2 \; \wIEEskew \; \vEBE = 2 \; \omegaE \lrsb{- \vEii \ \ \vEi \ \ 0}^T} \label{eq:EquationsMotion_equations_force6bis} \\
\nm{\vec J_{\ds{\; \vE}}^{\ds{\; \acorE}}} & = & \nm{2 \; \wIEEskew = 2 \; \omegaE \; \begin{bmatrix} 0 & - 1 & 0 \\ 1 & 0 & 0 \\ 0 & 0 & 0 \end{bmatrix}} \label{eq:EquationsMotion_jac_acorE_vE}
\end{eqnarray}

The introduction of these terms into (\ref{eq:EquationsMotion_equations_force1}) while converting to \nm{\FB}, \nm{\FN}, or \nm{\FE} results in:
\begin{eqnarray}
\nm{\sum \vec F^{\sss B}} & = & \nm{\vec g_{\ds{\vec \zeta_{{\sss IB}*}}}^{-1} \Big(\sum \vec F^{\sss I}\Big) = \FAERB + \FPROB + m \ \vec g_{\ds{\vec \zeta_{{\sss NB}*}}}^{-1} \big(\gN\big)} \nonumber \\
& = & \nm{ m \, \Big(\wEBBskew \, \vB + \vBdot + \vec g_{\ds{\vec \zeta_{{\sss NB}*}}}^{-1} \big(\atrsN\big) + \vec g_{\ds{\vec \zeta_{{\sss NB}*}}}^{-1} \big(\acorN\big) \Big)}\label{eq:EquationsMotion_equations_force7} \\
\nm{\sum \vec F^{\sss N}} & = & \nm{\vec g_{\ds{\vec \zeta_{{\sss IN}*}}}^{-1} \Big(\sum \vec F^{\sss I}\Big) = \vec g_{\ds{\vec \zeta_{{\sss NB}*}}} \big(\FAERB + \FPROB \big) + m \ \gN} \nonumber \\
& = & \nm{m \, \lrp{\wENNskew \, \vN + \vNdot + \atrsN + \acorN}}\label{eq:EquationsMotion_equations_force7_N} \\
\nm{\sum \vec F^{\sss E}} & = & \nm{\vec g_{\ds{\vec \zeta_{{\sss IE}*}}}^{-1} \Big(\sum \vec F^{\sss I}\Big) = \vec g_{\ds{\vec \zeta_{{\sss EB}*}}} \big(\FAERB + \FPROB \big) + m \ \vec g_{\ds{\vec \zeta_{{\sss EN}*}}} \big(\gN\big)} \nonumber \\
 & = & \nm{ m \, \Big(\vEdot + \vec g_{\ds{\vec \zeta_{{\sss EN}*}}} \big(\atrsN\big) + \vec g_{\ds{\vec \zeta_{{\sss EN}*}}} \big(\acorN\big) \Big)} \label{eq:EquationsMotion_equations_force7_E}
\end{eqnarray}

As \nm{\gc = \vec g + \ac = \vec g - \atrs} per (\ref{eq:EarthModel_GEOP_f2}) and \nm{\vEB = \vEN = \vvec} per (\ref{eq:EquationsMotion_auxiliary_ground_speed1}), the final expression for the time derivative of the aircraft linear velocity is the following:
\begin{eqnarray}
\nm{\vBdot} & = & \nm{\dfrac{\FAERB + \FPROB}{m} - \wEBBskew \; \vB + \vec g_{\ds{\vec \zeta_{{\sss NB}*}}}^{-1} \big(\gcNREAL - \acorN\big)} \label{eq:EquationsMotion_equations_force8} \\
\nm{\vNdot} & = & \nm{\vec g_{\ds{\vec \zeta_{{\sss NB}*}}} \big(\dfrac{\FAERB + \FPROB}{m}\big) - \wENNskew \; \vN + \gcNREAL - \acorN} \label{eq:EquationsMotion_equations_force8_N} \\
\nm{\vEdot} & = & \nm{\vec g_{\ds{\vec \zeta_{{\sss EB}*}}} \big(\dfrac{\FAERB + \FPROB}{m}\big) + \vec g_{\ds{\vec \zeta_{{\sss EN}*}}} \big(\gcNREAL - \acorN\big)} \label{eq:EquationsMotion_equations_force8_E}
\end{eqnarray}

Introducing into (\ref{eq:EquationsMotion_equations_force8}) the expressions for the Earth model (section \ref{cha:EarthModel}), the aircraft performances (section \ref{cha:AircraftModel}), and the velocities of section \ref{sec:EquationsMotion_velocity}, the dependencies of \nm{\vBdot} are the following:
\neweq{\vBdot = \vec f_{v,TR} \lrp{\TEgdt, \, \vB, \, \qNB, \, \wIBB, \, m, \, t, \, \deltaCNTR }}{eq:EquationsMotion_equations_force_depend}

On the other hand, the ground velocity time derivative viewed in \nm{\FN} provided by (\ref{eq:EquationsMotion_equations_force8_N}) is employed by the inertial navigation system described in chapter \ref{cha:nav}. Note that the relationship between \nm{\vNdot} and \nm{\vBdot}, which is key for the design of the navigation system, can be easily obtained from the comparison between (\ref{eq:EquationsMotion_equations_force8}) and (\ref{eq:EquationsMotion_equations_force8_N}), or independently obtained as follows. Both methods result in the same expression:
\neweq{\vN = \vec g_{\ds{\vec \zeta_{{\sss NB}*}}} \big(\vB\big) = \vec g_{\ds{\vec q_{{\sss NB}*}}} \big(\vB\big) = \RNB \, \vB \ \rightarrow \vNdot = \RNBdot \, \vB + \RNB \, \vBdot = \RNB \, \lrp{\wNBBskew \, \vB + \vBdot}}{eq:EquationsMotion_equations_vNdot_vBdot}


\subsubsection{Specific Force}\label{subsubsec:EquationsMotion_specific_force}

Making a parenthesis in the analysis of the equations of motion, the \emph{specific force} \nm{\fIB} is defined as the non-gravitational acceleration with respect to the inertial frame \nm{\FI} experienced by a given mass \cite{Groves2008}. This is a key concept for inertial navigation as it is the magnitude measured by the accelerometers (section \ref{sec:Sensors_Inertial}). 
\neweq{\fIB = \dfrac{\FAER + \FPRO}{m} = \vec a_{\sss IB} - \vec g}{eq:EquationsMotion_specificforce}

The specific force is generally measured by the accelerometers attached to the body frame \nm{\FB} but also employed when viewed in \nm{\FN}. Making use of (\ref{eq:EquationsMotion_equations_force7}, \ref{eq:EquationsMotion_equations_force7_N}) results in:
\begin{eqnarray}
\nm{\fIBB} & = & \nm{\dfrac{\FAERB + \FPROB}{m} = \wEBBskew \; \vB + \vBdot - \vec g_{\ds{\vec \zeta_{{\sss NB}*}}}^{-1} \big(\gcNREAL - \acorN\big)} \label{eq:EquationsMotion_speceficforce2} \\
\nm{\fIBN} & = & \nm{\vec g_{\ds{\vec \zeta_{{\sss NB}*}}} \big(\dfrac{\FAERB + \FPROB}{m}\big) = \wENNskew \; \vN + \vNdot - \gcNREAL + \acorN} \label{eq:EquationsMotion_speceficforce2_N} 
\end{eqnarray}


\subsection{Rotation Equations}\label{subsec:EquationsMotion_rotation_equations}

The rotation equations provide the time derivatives of the unit quaternion \nm{\qNB} describing the rotation between the \nm{\FB} and \nm{\FN} frames. Obtained in \cite{LIE}, its linear nature is the main benefit of the use of unit quaternions to describe the aircraft attitude within the equations of motion.
\neweq{\qNBdot = \dfrac{1}{2} \; \qNB \otimes \vec \wNBB}{eq::EquationsMotion_equations_rotation}

Making use of the section \ref{sec:EquationsMotion_velocity} relationships, the dependencies of \nm{\qNBdot} are the following:
\neweq{\qNBdot = \vec f_{q,TR}\lrp{\TEgdt, \, \vB, \, \qNB, \, \wIBB}}{eq:EquationsMotion_equations_rotation_depend}


\subsection{Moment Equations}\label{subsec:EquationsMotion_moment_equations}

The moment equations provide the time derivative of the inertial angular velocity \nm{\wIBB} based on the equilibrium between the moments applied to the aircraft (represented by its body frame \nm{\FB}) with respect to its center of mass and the product of the inertia matrix by its angular acceleration with respect to that point when expressed in an inertial frame \nm{\FI}:
\neweq{\sum {\vec M}^{\sss I} = \deriv{ \lrp{\vec I \; \wIB }^{\sss I}} = \deriv{\lrp{\vec I^{\sss I} \; \wIBI }}}{eq:EquationsMotion_equations_moment1}

(\ref{eq:EquationsMotion_equations_moment1}) can be transformed into the body frame \nm{\FB}:
\neweq{\sum \vec M^{\sss B} = \vec g_{\ds{\vec \zeta_{{\sss IB}*}}}^{-1} \Big(\sum \vec M^{\sss I}\Big) = \MAERB + \MPROB = \vec g_{\ds{\vec \zeta_{{\sss IB}*}}}^{-1} \Big(\deriv{\lrp{\vec I^{\sss I} \; \wIBI}} \Big) = \RBI \; \deriv{\lrp{\vec I^{\sss I} \; \wIBI }}}{eq:EquationsMotion_equations_moment2}

The product of the inertia matrix by the angular acceleration can also be transformed into \nm{\FB} and then time derivated:
\neweq{\lrp{\vec I \; \wIB}^{\sss B} = \vec I^{\sss B} \; \wIBB = \RBI \; \vec I^{\sss I} \; \wIBI = \RBI \; \lrp{\vec I \; \wIB }^{\sss I} \rightarrow \deriv{\lrp{\vec I \; \wIB }^{\sss B}} = \RBIdot \; \lrp{\vec I \; \wIB }^{\sss I} + \RBI \; \deriv{\lrp{\vec I \; \wIB }^{\sss I}}}{eq:EquationsMotion_equations_moment3}

Merging (\ref{eq:EquationsMotion_equations_moment2}) with (\ref{eq:EquationsMotion_equations_moment3}), and taking into account that the variation with time of the inertia matrix in \nm{\FB} can be considered quasi stationary (\nm{\dot {\vec I}^{\sss B} = \vec 0}) [A\ref{as:MOTION_inertia_matrix_stationary}]:
\neweq{\MAERB + \MPROB = \deriv{\lrp{\vec I \; \wIB }^{\sss B}} - \RBIdot \; \lrp{\vec I \; \wIB }^{\sss I} = \vec I^{\sss B} \; \wIBBdot - \RBIdot \; \lrp{\vec I \; \wIB }^{\sss I}}{eq:EquationsMotion_equations_moment4}

The rotation matrix time derivative \nm{\RBIdot} can be replaced by (\nm{\wBIBskew \; \RBI}):
\neweq{\MAERB + \MPROB = \vec I^{\sss B} \; \wIBBdot - \wBIBskew \; \RBI \; \lrp{\vec I \; \wIB}^{\sss I} = \vec I^{\sss B} \; \wIBBdot - \wBIBskew \; \vec I^{\sss B} \; \wIBB}{eq:EquationsMotion_equations_moment5}

Noting that (\nm{\wBI = - \wIB}), the final expression providing the time derivative of the angular velocity of \nm{\FB} with respect to \nm{\FI} viewed in \nm{\FB} is hence:
\neweq{\wIBBdot = {\vec I^{\sss B}}^{-1} \; \lrp{\MAERB + \MPROB - \wIBBskew \; \vec I^{\sss B} \; \wIBB }}{eq:EquationsMotion_equations_moment6}

When introducing the velocity definitions of section \ref{sec:EquationsMotion_velocity} together with the algebraic equations for the Earth model (chapter \ref{cha:EarthModel}) and the aircraft performances (chapter \ref{cha:AircraftModel}), the dependencies of \nm{\wIBBdot} are the following:
\neweq{\wIBBdot = \vec f_{\omega,TR} \lrp{\TEgdt, \, \vB, \, \qNB, \, \wIBB, \, m, \, t, \, \deltaCNTR }}{eq:EquationsMotion_equations_moment_depend}


\subsection{Mass Equation}\label{subsec:EquationsMotion_mass_equations}

The mass equation reflects the progressive diminution of the aircraft mass because of the fuel burnt at the power plant:
\neweq{\dot m = - F}{eq:EquationsMotion_equations_mass}

When considering the fuel consumption dependencies, this results in:
\neweq{\dot m = f_{m,TR} \lrp{\TEgdt, \, \deltaT, \, t}}{eq:EquationsMotion_equations_mass_depend}


\section{Integration of Equations of Motion}\label{sec:EquationsMotion_integration}
 
Adding together the expressions (\ref{eq:EquationsMotion_equations_kinematic_depend}, \ref{eq:EquationsMotion_equations_force_depend}, \ref{eq:EquationsMotion_equations_rotation_depend}, \ref{eq:EquationsMotion_equations_moment_depend}, \ref{eq:EquationsMotion_equations_mass_depend}), the final dependencies of the state vector time derivative \nm{\xvecdot = \xTRUTHdot} become:
\neweq{\xvecdot\lrp{t} = \xTRUTHdot\lrp{t} = \vec f_{TR} \lrp{\TEgdt, \, \vB, \, \qNB, \, \wIBB, \, m, \, t, \, \deltaCNTR } = \vec f \big(\xTRUTH\lrp{t}, \, t, \, \deltaCNTR\lrp{t}\big)}{eq:EquationsMotion_equations_depend}

When provided with the state vector initial value \nm{\xveczero = \xTRUTHzero} and the variation with time of the position of the different control parameters described in section \ref{sec:AircraftModel_Control_Parameters}\footnote{The position of the throttle, elevator, ailerons, and rudder is adjusted by the aircraft control system described in section \ref{sec:GNC_Control}.}, the system of differential equations shown in (\ref{eq:EquationsMotion_equations_motion01}) and (\ref{eq:EquationsMotion_equations_motion02}) can be integrated to obtain the state vector variation with time \nm{\xTRUTH\lrp{t}}, this is, the \emph{actual trajectory} or \emph{real trajectory} flown by the aircraft. The variation with time of any variable not contained in the state vector can be obtained through the algebraic equations of chapters \ref{cha:EarthModel}, \ref{cha:AircraftModel}, and \ref{cha:FlightPhysics}.

The integration of the equations of motion is executed based on a \fourth\ order Runge-Kutta method at a frequency of \nm{500 \ Hz} to obtain the highest possible accuracy in the results. This discrete integration can be achieved in different ways:


\subsection{Integration in Euclidean Space}\label{subsec:EquationsMotion_integration_euclidean}

The actual trajectory can be obtained by applying a standard \fourth\ order Runge-Kutta integration scheme \cite{LIE,Press2002} to the (\ref{eq:EquationsMotion_equations_depend}) system. Note however that the unit quaternion \nm{\qNB} represents the special orthogonal group \nm{\mathbb{SO}(3)}, which is not Euclidean and hence does not define the operation of addition. Integrating in this way implies considering unit quaternions as 4-vectors that belong to \nm{\mathbb{R}^4} instead of \nm{\mathbb{SO}(3)} and operating with the addition and scalar multiplication of the \nm{\mathbb{R}^4} vector space. The propagated 4-vectors hence do not represent \nm{\mathbb{SO}(3)}, this is, are not unit quaternions, and must be normalized after each step to eliminate the deviation from the three dimensional manifold \nm{\mathbb{SO}(3)} incurred by the \nm{\mathbb{R}^4} propagation. 


\subsection{Integration in Manifold of Rigid Body Rotations}\label{subsec:EquationsMotion_integration_so3}

Although the errors of employing the section \ref{subsec:EquationsMotion_integration_euclidean} scheme at \nm{500 \ Hz} are relatively small, it is more rigorous and precise to integrate the non-Euclidean components of the state vector respecting the constraints of the Lie group or manifold to which they belong, this is, ensuring that the unit quaternion \nm{\qNB} propagates without deviating from the \nm{\mathbb{SO}(3)} manifold. To do so, it is necessary to apply the modified \fourth\ order Runge-Kutta integration scheme and its particularization for rigid body rotations, as described in \cite{LIE}.
\neweq{\xvec = \xTRUTH = \lrsb{\vec y \ \ \wNBB \ \ \qNB}^T = \Big[\lrsb{\TEgdt \ \ \vB \ \ \wIBB \ \ m}^T \ \ \wNBB \ \ \qNB\Big]^T} {eq:EquationsMotion_state_vector_so3}

This scheme relies on the state vector \nm{\vec x}, composed by a vector \nm{\vec y}, the angular velocity \nm{\wNBB} representing the tangent space \nm{\mathfrak{so}(3)}, and the unit quaternion \nm{\qNB} representing the special orthogonal group \nm{\mathbb{SO}(3)}. In order to work with the same variables contained in the (\ref{eq:EquationsMotion_state_vector}) state vector, it is necessary to use \nm{\vec y = \lrsb{\TEgdt \ \ \vB \ \ \wIBB \ \ m}^T}, and obtain the body reference angular velocity \nm{\wNBB} from the remaining state variables by means of (\ref{eq:EquationsMotion_auxiliary_angular_velocities02}), (\ref{eq:EquationsMotion_auxiliary_angular_velocities_motion2}), and (\ref{eq:EquationsMotion_auxiliar_angular_velocities_earth3}) in each of the four intermediate steps of the Runge-Kutta scheme.

Note that this modified Runge-Kutta scheme employs (\ref{eq:EquationsMotion_equations_kinematic_depend}, \ref{eq:EquationsMotion_equations_force_depend}, \ref{eq:EquationsMotion_equations_moment_depend}, \ref{eq:EquationsMotion_equations_mass_depend}) but discards the (\ref{eq:EquationsMotion_equations_rotation_depend}) differential equation, as the unit quaternion is advanced based on its \nm{\oplus} operator and the angular velocity \nm{\wNBB} instead of its \nm{\qNBdot} time derivative.
 
This approach is more elegant, robust, and precise than the Euclidean one, and does not require periodic normalizations of the unit quaternion.


\subsection{Integration in Manifold of Rigid Body Motions}\label{subsec:EquationsMotion_integration_se3}

Euclidean integration schemes equal or similar to that of section \ref{subsec:EquationsMotion_integration_euclidean} are representative of what is employed in flight sciences and aircraft navigation, while imposing that the unit quaternion advances while respecting its manifold constraints, as shown in section \ref{subsec:EquationsMotion_integration_so3}, is more common in the robotics industry. The state vectors of these two approaches are nevertheless quite similar as both contain the geodetic coordinates and separate the aircraft attitude (rotation) from its translation (position), as well as the angular velocity from its linear one.

It is however possible to employ a more elegant and equally robust integration scheme, which the author is not aware of its use in navigation or flight simulators, that consists in treating the aircraft (its \nm{\FB} frame) as a rigid body moving with respect to the \nm{\FE} Earth frame while respecting the constraints of the special Euclidean group \nm{\mathbb{SE}(3)} manifold. This approach, more common when working with robots maneuvering in controlled environments, results in a more compact notation, the use of Cartesian coordinates instead of geodetic ones, and discards both the (\ref{eq:EquationsMotion_equations_kinematic_depend}) and (\ref{eq:EquationsMotion_equations_rotation_depend}) differential equations. This scheme is more compact than the \nm{\mathbb{SO}(3)} integration described in section \ref{subsec:EquationsMotion_integration_so3}, but also less intuitive; both are however equally correct, result in extremely similar results, and are superior in robustness and accuracy to the Euclidean integration of section \ref{subsec:EquationsMotion_integration_euclidean}.

This approach relies on the modified \fourth\ order Runge-Kutta integration scheme and its particularization for rigid body motions \cite{LIE}. The state vector contains the following members:
\neweq{\xvec = \xTRUTH = \lrsb{\vec y \ \ \xiEBB \ \ \zetaEB}^T = \lrsb{m \ \ \xiEBB \ \ \zetaEB}^T} {eq:EquationsMotion_state_vector_se3}
\begin{itemize}
\item The aircraft mass \emph{m}.

\item The twist or motion velocity of the aircraft body with respect to the Earth viewed in \nm{\FB}, \nm{\xiEBB = \lrsb{\nuEBB \ \ \wEBB}^T = \lrsb{\vB \ \ \wEBB}^T}, represents the rigid body motion tangent space \nm{\mathfrak{se}(3)} \cite{LIE}. Its linear component \nm{\nuEBB} coincides with the aircraft absolute velocity viewed in the body frame \nm{\vEBB = \vB}, and its angular one \nm{\wEBB} represents the sum of the motion and aircraft angular velocities, also viewed in \nm{\FB}.

\item The unit dual quaternion \nm{\zetaEB} \cite{LIE} parameterizing \nm{\mathbb{SE}(3)} represents the pose (position plus attitude) of the aircraft represented by \nm{\FB} with respect to \hypertt{ECEF}.
\end{itemize}

The components of the (\ref{eq:EquationsMotion_state_vector}) real trajectory \nm{\xTRUTH} not present in (\ref{eq:EquationsMotion_state_vector_se3}) can be obtained as follows:
\begin{itemize}
\item Geodetic coordinates \nm{\TEgdt}. Obtain the Cartesian coordinates \nm{\TEcar = \TEBE} from the unit dual quaternion \nm{\zetaEB} and then convert them to geodetic coordinates as explained in section \ref{subsubsec:RefSystems_E_GeodeticCoord}.
\item Unit quaternion \nm{\qNB}. Obtain the \nm{\qNE} rotation between the \hypertt{ECEF} and \hypertt{NED} frames as described in section \ref{subsec:RefSystems_N} and the \nm{\qEB} rotation from \nm{\zetaEB}, and then apply \nm{\qNB = \qNE \otimes \qEB}.
\item The body ground velocity \nm{\vB} coincides with the \nm{\nuEBB} linear component of the twist \nm{\xiEBB} \cite{LIE}.
\item Inertial angular velocity \nm{\wIBB}. Obtain \nm{\wEBB} from the twist \nm{\xiEBB} angular component and then proceed as described in section \ref{sec:EquationsMotion_velocity}.
\end{itemize}

The aircraft mass is advanced on Euclidean space within the \fourth\ order Runge-Kutta integration scheme based on its (\ref{eq:EquationsMotion_equations_mass_depend}) time derivative. The unit dual quaternion \nm{\zetaEB} is advanced relying on its \nm{\oplus} operator and the twist \nm{\xiEBB}, instead of its time derivative \nm{\zetaEBdot}, so the (\ref{eq:EquationsMotion_equations_kinematic_depend}) and (\ref{eq:EquationsMotion_equations_rotation_depend}) differential equations are not required. As noted above, the twist \nm{\xiEBB} linear component is in fact the aircraft ground velocity \nm{\vB}, and can be advanced in Euclidean space according to its (\ref{eq:EquationsMotion_equations_force_depend}) time derivative. The twist angular component \nm{\wEBB} is the difference between the inertial \nm{\wIBB} and Earth \nm{\wIEB} angular velocities described in section \ref{sec:EquationsMotion_velocity}, and hence its time derivative responds to:
\neweq{\wEBBdot = \wIBBdot - \wIEBdot}{eq:EquationsMotion_webb_time_deriv}

Of these two components, \nm{\wIBBdot} is in fact the (\ref{eq:EquationsMotion_equations_moment_depend}) time derivative, and \nm{\wIEBdot} is obtained below taking advantage of the Earth angular velocity being constant when viewed in \hypertt{ECEF} (\ref{eq:EquationsMotion_auxiliar_angular_velocities_earth1}):
\begin{eqnarray}
\nm{\wIEB} & = & \nm{\vec {Ad}_{\ds{\vec q_{\sss EB}}}^{-1} \ \wIEE = \qEBast \otimes \wIEE \otimes \qEB}\label{eq:EquationsMotion_wIEB} \\
\nm{\wIEBdot} & = & \nm{\qEBastdot \otimes \wIEE \otimes \qEB + \qEBast \otimes \wIEE \otimes \qEBdot} \label{eq:EquationsMotion_wIEBdot}
\end{eqnarray}

Note that the time derivatives \nm{\qEBastdot} and \nm{\qEBdot} can be obtained from \nm{\qEB} and \nm{\wEBB} \cite{LIE}.  


\section{Implementation, Validation, and Execution} \label{sec:EquationsMotion_implementation}

The integration of the equations of motion as described in sections \ref{sec:EquationsMotion_equations_motion} and \ref{subsec:EquationsMotion_integration_se3} has been implemented in \texttt{C++}, making use when needed of the implementations of the Earth model described in section \ref{sec:EarthModel_implementation} and that of the aircraft performances of section \ref{sec:AircraftModel_implementation}. It is important to emphasize the relevance of an efficient implementation given the elevated number of computations required to obtain the real aircraft trajectory. Note that a \fourth\ order Runge-Kutta integration requires the evaluation of the equations of motion four times per step, which at a frequency of \nm{500 \ Hz} results in a total of two thousand evaluations per second of real flight. Each \nm{\xTRUTHdot} evaluation involves computation intensive processes such as the obtainment of the aerodynamic actions, the low and high frequency wind, and the atmospheric properties, among others. The computation costs required to obtain the real aircraft trajectory constitute one of the main reasons for the selection of an efficient compiled language such as \texttt{C++} for the simulation. Although many processes have first been tested and validated in interpreted languages such as \texttt{MatLab}\textsuperscript{\textregistered} and \texttt{Python}, the use of these languages for the integration of any trajectory with a duration measured in minutes instead of seconds is impractical.

The implementation has been validated by ensuring the validity of the equations of motion based exclusively on the resulting state vector variation with time. On one hand, the right hand side of the equations of motion is evaluated based on \nm{\xTRUTH\lrp{t}}, and on the other, the left hand side is obtained by taking numeric differentials of \nm{\xTRUTH\lrp{t}}. Both should coincide.

 \cleardoublepage 
\chapter{Sensors}\label{cha:Sensors}

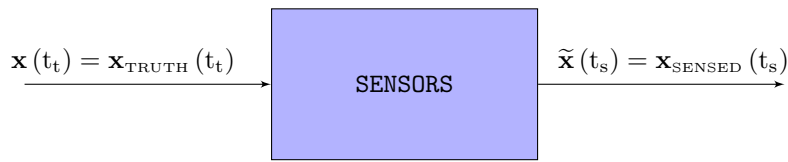
\begin{figure}[h]
\centering
\begin{tikzpicture}[auto, node distance=2cm,>=latex']
	\node [coordinate](input) {};
	\node [block, right of=input, minimum width=3.5cm, node distance=5.0cm, align=center, minimum height=2.0cm] (SENSORS) {\texttt{SENSORS}};	
	\node [coordinate, right of=SENSORS, node distance=5.0cm] (output) {};	
	\draw [->] (input) -- node[pos=0.4] {\nm{\xvec\lrp{t_t} = \xTRUTH\lrp{t_t}}} (SENSORS.west);
	\draw [->] (SENSORS.east) -- node[pos=0.55] {\nm{\xvectilde\lrp{t_s} = \xSENSED\lrp{t_s}}} (output);
\end{tikzpicture}
\caption{Sensors flow diagram}
\label{fig:Sensors_flow_diagram_generic}
\end{figure}

This chapter describes the different onboard sensors with which the aircraft is equipped to measure various aspects of the aircraft actual trajectory (section \ref{sec:EquationsMotion_AT}) and provide these measurements to the aircraft \hypertt{GNC} system described in chapters \ref{cha:gc} and \ref{cha:nav}. As a matter of fact, the outputs of these sensors, called the sensed state and described in section \ref{sec:Sensors_ST} below, represent the only link between the real but unknown actual state and the \hypertt{GNC} system in charge of achieving an actual trajectory that deviates as little as possible from the guidance objectives. The sensed trajectory, this is, the information provided by the sensors to the \hypertt{GNC} system, is not only incomplete, in the sense that it does not contain all the variables that make up the actual aircraft trajectory, but it is also partially incorrect, as the measured variables include the errors introduced by the sensors themselves. It is then the mission of the navigation system (chapter \ref{cha:nav}) to generate the observed or estimated state (section \ref{sec:GNC_OT}) by extracting as much information as possible from the sensed state, and pass it to the control system so it can adjust the throttle and aerodynamic controls to adhere to its guidance objectives. Figure \ref{fig:Sensors_flow_diagram_generic} depicts how the sensors measure the actual state \nm{\xvec\lrp{t_t} = \xTRUTH\lrp{t_t}} and obtain the sensed state \nm{\xvectilde\lrp{t_s} = \xSENSED\lrp{t_s}}, where \nm{t_t = t \cdot \DeltatTRUTH} and \nm{t_s = s \cdot \DeltatSENSED} are based on their respective frequencies of \nm{500} and \nm{100 \ Hz}.

The main objective of this chapter is to properly model the sources of error incurred by the different sensors, with special emphasis on the inertial ones (accelerometers and gyroscopes). The chapter begins with a description of the sensed trajectory and its components in section \ref{sec:Sensors_ST}. Section \ref{sec:Sensors_Inertial} provides a detailed description of the inertial sensors, its characteristics, its location and orientation with respect to the aircraft structure, its sources of error, and concludes with a model that properly captures their stochastic properties. A similar but less detailed approach is taken in section \ref{sec:Sensors_NonInertial} with the magnetometers, air data system, and \hypertt{GNSS} receiver. Section \ref{sec:Sensors_camera} describes the characteristics and mounting of the digital camera employed for visual navigation. Although the camera differs from all other sensors in that it does not provide a measurement or reading but a digital image, in this document it is indeed considered a sensor as it provides the \hypertt{GNC} system with information about its surroundings that can be employed for navigation. Last, section \ref{sec:Sensors_implementation} focuses on the implementation of the sensor models within the simulation.

This chapter does not only develop generic error models for the different aircraft sensors, but also provides the specific values on which the models rely for those sensors that could nowadays be installed onboard a small size fixed wing aircraft like the one described in chapter \ref{cha:AircraftModel}, and which from this point on are referred as the ``\emph{baseline}'' configuration. The models developed in this chapter assume that all sensors are fixed to the aircraft structure in a strapdown configuration [A\ref{as:SENSOR_strapdown}, A\ref{as:CAMERA_strapdown}]. They also assume that the measurement processes are instantaneous [A\ref{as:SENSOR_instantaneous}, A\ref{as:CAMERA_instantaneous}] and that all sensors are time synchronized with each other at a frequency of \nm{100 \ Hz}, with the exception of the digital camera and the \hypertt{GNSS} receiver, which operate at \nm{10} and \nm{1 \ Hz} respectively [A\ref{as:SENSOR_synchronize}].


\section{Sensed Trajectory}\label{sec:Sensors_ST}

The \emph{sensed trajectory} (\hypertt{ST}) can be defined as a compilation of the measurements provided by the different onboard sensors, and comprises the only view of the actual trajectory at the disposal of the \hypertt{GNC} system. It is a time stamped series of state vectors \nm{\xvectilde = \xSENSED} (\emph{sensed states}):
\neweq{\xvectilde_{\sss {GNSS\;ACTIVE}} = \xvec_{\sss {SENSED,GNSS\;ACTIVE}} = \lrsb{\fIBBtilde \ \ \wIBBtilde \ \ \BBtilde \ \ \widetilde{p} \ \ \widetilde{T} \ \ \vTASBtilde \ \ \vec I \ \ \widetilde{\xvec}^{\sss E,GDT} \ \ \vNtilde}^T}{eq:Sensors_state_vector_gnssactive}

Note that as the sensors work at a frequency of \nm{100 \ Hz} [A\ref{as:SENSOR_synchronize}] and the actual trajectory is integrated at \nm{500 \ Hz} (section \ref{sec:EquationsMotion_integration}), there is only one sensed state vector for every five actual state vectors. Another significant difference with the actual trajectory is that many useful variables describing the aircraft state, such as the attitude, can not be obtained from the sensed trajectory by means of the algebraic relationships included in chapters \ref{cha:EarthModel}, \ref{cha:AircraftModel}, and \ref{cha:FlightPhysics}, but require the integration of the chapter \ref{cha:FlightPhysics} equations (as shown in chapter \ref{cha:nav}), which induces the accumulation of errors. This fact, together with errors introduced by the sensors, is the main reason for the need for a navigation system capable of generating the observed trajectory (section \ref{sec:GNC_OT}), which is complete in the sense that it contains all the relevant variables describing the aircraft state, and is also less incorrect as a significant part of the errors introduced by the sensors have been removed.

Table \ref{tab:Sensors_ST} lists the variables that make up the sensed trajectory, together with the sensors where they are measured, their frequency, the section where they are defined, and the section where their measurement is described:
\begin{center}
\begin{tabular}{lclccc}
	\hline
	Components				  & Variable							& Measured by		& Frequency [Hz] & \multicolumn{2}{c}{Sections} \\
	\hline
	\rule[3pt]{0pt}{12pt}Specific force            & \nm{\fIBBtilde}                     & Accelerometers    & 100 & \ref{subsubsec:EquationsMotion_specific_force}  & \ref{sec:Sensors_Inertial} \\
	Inertial angular velocity & \nm{\wIBBtilde}                     & Gyroscopes        & 100 & \ref{sec:EquationsMotion_velocity} & \ref{sec:Sensors_Inertial} \\
	Magnetic field            & \nm{\BBtilde}                       & Magnetometers     & 100 & \ref{sec:EarthModel_WMM}                        & \ref{subsec:Sensors_NonInertial_Magnetometers} \\
	Air pressure              & \nm{\widetilde{p}}                  & Barometer         & 100 & \ref{subsec:EarthModel_ISA_Definitions}         & \ref{subsec:Sensors_NonInertial_ADS} \\
	Air temperature           & \nm{\widetilde{T}}                  & Thermometer       & 100 & \ref{subsec:EarthModel_ISA_Definitions}         & \ref{subsec:Sensors_NonInertial_ADS} \\
	Air velocity\footnotemark & \nm{\vTASBtilde}                    & Pitot tube, vanes & 100 & \ref{sec:EquationsMotion_velocity}  & \ref{subsec:Sensors_NonInertial_ADS} \\
	Ground position			  & \nm{\widetilde{\vec T}^{\sss E,GDT}} & \hypertt{GNSS} receiver     &   1 & \ref{subsubsec:RefSystems_E_GeodeticCoord} & \ref{subsec:Sensors_NonInertial_GNSS} \\
	Ground velocity			  & \nm{\vNtilde}                       & \hypertt{GNSS} receiver     &   1 & \ref{sec:EquationsMotion_velocity}  & \ref{subsec:Sensors_NonInertial_GNSS} \\
	Image                     & \nm{\vec I}                         & Digital camera    &  10 & \ref{sec:Sensors_camera}                        & \ref{sec:Sensors_camera} \\
	\hline
\end{tabular}
\end{center}
\captionof{table}{Components of sensed trajectory} \label{tab:Sensors_ST}
\footnotetext{Composed per (\ref{eq:EquationsMotion_auxiliary_airspeed03}) by the airspeed \nm{\vtastilde}, the angle of attack \nm{\widetilde{\alpha}}, and the angle of sideslip \nm{\widetilde{\beta}}.}


\section{Inertial Sensors}\label{sec:Sensors_Inertial}

Inertial sensors comprise \emph{accelerometers} and \emph{gyroscopes}, which measure specific force (\ref{eq:EquationsMotion_specificforce}) and inertial angular velocity (\ref{eq:EquationsMotion_auxiliary_angular_velocities01}) about a single axis, respectively \cite{Hibbeler2015}. An \emph{inertial measurement unit} or \hypertt{IMU} encompasses multiple accelerometers and gyroscopes, usually three of each, obtaining three dimensional measurements of the specific force and angular rate \cite{Farrell2008} viewed in the platform frame \nm{\FP} (section \ref{subsec:RefSystems_P}). The individual accelerometers and gyroscopes however are not aligned with the \nm{\FP} axes, but with those of the non-orthogonal \nm{\FA} and \nm{\FY} frames (sections \ref{subsec:RefSystems_A} and \ref{subsec:RefSystems_Y}), respectively. The output of the inertial sensors must hence first be transformed from the \nm{\FA} and \nm{\FY} frames to \nm{\FP}, and then to the body frame \nm{\FB}, where they can be employed by the navigation system. The accelerometers and gyroscopes are assumed to be infinitesimally small and located at the \hypertt{IMU} reference point [A\ref{as:SENSOR_small}], which coincides with the origin of these three frames \nm{\lrp{\OP = \OA = \OY}}.

The \hypertt{IMU} is a key input for inertial navigation and is generally physically attached to the aircraft structure in a strapdown configuration, so the relative position \nm{\TBPB} and attitude \nm{\phiBP = \lrsb{\psiP \ \ \thetaP \ \ \xiP}^T} between \nm{\FB} and \nm{\FP} are constant. It is important to point out that the specific force and angular velocity measured by the \hypertt{IMU} are those with respect to the inertial frame \nm{\FI} and not the Earth frame \nm{\FE}, so they include terms such as the transport and Coriolis accelerations in case of the accelerometers and the motion and Earth angular velocities for the gyroscopes (refer to chapter \ref{cha:FlightPhysics} for the meaning of these terms) that hinder the obtainment of the aircraft motion with respect to the Earth by the inertial navigation system.

Accelerometers can be divided by their underlying technology into pendulous and vibrating beams, while gyroscopes are classified into spinning mass, optical (ring laser or fiber optic), and vibratory \cite{Groves2008}. Current inertial sensor development is mostly focused on \emph{micro machined electromechanical system} (\hypertt{MEMS}) sensors\footnote{There exist both pendulous and vibrating beam \hypertt{MEMS} accelerometers, but all \hypertt{MEMS} gyroscopes are vibratory.}, which makes direct use of the chemical etching and batch processing techniques used by the electronics integrated circuit industry to obtain sensors with small size, low weight, rugged construction, low power consumption, low price, high reliability, and low maintenance \cite{Titterton2004}. On the negative side, the accuracy of \hypertt{MEMS} sensors is still low, although tremendous progress has been achieved in the last two decades and more is expected in the future.

There is no universal classification of inertial sensors according to their performance, although they can be broadly assigned into five different categories or grades: marine (submarines and spacecraft), aviation (commercial and military), intermediate (small aircraft and helicopters), tactical (\hypertt{UAV}s and guided weapons), and automotive (consumer) \cite{Groves2008}. The full range of grades covers approximately six orders of magnitude of gyroscope performance and only three for the accelerometers, but higher performance is always associated with  bigger size, weight, and cost. Tactical grade \hypertt{IMU}s cover a wide range of performance values, but can only provide a stand-alone navigation solution for a few minutes, while automotive grade \hypertt{IMU}s are unsuitable for navigation.

The different \hypertt{IMU} measurement errors are described in section \ref{subsec:Sensors_Inertial_ErrorSources}. Section \ref{subsec:Sensors_Inertial_ErrorModelSingleAxis} presents a model for the measurements of a single inertial sensor, while sections \ref{subsec:Sensors_Inertial_ObtainmentSystemNoise} and \ref{subsec:Sensors_Inertial_ObtainmentBiasDrift} focus on how to obtain from the documentation the specific values for white noise and bias on which the model relies. Additional errors appear when three sensors are employed together, and these are modeled in section \ref{subsec:Sensors_Accelerometer_Triad_ErrorModel} for accelerometers and section \ref{subsec:Sensors_Gyroscope_Triad_ErrorModel} for gyroscopes. While also depending on the relative position of the \hypertt{IMU} with respect to \nm{\FB} (section \ref{subsec:Sensors_Inertial_Mounting}), sections \ref{subsec:Sensors_Inertial_ErrorModel} and \ref{subsec:Sensors_Inertial_Selected_gyr_acc} provide comprehensive error models for the \hypertt{IMU} measurements, as well as the specific values of the \say{baseline} configuration.


\subsection{Inertial Sensor Error Sources}\label{subsec:Sensors_Inertial_ErrorSources}

In addition to the accelerometers and gyroscopes, an \hypertt{IMU} also contains a processor, storage for the calibration parameters, one or more temperature sensors, and a power supply. As described below, each sensor has several error sources, but each of them has four components: \emph{fixed} contribution, \emph{temperature} dependent variation, \emph{run-to-run} variation, and \emph{in-run} variation \cite{Groves2008,Grewal2010}. The first two can be measured at the laboratory (at different temperatures) and the calibration results stored in the \hypertt{IMU} so the processor can later compensate the sensor outputs based on the reading provided by the temperature sensor. Calibration however increases manufacturing costs so it may be partially or fully absent in the case of inexpensive sensors. The run-to-run variation results in a contribution to a given error source that varies every time the sensor is employed but remains constant within a given run. It can not be compensated by the \hypertt{IMU} processor but can be calibrated by the navigation system every time it is turned on with a process known as fine-alignment, described in section \ref{sec:PreFlight_FineAlignment}. The in-run contribution to the error sources slowly varies during execution and can not be calibrated in the laboratory nor by the navigation system.

The different error sources that influence an inertial sensor are the following \cite{Groves2008,Trusov2011,KVH2014,Chow2011}:
\begin{itemize}
\item The \emph{bias} is an error exhibited by all accelerometers and gyroscopes that is independent of the underlying specific force or angular rate being measured, and comprises the dominant contribution to the overall sensor error (refer to section \ref{subsec:Sensors_Inertial_ErrorModelSingleAxis} below). It can be defined as any nonzero output when the sensor input is zero \cite{Grewal2010}, and can be divided into its static and dynamic components.  The static part, also known as fixed bias, \emph{bias offset}, turn-on bias, or bias repeatability, comprises the run-to-run variation, while the dynamic component, known as in-run bias variation, \emph{bias drift}, bias instability (or stability), is typically about ten percent of the static part and slowly varies over periods of order one minute. As the bias is the main contributor to the overall sensor error, its value can be understood as a sensor quality measure. Table \ref{tab:Sensors_Inertial_bias} provides approximate values for the inertial sensor biases according to the \hypertt{IMU} grade \cite{Groves2008}:
\begin{center}
\begin{tabular}{lcc}
	\hline
	\hypertt{IMU} Grade & Accelerometer Bias \nm{\lrsb{m/s^2}} & Gyroscope Bias \nm{\lrsb{^{\circ}/hr}} \\
	\hline
	Marine			& \nm{10^{-4}}						& \nm{10^{-3}}		\\
	Aviation		& \nm{\lrsb{3 \cdot 10^{-4}, \ 10^{-3}}}	& \nm{10^{-2}}		\\
	Intermediate	& \nm{\lrsb{10^{-3}, \ 10 ^{-2}}}			& \nm{10^{-1}}		\\
	Tactical		& \nm{\lrsb{10^{-2}, \ 10 ^{-1}}}			& \nm{\lrsb{1, \ 100}}	\\
	Automotive		& \nm{> 10^{-1}}					& \nm{> 100}		\\
	\hline
\end{tabular}
\end{center}
\captionof{table}{Typical inertial sensor biases according to \texttt{IMU} grade}\label{tab:Sensors_Inertial_bias}

While the bias offset can be greatly reduced through fine-alignment (section \ref{sec:PreFlight_FineAlignment}), the bias drift can not be determined and needs to be modeled as a stochastic process. It is mostly a warm up effect that should be almost nonexistent after a few minutes of operation, and corresponds to the minimum point of the sensor Allan curve \cite{KVH2014,IEEE1998}. It is generally modeled as a random walk process, coupled with limits that represent the conclusion of the warm up process.

\item The \emph{scale factor error} is the departure of the input output gradient of the instrument from unity following unit conversion at the \hypertt{IMU} processor. It represents a varying relationship between sensor input and output caused by aging and manufacturing tolerances. As it is a combination of fixed contribution plus temperature dependent variation, most of it can be eliminated through calibration (section \ref{sec:PreFlight_Inertial_Calibration}).

\item The \emph{cross coupling error} or non-orthogonality error is a fixed contribution that arises from the misalignment of the sensors sensitive axes (\nm{\FY} and \nm{\FA}) with respect to the orthogonal axes of \nm{\FP} due to manufacturing limitations, and can also be highly reduced through calibration. The scale factor and cross coupling errors are in the order of \nm{10^{-4}} and \nm{10^{-3}} for most inertial sensors, although they can be higher for some low grade gyroscopes. The cross coupling error is equal to the sine of the misalignment, which is listed by some manufacturers.

\item \emph{System noise} or random noise is inherent to all inertial sensors and can combine electrical, mechanical, resonance, and quantization sources. It can originate at the sensor itself or at any other electronic equipment that interferes with it. It is a stochastic process usually modeled as white noise (section \ref{subsec:Sensors_Inertial_ErrorModelSingleAxis}) because its noise spectrum is approximately white, and can not be calibrated as there is no correlation between past and future values. A white noise process is characterized by its \emph{power spectral density} \hypertt{PSD} \cite{LIE}, which is constant as it does not depend on the signal frequency. It corresponds to the \nm{1 \ s} crossing of the sensor Allan curve \cite{KVH2014,IEEE1998}.
\begin{center}
\begin{tabular}{lcc}
	\hline
	\hypertt{IMU} Grade & Accelerometer Root \hypertt{PSD} \nm{\lrsb{m/s/hr^{0.5}}} & Gyroscope Root \hypertt{PSD} \nm{\lrsb{^{\circ}/hr^{0.5}}} \\
	\hline
	Aviation		& 0.012		& 0.002			\\
	Tactical		& 0.06		& \nm{\lrsb{0.03, \ 0.1}}	\\
	Automotive		& 0.6		& 1				\\
	\hline
\end{tabular}
\end{center}
\captionof{table}{Typical inertial sensor system noise according to \texttt{IMU} grade}\label{tab:Sensors_Inertial_noise}

System noise is sometimes referred to as random walk, which can generate confusion with the bias. The reason is that the inertial sensor outputs are always integrated to obtain ground velocity in the case of accelerometers and attitude in the case of gyroscopes. As the integration of a white noise process is indeed a random walk \cite{LIE}, the latter term is commonly employed to refer to system noise. Table \ref{tab:Sensors_Inertial_noise} contains typical values for accelerometer and gyroscope root \hypertt{PSD} according to sensor grade \cite{Groves2008}.

\item Other minor error sources not considered in this document are g-dependent bias (sensitivity of spinning mass and vibratory gyroscopes to specific force), scale factor nonlinearity, and higher order errors (spinning mass gyroscopes and pendulous accelerometers).
\end{itemize}


\subsection{Single Axis Inertial Sensor Error Model}\label{subsec:Sensors_Inertial_ErrorModelSingleAxis}

Since all sensors are required to provide measurements at equispaced discrete times \nm{t_s = s \cdot \DeltatSENSED = s \cdot \Deltat}, this section obtains a discrete model for the bias and white noise errors of a single axis inertial sensor. The results are employed in subsequent sections to generate a comprehensive \hypertt{IMU} model.

Consider a sensor in which the difference between its measurement at any given time \nm{\widetilde{x}\lrp{t}} and the real value of the physical magnitude being measured at that same time \nm{x\lrp{t}} can be represented by a zero mean white noise Gaussian process \nm{\eta_v\lrp{t}} with spectral density \nm{\sigma_v^2}:
\neweq{\widetilde{x}\lrp{t} = x\lrp{t} + \eta_v\lrp{t}}{eq:Sensor_SensorModel_wn01}

Dividing (\ref{eq:Sensor_SensorModel_wn01}) by \nm{\DeltatSENSED = \Deltat} and integrating results in:
\neweq{\dfrac{1}{\Deltat} \int_{t_0}^{t_0 + \Deltat} \widetilde{x}\lrp{t} \, dt = \dfrac{1}{\Deltat} \int_{t_0}^{t_0 + \Deltat} \lrsb{x\lrp{t} + \eta_v\lrp{t}} \, dt}{eq:Sensor_SensorModel_wn02}

Assuming that the measurement and real value are both constant over the integration interval\footnote{Note that the stochastic process \nm{\eta_v} can not be considered constant over any interval.} \cite{Crassidis2006} yields
\neweq{\widetilde{x}\lrp{t_0 + \Deltat} = x\lrp{t_0 + \Deltat} + \dfrac{1}{\Deltat} \int_{t_0}^{t_0 + \Deltat} \eta_v\lrp{t} \, dt}{eq:Sensor_SensorModel_wn03}

This expression results in the white noise sensor error \nm{w\lrp{t}}, which is the difference between the sensor measurement \nm{\widetilde{x}\lrp{t}} and the true value \nm{x\lrp{t}}. Its mean and variance can be readily computed:
\begin{eqnarray}
\nm{w\lrp{t_0 + \Deltat}} & = & \nm{\dfrac{1}{\Deltat} \int_{t_0}^{t_0 + \Deltat} \eta_v\lrp{t} \, dt} \label{eq:Sensor_SensorModel_wnerror_cont} \\
\nm{E\lrsb{w\lrp{t_0 + \Deltat}}} & = & \nm{0} \label{eq:Sensor_SensorModel_wnerror_mean} \\
\nm{Var\lrsb{w\lrp{t_0 + \Deltat}}} & = & \nm{\dfrac{\sigma_v^2}{\Deltat}} \label{eq:Sensor_SensorModel_wnerror_variance}
\end{eqnarray}

Based on these results, the white noise error can be modeled by a discrete random variable identically distributed to the above continuous white noise error, this is, one that results in the same mean and variance:
\neweq{w\lrp{s \, \DeltatSENSED} = w\lrp{s \, \Deltat} = \dfrac{\sigma_v}{\Deltat^{1/2}} \, \Nvs}{eq:Sensor_SensorModel_wnerror}

where \nm{\Nv \sim N\lrp{0, \, 1}} is a standard normal random variable. Consider now a different model in which the measurement error or bias is given by a first order random walk process \cite{LIE}, this is, the integration of a zero mean white noise Gaussian process \nm{\eta_u\lrp{t}} with spectral density \nm{\sigma_u^2}. It is also possible to compute its mean and variance:
\begin{eqnarray}
\nm{\dot b\lrp{t}} & = & \nm{\eta_u\lrp{t}} \label{eq:Sensor_SensorModel_bias01a} \\
\nm{b\lrp{t_0 + \Deltat}} & = & \nm{b\lrp{t_0} + \int_{t_0}^{t_0 + \Deltat} \eta_u\lrp{t} \, dt} \label{eq:Sensor_SensorModel_bias01b} \\
\nm{E\lrsb{b\lrp{t_0 + \Deltat}}} & = & \nm{E\lrsb{b\lrp{t_0}}} \label{eq:Sensor_SensorModel_bias_mean} \\
\nm{Var\lrsb{b\lrp{t_0 + \Deltat}}} & = & \nm{\sigma_u^2 \, \Deltat} \label{eq:Sensor_SensorModel_bias_variance}
\end{eqnarray}

These results indicate that the bias can be modeled by a discrete random variable identically distributed (this is, resulting in the same expected value and variance) to the continuous random walk above:
\neweq{b\lrp{t_0 + \Deltat} = b\lrp{t_0} + \sigma_u \, \Deltat^{1/2} \, \Nu} {eq:Sensor_SensorModel_bias_disc1}

where \nm{\Nu \sim N\lrp{0, \, 1}} is also a standard normal random variable. The final expressions for the discrete bias, as well as its mean and variance, are the following:
\begin{eqnarray}
\nm{b\lrp{s \, \Deltat}} & = & \nm{B_0 \, \Nuzero + \sigma_u \, \Deltat^{1/2} \, \sum_{i=1}^s \Nui} \label{eq:Sensor_SensorModel_bias_disc} \\
\nm{E\lrsb{b\lrp{s \, \Deltat}}} & = & \nm{0} \label{eq:Sensor_SensorModel_bias_disc_mean} \\
\nm{Var\lrsb{b\lrp{s \, \Deltat}}} & = & \nm{B_0^2 + \sigma_u^2 \, s \, \Deltat} \label{eq:Sensor_SensorModel_bias_discvariance}
\end{eqnarray}

A comprehensive single axis sensor error model without scale factor can hence be constructed by adding together the influence of the system noise provided by (\ref{eq:Sensor_SensorModel_wnerror}) and the bias given by (\ref{eq:Sensor_SensorModel_bias_disc}) \cite{Woodman2007}, while assuming that the standard normal random variables \nm{\Nu} and \nm{\Nv} are uncorrelated\footnote{Note that the expected value and variance of each of the two discrete components of this sensor model coincide with those of their continuous counterparts, but their combined mean and variance provided by expressions (\ref{eq:Sensor_SensorModel_error_mean}) and (\ref{eq:Sensor_SensorModel_error_variance}) differ from that of the combination of the two continuous error models given by (\ref{eq:Sensor_SensorModel_wnerror_cont}) and (\ref{eq:Sensor_SensorModel_bias01b}). This is the case even if considering that the two zero mean white noise Gaussian processes \nm{\eta_u} and \nm{\eta_v} are independent and hence uncorrelated. It is however possible to obtain a discrete model whose discrete bias and white noise components are not only identically distributed that those of their continuous counterparts \cite{Crassidis2006}, even adding the equivalence of covariance between the bias and the sensor error, but this results in a significantly more complex model that behaves similarly to the one above at all but the shortest time samples after sensor initialization. As this document relies on obtaining measurements at high frequencies for long periods of time, the author has chosen to reduce complexity with little or no loss of realism.}:
\begin{eqnarray}
\nm{e_{\sss BW}\lrp{s \, \Deltat}} & = & \nm{\widetilde{x}\lrp{s \, \Deltat} - x\lrp{s \, \Deltat} } \nonumber \\
 & = & \nm{B_0 \, \Nuzero + \sigma_u \, \Deltat^{1/2} \, \sum_{i=1}^s \Nui + \dfrac{\sigma_v}{\Deltat^{1/2}} \, \Nvs} \label{eq:Sensor_SensorModel_error} \\
\nm{E\lrsb{e_{\sss BW}\lrp{s \, \Deltat}}} & = & \nm{0} \label{eq:Sensor_SensorModel_error_mean} \\
\nm{Var\lrsb{e_{\sss BW}\lrp{s \, \Deltat}}} & = & \nm{B_0^2 + \sigma_u^2 \, s \, \Deltat + \dfrac{\sigma_v^2}{\Deltat}} \label{eq:Sensor_SensorModel_error_variance}
\end{eqnarray}

The \emph{discrete sensor error} or difference between the measurement provided by the sensor \nm{\widetilde{x}\lrp{s \, \DeltatSENSED} = \widetilde{x}\lrp{s \, \Deltat}} at any given discrete time \nm{s \, \Deltat} and the real value of the physical variable being measured at that same discrete time \nm{x\lrp{s \, \Deltat}} is the combination of a bias or first order random walk and a white noise process, and depends on three parameters: the bias offset \nm{B_0}, the bias instability \nm{\sigma_u}, and the white noise \nm{\sigma_v}. The contributions of these three different sources to the sensor error as well as to its first and second integrals\footnote{Gyroscopes measure angular velocity and their output needs to be integrated once to obtain attitude, while accelerometers measure specific force and are integrated once to obtain velocity and twice to obtain position.} are very different and inherent to many of the challenges encountered when employing accelerometers and gyroscopes for inertial navigation, as explained below.

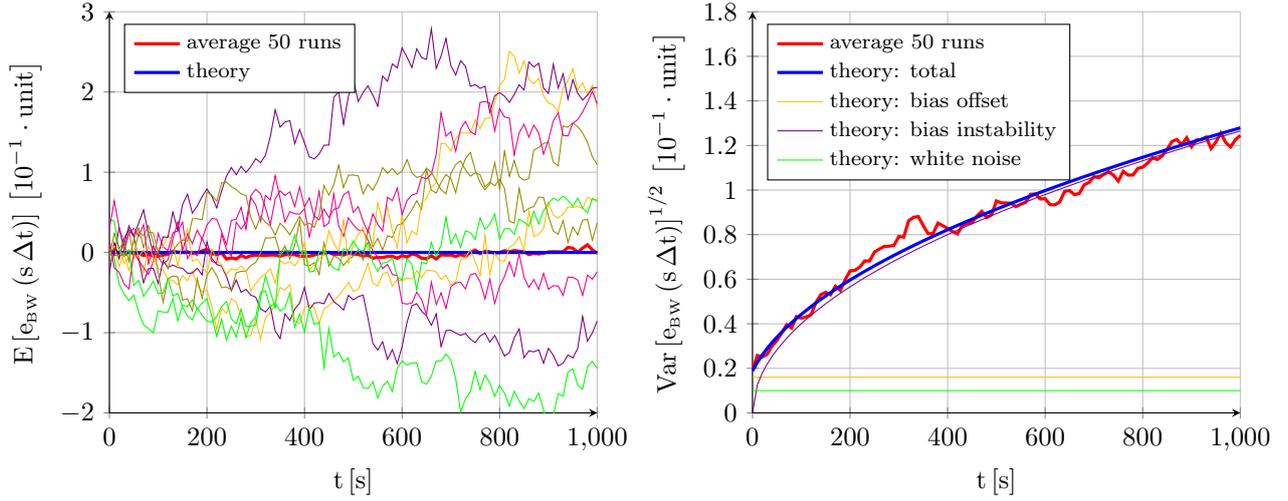
\begin{figure}[h]
\centering
\begin{tikzpicture}
\begin{axis}[
cycle list={{red,no markers,very thick},{blue,no markers,very thick},{orange!50!yellow, no markers},{violet, no markers},{green, no markers},{magenta, no markers},{olive,no markers}},
width=8.0cm, 
xmin=0, xmax=1000, xtick={0,200,400,600,800,1000},
xlabel={\nm{t \left[s\right]}},
xmajorgrids,
ymin=-2, ymax=3, ytick={-2,-1,0,1,2,3},
ylabel={\nm{E\left[e_{\sss BW}\left(s \,  \Deltat\right)\right] \, \left[10^{-1} \cdot unit\right]}},
ymajorgrids,
axis lines=left,
axis line style={-stealth},
legend entries={average 50 runs, theory},
legend pos=north west,
legend style={font=\footnotesize},
legend cell align=left,
]
\pgfplotstableread{figs/ch05_sensors/error_mean.txt}\mytable
\addplot table [header=false, x index=0,y index=1] {\mytable};
\addplot table [header=false, x index=0,y index=2] {\mytable};
\addplot table [header=false, x index=0,y index=3] {\mytable};
\addplot table [header=false, x index=0,y index=4] {\mytable};
\addplot table [header=false, x index=0,y index=5] {\mytable};
\addplot table [header=false, x index=0,y index=6] {\mytable};
\addplot table [header=false, x index=0,y index=7] {\mytable};
\pgfplotsset{cycle list shift = 2}
\addplot table [header=false, x index=0,y index=8] {\mytable};
\addplot table [header=false, x index=0,y index=9] {\mytable};
\addplot table [header=false, x index=0,y index=10] {\mytable};
\addplot table [header=false, x index=0,y index=11] {\mytable};
\addplot table [header=false, x index=0,y index=12] {\mytable};
\end{axis}	
\end{tikzpicture}%
\hskip 1pt
\begin{tikzpicture}
\begin{axis}[
cycle list={{red,no markers,very thick},{blue,no markers,very thick},{orange!50!yellow, no markers},{violet, no markers},{green, no markers}},
width=8.0cm,
xmin=0, xmax=1000, xtick={0,200,400,600,800,1000},
xlabel={\nm{t \left[s\right]}},
xmajorgrids,
ymin=0, ymax=1.8, ytick={0,0.2,0.4,0.6,0.8,1,1.2,1.4,1.6,1.8},
ylabel={\nm{Var\left[e_{\sss BW}\left(s \,  \Deltat\right)\right]^{1/2} \, \left[10^{-1} \cdot unit\right]}},
ymajorgrids,
axis lines=left,
axis line style={-stealth},
legend entries={average 50 runs, theory: total, theory: bias offset, theory: bias instability, theory: white noise},
legend pos=north west,
legend style={font=\footnotesize},
legend cell align=left,
]
\pgfplotstableread{figs/ch05_sensors/error_std.txt}\mytable
\addplot table [header=false, x index=0,y index=1] {\mytable};
\addplot table [header=false, x index=0,y index=2] {\mytable};
\addplot table [header=false, x index=0,y index=3] {\mytable};
\addplot table [header=false, x index=0,y index=4] {\mytable};
\addplot table [header=false, x index=0,y index=5] {\mytable};
\end{axis}		
\end{tikzpicture}%
\caption{Propagation with time of sensor error mean and standard deviation}
\label{fig:sensors_error}
\end{figure}

Figure \ref{fig:sensors_error} represents the performance of a fictitious sensor of \nm{B_0 = 1.6 \cdot 10^{-2}}, \nm{\sigma_u = 4 \cdot 10^{-3}}, and \nm{\sigma_v = 1 \cdot 10^{-3}} working at a frequency of \nm{100 \ Hz \ \lrp{\Deltat = 0.01 \ s}}, and is intended to showcase the different behavior and relative influence on the total error of each of its three components. The figures show the theoretical variation with time of the sensor error mean (left side) and standard deviation (right side) given by (\ref{eq:Sensor_SensorModel_error_mean}) and (\ref{eq:Sensor_SensorModel_error_variance}) together with the average of fifty different runs. In addition, the left figure also includes ten of those runs to showcase the variability in results implicit to the random variables\footnote{Although the data is generated at \nm{100 \ Hz}, the figure only employs 1 out of every 1000 points, so it appears less noisy.}, while the right side shows the theoretical contribution to the standard deviation of each of the three components. In addition to the near equivalence between the theory and the average of several runs, the figure shows that the bias instability is the commanding long term factor in the deviation between the sensor measurement and its zero mean (the standard deviation of the bias instability grows with the square root of time while the other two components are constant). As discussed in section \ref{subsec:Sensors_Inertial_ErrorSources}, the bias drift or bias instability is indeed the most important quality parameter of an inertial sensor. 

It is possible to integrate the sensor error over a timespan \nm{s \, \Deltat} to evaluate the growth with time of both its expected value and its variance\footnote{As the interest lies primarily in \nm{s \gg 1}, a simple integration method such as the rectangular rule is employed.}:
\begin{eqnarray}
\nm{f_{\sss BW}\lrp{s \, \Deltat}} & = & \nm{f_{\sss BW}\lrp{0} + \int_0^{s \, \Deltat} e_{\sss BW}\lrp{\tau} \, d\tau = f_{\sss BW}\lrp{0} + \Deltat \sum_{i=1}^s e_{\sss BW}\lrp{i \, \Deltat}} \nonumber \\
& = & \nm{f_{\sss BW}\lrp{0} + B_0 \, \Nuzero \, s \, \Deltat + \sigma_u \, \Deltat^{3/2} \sum_{i=1}^s \lrp{s-i+1} \, \Nui + \sigma_v \, \Deltat^{1/2} \sum_{i=1}^s \Nvi} \label{eq:Sensor_SensorModel_1st} \\
\nm{E\lrsb{f_{\sss BW}\lrp{s \, \Deltat}}} & = & \nm{f_{\sss BW}\lrp{0}} \label{eq:Sensor_SensorModel_1st_mean} \\
\nm{Var\lrsb{f_{\sss BW}\lrp{s \, \Deltat}}} & = & \nm{B_0^2 \, \lrp{s \, \Deltat}^2 + \dfrac{\sigma_u^2}{6} \, \Deltat^3 \, s \, \lrp{s + 1} \, \lrp{2 \, s + 1} + \sigma_v^2 \, \lrp{s \, \Deltat}} \nonumber \\
& \nm{\approx} & \nm{B_0^2 \, \lrp{s \, \Deltat}^2 + \dfrac{\sigma_u^2}{3} \, \lrp{s \, \Deltat}^3 + \sigma_v^2 \, \lrp{s \, \Deltat}} \label{eq:Sensor_SensorModel_1st_variance}
\end{eqnarray}
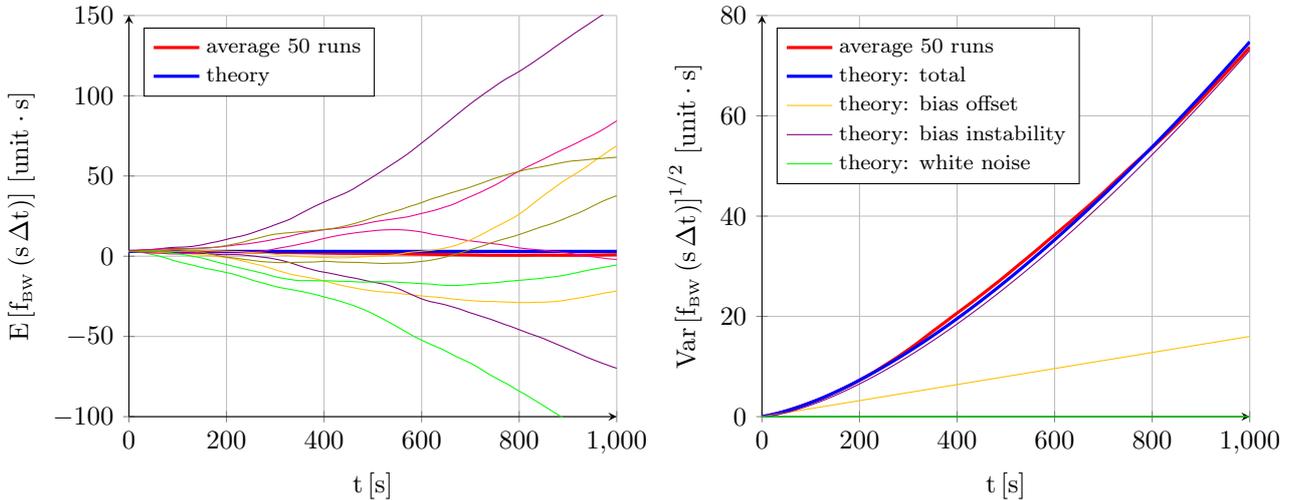
\begin{figure}[h]
\centering
\begin{tikzpicture}
\begin{axis}[
cycle list={{red,no markers,very thick},{blue,no markers,very thick},{orange!50!yellow, no markers},{violet, no markers},{green, no markers},{magenta, no markers},{olive,no markers}},
width=8.0cm, 
xmin=0, xmax=1000, xtick={0,200,400,600,800,1000},
xlabel={\nm{t \left[s\right]}},
ymin=-100, ymax=150, ytick={-100,-50,0,50,100,150},
xmajorgrids,
ylabel={\nm{E\left[f_{\sss BW}\left(s \,  \Deltat\right)\right] \, \left[unit \cdot s\right]}},
ymajorgrids,
axis lines=left,
axis line style={-stealth},
legend entries={average 50 runs, theory},
legend pos=north west,
legend style={font=\footnotesize},
legend cell align=left,
]
\pgfplotstableread{figs/ch05_sensors/error_1st_integral_mean.txt}\mytable
\addplot table [header=false, x index=0,y index=1] {\mytable};
\addplot table [header=false, x index=0,y index=2] {\mytable};
\addplot table [header=false, x index=0,y index=3] {\mytable};
\addplot table [header=false, x index=0,y index=4] {\mytable};
\addplot table [header=false, x index=0,y index=5] {\mytable};
\addplot table [header=false, x index=0,y index=6] {\mytable};
\addplot table [header=false, x index=0,y index=7] {\mytable};
\pgfplotsset{cycle list shift = 2}
\addplot table [header=false, x index=0,y index=8] {\mytable};
\addplot table [header=false, x index=0,y index=9] {\mytable};
\addplot table [header=false, x index=0,y index=10] {\mytable};
\addplot table [header=false, x index=0,y index=11] {\mytable};
\addplot table [header=false, x index=0,y index=12] {\mytable};
\end{axis}	
\end{tikzpicture}%
\hskip 1pt
\begin{tikzpicture}
\begin{axis}[
cycle list={{red,no markers,very thick},{blue,no markers,very thick},{orange!50!yellow, no markers},{violet, no markers},{green, no markers}},
width=8.0cm,
xmin=0, xmax=1000, xtick={0,200,400,600,800,1000},
xlabel={\nm{t \left[s\right]}},
xmajorgrids,
ymin=0, ymax=80, ytick={0,20,40,60,80},
ylabel={\nm{Var\left[f_{\sss BW}\left(s \, \Deltat\right)\right]^{1/2} \, \left[unit \cdot s\right]}},
ymajorgrids,
axis lines=left,
axis line style={-stealth},
legend entries={average 50 runs, theory: total, theory: bias offset, theory: bias instability, theory: white noise},
legend pos=north west,
legend style={font=\footnotesize},
legend cell align=left,
]
\pgfplotstableread{figs/ch05_sensors/error_1st_integral_std.txt}\mytable
\addplot table [header=false, x index=0,y index=1] {\mytable};
\addplot table [header=false, x index=0,y index=2] {\mytable};
\addplot table [header=false, x index=0,y index=3] {\mytable};
\addplot table [header=false, x index=0,y index=4] {\mytable};
\addplot table [header=false, x index=0,y index=5] {\mytable};
\end{axis}		
\end{tikzpicture}%
\caption{Propagation with time of first integral of sensor error mean and standard deviation}
\label{fig:sensors_1st_integral}
\end{figure}

Figure \ref{fig:sensors_1st_integral} follows the same pattern as figure \ref{fig:sensors_error} but applied to the error integral instead of to the error itself. It would represent the attitude error resulting from integrating the gyroscope output, or the velocity error expected when integrating the specific force measured by an accelerometer. The conclusions are the same as before but significantly more accentuated. Not only is the expected value of the error constant instead of zero (\nm{f_{\sss BW}\lrp{0} = 3} has been employed in the experiment), but the growth in the standard deviation (over a nonzero mean) is much quicker than before. The bias instability continues to be the dominating factor but now increases with a power of \nm{t^{3/2}}, while the bias offset and white noise contributions also increase with time, although with powers of \nm{t} and \nm{t^{1/2}} respectively. This justifies the need for robust inertial navigation systems, as otherwise the errors in attitude and velocity continuously grow with time and preclude any type of effective navigation. The error can be integrated a second time:
\begin{eqnarray}
\nm{g_{\sss BW}\lrp{s \, \Deltat}} & = & \nm{g_{\sss BW}\lrp{0} + \int_0^{s \, \Deltat} f_{\sss BW}\lrp{\tau} \, d\tau = g_{\sss BW}\lrp{0} + \Deltat \sum_{i=1}^s f_{\sss BW}\lrp{i \, \Deltat}} \nonumber \\
& = & \nm{g_{\sss BW}\lrp{0} + f_{\sss BW}\lrp{0} \, s \, \Deltat + \dfrac{B_0}{2} \, \Nuzero \, \lrp{s \, \Deltat}^2} \nonumber \\
& & \nm{+ \, \sigma_u \, \Deltat^{5/2} \sum_{i=1}^s \sum_{j=1}^{s-i+1} j \, \Nui + \sigma_v \, \Deltat^{3/2} \sum_{i=1}{s} \lrp{s-i+1} \, \Nvi} \label{eq:Sensor_SensorModel_2nd} \\
\nm{E\lrsb{g_{\sss BW}\lrp{s \, \Deltat}}} & = & \nm{g_{\sss BW}\lrp{0} + f_{\sss BW}\lrp{0} \, \lrp{s \, \Deltat}} \label{eq:Sensor_SensorModel_2nd_mean} \\
\nm{Var\lrsb{g_{\sss BW}\lrp{s \,  \Deltat}}} & = & \nm{\dfrac{B_0^2}{4} \, \lrp{s \, \Deltat}^4 + \sigma_u^2 \, \Deltat^5 \, \sum_{i=1}^s \lrp{\sum_{j=1}^{s-i+1} j}^2 + \dfrac{\sigma_v^2}{6} \, \Deltat^3 \, s \, \lrp{s + 1} \, \lrp{2 \, s + 1}} \nonumber \\
& \nm{\approx} & \nm{\dfrac{B_0^2}{4} \, \lrp{s \, \Deltat}^4 + \dfrac{\sigma_u^2}{20} \, \lrp{s \, \Deltat}^5 + \dfrac{\sigma_v^2}{3} \, \lrp{s \, \Deltat}^3} \label{eq:Sensor_SensorModel_2nd_variance}
\end{eqnarray}

\begin{figure}[h]
\centering
\begin{tikzpicture}
\begin{axis}[
cycle list={{red,no markers,very thick},{blue,no markers,very thick},{orange!50!yellow, no markers},{violet, no markers},{green, no markers},{magenta, no markers},{olive,no markers}},
width=8.0cm, 
xmin=0, xmax=1000, xtick={0,200,400,600,800,1000},
xlabel={\nm{t \left[s\right]}},
xmajorgrids,
ymin=-4, ymax=6, ytick={-4,-2,0,2,4,6},
ylabel={\nm{E\left[g_{\sss BW}\left(s \, \Deltat\right)\right] \, \left[10^4 \cdot unit \cdot s^2\right]}},
ymajorgrids,
axis lines=left,
axis line style={-stealth},
legend entries={average 50 runs, theory},
legend pos=north west,
legend style={font=\footnotesize},
legend cell align=left,
]
\pgfplotstableread{figs/ch05_sensors/error_2nd_integral_mean.txt}\mytable
\addplot table [header=false, x index=0,y index=1] {\mytable};
\addplot table [header=false, x index=0,y index=2] {\mytable};
\addplot table [header=false, x index=0,y index=3] {\mytable};
\addplot table [header=false, x index=0,y index=4] {\mytable};
\addplot table [header=false, x index=0,y index=5] {\mytable};
\addplot table [header=false, x index=0,y index=6] {\mytable};
\addplot table [header=false, x index=0,y index=7] {\mytable};
\pgfplotsset{cycle list shift = 2}
\addplot table [header=false, x index=0,y index=8] {\mytable};
\addplot table [header=false, x index=0,y index=9] {\mytable};
\addplot table [header=false, x index=0,y index=10] {\mytable};
\addplot table [header=false, x index=0,y index=11] {\mytable};
\addplot table [header=false, x index=0,y index=12] {\mytable};
\end{axis}	
\end{tikzpicture}%
\hskip 1pt
\begin{tikzpicture}
\begin{axis}[
cycle list={{red,no markers,very thick},{blue,no markers,very thick},{orange!50!yellow, no markers},{violet, no markers},{green, no markers}},
width=8.0cm,
xmin=0, xmax=1000, xtick={0,200,400,600,800,1000},
xlabel={\nm{t \left[s\right]}},
xmajorgrids,
ymin=0, ymax=3.0, ytick={0,0.5,1,1.5,2,2.5,3},
ylabel={\nm{Var\left[g_{\sss BW}\left(s \, \Deltat\right)\right]^{1/2} \, \left[10^4 \cdot unit \cdot s^2\right]}},
ymajorgrids,
axis lines=left,
axis line style={-stealth},
legend entries={average 50 runs, theory: total, theory: bias offset, theory: bias instability, theory: white noise},
legend pos=north west,
legend style={font=\footnotesize},
legend cell align=left,
]
\pgfplotstableread{figs/ch05_sensors/error_2nd_integral_std.txt}\mytable
\addplot table [header=false, x index=0,y index=1] {\mytable};
\addplot table [header=false, x index=0,y index=2] {\mytable};
\addplot table [header=false, x index=0,y index=3] {\mytable};
\addplot table [header=false, x index=0,y index=4] {\mytable};
\addplot table [header=false, x index=0,y index=5] {\mytable};
\end{axis}		
\end{tikzpicture}%
\caption{Propagation with time of second integral of sensor error mean and standard deviation}
\label{fig:sensors_2nd_integral}
\end{figure}
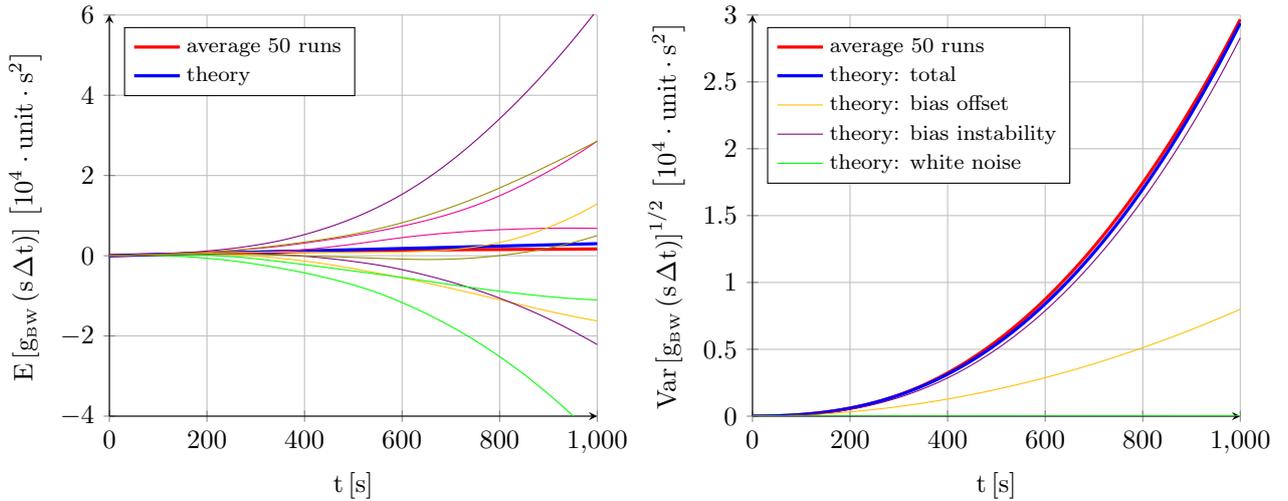

Figure \ref{fig:sensors_2nd_integral} shows the same type of figures but applied to the second integral of the error (\nm{g_{\sss BW}\lrp{0} = 1.5 } has been employed in the experiment). In this case the degradation of the results with time is even more intense to the point were the measurements are useless after a very short period of time. Unless corrected by the navigation system, this is equivalent to the error in position obtained by double integrating the output of the accelerometers.

This paragraph summarizes the key points of the single axis inertial sensor discrete error model developed in this section. The error \nm{e_{\sss BW}\lrp{s \, \Deltat}}, which applies to specific force for accelerometers and inertial angular velocity in the case of gyroscopes, depends on three factors: bias offset \nm{B_0}, bias drift \nm{\sigma_u}, and white noise \nm{\sigma_v}. Its mean is always zero, but the error standard deviation grows with time (\nm{\propto \, t^{1/2}}) due to the bias drift with constant contributions from the bias offset and the white noise. When integrating the error to obtain \nm{f_{\sss BW}\lrp{s \, \Deltat}}, equivalent to ground velocity for accelerometers and attitude for gyroscopes, the initial speed error or initial attitude error \nm{f_{\sss BW}\lrp{0}} becomes the fourth contributor, and an important one indeed, as it becomes the mean of the first integral error at any time. The standard deviation, which measures the spread over the nonzero mean, increases very quickly with time because of the bias instability (\nm{\propto \, t^{3/2}}), with contributions from the offset (\nm{\propto \, t}) and the white noise (\nm{\propto \, t^{1/2}}). If integrating a second time to obtain \nm{g_{\sss BW}\lrp{s \, \Deltat}}, or position in case of the accelerometer, the initial position error \nm{g_{\sss BW}\lrp{0}} turns into the fifth contributor. The expected value of the position error grows linearly with time due to the initial velocity error with an additional constant contribution from the initial position error, while the position standard deviation (measuring spread over a growing average value) grows extremely quick due mostly to the bias instability (\nm{\propto \, t^{5/2}}), but also because of the bias offset (\nm{\propto \, t^{2}}) and the white noise (\nm{\propto \, t^{3/2}}). Table \ref{tab:Sensors_Inertial_units} shows the standard units of the different sources of error for both accelerometers and gyroscopes.
\begin{center}
\begin{tabular}{lccccc}
	\hline
	Units			& \nm{B_0} & \nm{\sigma_u} & \nm{\sigma_v} & \nm{f_{\sss BW}\lrp{0}} & \nm{g_{\sss BW}\lrp{0}} \\
	\hline
	Accelerometer	& \nm{m/s^2}	  & \nm{m/s^{2.5}}		  & \nm{m/s^{1.5}}		  & \nm{m/s}	  & \nm{m}	\\
	Gyroscope		& \nm{^{\circ}/s} & \nm{^{\circ}/s^{1.5}} & \nm{^{\circ}/s^{0.5}} & \nm{^{\circ}} & N/A		\\
	\hline
\end{tabular}
\end{center}
\captionof{table}{Units for single axis inertial sensor error sources}\label{tab:Sensors_Inertial_units}

Sections \ref{subsec:Sensors_Inertial_ObtainmentSystemNoise} and \ref{subsec:Sensors_Inertial_ObtainmentBiasDrift} obtain the bias instability \nm{\sigma_u} and system noise \nm{\sigma_v} from the sensor specifications, and show that the value at \nm{100 \ s} of the bias instability line (purple color) at the right hand side of figure \ref{fig:sensors_error} (sensor error) is equal to ten times \nm{\sigma_u}, while the value at \nm{1000 \ s} of the white noise line (green color) at the right hand side of figure \ref{fig:sensors_1st_integral} (integral of sensor error) is equal to \nm{1000^{1/2} \approx 31.6} times \nm{\sigma_v}.


\subsection{Obtainment of System Noise Values}\label{subsec:Sensors_Inertial_ObtainmentSystemNoise}

This section focuses on the significance of system or white noise error \nm{\sigma_v} and how to obtain it from sensor specifications, which often refer to the integral of the output instead of the output itself. As the integral of white noise is a random walk process, the angle random walk of a gyroscope is equivalent to white noise in the angular rate output, while velocity random walk refers to specific force white noise in accelerometers \cite{Grewal2010}. The discussion that follows applies to gyroscopes but is fully applicable to accelerometers if replacing angular rate by specific force and attitude or angle by ground velocity.

Angle random walk, measured in [\nm{rad/s^{1/2}}], [\nm{^{\circ}/hr^{1/2}}], or equivalent, describes the average deviation or error that occurs when the sensor output signal is integrated due to system noise exclusively, without considering other error sources such as bias or scale factor \cite{Stockwell}. If integrating multiple times and obtaining a distribution of end points at a given final time \nm{s \, \Deltat}, the standard deviation of this distribution, containing the final angles at the final time, scales linearly with the white noise level \nm{\sigma_v}, the square root of the integration step size \nm{\Deltat}, and the square root of the number of steps \emph{s}, as noted by the last term of (\ref{eq:Sensor_SensorModel_1st_variance}). This means that an angle random walk of \nm{1^{\circ}/s^{1/2}} translates into an standard deviation for the error of \nm{1^{\circ}} after \nm{1 \ s}, \nm{10^{\circ}} after \nm{100 \ s}, and \nm{1000^{1/2} \approx 31.6^{\circ}} after \nm{1000 \ s}.

Manufacturers often provide this information as the \hypertt{PSD} of the white noise process in [\nm{^{\circ 2}/hr^2/Hz}] or equivalent, where it is necessary to take its square root to obtain \nm{\sigma_v}, or as the root \hypertt{PSD} in [\nm{^{\circ}/hr/Hz^{1/2}}] that is equivalent to \nm{\sigma_v}. Sometimes it is even provided as the \hypertt{PSD} of the random walk process, not the white noise, in units [\nm{^{\circ}/hr}] or equivalent. It is then necessary to multiply this number by the square root of the sampling interval \nm{\Deltat} or divide it by the square root of the sampling frequency to obtain the desired \nm{\sigma_v}.


\subsection{Obtainment of Bias Drift Values}\label{subsec:Sensors_Inertial_ObtainmentBiasDrift}

This section describes the meaning of bias instability \nm{\sigma_u} (also known as bias stability or bias drift) and how to obtain it from sensor specifications. It is also centered on gyroscopes, although applicable to accelerometers as well. Bias instability can be defined as the potential of the sensor error to stay within a certain range for a certain time \cite{Renaut2013}.  A few manufacturers directly provide sensor output changes over time, which relates with the bias instability (also known as in-run bias variation, bias drift, or rate random walk) per the second term of (\ref{eq:Sensor_SensorModel_error_variance}). If provided with an angular rate change of x [\nm{^{\circ}/s}] (\nm{1\sigma}) in t [\nm{s}], then \nm{\sigma_u} can be obtained as follows \cite{Crassidis2006,Farrenkopf1974}:
\neweq{\sigma_u = \dfrac{x}{t^{1/2}}}{eq:Sensor_SensorModel_sigmau}

As the bias drift is responsible for the growth of sensor error with time (figure \ref{fig:sensors_error}), manufacturers more commonly provide bias stability measurements that describe how the bias of a device may change over a specified period of time \cite{Woodman2007}, typically around \nm{100 \ s}. Bias stability is usually specified as a \nm{1\sigma} value with units [\nm{^{\circ}/hr}] or [\nm{^{\circ}/s}], which can be interpreted as follows according to (\ref{eq:Sensor_SensorModel_error}, \ref{eq:Sensor_SensorModel_error_mean}, \ref{eq:Sensor_SensorModel_error_variance}). If the sensor error (or bias) is known at a given time \emph{t}, then a \nm{1\sigma} bias stability of \nm{0.01^{\circ}/hr} over \nm{100 \ s} means that the bias at time \nm{t + 100} [\nm{s}] is a random variable with mean the error at time \emph{t} and standard deviation \nm{0.01^{\circ}/hr}, and expression (\ref{eq:Sensor_SensorModel_sigmau}) can be used to obtain \nm{\sigma_u}. As the bias behaves as a random walk over time whose standard deviation grows proportionally to the square root of time, the bias stability is sometimes referred as a bias random walk.

Bias fluctuations do not behave as random walk for long periods of time. If they did, the uncertainty in the bias of a device would grow without bounds as the timespan increased, which is not the case. In practice the bias is constrained to be within some range, and therefore the random walk model is only a good approximation to the true process for short periods of time \cite{Woodman2007}.


\subsection{Accelerometer Triad Sensor Error Model}\label{subsec:Sensors_Accelerometer_Triad_ErrorModel}

An \hypertt{IMU} is equipped with an accelerometer triad composed by three individual accelerometers, each of which measures the projection of the specific force over its sensing axis as described in section \ref{subsec:Sensors_Inertial_ErrorModelSingleAxis} while incurring in an error \nm{e_{\sss BW,ACC}} that can be modeled as a combination of bias offset, bias drift, and white noise (\ref{eq:Sensor_SensorModel_error}). The three accelerometers can be considered infinitesimally small [A\ref{as:SENSOR_small}] and located at the \hypertt{IMU} \emph{reference point}, which is defined as the intersection between the sensing axes of the three accelerometers. As the accelerometer frame \nm{\FA} is centered at the \hypertt{IMU} reference point and its three non-orthogonal axes coincide with its sensing axes, (\ref{eq:Sensor_Inertial_acc_error_FA}) joins together the measurements of the three individual accelerometers:
\neweq{\fIAAtilde = \SACC \ \lrp{\fIAA + \vec e_{\sss BW,ACC}^{\sss A}}}{eq:Sensor_Inertial_acc_error_FA}

where \nm{\fIAA} is the specific force (section \ref{subsubsec:EquationsMotion_specific_force}) viewed in \nm{\FA}, \nm{\fIAAtilde} represents its measurement also viewed in \nm{\FA}, \nm{\vec e_{\sss BW,ACC}^{\sss A}} is the error introduced by each accelerometer (\ref{eq:Sensor_SensorModel_error}), and \nm{\SACC} is a square diagonal matrix containing the scale factor errors \nm{\lrb{\sACCi, \ \sACCii, \ \sACCiii}} for each accelerometer (section \ref{subsec:Sensors_Inertial_ErrorSources}). It is however preferred to obtain an expression in which the specific forces are viewed in the orthogonal platform frame \nm{\FP} instead of \nm{\FA}. As both share the same origin \nm{\lrp{\OA = \OP}},
\neweq{\fIPPtilde = \RPA \ \SACC \ \lrp{\RAP \ \fIPP + \vec e_{\sss BW,ACC}^{\sss A}}}{eq:Sensor_Inertial_acc_error_FP_bis}

where \nm{\RPA} and \nm{\RAP}, defined by (\ref{eq:RefSystems_PA_AP}), contain the cross coupling errors \nm{\lrb{\alphaACCi, \ \alphaACCii, \ \alphaACCiii}} generated by the misalignment of the accelerometer sensing axes. The scale factor and cross coupling errors contain fixed and temperature dependent error contributions (section \ref{subsec:Sensors_Inertial_ErrorSources}) that can be modeled as normal random variables:
\begin{eqnarray}
\nm{\sACCXi}     & = & \nm{N\lrp{1, \ \sACC^2}  \; \hspace{20pt} \forall \ i \in \lrb{1,\ 2, \ 3}}\label{eq:Sensors_Inertial_acc_error_scale} \\
\nm{\alphaACCXi} & = & \nm{N\lrp{0, \ \alphaACC^2} \hspace{20pt} \forall \ i \in \lrb{1,\ 2, \ 3}}\label{eq:Sensors_Inertial_acc_error_cross}
\end{eqnarray}

where \nm{\sACC} and \nm{\alphaACC} can be obtained from the sensor specifications. Equation (\ref{eq:Sensor_Inertial_acc_error_FP_bis}) can be transformed by defining the accelerometer scale and cross coupling error matrix \nm{\MACC}:
\begin{eqnarray}
\nm{\MACC} & = & \nm{\RPA \ \SACC \ \RAP = \begin{bmatrix} \nm{m_{\sss{ACC,11}}} & 0 & 0 \\ \nm{m_{\sss{ACC,21}}} & \nm{m_{\sss{ACC,22}}} & 0 \\ \nm{m_{\sss{ACC,31}}} & \nm{m_{\sss{ACC,32}}} & \nm{m_{\sss{ACC,33}}} \end{bmatrix}}\nonumber \\
& \nm{\approx} & \nm{\begin{bmatrix} \nm{\sACCi} & 0 & 0 \\ \nm{\alphaACCiii \ \lrp{\sACCi - \sACCii}} & \nm{\sACCii} & 0 \\ \nm{\alphaACCii \ \lrp{\sACCiii - \sACCi}} & \nm{\alphaACCi \ \lrp{\sACCii - \sACCiii}} & \nm{\sACCiii} \end{bmatrix}}\label{eq:Sensors_Inertial_acc_M}
\end{eqnarray}

Considering that the scale and cross coupling errors are uncorrelated and very small, and relying on expressions derived in \cite{LIE} for the expected value and standard deviation for the sum and product of two random variables, the different components \nm{\mACCXij} of \nm{\MACC} can be obtained as follows \nm{\forall \ i, \ j \in \lrb{1,\ 2, \ 3}}:
\begin{eqnarray}
\nm{\mACCXij} & = & \nm{N\lrp{1, \ \sACC^2}                                                   \hspace{147pt} i = j}\label{eq:Sensors_Inertial_acc_error_scale_cross1} \\
\nm{\mACCXij} & = & \nm{N\lrp{0, \ \lrsb{\sqrt{2} \ \alphaACC \ \sACC}^2} = N\lrp{0, \mACC^2} \hspace{20pt} i > j}\label{eq:Sensors_Inertial_acc_error_scale_cross2} \\
\nm{\mACCXij} & = & \nm{0                                                                     \hspace{191pt} i < j}\label{eq:Sensors_Inertial_acc_error_scale_cross3}
\end{eqnarray}

The accelerometer error transformation matrix \nm{\NACC} is defined as:
\neweq{\NACC = \RPA \ \SACC = \begin{bmatrix} \nm{n_{\sss{ACC,11}}} & 0 & 0 \\ \nm{n_{\sss{ACC,21}}} & \nm{n_{\sss{ACC,22}}} & 0 \\ \nm{n_{\sss{ACC,31}}} & \nm{n_{\sss{ACC,32}}} & \nm{n_{\sss{ACC,33}}} \end{bmatrix} = \begin{bmatrix} \nm{\sACCi} & 0 & 0 \\ \nm{\alphaACCiii \ \sACCi} & \nm{\sACCii} & 0 \\ - \nm{\alphaACCii \ \sACCi} & \nm{\alphaACCi \ \sACCii} & \nm{\sACCiii} \end{bmatrix}}{eq:Sensors_Inertial_acc_N}

A process similar to that employed above leads to:
\begin{eqnarray}
\nm{\nACCXij} & = & \nm{N\lrp{1, \ \sACC^2}                                                     \hspace{133pt} i = j}\label{eq:Sensors_Inertial_acc_errorN_scale_cross1} \\
\nm{\nACCXij} & = & \nm{N\lrp{0, \ \alphaACC^2 \lrp{1 + \sACC^2}} \approx N\lrp{0, \alphaACC^2} \hspace{20pt} i > j}\label{eq:Sensors_Inertial_acc_errorN_scale_cross2} \\
\nm{\nACCXij} & = & \nm{0                                                                       \hspace{178pt} i < j}\label{eq:Sensors_Inertial_acc_errorN_scale_cross3}
\end{eqnarray}

Knowing that the cross coupling errors are very small \nm{\lrp{1 + \alphaACC^2 \approx 1}}, and relying again on the mean and standard deviation expressions employed above, it can be proven that the bias and white noise errors viewed in \nm{\FP} respond to a expression similar to (\ref{eq:Sensor_SensorModel_error}):
\begin{eqnarray}
\nm{\vec e_{\sss BW,ACC}^{\sss P}} & = & \nm{\vec e_{\sss BW,ACC}^{\sss P} \lrp{s \, \DeltatSENSED} = \vec e_{\sss BW,ACC}^{\sss P} \lrp{s \, \Deltat} = \NACC \ \vec e_{\sss BW,ACC}^{\sss A}} \nonumber \\
& = & \nm{\BzeroACC \, \NuzeroACC + \sigmauACC \, \Deltat^{1/2} \, \sum_{i=1}^s \NuiACC + \dfrac{\sigmavACC}{\Deltat^{1/2}} \, \NvsACC} \label{eq:Sensors_Inertia_acc_error}
\end{eqnarray}

where each \nm{\NuACC} and \nm{\NvACC} is a random vector composed by three independent standard normal random variables \nm{N\lrp{0, \, 1}}. Note that as the bias drift is mostly a warm up process that stabilizes itself after a few minutes of operation, the random walk within (\ref{eq:Sensors_Inertia_acc_error}) is not allowed to vary freely but is restricted to within a band of width \nm{\pm \ 100 \, \sigmauACC \, \Deltat^{1/2}}. The final \nm{\FP} accelerometer model is hence the following:
\neweq{\fIPPtilde = \MACC \ \fIPP + \vec e_{\sss BW,ACC}^{\sss P}}{eq:Sensor_Inertial_acc_error_FP}

where \nm{\MACC} is described in (\ref{eq:Sensors_Inertial_acc_M}) through (\ref{eq:Sensors_Inertial_acc_error_scale_cross3}) and \nm{\vec e_{\sss BW,ACC}^{\sss P}} is provided by (\ref{eq:Sensors_Inertia_acc_error}). This model relies on inputs for the bias offset \nm{\BzeroACC}, bias drift \nm{\sigmauACC}, white noise \nm{\sigmavACC}, scale factor error \nm{\sACC}, and cross coupling error \nm{\mACC}, which are obtained from the accelerometer specifications shown in table \ref{tab:Sensors_acc} within section \ref{subsec:Sensors_Inertial_Selected_gyr_acc}.


\subsection{Gyroscope Triad Sensor Error Model}\label{subsec:Sensors_Gyroscope_Triad_ErrorModel}

The \hypertt{IMU} is also equipped with a triad of gyroscopes, each of which measures the projection of the inertial angular velocity over its sensing axis as described in section \ref{subsec:Sensors_Inertial_ErrorModelSingleAxis}. The gyroscope triad model is analogous to that of the accelerometers, with the added difficulty that the transformation between the gyroscope frame \nm{\FY} and \nm{\FP} relies on six small angles instead of three, as described in section \ref{subsec:RefSystems_Y}. The starting point hence is:
\neweq{\wIPPtilde = \RPY \ \SGYR \ \lrp{\RYP \ \wIPP + \vec e_{\sss BW,GYR}^{\sss Y}}}{eq:Sensor_Inertial_gyr_error_FP_bis}

where \nm{\wIPP} is the inertial angular velocity (section \ref{sec:EquationsMotion_velocity}) viewed in \nm{\FP}, \nm{\wIPPtilde} represents its measurement also viewed in \nm{\FP}, \nm{\vec e_{\sss BW,GYR}^{\sss Y}} is the error introduced by each gyroscope (\ref{eq:Sensor_SensorModel_error}), \nm{\SGYR} is a square diagonal matrix containing the scale factor errors \nm{\lrb{\sGYRi, \ \sGYRii, \ \sGYRii}}, and \nm{\RPY} and \nm{\RYP}, defined by (\ref{eq:RefSystems_PY_YP}), contain the cross coupling errors \nm{\alphaGYRiXii, \ \alphaGYRiiXi, \ \alphaGYRiXiii, \ \alphaGYRiiiXi, \ \alphaGYRiiXiii, \ \alphaGYRiiiXii} generated by the misalignment of the gyroscope sensing axes.

Operating in the same way as in section \ref{subsec:Sensors_Accelerometer_Triad_ErrorModel} leads to:
\begin{eqnarray}
\nm{\vec e_{\sss BW,GYR}^{\sss P}} & = & \nm{\vec e_{\sss BW,GYR}^{\sss P} \lrp{s \, \DeltatSENSED} = \vec e_{\sss BW,GYR}^{\sss P} \lrp{s \, \Deltat}}\nonumber \\
& = & \nm{\BzeroGYR \, \NuzeroGYR + \sigmauGYR \, \Deltat^{1/2} \, \sum_{i=1}^s \NuiGYR + \dfrac{\sigmavGYR}{\Deltat^{1/2}} \, \NvsGYR}\label{eq:Sensors_Inertia_gyr_error} \\
\nm{\wIPPtilde} & = & \nm{\MGYR \ \wIPP + \vec e_{\sss BW,GYR}^{\sss P}}\label{eq:Sensor_Inertial_gyr_error_FP}
\end{eqnarray}

where each \nm{\NuiGYR} and \nm{\NvGYR} is a random vector composed by three independent standard normal random variables \nm{N\lrp{0, \, 1}}. As in the case of the accelerometers, the random walk within (\ref{eq:Sensors_Inertia_gyr_error}) representing the bias drift is not allowed to vary freely but is restricted to within a band of width \nm{\pm \ 100 \, \sigmauGYR \, \Deltat^{1/2}}. This model relies on inputs for the bias offset \nm{\BzeroGYR}, bias drift \nm{\sigmauGYR}, white noise \nm{\sigmavGYR}, scale factor error \nm{\sGYR}, and cross coupling error \nm{\mGYR}, which are obtained from the gyroscope specifications shown in table \ref{tab:Sensors_gyr} within section \ref{subsec:Sensors_Inertial_Selected_gyr_acc}. The gyroscope scale and cross coupling error matrix \nm{\MGYR} responds to:
\begin{eqnarray}
\nm{\MGYR} & = & \nm{\RPY \ \SGYR \ \RYP = \begin{bmatrix} \nm{m_{\sss{GYR,11}}} & \nm{m_{\sss{GYR,12}}} & \nm{m_{\sss{GYR,13}}} \\ \nm{m_{\sss{GYR,21}}} & \nm{m_{\sss{GYR,22}}} & \nm{m_{\sss{GYR,23}}} \\ \nm{m_{\sss{GYR,31}}} & \nm{m_{\sss{GYR,32}}} & \nm{m_{\sss{GYR,33}}} \end{bmatrix}}\nonumber \\
& \nm{\approx} & \nm{\begin{bmatrix} \nm{\sGYRi} & \nm{\alphaGYRiiXiii \ \lrp{\sGYRi - \sGYRii}} & \nm{\alphaGYRiiiXii \ \lrp{\sGYRiii - \sGYRi}} \\ \nm{\alphaGYRiXiii \ \lrp{\sGYRi - \sGYRii}} & \nm{\sGYRii} & \nm{\alphaGYRiiiXi \ \lrp{\sGYRii - \sGYRiii}} \\ \nm{\alphaGYRiXii \ \lrp{\sGYRiii - \sGYRi}} & \nm{\alphaGYRiiXi \ \lrp{\sGYRii - \sGYRiii}}  & \nm{\sGYRiii} \end{bmatrix}}\label{eq:Sensors_Inertial_gyr_M} \\
\nm{\mGYRXij} & = & \nm{N\lrp{1, \ \sGYR^2}                                         \hspace{122pt} i = j}  \label{eq:Sensors_Inertial_gyr_error_scale_cross1} \\
\nm{\mGYRXij} & = & \nm{N\lrp{0, \ 2 \ \alphaGYR^2 \ \sGYR^2} = N\lrp{0, \ \mGYR^2} \hspace{20pt} i \neq j}\label{eq:Sensors_Inertial_gyr_error_scale_cross2}
\end{eqnarray}


\subsection{Mounting of Inertial Sensors}\label{subsec:Sensors_Inertial_Mounting}

Equations (\ref{eq:Sensor_Inertial_acc_error_FP}) and (\ref{eq:Sensor_Inertial_gyr_error_FP}) contain the relationships between the specific force \nm{\fIPP} and inertial angular velocity \nm{\wIPP} and their measurements \nm{\lrp{\fIPPtilde, \ \nm{\wIPPtilde}}}, when evaluated and viewed in \nm{\FP}. However, from the point of view of the navigation system, both vectors need to be evaluated and viewed at the body frame \nm{\FB}, not \nm{\FP}. These equations thus need to be modified so they relate \nm{\fIBB} with \nm{\fIBBtilde} as well as \nm{\wIBB} with \nm{\wIBBtilde}, respectively, as described in section \ref{subsec:Sensors_Inertial_ErrorModel} below. To do that, it is necessary to define the relative pose (position plus attitude) between the \nm{\FP} and \nm{\FB} frames.
\begin{figure}[h]
\centering
\begin{tikzpicture}[auto,>=latex',scale=1.4]
	\coordinate (OB)    at (+0.0,+0.0);
	\coordinate (OR)    at ($(OB) + (+2.1,-0.2)$);
	\coordinate (OP)    at ($(OB) + (-1.0,-1.0)$);
	\coordinate (OC)    at ($(OB) + (-3.0,-1.2)$);
	
	\coordinate (iBi)   at  ($(OB) + (180:0.6cm)$);
	\coordinate (iBiii) at  ($(OB) + (-90:0.6cm)$);
	\node [left of=iBi, node distance=0.2cm] () {\nm{\iBi}};
	\node [right of=iBiii, node distance=0.3cm] () {\nm{\iBiii}};
	\draw [ultra thick] [-{Stealth[length=3mm]}] (OB) -- (iBi) {};
	\draw [ultra thick] [-{Stealth[length=3mm]}] (OB) -- (iBiii) {};		
	\filldraw [black] (OB) circle [radius=2pt] node [above=5pt] {\nm{\OB}};
		
	\coordinate (iRi)   at  ($(OR) + (180:0.6cm)$);
	\coordinate (iRiii) at  ($(OR) + (-90:0.6cm)$);
	\node [left of=iRi, node distance=0.2cm] () {\nm{\iRi}};
	\node [right of=iRiii, node distance=0.3cm] () {\nm{\iRiii}};
	\draw [ultra thick] [-{Stealth[length=3mm]}] (OR) -- (iRi) {};
	\draw [ultra thick] [-{Stealth[length=3mm]}] (OR) -- (iRiii) {};		
	\filldraw [black] (OR) circle [radius=2pt] node [right=2pt] {};	
	\coordinate (temp)   at  ($(OR) + (-0.3,-0.3)$);
	\node [left of=temp, node distance=1pt] () {\nm{\OR}};

	\coordinate (iPi)   at  ($(OP) + (165:0.6cm)$);
	\coordinate (iPiii) at  ($(OP) + (-105:0.6cm)$);
	\node [left of=iPi, node distance=0.2cm] () {\nm{\iPi}};
	\node [left of=iPiii, node distance=0.3cm] () {\nm{\iPiii}};
	\draw [ultra thick] [-{Stealth[length=3mm]}] (OP) -- (iPi) {};
	\draw [ultra thick] [-{Stealth[length=3mm]}] (OP) -- (iPiii) {};		
	\filldraw [black] (OP) circle [radius=2pt] node [above=3pt] {\nm{\OP}};	
	
	\coordinate (iCii)  at  ($(OC) + (15:0.6cm)$);
	\coordinate (iCiii) at  ($(OC) + (-75:0.6cm)$);
	\node [above of=iCii, node distance=0.2cm] () {\nm{\iCii}};
	\node [right of=iCiii, node distance=0.3cm] () {\nm{\iCiii}};
	\draw [ultra thick] [-{Stealth[length=3mm]}] (OC) -- (iCii) {};
	\draw [ultra thick] [-{Stealth[length=3mm]}] (OC) -- (iCiii) {};		
	\filldraw [black] (OC) circle [radius=2pt] node [left=5pt] {\nm{\OC}};
	
	\draw [ultra thin] [{Stealth[length=1mm]}-{Stealth[length=1mm]}] ($(OC) + (+0.0,+1.7)$) -- node[above=1mm] {500 mm} ($(OR) + (+0.0,+0.7)$);
	\draw [ultra thin] [{Stealth[length=1mm]}-{Stealth[length=1mm]}] ($(OP) + (+0.0,-0.9)$) -- node[below=1mm] {300 mm} ($(OR) + (+0.0,-1.7)$);
	\draw [ultra thin] [{Stealth[length=1mm]}-{Stealth[length=1mm]}] ($(OB) + (+0.0,-1.7)$) -- node[above=1mm] {207-219 mm} ($(OR) + (+0.0,-1.5)$);
	
	\draw [ultra thin] [{Stealth[length=1mm]}-{Stealth[length=1mm]}] ($(OR) + (+0.5,+0.0)$) -- node[right=1mm] {100 mm} ($(OP) + (+3.6,+0.0)$);
	\draw [ultra thin] [{Stealth[length=1mm]}-{Stealth[length=1mm]}] ($(OR) + (+0.5,+0.0)$) -- node[right=1mm] {5-6 mm} ($(OB) + (+2.6,+0.0)$);
	\draw [ultra thin] [{Stealth[length=1mm]}-{Stealth[length=1mm]}] ($(OR) + (+2.2,+0.0)$) -- node[right=1mm] {120 mm} ($(OC) + (+7.3,+0.0)$);
	
	\draw [ultra thin, dashed] [-] ($(OC) + (+0.0,+1.7)$) -- (OC);
	\draw [ultra thin, dashed] [-] ($(OP) + (+0.0,-0.9)$) -- (OP);
	\draw [ultra thin, dashed] [-] ($(OB) + (+0.0,-1.7)$) -- (OB);
	\draw [ultra thin, dashed] [-] ($(OR) + (+0.0,+0.7)$) -- (OR);
	\draw [ultra thin, dashed] [-] ($(OR) + (+0.0,-1.5)$) -- (OR);
	
	\draw [ultra thin, dashed] [-] ($(OC) + (+7.3,+0.0)$) -- (OC);
	\draw [ultra thin, dashed] [-] ($(OP) + (+3.6,+0.0)$) -- (OP);
	\draw [ultra thin, dashed] [-] ($(OB) + (+2.6,+0.0)$) -- (OB);
	\draw [ultra thin, dashed] [-] ($(OR) + (+2.2,+0.0)$) -- (OR);
		
\end{tikzpicture}
\caption{Relative position within aircraft symmetry plane of the \nm{\FB}, \nm{\FR}, \nm{\FP} and \nm{\FC} frames}
\label{fig:Sensor_symmetry_plane}
\end{figure}

The \hypertt{IMU}, represented by the platform frame \nm{\FP}, should be mounted inside the aircraft as close as possible to its center of mass (this reduces errors, as described in section \ref{subsec:Sensors_Inertial_ErrorModel}), and loosely aligned with the \nm{\FB} axes, as depicted in figure \ref{fig:Sensor_symmetry_plane}. In the simulation, the location of the \hypertt{IMU} is deterministic but its orientation stochastic. The \nm{\FP} origin \nm{\OP} is located \nm{300 \ mm} forward and \nm{100 \ mm} below the aircraft reference point \nm{\OR}. Considering (\ref{eq:AircraftModel_Trbb_full}), (\ref{eq:AircraftModel_Trbb_empty}), and (\ref{eq:AircraftModel_Trbb}), this results in:
\begin{eqnarray}
\nm{\TBPBfull}  & = & \nm{\begin{bmatrix} 0.093 & 0 & 0.105 \end{bmatrix}^T \ m} \label{eq:Sensors_Inertial_Mounting_Tbpb_full} \\
\nm{\TBPBempty} & = & \nm{\begin{bmatrix} 0.081 & 0 & 0.106 \end{bmatrix}^T \ m} \label{eq:Sensors_Inertial_Mounting_Tbpb_empty} \\
\nm{\TBPB}      & = & \nm{\vec f_{P}\lrp{m} = \TBPBfull + \dfrac{m_{full} - m}{m_{full} - m_{empty}} \; \lrp{\TBPBempty - \TBPBfull}} \label{eq:Sensors_Inertial_Mounting_Tbpb}
\end{eqnarray}

The rotation or lack of complete alignment between the \nm{\FB} and \nm{\FP} frames is represented by means of the platform Euler angles \nm{\phiBP = \lrsb{\psiP \ \ \thetaP \ \ \xiP}^T} defined in section \ref{subsec:RefSystems_P}. The author has arbitrarily established the stochastic model provided by (\ref{eq:Sensors_Inertial_Mounting_eulerBP}), in which each specific Euler angle is obtained as the product of the standard deviations (\nm{\sigmapsiP}, \nm{\sigmathetaP}, \nm{\sigmaxiP}) contained in table \ref{tab:Sensor_Inertial_mounting} by a single realization of a standard normal random variable \nm{N\lrp{0, \, 1}} (\nm{\NpsiP}, \nm{\NthetaP}, and \nm{\NxiP}):
\neweq{\phiBP = \lrsb{\sigmapsiP \, \NpsiP \ \ \ \ \sigmathetaP \, \NthetaP \ \ \ \ \sigmaxiP \NxiP}^T}{eq:Sensors_Inertial_Mounting_eulerBP}
\begin{center}
\begin{tabular}{lccc}
	\hline
	Concept							   & Variable			 & Value & Unit \\
	\hline
	True yaw error 	   	               & \nm{\sigmapsiP}     & \nm{0.5}  & \nm{^{\circ}}  \\
	True pitch error                   & \nm{\sigmathetaP}   & \nm{2.0}  & \nm{^{\circ}}  \\
	True bank error 		           & \nm{\sigmaxiP}      & \nm{0.1}  & \nm{^{\circ}}  \\
	Platform position estimation error & \nm{\sigmaTBPBest}  & \nm{0.01} & \nm{m} \\
	Platform angular estimation error  & \nm{\sigmaphiBPest}	& \nm{0.03} & \nm{^{\circ}}  \\
	\hline
\end{tabular}
\end{center}
\captionof{table}{\texttt{IMU} mounting accuracy values}\label{tab:Sensor_Inertial_mounting}

The estimation of the relative pose between the \nm{\FB} and \nm{\FP} frames is discussed in section \ref{sec:PreFlight_platform_frame}. Stochastic models are considered for both the translation \nm{\TBPBest} and rotation \nm{\phiBPest}, changing their values from one simulation run to another:
\begin{eqnarray}
\nm{\TBPBest} & = & \nm{\TBPB + \lrsb{\sigmaTBPBest \, \NTBPBiest \ \ \ \ \sigmaTBPBest \, \NTBPBiiest \ \ \ \ \sigmaTBPBest \NTBPBiiiest}^T} \label{eq:Sensors_Inertial_Mounting_TBPBest} \\
\nm{\phiBPest} & = & \nm{\phiBP + \lrsb{\sigmaphiBPest \, \NpsiPest \ \ \ \ \sigmaphiBPest \, \NthetaPest \ \ \ \ \sigmaphiBPest \NxiPest}^T} \label{eq:Sensors_Inertial_Mounting_eulerBPest}
\end{eqnarray}

where the standard deviations \nm{\sigmaTBPBest} and \nm{ \sigmaphiBPest} are shown in table \ref{tab:Sensor_Inertial_mounting}, and \nm{\NpsiPest}, \nm{\NthetaPest}, \nm{\NxiPest}, \nm{\NTBPBiest}, \nm{\NTBPBiiest}, \nm{\NTBPBiiiest} are six realizations of a standard normal random variable \nm{N\lrp{0, \, 1}}.

The translation \nm{\TBP} between the origins of the \nm{\FB} and \nm{\FP} frames can be considered quasi stationary as it slowly varies based on the aircraft mass (\ref{eq:Sensors_Inertial_Mounting_Tbpb}), and the relative position of their axes \nm{\phiBP} remains constant because the \hypertt{IMU} is rigidly attached to the aircraft structure (\ref{eq:Sensors_Inertial_Mounting_eulerBP}):
\neweq{\dot{\vec T}_{\sss BP} = \dot{\vec q}_{\sss BP} = \dot{\vec \zeta}_{\sss BP} = \vec 0 \ \longrightarrow \ \vBP = \vec a_{\sss BP} = \wBP = \vec \alpha_{\sss BP} = \vec 0}{eq:Sensor_Inertial_BP_diff}


\subsection{Comprehensive Inertial Sensor Error Model}\label{subsec:Sensors_Inertial_ErrorModel}

Two considerations are required to establish the measurement equations for the inertial sensors. First, the application of the composition rules discussed in \cite{LIE} considering \nm{\FI} as \nm{F_{\sss0}}, \nm{\FB} as \nm{F_{\sss1}}, and \nm{\FP} as \nm{F_{\sss2}}, results in:
\begin{eqnarray}
\nm{\wIP} & = & \nm{\wIB}\label{eq:Sensor_Inertial_BP_omega} \\
\nm{\vec \alpha_{\sss IP}} & = & \nm{\vec \alpha_{\sss IB}}\label{eq:Sensor_Inertial_BP_alpha} \\
\nm{\vIP} & = & \nm{\vIB + \wIBskew \ \TBP}\label{eq:Sensor_Inertial_BP_v} \\
\nm{\vec a_{\sss IP}} & = & \nm{\vec a_{\sss IB} + \alphaIBskew \ \TBP + \wIBskew \ \wIBskew \ \TBP}\label{eq:Sensor_Inertial_BP_a}
\end{eqnarray}

Second, it is also necessary to consider that as all rows and columns within \nm{\RBPest \equiv \qBPest} are unitary vectors, the projection of \nm{\vec e_{\sss BW,ACC}^{\sss P}} and \nm{\vec e_{\sss BW,GYR}^{\sss P}} from \nm{\FP} onto \nm{\FB} does not change their stochastic properties:
\begin{eqnarray}
\nm{\vec e_{\sss BW,ACC}^{\sss B} \lrp{s \, \Deltat}} & = & \nm{\vec g_{\ds{\hat{\vec \zeta}_{{\sss BP}*}}} \big(\vec e_{\sss BW,ACC}^{\sss P} \big) = \BzeroACC \, \NuzeroACC + \sigmauACC \, \Deltat^{1/2} \, \sum_{i=1}^s \NuiACC + \dfrac{\sigmavACC}{\Deltat^{1/2}} \, \NvsACC}\label{eq:Sensors_Inertia_acc_bw_error} \\
\nm{\vec e_{\sss BW,GYR}^{\sss B} \lrp{s \, \Deltat}} & = & \nm{\vec g_{\ds{\hat{\vec \zeta}_{{\sss BP}*}}} \big(\vec e_{\sss BW,GYR}^{\sss P} \big) = \BzeroGYR \, \NuzeroGYR + \sigmauGYR \, \Deltat^{1/2} \, \sum_{i=1}^s \NuiGYR + \dfrac{\sigmavGYR}{\Deltat^{1/2}} \, \NvsGYR}\label{eq:Sensors_Inertia_gyr_bw_error}
\end{eqnarray}

As the inertial angular velocity does not change when evaluated in the \nm{\FB} and \nm{\FP} frames (\ref{eq:Sensor_Inertial_BP_omega}), its measurement in \nm{\FB} can be derived from (\ref{eq:Sensor_Inertial_gyr_error_FP}) by first projecting it from \nm{\FB} to \nm{\FP} based on the real attitude \nm{\qBP} and then projecting back the measurement into \nm{\FB} based on the estimated attitude \nm{\qBPest}. The bias and white noise error is also projected according to (\ref{eq:Sensors_Inertia_gyr_bw_error}):
\neweq{\wIBBtilde = \vec {Ad}_{\ds{\hat{\vec q}_{\sss BP}}} \ \MGYR \ \vec {Ad}_{\ds{\vec q_{\sss BP}}}^{-1} \ \wIBB + \vec e_{\sss BW,GYR}^{\sss B}}{eq:Sensor_Inertial_gyr_error_final}

The expression for the specific force measurement is significantly more complex because the back and forth transformations of the specific force between the \nm{\FB} and \nm{\FP} frames need to consider the influence of the lever arm \nm{\TBP}, as indicated in (\ref{eq:Sensor_Inertial_BP_a}). The additional terms introduce errors in the measurements, so it is desirable to locate the \hypertt{IMU} as close as possible to the aircraft center of mass.
\neweq{\fIBBtilde = \vec g_{\ds{\hat{\vec \zeta}_{{\sss BP}*}}} \bigg(\MACC \Big(\vec g_{\ds{\vec \zeta_{{\sss BP}*}}}^{-1} \big(\fIBB + \alphaIBBskew \ \TBPB + \wIBBskew \ \wIBBskew \ \TBPB\big)\Big)\bigg) - \alphaIBBestskew \ \TBPBest - \wIBBestskew \ \wIBBestskew \ \TBPBest + \vec e_{\sss BW,ACC}^{\sss B}}{eq:Sensor_Inertial_acc_error_prefinal}

Note that this expression can not be evaluated as the estimated values for the inertial angular velocity and acceleration (\nm{\wIBBest, \ \alphaIBBest}) are unknown by the \hypertt{IMU} until obtained by the inertial navigation filter (chapter \ref{cha:nav}). The \hypertt{IMU} can however rely on the gyroscope readings, directly replacing \nm{\wIBBest} with \nm{\wIBBtilde} and computing \nm{\alphaIBBtilde} based on the difference between the present and previous \nm{\wIBBtilde} readings, resulting in:
\neweq{\fIBBtilde = \vec g_{\ds{\hat{\vec \zeta}_{{\sss BP}*}}} \bigg(\MACC \Big(\vec g_{\ds{\vec \zeta_{{\sss BP}*}}}^{-1} \big(\fIBB + \alphaIBBskew \ \TBPB + \wIBBskew \ \wIBBskew \ \TBPB\big)\Big)\bigg) - \alphaIBBtildeskew \ \TBPBest - \wIBBtildeskew \ \wIBBtildeskew \ \TBPBest + \vec e_{\sss BW,ACC}^{\sss B}}{eq:Sensor_Inertial_acc_error_final}

Table \ref{tab:Sensors_Inertial_error_sources} lists the error sources contained in the comprehensive inertial sensor error model represented by (\ref{eq:Sensor_Inertial_gyr_error_final}, \ref{eq:Sensor_Inertial_acc_error_final}) [A\ref{as:SENSOR_ACC}, A\ref{as:SENSOR_GYR}]. The first two columns list the different error sources, while the third column specifies their origin according to the criterion established in the first paragraph of section \ref{subsec:Sensors_Inertial_ErrorSources}. Last, the section where each error is described appears on the fourth column:
\begin{center}
\begin{tabular}{lccc}
	\hline
	Error				& Symbol						& Main Source	& Description \\
	\hline
	Bias offset			& \nm{\BzeroACC, \ \BzeroGYR}	& run-to-run	& \ref{subsec:Sensors_Inertial_ErrorModelSingleAxis}													\\
	Bias drift			& \nm{\sigmauACC, \ \sigmauGYR} & in-run		& \ref{subsec:Sensors_Inertial_ErrorModelSingleAxis}													\\
	System noise		& \nm{\sigmavACC, \ \sigmavGYR}	& in-run		& \ref{subsec:Sensors_Inertial_ErrorModelSingleAxis}													\\
	Scale factor		& \nm{\sACC, \ \sGYR}			& fixed \& T	& \ref{subsec:Sensors_Accelerometer_Triad_ErrorModel}, \ref{subsec:Sensors_Gyroscope_Triad_ErrorModel}	\\
	Cross coupling		& \nm{\mACC, \ \mGYR}			& fixed			& \ref{subsec:Sensors_Accelerometer_Triad_ErrorModel}, \ref{subsec:Sensors_Gyroscope_Triad_ErrorModel}	\\
	Lever arm			& \nm{\TBP, \ \sigmaTBPBest}	& fixed			& \ref{subsec:Sensors_Inertial_Mounting}																\\
	\hypertt{IMU} Attitude		& \nm{\sigmapsiP, \ \sigmathetaP, \ \sigmaxiP, \ \sigmaphiBPest} & fixed & \ref{subsec:Sensors_Inertial_Mounting}										\\
	\hline
\end{tabular}
\end{center}
\captionof{table}{Inertial sensor error sources}\label{tab:Sensors_Inertial_error_sources}

Note that \nm{\BzeroACC}, \nm{\BzeroGYR}, \nm{\sigmauACC}, \nm{\sigmauGYR}, \nm{\sigmavACC}, \nm{\sigmavGYR}, \nm{\sACC}, \nm{\sGYR}, \nm{\mACC}, and \nm{\mGYR} are taken from the right hand column of tables \ref{tab:Sensors_gyr} and \ref{tab:Sensors_acc} (section \ref{subsec:Sensors_Inertial_Selected_gyr_acc}), while the remaining error sources come from section \ref{subsec:Sensors_Inertial_Mounting}. It is worth pointing out that all errors are modeled as stochastic variables or processes (with the exception of the \nm{\TBP} displacement between the body and platform frames, which is deterministic), as expressions (\ref{eq:Sensor_Inertial_gyr_error_final}, \ref{eq:Sensor_Inertial_acc_error_final}) rely on the errors provided by (\ref{eq:Sensors_Inertia_acc_bw_error}, \ref{eq:Sensors_Inertia_gyr_bw_error}), the scale and cross coupling matrices given by (\ref{eq:Sensors_Inertial_acc_M}, \ref{eq:Sensors_Inertial_gyr_M}), and the transformations given by (\ref{eq:Sensors_Inertial_Mounting_Tbpb}, \ref{eq:Sensors_Inertial_Mounting_eulerBP}, \ref{eq:Sensors_Inertial_Mounting_TBPBest}, \ref{eq:Sensors_Inertial_Mounting_eulerBPest}).

In the case of the accelerometer triad, the stochastic nature of the fixed and run-to-run error contributions to the model relies on three realizations of normal distributions for the bias offset, three for the scale factor errors, three for the cross coupling errors, and nine for the mounting errors, while the in-run error contributions require three realizations each for the bias drift and system noise at every discrete sensor measurement. The gyroscope triad is similar but requires six realizations to model the cross coupling errors instead of three, while using the same six realizations that the accelerometer triad to model the true and estimated rotation between the \nm{\FB} and \nm{\FP} frames. Refer to section \ref{sec:Sensors_implementation} for an explanation of how the different stochastic variables present in the models are evaluated during simulation.

Expressions (\ref{eq:Sensor_Inertial_gyr_error_final}, \ref{eq:Sensor_Inertial_acc_error_final}) can be rewritten to show the measurements as functions of the full errors (\nm{\EACC, \, \EGYR}), which represent all the errors introduced by the inertial sensors with the exception of white noise. These expressions are adopted in the inertial navigation algorithms of chapter \ref{cha:nav}:
\begin{eqnarray}
\nm{\fIBBtilde \lrp{s \, \Deltat}} & = & \nm{\fIBB \lrp{s \, \Deltat} + \EACC \lrp{s \, \Deltat} + \dfrac{\sigmavACC}{\Deltat^{1/2}} \, \NvsACC} \label{eq:Sensor_Inertial_acc_error_filter} \\
\nm{\wIBBtilde \lrp{s \, \Deltat}} & = & \nm{\wIBB \lrp{s \, \Deltat} + \EGYR \lrp{s \, \Deltat} + \dfrac{\sigmavGYR}{\Deltat^{1/2}} \, \NvsGYR} \label{eq:Sensor_Inertial_gyr_error_filter}
\end{eqnarray}


\subsection{Selected Inertial Sensors}\label{subsec:Sensors_Inertial_Selected_gyr_acc}

The default gyroscopes employed in the simulation, denoted as \say{baseline}, correspond to the \hypertt{MEMS} gyroscopes installed inside the Analog Devices \texttt{ADIS16488A} \hypertt{IMU} \cite{ADIS16488A}. Table \ref{tab:Sensors_gyr} shows the baseline performances, which have been taken from the data sheet when possible, and corrected when suspicious. A calibration process as that described in section \ref{sec:PreFlight_Inertial_Calibration} is assumed to eliminate \nm{95\%} of the scale factor and cross coupling errors.
\begin{center}
\begin{tabular}{lrc|lllc}
	\hline
	\hypertt{GYR} Baseline & \multicolumn{1}{c}{Spec} & Unit & \multicolumn{1}{c}{Variable} & \multicolumn{1}{c}{Value} & \multicolumn{1}{c}{Calibration} & Unit \\
	\hline
	In-run bias stability (1 \nm{\sigma})	& 5.10			& \nm{^{\circ}/hr}			& \nm{\sigmauGYR}	& \nm{1.42 \cdot 10^{-4}}	& \nm{1.42 \cdot 10^{-4}}	& \nm{^{\circ}/s^{1.5}} \\
	Angle random walk (1 \nm{\sigma})		& 0.26			& \nm{^{\circ}/hr^{0.5}}	& \nm{\sigmavGYR}	& \nm{4.30 \cdot 10^{-3}}	& \nm{4.30 \cdot 10^{-3}}	& \nm{^{\circ}/s^{0.5}} \\
	Nonlinearity\footnotemark				& 0.01			& \nm{\%}		 			& \nm{\sGYR}		& \nm{3.00 \cdot 10^{-4}}	& \nm{1.50 \cdot 10^{-5}}	& - \\
	Misalignment							& \nm{\pm} 0.05	& \nm{^{\circ}}				& \nm{\mGYR}		& \nm{8.70 \cdot 10^{-4}}	& \nm{4.35 \cdot 10^{-5}}	& - \\
	Bias repeatability (1 \nm{\sigma})		& \nm{\pm} 0.2	& \nm{^{\circ}/s}			& \nm{\BzeroGYR}	& \nm{2.00 \cdot 10^{-1}}	& \nm{2.00 \cdot 10^{-1}}	& \nm{^{\circ}/s} \\
	\hline
\end{tabular}
\captionof{table}{Performance of ``baseline'' gyroscopes} \label{tab:Sensors_gyr}
\end{center}
\footnotetext{The \nm{0.01\ \%} scale factor error obtained in \cite{ADIS16488A} is considered too optimistic and hence modified to \nm{0.03 \ \% = 3.00 \cdot 10^{-4}}.}

The table contains three data columns. The left most column (\say{Spec}) corresponds to data taken directly from the sensors specifications, which get converted in the middle column (\say{Value}) to the parameters shown in sections \ref{subsec:Sensors_Inertial_ErrorModelSingleAxis} through \ref{subsec:Sensors_Inertial_ErrorModel}\footnote{The conversion between bias instability and \nm{\sigma_u} uses a period of \nm{100 \ s}, as noted in section \ref{subsec:Sensors_Inertial_ObtainmentBiasDrift}.}. The right column (\say{Calibration}) contains the specifications employed in the simulation after the calibration process, which reduces the scale factor and cross coupling errors by \nm{95\%}.
\begin{center}
\begin{tabular}{lrc|lllc}
	\hline
	\hypertt{ACC} Baseline & \multicolumn{1}{c}{Spec} & Unit & \multicolumn{1}{c}{Variable} & \multicolumn{1}{c}{Value} & \multicolumn{1}{c}{Calibration} & Unit \\
	\hline
	In-run bias stability (1 \nm{\sigma})	& 0.07 				& \nm{mg^{\prime}}\footnotemark	& \nm{\sigmauACC}	& \nm{6.86 \cdot 10^{-5}}	& \nm{6.86 \cdot 10^{-5}}	& \nm{m/s^{2.5}} \\
	Velocity random walk (1 \nm{\sigma})	& 0.029				& \nm{m/s/hr^{0.5}}				& \nm{\sigmavACC}	& \nm{4.83 \cdot 10^{-4}}	& \nm{4.83 \cdot 10^{-4}}	& \nm{m/s^{1.5}} \\
	Nonlinearity							& 0.1				& \nm{\%}		 				& \nm{\sACC}		& \nm{1.00 \cdot 10^{-3}}   & \nm{5.00 \cdot 10^{-5}}	& - \\
	Misalignment							& \nm{\pm} 0.035	& \nm{^{\circ}}					& \nm{\mACC}		& \nm{6.11 \cdot 10^{-4}}	& \nm{3.05 \cdot 10^{-5}}	& - \\
	Bias repeatability (1 \nm{\sigma})		& \nm{\pm} 16		& \nm{mg^{\prime}}				& \nm{\BzeroACC}	& \nm{1.57 \cdot 10^{-1}}	& \nm{1.57 \cdot 10^{-1}}	& \nm{m/s^2} \\
	\hline
\end{tabular}
\captionof{table}{Performance of ``baseline'' accelerometers} \label{tab:Sensors_acc}
\end{center}
\footnotetext{A \nm{\lrsb{g^{\prime}}} stands for the value of the sea level gravitational acceleration.}

The \hypertt{MEMS} accelerometers installed inside the Analog Devices \texttt{ADIS16488A} \hypertt{IMU} \cite{ADIS16488A} also constitute the \say{baseline} configuration. Table \ref{tab:Sensors_acc} obtains all its values from the data sheet, and considers that a calibration process as that described in section \ref{sec:PreFlight_Inertial_Calibration} eliminates \nm{95\%} of the scale factor and cross coupling errors\footnote{The conversion between bias instability and \nm{\sigma_u} uses a period of \nm{100 \ s}, as noted in section \ref{subsec:Sensors_Inertial_ObtainmentBiasDrift}.}.


\section{Non-Inertial Sensors}\label{sec:Sensors_NonInertial}

In addition to the \hypertt{IMU}, the aircraft is equipped with other sensors that provide useful information to its navigation system, such as the \emph{magnetometers}, which assist in the establishment of the aircraft heading, the \emph{air data system}, which in addition of the pressure altitude and temperature also provides a measurement of the air velocity, and the \hypertt{GNSS} receiver, which obtains absolute ground velocity and position observations.


\subsection{Magnetometers} \label{subsec:Sensors_NonInertial_Magnetometers}

Magnetometers measure magnetic field intensity along a given direction and are very useful for estimating the aircraft heading. Although other types exist, magnetoinductive and magnetoresistive sensors are generally employed for navigation due to their accuracy and small size \cite{Groves2008,Titterton2004}. As with the inertial sensors, three orthogonal magnetometers are usually employed in a strapdown configuration to measure the magnetic field with respect to \nm{\FB}.

Unfortunately magnetometers not only measure the Earth magnetic field \nm{\Bvec} (section \ref{sec:EarthModel_WMM}), but also that generated by the aircraft permanent magnets and electrical equipment (known as hard iron magnetism), as well as the magnetic field disturbances generated by the aircraft ferrous materials (soft iron magnetism). For that reason, the magnetometers should be placed in a location inside the aircraft that minimizes these errors. On the positive side, magnetometers do not exhibit the bias instability present in inertial sensors, and the error of an individual sensor can be properly modeled by the combination of bias offset and white noise. A triad of magnetometers capable of measuring the magnetic field in three directions adds the same scale factor (nonlinearity) and cross coupling (misalignment) errors as those present in the inertial sensors, together with the transformation between the magnetic axes and the body ones.

Modeling the behavior of a triad of magnetometers is simpler but less precise than that of inertial sensors, as the effect of the fixed hard iron magnetism is indistinguishable from that of the run-to-run bias offset, while the fixed effect of soft iron magnetism is indistinguishable from that of the scale factor and cross coupling error matrix. This has several consequences. First is that magnetometers can not be calibrated at the laboratory before being mounted in the aircraft as in the case of inertial sensors (section \ref{sec:PreFlight_Inertial_Calibration}), but are instead calibrated once attached to the aircraft by a process known as swinging (section \ref{sec:PreFlight_swinging}), which is less precise as the aircraft attitude during swinging can not be determined with so much accuracy as it would be in a laboratory setting. Second is that defining a magnetic platform frame to then transform the results into body axes serves no purpose, as the magnetometer readings are only valid, this is, contain the effects of hard and soft iron magnetism, once they are attached to the aircraft, and then they can be directly measured in body axes. And third is that percentage wise the errors induced by the magnetometers are bigger than those of the inertial sensors. The simulation relies on the following model:
\neweq{\BBtilde = \BzeroMAGvec + \BhiMAGvec + \MMAG \, \Big(\vec g_{\ds{\vec \zeta_{{\sss NB}*}}}^{-1} \big(\BNREAL\big)\Big) + \vec e_{\sss W,MAG}^{\sss B}}{eq:Sensor_NonInertial_mag_error}

where \nm{\BBtilde} is the measurement viewed in \nm{\FB}, \nm{\BzeroMAGvec} is the run-to-run bias offset, \nm{\BhiMAGvec} is the fixed hard iron magnetism, \nm{\MMAG} is a fixed matrix combining the effects of soft iron magnetism with the scale factor and cross coupling errors, and \nm{\BNREAL} is the real magnetic field including local anomalies (section \ref{sec:EarthModel_accuracy}).

The \say{baseline} magnetometer features are shown in table \ref{tab:Sensors_mag}, where the white noise has been taken from \cite{Groves2008} and the rest of the parameters correspond to the magnetometers present in the Analog Devices \texttt{ADIS16488A} \hypertt{IMU} \cite{ADIS16488A}. Although the value of hard and soft iron magnetism in aircraft is rather small, the author has not been able to obtain trusted values for them. To avoid eliminating sources of error, the author has decided to increase by 50\% the values for bias offset, scale factor, and cross coupling errors found in the literature, as shown in the column named \say{Compensation}. As both result in a similar effect, the author expects that the realism of the results is not adversely affected. In the case of the bias, most of the error has been assigned to \nm{\BhiMAG} and the remaining to the run-to-run \nm{\BzeroMAG}.

A swinging process as that described in section \ref{sec:PreFlight_swinging} is assumed to eliminate 90\% of the fixed error contributions, this is, the hard iron magnetism, the scale factor, and the cross coupling error\footnote{The soft iron effect is combined with the scale factor and cross coupling errors.}. This number is inferior to the 95\% reduction achieved by the calibration of inertial sensors because although the lack of magnetometer bias drift facilitates calibration, the determination of the body attitude during calibration is inherently less accurate working with the complete aircraft than just with the \hypertt{IMU} in a laboratory setting. 
\begin{center}
\begin{tabular}{lrc|llllc}
	\hline
	\hypertt{MAG} Baseline & \multicolumn{1}{c}{Spec} & Unit & \multicolumn{1}{c}{Variable} & \multicolumn{1}{c}{Value} & \multicolumn{1}{c}{Compensation} & \multicolumn{1}{c}{Swinging} & Unit \\
	\hline
	Output noise		& 5				& \nm{nT \cdot s^{0.5}}	& \nm{\sigmavMAG}	& \nm{5.00 \cdot 10^{0}}	& \nm{5.00 \cdot 10^{0}}	& \nm{5.00 \cdot 10^{0}}	& \nm{nT \cdot s^{0.5}} \\
	Nonlinearity		& 0.5			& \nm{\%}						& \nm{\sMAG}		& \nm{5.00 \cdot 10^{-3}}   & \nm{7.50 \cdot 10^{-3}}	& \nm{7.50 \cdot 10^{-4}}	& - \\
	Misalignment		& \nm{\pm} 0.35	& \nm{^{\circ}}		& \nm{\mMAG}		& \nm{6.11 \cdot 10^{-3}}	& \nm{9.16 \cdot 10^{-3}}	& \nm{9.16 \cdot 10^{-4}}	& - \\
	Bias (1 \nm{\sigma})& \multirow{2}{*}{\nm{\pm} 1500} & \multirow{2}{*}{\nm{nT}} 	& \nm{\BhiMAG}		& \multirow{2}{*}{\nm{1.50 \cdot 10^{3}}} & \nm{1.75 \cdot 10^{3}}	& \nm{1.75 \cdot 10^{2}} & \nm{nT} \\
	Repeatability		& 								 &								& \nm{\BzeroMAG}	&  & \nm{5.00 \cdot 10^{2}}	& \nm{5.00 \cdot 10^{2}} & \nm{nT} \\
	\hline
\end{tabular}
\captionof{table}{Performance of ``baseline'' magnetometers} \label{tab:Sensors_mag}
\end{center}

The final model for a triaxial magnetometer thus includes contributions from hard iron magnetism, bias offset, system noise, soft iron magnetism, scale factor, and cross coupling errors [A\ref{as:SENSOR_MAG}]:
\neweq{\BBtilde \lrp{s \, \Deltat} = \BhiMAG \, \NhiMAG + \BzeroMAG \, \NuzeroMAG + \MMAG \, \Big(\vec g_{\ds{\vec \zeta_{{\sss NB}*}}}^{-1} \big(\BNREAL\big)\Big) + \dfrac{\sigmavMAG}{\Deltat^{1/2}} \, \NvsMAG}{eq:Sensor_NonInertial_mag_error_final}

where \nm{\NhiMAG}, \nm{\NuzeroMAG} and \nm{\NvsMAG} are uncorrelated normal vectors of size three each composed of three uncorrelated standard normal random variables \nm{N\lrp{0, \, 1}}. The soft iron, scale factor and cross coupling matrix \nm{\vec M_{\sss MAG}} does not vary with time and is computed as follows:
\neweq{\vec M_{\sss MAG} = \begin{bmatrix} \nm{1 + s_{\sss MAG}} & \nm{m_{\sss MAG}} & \nm{m_{\sss MAG}} \\ \nm{m_{\sss MAG}} & \nm{1 + s_{\sss MAG}} & \nm{m_{\sss MAG}} \\ \nm{m_{\sss MAG}} & \nm{m_{\sss MAG}} & \nm{1 + s_{\sss MAG}} \end{bmatrix} \circ \vec{N}_{m,\sss MAG}} {eq:Sensor_NonInertial_summary_scale_factor}

In this expression \nm{\vec{N}_{m,\sss MAG}} contains nine outputs of a standard normal random variable \nm{N\lrp{0, \, 1}}, and the symbol \nm{\circ} represents the Hadamart or element-wise matrix product.

Table \ref{tab:Sensors_NonInertial_mag_error_sources} lists the error sources contained in the magnetometer model represented by (\ref{eq:Sensor_NonInertial_mag_error_final}) [A\ref{as:SENSOR_MAG}], noting that soft iron magnetism is included in both the scale factor and cross coupling errors. The first two columns list the different error sources, while the third one specifies their origin according to the criterion established in the first paragraph of section \ref{subsec:Sensors_Inertial_ErrorSources}.
\begin{center}
\begin{tabular}{lcc}
	\hline
	\multicolumn{2}{c}{Error} & Main Source \\
	\hline
	Hard iron			& \nm{\BhiMAG}		& fixed			\\
	Bias offset			& \nm{\BzeroMAG}	& run-to-run	\\
	System noise		& \nm{\sigmavMAG}	& in-run		\\
	Scale factor		& \nm{\sMAG}		& fixed			\\
	Cross coupling		& \nm{\mMAG}		& fixed			\\
	\hline
\end{tabular}
\end{center}
\captionof{table}{Magnetometer error sources}\label{tab:Sensors_NonInertial_mag_error_sources}

Note that \nm{\BhiMAG}, \nm{\BzeroMAG}, \nm{\sigmavMAG}, \nm{\sMAG}, and \nm{\mMAG} are taken from the right column of the specs listed above, and that all errors are modeled as stochastic variables or processes. The stochastic nature of the fixed and run-to-run error contributions to the magnetometer model relies on three realizations of normal distributions for the hard iron magnetism, three for the bias offset, three for the scale factor errors, and six for the cross coupling errors, while the in-run error contributions require three realizations for system noise at every discrete sensor measurement. Refer to section \ref{sec:Sensors_implementation} for an explanation of how the different stochastic variables present in the models are evaluated during simulation.

Expression (\ref{eq:Sensor_NonInertial_mag_error_final}) can be rewritten to show the measurements as functions of the magnetometer full error \nm{\EMAG}, which represents all the errors introduced by the magnetometers with the exception of white noise. This expression is adopted in the inertial navigation algorithms of chapter \ref{cha:nav}.
\neweq{\BBtilde \lrp{s \, \Deltat} = \BBREAL \lrp{s \, \Deltat} + \EMAG \lrp{s \, \Deltat} + \dfrac{\sigmavMAG}{\Deltat^{1/2}} \, \NvsMAG} {eq:Sensor_NonInertial_mag_error_filter}


\subsection{Air Data System} \label{subsec:Sensors_NonInertial_ADS}

The mission of the \emph{air data system} (\hypertt{ADS}) is to measure the aircraft pressure altitude \nm{\Hp} (section \ref{subsec:EarthModel_ISA_Definitions}) by means of the atmospheric pressure \emph{p}, the outside air temperature T, and the body air velocity \nm{\vTASB}, composed by the airspeed \nm{\vtas} plus the angles of attack \nm{\alpha} and sideslip \nm{\beta} (section \ref{sec:EquationsMotion_velocity}).

A barometer or static pressure sensor, generally part of the Pitot tube as explained below \cite{Titterton2004}, measures atmospheric pressure, which can be directly translated into pressure altitude by (\ref{eq:EarthModel_ISA_p_Hp}). The \nm{\sigmaOSP} value shown in table \ref{tab:Sensors_air} below, where \hypertt{OSP} stands for outside static pressure, comes from the \nm{\pm \, 10 \ m} altitude error listed in the specifications of the \texttt{Aeroprobe} air data system \cite{Aeroprobe}, which translates into \nm{\pm \, 100 \ Pa} at a pressure altitude of \nm{1500 \ m}  per (\ref{eq:EarthModel_ISA_p_Hp}). Although not present in the documentation, the author has decided to also include a bias offset \nm{\BzeroOSP} for added realism [A\ref{as:SENSOR_OSP}], so the error introduced by the barometer is modeled as a combination of bias offset and random noise:
\neweq{e_{\sss OSP} \lrp{s \, \DeltatSENSED} = e_{\sss OSP} \lrp{s \, \Deltat} = \widetilde{p}\lrp{s \, \Deltat} - p\lrp{s \, \Deltat} = \BzeroOSP \, \NzeroOSP +  \sigmaOSP \, \NsOSP}{eq:Sensor_NonInertial_OSP}

where \nm{\NzeroOSP} and \nm{\NsOSP} are uncorrelated standard normal random variables \nm{N\lrp{0, \, 1}}.

Air data systems are also equipped with a thermometer to measure the external air temperature T. As in the case of the barometer above, the specifications of the Analog Devices \texttt{ADT7420} temperature sensor \cite{ADT7420} only include system noise, but the author has decided to also include a bias offset for added realism [A\ref{as:SENSOR_OAT}]. The model contained in (\ref{eq:Sensor_NonInertial_OAT}) hence relies on the bias offset \nm{\BzeroOAT} and system noise \nm{\sigmaOAT} shown in table \ref{tab:Sensors_air}, where \hypertt{OAT} stands for outside air temperature, plus two uncorrelated standard normal random variables \nm{N\lrp{0, \, 1}} (\nm{\NzeroOAT} and \nm{\NsOAT}):
\neweq{e_{\sss OAT} \lrp{s \, \DeltatSENSED} = e_{\sss OAT} \lrp{s \, \Deltat} = \widetilde{T}\lrp{s \, \Deltat} - T\lrp{s \, \Deltat} = \BzeroOAT \, \NzeroOAT + \sigmaOAT \, \NsOAT}{eq:Sensor_NonInertial_OAT}
\begin{center}
\begin{tabular}{lrc|lrc}
	\hline
	\hypertt{OSP}-\hypertt{OAT} Baseline & \multicolumn{1}{c}{Spec} & Unit & \multicolumn{1}{c}{Variable} & \multicolumn{1}{c}{Value} & Unit \\
	\hline
	Altitude error			         & \nm{\pm \, 10}	& \nm{m}		   & \nm{\sigmaOSP}	& \nm{1.00 \cdot 10^{+2}}	& \nm{Pa} \\
	                    	         & 					&				   & \nm{\BzeroOSP}	& \nm{1.00 \cdot 10^{+2}}	& \nm{Pa} \\
	Temperature error (\nm{3 \sigma}) & \nm{\pm \, 0.15} & \nm{K} & \nm{\sigmaOAT}	& \nm{5.00 \cdot 10^{-2}}   & \nm{K} \\
	                    	         & 					&				   & \nm{\BzeroOAT}	& \nm{5.00 \cdot 10^{-2}}	& \nm{K} \\
	\hline
\end{tabular}
\captionof{table}{Performance of ``baseline'' atmospheric sensors} \label{tab:Sensors_air}
\end{center}

A \emph{Pitot probe} is a tube with no outlet pointing directly into the undisturbed air stream, where the values of the air variables (temperature, pressure, and density) at its dead end resemble the total or stagnation variables of the atmosphere prior to its deceleration inside the Pitot \cite{Eshelby2000}. \emph{Total atmospheric variables} (temperature \nm{T_t}, pressure \nm{p_t}, and density \nm{\rho_t}) are those obtained if a moving fluid with static atmospheric variables (T, p, \nm{\rho}) and speed \nm{v} decelerates until it has no speed through a process that is stationary, has no viscosity nor gravity (gravitation plus inertial) accelerations, is adiabatic, and presents fixed boundaries for the analyzed control volume \cite{Oates1989,Batchelor2000}. Such a process maintains the fluid total enthalpy, as well as its entropy, and hence complies with the Bernoulli equation \cite{Batchelor2000}:
\neweq{\dfrac{d}{dt}\lrp{\dfrac{\kappa}{\kappa - 1} \dfrac{p}{\rho} + \dfrac{1}{2}\, v^2} = 0}{eq:Sensor_NonInertial_Bernoulli}

where \nm{\kappa} is the air adiabatic index (appendix \ref{cha:PhysicalConstants}). In addition to the static pressure and temperature sensors discussed above, a Pitot tube is also equipped with a dynamic pressure sensor located at the dead end to measure the air flow total pressure \nm{p_t}. The air data system then estimates the aircraft airspeed based on the following expression, which results from applying (\ref{eq:Sensor_NonInertial_Bernoulli}) at the Pitot dead end as well as at the static ports:
\neweq{\vtas = \sqrt{\frac{2 \, \kappa}{\kappa - 1}\frac{p}{\rho}\lrsb{\lrp{\frac{p_t - p}{p} + 1}^{\textstyle \frac{\kappa - 1}{\kappa}}-1}}}{eq:Sensor_NonInertial_tas}

The errors induced when measuring the airspeed this way can also be modeled by a combination of bias offset and system noise [A\ref{as:SENSOR_TAS}], where \nm{\sigmaTAS} is the error standard deviation taken from table \ref{tab:Sensors_vtasb}, \nm{\BzeroTAS} has been added by the author for increased realism, and \nm{\NzeroTAS} and \nm{\NsTAS} are two uncorrelated standard normal random variables \nm{N\lrp{0, \, 1}}:
\neweq{e_{\sss TAS} \lrp{s \, \DeltatSENSED} = e_{\sss TAS} \lrp{s \, \Deltat} = \vtastilde\lrp{s \, \Deltat} - \vtas\lrp{s \, \Deltat} = \BzeroTAS \, \NzeroTAS + \sigmaTAS \, \NsTAS}{eq:Sensor_NonInertial_TAS}

The \texttt{Aeroprobe} air data system specifications \cite{Aeroprobe} list a maximum airspeed error of \nm{1 \ m/s}, which can be interpreted as \nm{3 \, \sigma}, and hence results in the \nm{\sigmaTAS} value shown below:
\begin{center}
\begin{tabular}{lrc|lrc}
	\hline
	\hypertt{TAS}-\hypertt{AOA}-\hypertt{AOS} Baseline & \multicolumn{1}{c}{Spec} & Unit & \multicolumn{1}{c}{Variable} & \multicolumn{1}{c}{Value} & Unit \\
	\hline
	Airspeed error (max)	         & \nm{1}			& \nm{m/s}	   & \nm{\sigmaTAS}	& \nm{3.33 \cdot 10^{-1}}	& \nm{m/s} \\
	                    	         & 					&				   & \nm{\BzeroTAS}	& \nm{3.33 \cdot 10^{-1}}	& \nm{m/s} \\
	Flow angle error (max)	         & \nm{\pm \, 1.0}	& \nm{^{\circ}}	   & \nm{\sigmaAOA}	& \nm{3.33 \cdot 10^{-1}}	& \nm{^{\circ}} \\
	                    	         & 					&				   & \nm{\BzeroAOA}	& \nm{3.33 \cdot 10^{-1}}	& \nm{^{\circ}} \\
	Flow angle error (max)	         & \nm{\pm \, 1.0}	& \nm{^{\circ}}   & \nm{\sigmaAOS}	& \nm{3.33 \cdot 10^{-1}}	& \nm{^{\circ}} \\
							         &					& 				   & \nm{\BzeroAOS}	& \nm{3.33 \cdot 10^{-1}}	& \nm{^{\circ}} \\
	\hline
\end{tabular}
\captionof{table}{Performance of ``baseline'' Pitot tube and air vanes} \label{tab:Sensors_vtasb}
\end{center}

The air data system is also capable of measuring the direction of the air stream with respect to the aircraft, represented by the angles of attack and sideslip (section \ref{subsec:RefSystems_W}). To do so, it can be equipped with two air vanes that align themselves with the unperturbed air stream or with a more complex multi hole Pitot probe. The latter is the case of the \texttt{Aeroprobe} air data system \cite{Aeroprobe} employed in the simulation, which measures both angles with a maximum error of \nm{\pm \, 1.0^{\circ}}. If interpreted as \nm{3 \, \sigma}, this results in standard deviations \nm{\sigmaAOA} and \nm{\sigmaAOS} of \nm{0.33^{\circ}}, where \hypertt{AOA} and \hypertt{AOS} stand for angles of attack and sideslip, respectively. Although not present in the documentation, the author has decided to also include bias offsets \nm{\BzeroAOA} and \nm{\BzeroAOS} to provide more realism to the sensors [A\ref{as:SENSOR_AOA_AOS}], specially in the case of air vanes. The final expressions for the angles of attack and sideslip measurement errors are the following:
\begin{eqnarray}
\nm{e_{\sss AOA} \lrp{s \, \DeltatSENSED} = e_{\sss AOA} \lrp{s \, \Deltat}} & = & \nm{\widetilde\alpha\lrp{s \, \Deltat} - \alpha\lrp{s \, \Deltat} = \BzeroAOA \, \NzeroAOA + \sigmaAOA \, \NsAOA}\label{eq:Sensor_NonInertial_AOA} \\
\nm{e_{\sss AOS} \lrp{s \, \DeltatSENSED} = e_{\sss AOS} \lrp{s \, \Deltat}} & = & \nm{\widetilde\beta \lrp{s \, \Deltat} - \beta \lrp{s \, \Deltat} = \BzeroAOS \, \NzeroAOS + \sigmaAOS \, \NsAOS}\label{eq:Sensor_NonInertial_AOS}
\end{eqnarray}

where \nm{\NzeroAOA}, \nm{\NsAOA}, \nm{\NzeroAOS}, and \nm{\NsAOS} are uncorrelated standard normal random variables \nm{N\lrp{0, \, 1}}.

Table \ref{tab:Sensors_NonInertial_air_data_sources} lists the error sources contained in the air data sensor model represented by (\ref{eq:Sensor_NonInertial_OSP}, \ref{eq:Sensor_NonInertial_OAT}, \ref{eq:Sensor_NonInertial_TAS}, \ref{eq:Sensor_NonInertial_AOA}, \ref{eq:Sensor_NonInertial_AOS}). The specific errors are contained in the first two columns, while their origin (according to the criterion established in the first paragraph of section \ref{subsec:Sensors_Inertial_ErrorSources}) is listed in the third column.
\begin{center}
\begin{tabular}{lcc}
	\hline
	\multicolumn{2}{c}{Error} & Main Source \\
	\hline
	Bias offset		& \nm{\BzeroOSP, \, \BzeroOAT, \, \BzeroTAS, \, \BzeroAOA, \, \BzeroAOS}	& run-to-run	\\
	System noise	& \nm{\sigmaOSP, \, \sigmaOAT, \, \sigmaTAS, \, \sigmaAOA, \, \sigmaAOS}	& in-run		\\
	\hline
\end{tabular}
\end{center}
\captionof{table}{Air data sensor error sources}\label{tab:Sensors_NonInertial_air_data_sources}

Note that all errors are modeled as stochastic variables or processes. The stochastic nature of the run-to-run error contributions to the models relies on five realizations of normal distributions for the bias offsets, while the in-run error contributions require five realizations for the system noises at every discrete sensor measurement.


\subsection{Global Navigation Satellite System Receiver}\label{subsec:Sensors_NonInertial_GNSS}

A \hypertt{GNSS} receiver enables the determination of the aircraft position and absolute velocity based on signals obtained from various constellations of satellites, such as \hypertt{GPS}, \hypertt{GLONASS}, \texttt{BeiDou}, and \texttt{Galileo}. The position is obtained by triangulation based on the accurate satellite position and time contained within each signal. Instead of derivating the position with respect to time, which introduces noise, \hypertt{GNSS} receivers obtain the vehicle ground velocity by measuring the Doppler shift between the constant satellite frequencies and those measured by the receiver.

It is important to note that because of the heavy processing required to fix a position based on the satellite signals, \hypertt{GNSS} receivers are not capable of working at the high frequencies characteristic of inertial and air data sensors, with \nm{1 \ Hz} employed in the simulation (\nm{\DeltatGNSS = 1 \ s}). The position error of a \hypertt{GNSS} receiver can be modeled as the sum of a zero mean white noise process plus slow varying ionospheric effects \cite{Kayton1997} [A\ref{as:SENSOR_GNSS}] modeled as the sum of the bias offset plus a random walk. This random walk is modeled with a frequency of \nm{1/60 \ Hz} (\nm{\DeltatION = 60 \ s}), and linearly interpolated in between. The ground velocity error is modeled exclusively with a white noise process [A\ref{as:SENSOR_GNSS}].
\begin{eqnarray}
\nm{\vec e_{\sss GNSS,POS} \lrp{g \, \DeltatGNSS}} & = & \nm{\TEgdttilde - \TEgdt = \sigmaGNSSPOS \, \NgGNSSPOS + \vec e_{\sss GNSS,ION} \lrp{g \, \DeltatGNSS}} \label{eq:Sensor_NonInertial_GNSS_pos_error} \\
\nm{\vec e_{\sss GNSS,VEL} \lrp{g \, \DeltatGNSS}} & = & \nm{\vNtilde - \vN = \sigmaGNSSVEL \, \NgGNSSVEL} \label{eq:Sensor_NonInertial_GNSS_vel_error} \\
\nm{\vec e_{\sss GNSS,ION} \lrp{g \, \DeltatGNSS}} & = & \nm{\vec e_{\sss GNSS,ION} \lrp{i \, \DeltatION}} \nonumber \\
 & & \nm{+ \, \dfrac{r}{f_{\sss ION}} \Big(\vec e_{\sss GNSS,ION} \big(\lrp{i+1} \DeltatION\big) - \vec e_{\sss GNSS,ION} \lrp{i \, \DeltatION}\Big)} \label{eq:Sensor_NonInertial_GNSS_ion_error} \\
\nm{g} & = & \nm{f_{\sss ION} \cdot i + r \ \ \ \ \ \ \ \ \ 0 \le r < f_{\sss ION}} \label{eq:Sensor_NonInertial_GNSS_ion_error2} \\
\nm{\vec e_{\sss GNSS,ION} \lrp{i \, \DeltatION}} & = & \nm{\BzeroGNSSION \, \NzeroGNSSION + \sigmaGNSSION \, \sum_{j=1}^i \NjGNSSION} \label{eq:Sensor_NonInertial_GNSS_ion_error3} \\
\nm{f_{\sss ION}} & = & \nm{\DeltatION / \DeltatGNSS = 60}\label{eq:Sensor_NonInertial_GNSS_ion_error4}
\end{eqnarray}

where \nm{\sigmaGNSSPOS}, \nm{\sigmaGNSSION}, \nm{\BzeroGNSSION}, and \nm{\sigmaGNSSVEL} are taken from the table below, and \nm{\NgGNSSPOS}, \nm{\NgGNSSVEL}, \nm{\NzeroGNSSION}, and \nm{\NjGNSSION} are uncorrelated normal vectors of size three each composed of three uncorrelated standard normal random variables \nm{N\lrp{0, \, 1}}. Also note that as both \emph{g} and \nm{f_{\sss ION}} are integers, the quotient remainder theorem guarantees that there exist unique integers \emph{i} and \emph{r} that comply with (\ref{eq:Sensor_NonInertial_GNSS_ion_error2}) \cite{Pinter1990}.
\begin{center}
\begin{tabular}{lrc|lrc}
	\hline
	\hypertt{GNSS} & \multicolumn{1}{c}{Spec} & Unit & \multicolumn{1}{c}{Variable} & \multicolumn{1}{c}{Value} & Unit \\
  	\hline
	Horizontal position accuracy (\hypertt{CEP} 50\%) & 2.50 & \nm{m}		& \nm{\sigmaGNSSPOSHOR} & \nm{2.12 \cdot 10^{0}}	& \nm{m} \\
	Vertical position accuracy (\hypertt{CEP} 50\%)	& N/A	& 				& \nm{\sigmaGNSSPOSVER} & \nm{4.25 \cdot 10^{0}}	& \nm{m} \\
	Ionospheric random walk \nm{1/60 \ Hz}			& N/A	& 				& \nm{\sigmaGNSSION}	& \nm{1.60 \cdot 10^{-1}}	& \nm{m} \\
	Ionospheric bias offset                         & N/A   &               & \nm{\BzeroGNSSION}    & \nm{8.00 \cdot 10^{0}}    & \nm{m} \\
	Velocity accuracy (50\%) 						& 0.05	& \nm{m/s}		& \nm{\sigmaGNSSVEL}	& \nm{7.41 \cdot 10^{-2}}	& \nm{m/s} \\
  	\hline
\end{tabular}
\captionof{table}{Performances of \texttt{GNSS} receiver} \label{tab:Sensors_gnss}
\end{center}

The horizontal position accuracy in the table \ref{tab:Sensors_gnss} corresponds to the U-blox \texttt{NEO-M8} receiver data sheet \cite{NEOM8}, where \hypertt{CEP} stands for \emph{circular error probability}. As \hypertt{CEP} is equivalent to 1.18 standard deviations \cite{NOVATELGPS}, it enables the obtainment of \nm{\sigmaGNSSPOSHOR}. The author has not found any reference for \hypertt{GNSS} vertical accuracy, but has determined through conversations with several knowledgeable individuals that it is at least 50\% higher than the horizontal one. A conservative value for \nm{\sigmaGNSSPOSVER} of twice that of \nm{\sigmaGNSSPOSHOR} has been selected. The ionospheric effects have also been obtained from these conversations. The velocity accuracy also originates at the U-blox \texttt{NEO-M8} receiver data sheet \cite{NEOM8}. Assuming that it corresponds to a per axis error of \nm{\pm 0.05 \ m/s} instead of \hypertt{CEP}, and knowing that the 50\% mark of a normal distribution lies at 0.67448 standard deviations, it is possible to obtain \nm{\sigma_{\sss GNSS,VEL}}.

Table \ref{tab:Sensors_NonInertial_GNSS_error_sources} lists the error sources contained in the \hypertt{GNSS} receiver model represented by (\ref{eq:Sensor_NonInertial_GNSS_pos_error}, \ref{eq:Sensor_NonInertial_GNSS_vel_error}). The first two columns list the different error sources, while the third one specifies their origin according to the criterion established in the first paragraph of section \ref{subsec:Sensors_Inertial_ErrorSources}.
\begin{center}
\begin{tabular}{lcc}
	\hline
	\multicolumn{2}{c}{Error} & Main Source \\
	\hline
	Bias offset			& \nm{\BzeroGNSSION}										& run-to-run	\\
	System noise		& \nm{\sigmaGNSSPOS, \, \sigmaGNSSVEL, \, \sigmaGNSSION}	& in-run		\\
	\hline
\end{tabular}
\end{center}
\captionof{table}{\texttt{GNSS} receiver error sources}\label{tab:Sensors_NonInertial_GNSS_error_sources}

Note that all errors are modeled as stochastic variables or processes. Three realizations of a normal distribution are required for the run-to-run error contributions, while the in-run error contributions require three realizations each for position and velocity at every discrete sensor measurement, plus an extra three when corresponding for the ionospheric error.


\section{Camera} \label{sec:Sensors_camera}

Image generation is a power and data intensive process that can not work at the high frequencies characteristic of inertial and air data sensors. A frequency of \nm{10 \ Hz} is employed in the simulation (\nm{\DeltatIMG = 0.1 \ s}), although there are cameras available capable of working significantly faster. The camera is considered rigidly attached to the aircraft structure [A\ref{as:CAMERA_strapdown}]. The simulation considers that the shutter speed is sufficiently high so that all images are equally sharp [A\ref{as:CAMERA_sharp}], and that the image generation process is instantaneous [A\ref{as:CAMERA_instantaneous}]. In addition, the camera \hypertt{ISO} setting remains constant during the flight [A\ref{as:CAMERA_iso}], and all generated images are noise free [A\ref{as:CAMERA_noise}].
\begin{center}
\begin{tabular}{lcrc}
	\hline
	Parameter & Symbol & Value & Unit \\
	\hline
	Focal length						& f            & 19.0			       & mm   \\
	Image width                         & \nm{\Sh}   & 768                   & px    \\
	Image height                        & \nm{\Sv}   & 1024                  & px    \\
	Pixel size                          & \nm{\sPX}    & \nm{17 \cdot 10^{-3}} & mm/px \\
	Principal point horizontal location & \nm{\cIMGi}  & 384.5                 & px    \\
	Principal point vertical location   & \nm{\cIMGii} & 511.5                 & px    \\
	Horizontal field of view            & \nm{\fovh}   & 37.923               & \nm{^{\circ}}   \\
	Vertical field of view              & \nm{\fovv}   & 49.226               & \nm{^{\circ}}   \\
	\hline
\end{tabular}
\captionof{table}{Camera parameters} \label{tab:Sensors_cam}
\end{center}

It is also assumed that the visible spectrum radiation reaching all patches of the Earth surface remains constant [A\ref{as:CAMERA_luminosity}], and in addition the terrain is considered Lambertian (section \ref{subsec:Vis_Camera}), so its appearance at any given time does not vary with the viewing direction [A\ref{as:CAMERA_lambertian}]. The combined use of [A\ref{as:CAMERA_iso}], [A\ref{as:CAMERA_luminosity}], and [A\ref{as:CAMERA_lambertian}] implies that a given terrain object is represented with the same luminosity in all images, even as its relative pose (position and attitude) with respect to the camera varies. 

Geometrically, the simulation adopts a perspective projection or pinhole camera model [A\ref{as:CAMERA_pinhole}], as described in section \ref{subsec:Vis_Camera}, which in addition is perfectly calibrated and hence shows no distortion [A\ref{as:CAMERA_calibration}]. Table \ref{tab:Sensors_cam} contains the parameters of the employed camera, all of them defined in section \ref{subsec:Vis_Camera}:

It is important to remark that as the camera calibration discussed in section \ref{sec:PreFlight_Camera_Calibration} is considered perfect [A\ref{as:CAMERA_calibration}], the camera perspective map \nm{\vec{\Pi}} provided by (\ref{eq:Vis_camera_pIMGi}, \ref{eq:Vis_camera_pIMGii}) that is employed by the \texttt{Earth Viewer} application to generate the images (section \ref{subsec:Sensors_camera_earth_viewer}) coincides with the estimated map \nm{\hat{\vec{\Pi}}} employed by visual navigation algorithms. As both maps coincide, the symbol \nm{\vec{\Pi}} is the only one employed in this document, whether it applies to the real camera (\nm{\vec{\Pi}}) or to the knowledge of the camera available to the navigation system (\nm{\hat{\vec{\Pi }}}). 


\subsection{Image Generation, Pinhole Camera Model, and Image Frame} \label{subsec:Vis_Camera} 

An eight-bit monochrome \emph{digital image} is a two dimensional brightness array \nm{\lrb{\vec I : \Omega \subset \mathbb{Z}^2 \rightarrow \mathbb{Z}_+}} that assigns a positive integer \nm{\vec I\lrp{\pIMG} \in \lrsb{0, \, 255} \subset \mathbb{Z}_+} to each  discrete location within its domain \nm{\pIMG = \lrsb{\pIMGi \ \ \pIMGii}^T \in \Omega \ | \ \Omega = \lrsb{1, \Sh} \times \lrsb{1, \Sv} \subset \mathbb{Z}^2}, where \nm{\Sh} and \nm{\Sv} are the image width and height measured in pixels. The image generation process within a \emph{digital camera} is composed by two maps that, when in view of an object, determine which pixels to draw based on the object geometry with respect to the camera (first map or camera model), and which brightness or intensity to assign to each pixel (second map or luminosity assumption) \cite{Soatto2001}.
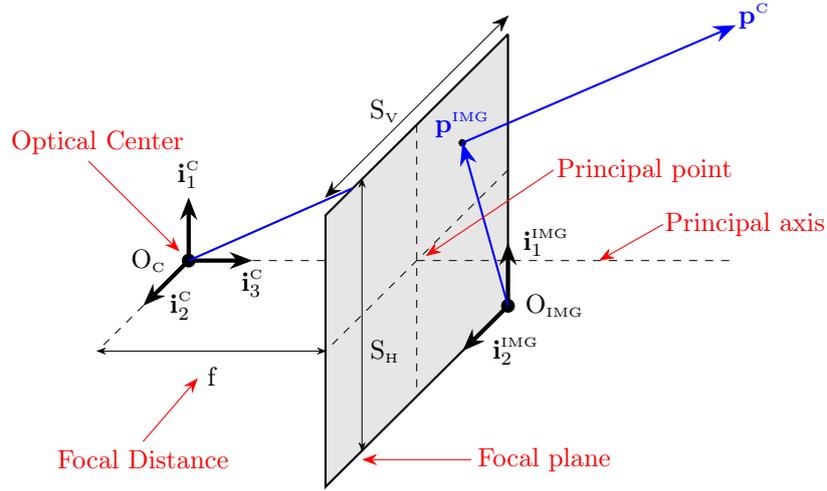
\begin{figure}[h]
\centering
\begin{tikzpicture}[auto,>=latex',scale=1.2]
	\coordinate (OC)    at (+0.0,+0.0);
	\coordinate (iCi)   at  ($(OC) + (+0.0,+0.7)$);
	\coordinate (iCii)  at  ($(OC) + (-0.5,-0.5)$);
	\coordinate (iCiii) at  ($(OC) + (+0.7,+0.0)$);
	\node [above of=iCi,   node distance=0.3cm] () {\nm{\iCi}};
	\node [right of=iCii,   node distance=0.5cm] () {\nm{\iCii}};
	\node [below of=iCiii, node distance=0.3cm] () {\nm{\iCiii}};
	\draw [ultra thick] [-{Stealth[length=3mm]}] (OC) -- (iCi) {};
	\draw [ultra thick] [-{Stealth[length=3mm]}] (OC) -- (iCii) {};
	\draw [ultra thick] [-{Stealth[length=3mm]}] (OC) -- (iCiii) {};		
	\filldraw [black] (OC) circle [radius=2pt] node [left=5pt] {\nm{\OC}};	
	
	\coordinate (PP)    at ($(OC) + (+2.5,+0.0)$);	
	\coordinate (ul)    at ($(PP) + (+1.0,+2.5)$);
	\coordinate (ur)    at ($(PP) + (-1.0,+0.5)$);
	\coordinate (ll)    at ($(PP) + (+1.0,-0.5)$);
	\coordinate (lr)    at ($(PP) + (-1.0,-2.5)$);
	\fill [color=gray!20] (ul) -- (ur) -- (lr) -- (ll) -- cycle;
		
	\coordinate (OIMG)  at ($(PP) + (+1.0,-0.5)$);	
	\coordinate (iIMGi)   at ($(OIMG) + (+0.0,+0.7)$);
	\coordinate (iIMGii)  at ($(OIMG) + (-0.5,-0.5)$);
	\node [right of=iIMGi,   node distance=0.5cm] () {\nm{\iIMGi}};
	\node [right of=iIMGii,   node distance=0.7cm] () {\nm{\iIMGii}};
	\draw [ultra thick] [-{Stealth[length=3mm]}] (OIMG) -- (iIMGi) {};
	\draw [ultra thick] [-{Stealth[length=3mm]}] (OIMG) -- (iIMGii) {};
	\filldraw [black] (OIMG) circle [radius=2pt] node [right=3pt] {\nm{\OIMG}};

	\draw [thick] [-] (ul) -- (ur) -- (lr) -- (ll) -- (ul);
	\draw [ultra thin, dashed] [-] (PP) -- ($(PP) + (+3.5,+0.0)$);
	\draw [ultra thin, dashed] [-] (OC) -- ($(OC) + (+1.5,+0.0)$);
	\draw [ultra thin, dashed] [-] (OC) -- ($(OC) + (-1.0,-1.0)$);
	\draw [ultra thin, dashed] [-] ($(PP) + (+1.0,+1.0)$) -- ($(PP) + (-1.0,-1.0)$);
	\draw [ultra thin, dashed] [-] ($(PP) + (+0.0,+1.5)$) -- ($(PP) + (-0.0,-1.5)$);
	\draw [ultra thin] [{Stealth[length=1mm]}-{Stealth[length=1mm]}] ($(OC) + (-1.0,-1.0)$) -- node[below=1mm] {f} ($(PP) + (-1.0,-1.0)$);
		
	\draw [ultra thin] [{Stealth[length=1mm]}-{Stealth[length=1mm]}] ($(lr) + (+0.4,+0.4)$) -- ($(ur) + (+0.4,+0.4)$);		
	\node at ($(PP) + (-0.35,-1.0)$) {\nm{\Sh}};		
	
	\draw [ultra thin] [{Stealth[length=2mm]}-{Stealth[length=2mm]}] ($(ul) + (+0.0,+0.2)$) -- ($(ur) + (+0.0,+0.2)$);		
	\node at ($(PP) + (-0.35,+1.65)$) {\nm{\Sv}};		
	
	\coordinate (PIMG)    at ($(OC) + (+3.0,+1.3)$);
	\filldraw [black] (PIMG) circle [radius=1pt] node [left=5pt] {};	
	\draw [thick, blue] [-{Stealth[length=3mm]}] (PIMG) -- ($(PIMG) + (+3.0,+1.3)$);
	\draw [thick, blue] [-] (OC) -- ($(OC) + (+1.8,+0.8)$);
	\draw [thick, blue] [-{Stealth[length=3mm]}] (OIMG) -- (PIMG);

	\node [blue] at ($(PIMG) + (+90:0.2cm)$) {\nm{\pIMG}};		
	\node [blue] at ($(PIMG) + (+3.2,+1.4)$) {\nm{\pC}};		
		
	\draw [red] [-{Stealth[length=2mm]}] ($(OC) + (+3.1,-2.2)$) -- ($(OC) + (+1.9,-2.2)$);		
	\node [red] at ($(OC) + (+3.9,-2.2)$) {Focal plane};		

	\draw [red] [-{Stealth[length=2mm]}] ($(PP) + (+1.5,+1.0)$) -- ($(PP) + (+0.1,+0.05)$);		
	\node [red] at ($(PP) + (+2.5,1.0)$) {Principal point};		
		
	\draw [red] [-{Stealth[length=2mm]}] ($(PP) + (+2.7,+0.4)$) -- ($(PP) + (+2.0,+0.0)$);		
	\node [red] at ($(PP) + (+3.6,0.4)$) {Principal axis};				
		
	\draw [red] [-{Stealth[length=2mm]}] ($(OC) + (-1.1,+1.1)$) -- ($(OC) + (-0.1,+0.1)$);		
	\node [red] at ($(OC) + (-1.0,1.3)$) {Optical Center};			
				
	\draw [red] [-{Stealth[length=2mm]}] ($(OC) + (-0.5,-2.0)$) -- ($(OC) + (+0.1,-1.3)$);		
	\node [red] at ($(OC) + (-0.5,-2.2)$) {Focal Distance};		
	
\end{tikzpicture}
\caption{Frontal pinhole camera model}
\label{fig:Vis_camera_pinhole}
\end{figure}

From a light intensity point of view, it is customary to consider that all surfaces viewed by the camera are Lambertian and hence maintain their appearance independent of the viewing direction, this is, they diffuse light uniformly in all directions\footnote{This is a good approximation for matte surfaces, and not so good for shiny ones.}. This being the case, the intensity measured at a pixel is identical to the amount of energy radiated at the corresponding point in space, independent of the vantage point. In these conditions the image generation process is reduced to tracing rays from space points to image points, a purely geometric problem \cite{Soatto2001}.

A pinhole camera model is a geometric model of the projection of three dimensional objects into the image that relies on a \emph{perspective projection}, which is that in which collinearity  is maintained (world straight lines in \nm{\mathbb{R}^3} are imaged as straight lines in \nm{\mathbb{R}^2}), but does not respect parallelism or angles between lines and/or planes, nor distances between points \cite{Hartley2003}. As depicted in figure \ref{fig:Vis_camera_pinhole}, the projection \nm{\lrb{\vec{\Pi} : \mathbb{R}^3 \rightarrow \mathbb{R}^2 \ | \ \pC \rightarrow \pIMG = \vec{\Pi}\lrp{\pC}}} considers that all light rays are straight and travel through a point named the camera \emph{optical center} (\nm{\OC}). The image \nm{\pIMG} of a point \nm{\pC} is given by the intersection of the light ray that connects them with a \emph{focal plane} that lies at a \emph{focal distance} \emph{f} from the optical center. The \emph{principal point} is the closest point within the focal plane to the optical center, while the camera \emph{principal axis} joins the optical center and the principal point.

In addition to the focal length, a pinhole camera is defined by the number of pixels in the sensor or image domain (\nm{\Sh} horizontally and \nm{\Sv} vertically), the size of each pixel\footnote{Pixels are considered square in this document.} \nm{\sPX}, and the location of the principal point with respect to the upper left corner of the image\footnote{In a perfectly aligned pinhole camera, \nm{\cIMGi = 0.5 \, \Sh - 0.5}, and \nm{\cIMGii = 0.5 \, \Sv - 0.5}.}, \nm{\cIMG = \lrsb{\cIMGi \ \ \cIMGii}^T}. The resulting \emph{field of view} is the angle subtended by the spatial extent of the sensor as seen from the optical center, and it is different if viewed horizontally or vertically: 
\begin{eqnarray}
\nm{\tan\dfrac{\fovh}{2}} & = & \nm{\dfrac{\Sh}{2 \, f} \cdot \sPX} \label{eq:Vis_camera_fovh} \\
\nm{\tan\dfrac{\fovv}{2}} & = & \nm{\dfrac{\Sv}{2 \, f} \cdot \sPX} \label{eq:Vis_camera_fovv}
\end{eqnarray}

The \emph{image frame} is a two-dimensional Cartesian reference frame \nm{\FIMG = \{\OIMG ,\, \iIMGi, \, \iIMGii\}} whose axes are parallel to those of the \nm{\FC} camera frame defined in section \ref{subsec:RefSystems_C} (\nm{\iIMGi \parallel \iCi, \, \iIMGii \parallel \iCii}), and whose origin \nm{\OIMG} is located on the focal plane displaced a distance \nm{\cIMG} from the principal point so the \nm{\FIMG} coordinates \nm{\pIMGi} and \nm{\pIMGii} of any point in the image domain \nm{\Omega} are always positive. The perspective projection map \nm{\pIMG = \vec{\Pi}\lrp{\pC}} that converts points viewed in \nm{\FC} into \nm{\FIMG} is hence the following:
\begin{eqnarray}
\nm{\pIMGi}  & = & \nm{\dfrac{f}{\sPX} \ \dfrac{\pCi}{\pCiii} + \cIMGi} \label{eq:Vis_camera_pIMGi} \\
\nm{\pIMGii} & = & \nm{\dfrac{f}{\sPX} \ \dfrac{\pCii}{\pCiii} + \cIMGii} \label{eq:Vis_camera_pIMGii} 
\end{eqnarray}

The \nm{\pIMG} and \nm{\pC} image and camera coordinates can be expressed both in metric units or in pixels, with their ratio being \nm{\sPX}, although it is customary to employ pixels for the former and metric units for the latter. Note that the three camera coordinates \nm{\pC} can only be obtained up to scale from the two image ones \nm{\pIMG}, resulting in the inverse projection \nm{\pC = \vec{\Pi}^{-1}\lrp{\pIMG, \, \pCiii}}:
\begin{eqnarray}
\nm{\pCi}  & = & \nm{\pCiii \, \dfrac{\sPX}{f} \, \lrp{\pIMGi - \cIMGi}} \label{eq:Vis_camera_pCi} \\
\nm{\pCii} & = & \nm{\pCiii \, \dfrac{\sPX}{f} \, \lrp{\pIMGii - \cIMGii}} \label{eq:Vis_camera_pCii} 
\end{eqnarray}

Because of this, it is common to work with the homogeneous camera coordinates \nm{\pCbar} or the unitary ones \nm{\pCunit}, which are equivalent to working with a unitary focal length, or to projecting on a unitary radius sphere instead of a focal plane, respectively. The homogeneous and unitary camera coordinates contain the same information as the image ones, and it is necessary to multiply them by the point depth \nm{\pCiii} or its distance \nm{\| \pC \|} to obtain the real camera coordinates \nm{\pC}.
\begin{eqnarray}
\nm{\pCbar} & = & \nm{\dfrac{\pC}{\pCiii} = \lrsb{\dfrac{\pCi}{\pCiii} \ \ \dfrac{\pCii}{\pCiii} \ \ 1}^T} \label{eq:Vis_camera_pCbar} \\
\nm{\pCunit} & = & \nm{\dfrac{\pC}{\| \pC \|}} \label{eq:Vis_camera_pCunit} 
\end{eqnarray}


\subsection{Mounting of Camera}\label{subsec:Sensors_Camera_Mounting}

The digital camera can be located anywhere on the aircraft structure as long as its view of the terrain is unobstructed by other platform elements. It is desirable that the lever arm or distance between the camera optical center \nm{\OC} and the aircraft center of mass \nm{\OB} is as small as possible to reduce the negative effects of any camera alignment error. With respect to its orientation, the camera should be facing down to show a balanced view of the ground during level flight, but minor deviations are not problematic. Figure \ref{fig:Sensor_symmetry_plane} shows a schematic view of the camera frame \nm{\FC} with relation to other aircraft related frames introduced in appendix \ref{cha:RefSystems}.

The simulation considers that the location of the camera is deterministic but its orientation stochastic, like that of the \hypertt{IMU} described in section \ref{subsec:Sensors_Inertial_Mounting}. The origin of the camera frame \nm{\OC} is located \nm{500 \ mm} forward and \nm{120 \ mm} below the aircraft reference point \nm{\OR}. Considering (\ref{eq:AircraftModel_Trbb_full}), (\ref{eq:AircraftModel_Trbb_empty}), and (\ref{eq:AircraftModel_Trbb}), this results in:
\begin{eqnarray}
\nm{\TBCBfull}  & = & \nm{\begin{bmatrix} 0.293 & 0 & 0.125 \end{bmatrix}^T \ m} \label{eq:Sensors_Camera_Mounting_Tbcb_full} \\
\nm{\TBCBempty} & = & \nm{\begin{bmatrix} 0.281 & 0 & 0.126 \end{bmatrix}^T \ m} \label{eq:Sensors_Camera_Mounting_Tbcb_empty} \\
\nm{\TBCB}      & = & \nm{\vec f_C\lrp{m} = \TBCBfull + \dfrac{m_{full} - m}{m_{full} - m_{empty}} \; \lrp{\TBCBempty - \TBCBfull}} \label{eq:Sensors_Camera_Mounting_Tbcb}
\end{eqnarray}

The rotation between \nm{\FB} and \nm{\FC} is represented by means of the camera Euler angles \nm{\phiBC = \lrsb{\psiC \ \ \thetaC \ \ \xiC}^T} defined in section \ref{subsec:RefSystems_C}. The author has arbitrarily established the stochastic model provided by (\ref{eq:Sensors_Camera_Mounting_eulerBC}), in which each specific Euler angle is obtained as the product of the standard deviations (\nm{\sigmapsiC}, \nm{\sigmathetaC}, \nm{\sigmaxiC}) contained in table \ref{tab:Sensor_Camera_mounting} by a single realization of a standard normal random variable \nm{N\lrp{0, \, 1}} (\nm{\NpsiC}, \nm{\NthetaC}, and \nm{\NxiC}):
\neweq{\phiBC = \lrsb{90^{\circ} + \sigmapsiC \, \NpsiC \ \ \ \ \sigmathetaC \, \NthetaC \ \ \ \ \sigmaxiC \NxiC}^T}{eq:Sensors_Camera_Mounting_eulerBC}

The problem however is not the real translation and rotation between \nm{\FB} and \nm{\FC} given by the previous equations, but the accuracy with which they are known to the visual navigation system. The determination of the camera position \nm{\TBCBest} and rotation \nm{\phiBCest = \lrsb{\psiCest \ \ \thetaCest \ \ \xiCest}^T} is discussed in section \ref{sec:PreFlight_camera_frame}. As in previous cases, stochastic models are considered for both the position \nm{\TBCBest} and attitude \nm{\phiBCest}, changing their values from one simulation run to another:
\begin{eqnarray}
\nm{\TBCBest} & = & \nm{\TBCB + \lrsb{\sigmaTBCBest \, \NTBCBiest \ \ \ \ \sigmaTBCBest \, \NTBCBiiest \ \ \ \ \sigmaTBCBest \NTBCBiiiest}^T} \label{eq:Sensors_Camera_Mounting_TBCBest} \\
\nm{\phiBCest} & = & \nm{\phiBC + \lrsb{\sigmaphiBCest \, \NpsiCest \ \ \ \ \sigmaphiBCest \, \NthetaCest \ \ \ \ \sigmaphiBCest \NxiCest}^T} \label{eq:Sensors_Camera_Mounting_eulerBCest}
\end{eqnarray}

where the standard deviations \nm{\sigmaTBCBest} and \nm{ \sigmaphiBCest} are shown in table \ref{tab:Sensor_Camera_mounting}, and \nm{\NpsiCest}, \nm{\NthetaCest}, \nm{\NxiCest}, \nm{\NTBCBiest}, \nm{\NTBCBiiest}, \nm{\NTBCBiiiest} are six realizations of a standard normal random variable \nm{N\lrp{0, \, 1}}.
\begin{center} 
\begin{tabular}{lccc}
	\hline
	Concept & Variable & Value & Unit \\
	\hline
	True yaw error 	   	             & \nm{\sigmapsiC}     & \nm{0.1}  & \nm{^{\circ}}  \\
	True pitch error                 & \nm{\sigmathetaC}   & \nm{0.1}  & \nm{^{\circ}}  \\
	True bank error                  & \nm{\sigmaxiC}      & \nm{0.1}  & \nm{^{\circ}}  \\
	Camera position estimation error & \nm{\sigmaTBCBest}  & \nm{0.002} & \nm{m} \\
	\rule[-7pt]{0pt}{12pt}Camera angular estimation error  & \nm{\sigmaphiBCest} & \nm{0.01} & \nm{^{\circ}}  \\
	\hline
\end{tabular}
\end{center}
\captionof{table}{Camera mounting accuracy values}\label{tab:Sensor_Camera_mounting}

The translation \nm{\TBC} between the origins of the \nm{\FB} and \nm{\FC} frames can be considered quasi stationary as it slowly varies based on the aircraft mass (\ref{eq:Sensors_Camera_Mounting_Tbcb}), and the relative position of their axes \nm{\phiBC} remains constant because the camera is rigidly attached to the aircraft structure (\ref{eq:Sensors_Camera_Mounting_eulerBC}):
\neweq{\dot{\vec T}_{\sss BC} = \dot{\vec q}_{\sss BC} = \dot{\vec \zeta}_{\sss BC} = \vec 0 \ \longrightarrow \ \vBC = \vec a_{\sss BC} = \wBC = \vec \alpha_{\sss BC} = \vec 0}{eq:Sensor_Camera_BC_diff}


\subsection{Earth Viewer}\label{subsec:Sensors_camera_earth_viewer}

The camera model differs from all other sensor models in that it does not return a sensed variable \nm{\xvectilde} consisting of its real value \nm{\xvec} plus a sensor error \nm{\vec E}, but instead generates a digital image simulating what a real camera would record based on the aircraft position and attitude as given by the actual or real trajectory \nm{\xvec = \xTRUTH} described in section \ref{sec:EquationsMotion_AT}. The scene viewed by the camera at a time \nm{t_i = i \cdot \DeltatIMG} depends on the \nm{\zetaEC} pose of \nm{\FC} with respect to the \hypertt{ECEF} frame \nm{\FE}:
\neweq{\vec I\lrp{t_i} = \vec I\lrp{i \, \DeltatIMG} = \vec I\lrp{\zetaEC} = \vec I\lrp{\zetaEN \otimes \zetaNB \otimes \zetaBC} = \vec I\lrp{\qEC + \dfrac{\epsilon}{2} \, \TECE \otimes \qEC}} {eq:Sensors_Camera_earth_I}

\begin{figure}[h]
\centering
\includegraphics[width=6.5cm]{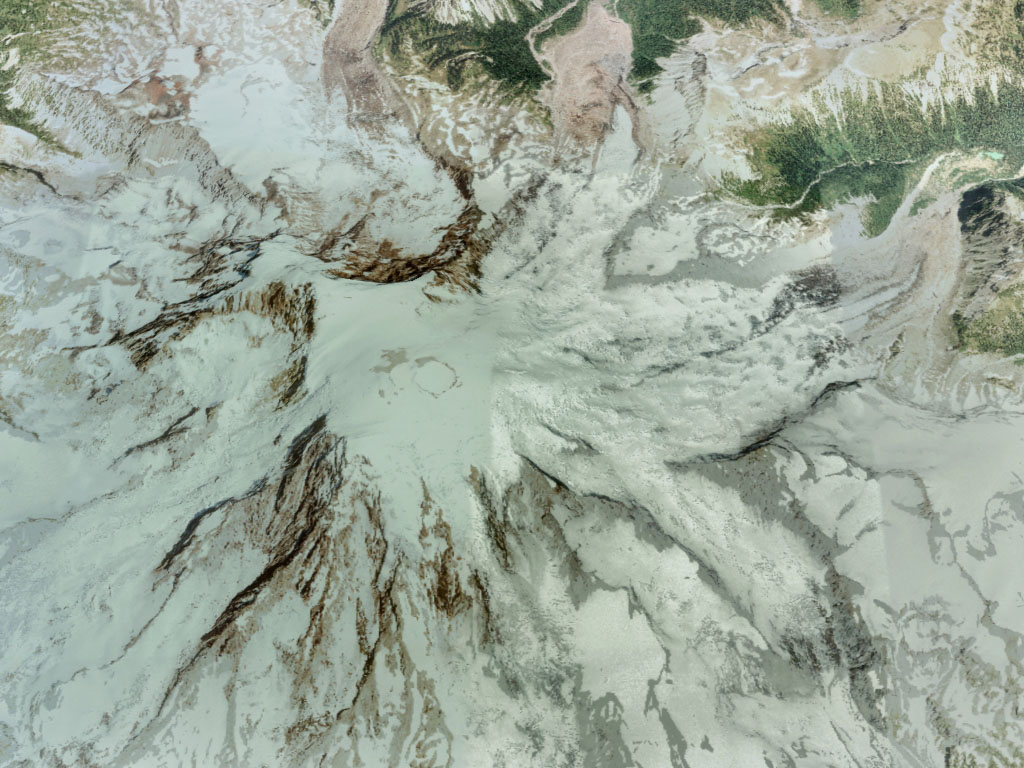}
\hskip 10pt
\includegraphics[width=6.5cm]{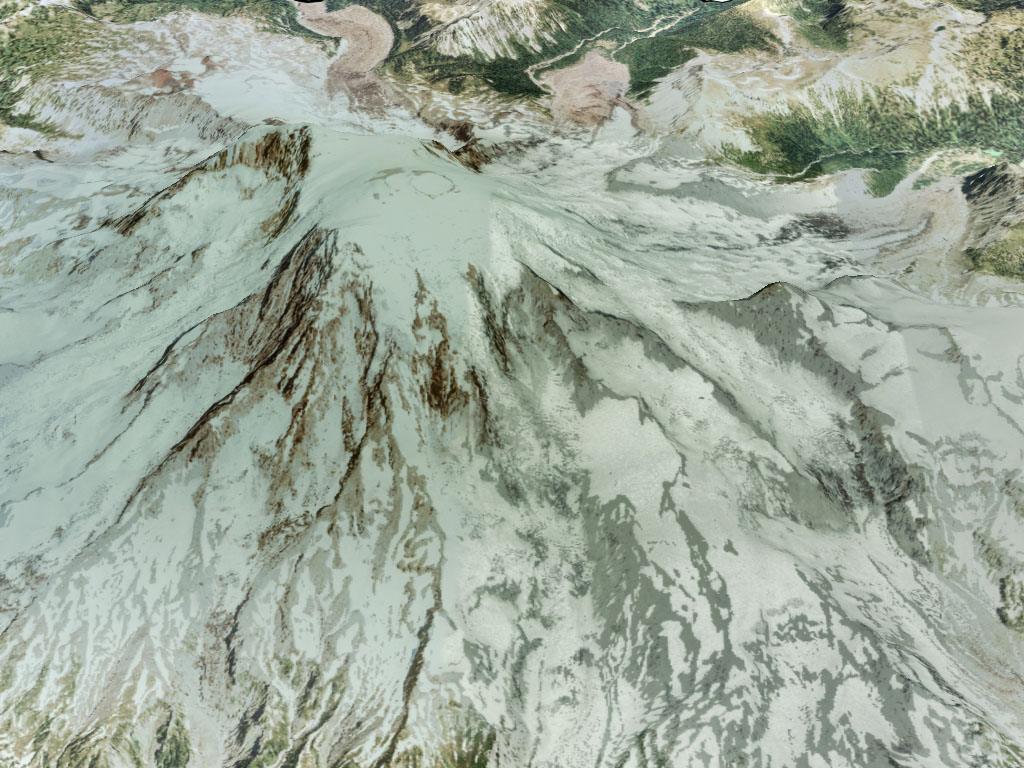}
\caption{Example of \texttt{Earth Viewer} images}
\label{fig:EarthViewer_photo}
\end{figure}

When provided with the camera pose \nm{\zetaEC} at equally time spaced intervals, the simulation is capable of generating images that resemble the view of the Earth surface that the camera would record if located at that particular pose. To do so, it relies on three software libraries:
\begin{itemize}

\item \texttt{OpenSceneGraph} \cite{OpenSceneGraph} is an open source high performance 3D graphics toolkit written in \texttt{C++} and \texttt{OpenGL}, used by application developers in fields such as visual simulation, games, virtual reality, scientific visualization and modeling. The library enables the representation of objects in a scene by means of a graph data structure, which allows grouping objects that share some properties to automatically manage rendering properties such as the level of detail necessary to faithfully draw the scene, but without considering the unnecessary detail that slows down the graphics hardware drawing the scene.

\item \texttt{osgEarth} \cite{osgEarth} is a dynamic and scalable 3D Earth surface rendering toolkit that relies on \texttt{OpenSceneGraph}, and is based on publicly available \emph{orthoimages} of the area flown by the aircraft. Orthoimages consist of aerial or satellite imagery geometrically corrected such that the scale is uniform; they can be used to measure true distances as they are accurate representations of the Earth surface, having been adjusted for topographic relief, lens distortion, and camera tilt. When coupled with a \emph{terrain elevation model}, \texttt{osgEarth} is capable of generating realistic images based on the camera position as well as its yaw and pitch, but does not accept the camera roll (in other words, the \texttt{osgEarth} images are always aligned with the horizon).

\item \texttt{Earth Viewer} is a modification to \texttt{osgEarth} implemented by the author so it is also capable of accepting the bank angle of the camera with respect to the \hypertt{NED} axes. \texttt{Earth Viewer} is capable of generating realistic Earth images as long as the camera height over the terrain is significantly higher than the vertical relief present in the image. As an example, figure \ref{fig:EarthViewer_photo} shows two different views of a volcano in which the dome of the mountain, having very steep slopes, is properly rendered.
\end{itemize}


\section{Implementation, Validation, and Execution} \label{sec:Sensors_implementation}

The sensor models have been implemented in \texttt{C++} with the objective of maintaining the stochastic nature of the different random processes involved while at the same time ensuring that the time variation of the errors generated in the different sensors can be repeated if so desired. The implementation hence follows these steps:
\begin{enumerate}

\item Initialize a discrete uniform distribution with any seed (any value is valid, so 1 is employed by the author), which produces random integers where each possible value has an equal likelihood of being produced. Call this distribution a number of times equal or higher than twice the maximum number of flights employed to evaluate the navigation solutions, divide them into two groups of the same size, and store the results for later use. These values, called \nm{\seedA} and \nm{\seedR}, are respectively the \emph{aircraft seeds} and \emph{flight seeds}, where the aircraft \emph{i} identifies the fixed error realizations, and the flight \emph{j} represents the run-to-run and in-run error realizations. The stored aircraft and flight seeds become the initialization seeds for each of the flight simulation executions, so this step does not need to be repeated.

\item Every Monte Carlo simulation run is initialized with given aircraft and flight seeds (\nm{\seedA} and \nm{\seedR}). As these seeds are the only inputs required for all the stochastic processes within the sensors, the results of a given run can always be repeated by employing the same seeds. \nm{\seedA} and \nm{\seedR} are employed to initialize two different discrete uniform distributions. One is executed five times to provide the \emph{fixed sensor seeds} (\nm{\seedAACC}, \nm{\seedAGYR}, \nm{\seedAMAG}, \nm{\seedAPLAT}, and \nm{\seedACAM}), while the other is realized nine times to obtain the \emph{run sensor seeds} (\nm{\seedRACC}, \nm{\seedRGYR}, \nm{\seedRMAG}, \nm{\seedROSP}, \nm{\seedROAT}, \nm{\seedRTAS}, \nm{\seedRAOA}, \nm{\seedRAOS}, and \nm{\seedRGNSS}). These seeds hence become the initialization seeds for each of the different sensors. Note that in addition to the sensor seeds, the second distribution is also employed to obtain additional seeds for the Earth model, the fine alignment process, and the guidance parameters, as described in sections \ref{sec:EarthModel_implementation}, \ref{sec:PreFlight_implementation}, and \ref{sec:gc_implementation}.
\begin{center}
\begin{tabular}{lccc}
	\hline
	Type & Group & Error Sources & Seeds \\
	\hline
	Aircraft	& \nm{\seedA}	& fixed 				& \nm{\seedAACC, \, \seedAGYR, \, \seedAMAG, \, \seedAPLAT, \, \seedACAM} \\
	Flight		& \nm{\seedR}	& run-to-run \& in-run	& \nm{\seedRACC, \, \seedRGYR, \, \seedRMAG, \, \seedROSP, \, \seedROAT} \\
	\rule[-7pt]{0pt}{12pt}     & 					& & \nm{\seedRTAS, \, \seedRAOA, \, \seedRAOS, \, \seedRGNSS}  \\
	\hline
\end{tabular}
\end{center}
\captionof{table}{Sensor seeds} \label{tab:Sensors_seeds}

\item Each sensor relies on either one or two standard normal distributions \nm{N\lrp{0, \, 1}}, depending on whether the sensor error model is based exclusively on run-to-run and in-run error contributions (flight seed) or it also contains fixed error sources (aircraft seed). The normal distributions of every sensor are initialized with the corresponding seeds (\nm{\upsilon_{i,\sss A}, \, \upsilon_{j,\sss F}}) for that sensor.

\item Upon initialization, the aircraft normal distributions of every sensor are employed to generate all the values corresponding to scale factors, cross couplings, hard iron magnetism, and mounting errors. The flight normal distributions in turn are employed to generate the required bias offsets.

\item Every time a sensor is called to provide a measurement, its already initialized and used flight normal distribution is called to generate the corresponding random walk increments and white noises.
\end{enumerate}
\begin{figure}[h]
\centering
\begin{tikzpicture}[auto, node distance=2cm,>=latex']
	\node [coordinate](fIBinput) {};
	\node [coordinate, below of=fIBinput, node distance=1.0cm] (wBIBinput){};
	\node [coordinate, below of=wBIBinput, node distance=1.0cm] (BBinput){};	
	\node [coordinate, below of=wBIBinput, node distance=0.7cm] (Zinput){};
	\node [coordinate, below of=Zinput, node distance=0.4cm] (Yinput){};
	\node [coordinate, below of=BBinput, node distance=0.85cm] (xEGDTinput){};		
	\node [coordinate, below of=xEGDTinput, node distance=0.35cm] (Ainput){};		
	\node [coordinate, below of=xEGDTinput, node distance=0.7cm] (vNinput){};	
	\node [coordinate, below of=BBinput, node distance=2.6cm] (Cinput){};	
	\node [coordinate, above of=Cinput, node distance=0.35cm] (xEGDTinputbis){};	
	\node [coordinate, below of=Cinput, node distance=0.35cm] (qNBinput){};	
	
	\node [block, right of=fIBinput, minimum width=2.0cm, node distance=3.0cm, align=center, minimum height=0.8cm] (ACC) {\hypertt{ACC}};
	\node [block, right of=wBIBinput, minimum width=2.0cm, node distance=3.0cm, align=center, minimum height=0.8cm] (GYR) {\hypertt{GYR}};
	\node [block, right of=BBinput, minimum width=2.0cm, node distance=3.0cm, align=center, minimum height=0.8cm] (MAG) {\hypertt{MAG}};	
	\node [block, right of=Ainput, minimum width=2.0cm, node distance=3.0cm, align=center, minimum height=1.2cm] (GNSS) {\hypertt{GNSS}};			
	\node [block, right of=Cinput, minimum width=2.0cm, node distance=3.0cm, align=center, minimum height=1.2cm] (CAM) {\hypertt{CAM}};
		
	\node [coordinate, right of=ACC, node distance=3.0cm] (fIBoutput) {};	
	\node [coordinate, right of=GYR, node distance=3.0cm] (wBBIoutput){};
	\node [coordinate, right of=MAG, node distance=3.0cm] (BBoutput){};
	\node [coordinate, right of=GNSS, node distance=3.0cm] (Aoutput){};
	\node [coordinate, above of=Aoutput, node distance=0.35cm] (xEGDToutput){};	
	\node [coordinate, below of=Aoutput, node distance=0.35cm] (vNoutput){};	
	\node [coordinate, right of=CAM, node distance=3.0cm] (Ioutput){};

	\draw [->] (fIBinput) -- node[pos=0.2] {\nm{\fIBB}} (ACC.west);
	\draw [->] (wBIBinput) -- node[pos=0.2] {\nm{\wIBB}} (GYR.west);
	\draw [->] (BBinput) -- node[pos=0.2] {\nm{\BBREAL}} (MAG.west);	
	\draw [->] (xEGDTinput) -- node[pos=0.35] {\nm{\TEgdt}} ($(GNSS.west)+(0cm,0.35cm)$);
	\draw [->] (vNinput) -- node[pos=0.2] {\nm{\vN}} ($(GNSS.west)-(0cm,0.35cm)$);
	\draw [->] (xEGDTinputbis) -- node[pos=0.35] {\nm{\TEgdt}} ($(CAM.west)+(0cm,0.35cm)$);
	\draw [->] (qNBinput) -- node[pos=0.2] {\nm{\qNB}} ($(CAM.west)-(0cm,0.35cm)$);
		
	\draw [->] (ACC.east) -- node[pos=0.8] {\nm{\fIBBtilde}} (fIBoutput);
	\draw [->] (GYR.east) -- node[pos=0.8] {\nm{\wIBBtilde}} (wBBIoutput);
	\draw [->] (MAG.east) -- node[pos=0.8] {\nm{\BBtilde}} (BBoutput);	
	\draw [->] ($(GNSS.east)+(0cm,0.35cm)$) -- node[pos=0.65] {\nm{\widetilde{\vec T}^{\sss E,GDT}}} (xEGDToutput);	
	\draw [->] ($(GNSS.east)-(0cm,0.35cm)$) -- node[pos=0.8] {\nm{\vNtilde}} (vNoutput);	
	\draw [->] (CAM.east) -- node[pos=0.8] {\nm{\vec I}} (Ioutput);
		
	\node [coordinate, right of=Yinput,    node distance=7.0cm] (vtasinput){};	
	\node [coordinate, above of=vtasinput, node distance=1.6cm] (pinput){};
	\node [coordinate, above of=vtasinput, node distance=0.8cm] (Tinput){};
	\node [coordinate, below of=vtasinput, node distance=0.8cm] (alphainput){};		
	\node [coordinate, below of=vtasinput, node distance=1.6cm] (betainput){};		
		
	\node [block, right of=vtasinput, minimum width=2.0cm, node distance=3.0cm, align=center, minimum height=4.7cm] (ADS) {\hypertt{ADS}};
		
	\node [coordinate, right of=ADS, node distance=3.0cm] (vtasoutput){};		
	\node [coordinate, above of=vtasoutput, node distance=1.6cm] (poutput){};
	\node [coordinate, above of=vtasoutput, node distance=0.8cm] (Toutput){};
	\node [coordinate, below of=vtasoutput, node distance=0.8cm] (alphaoutput){};		
	\node [coordinate, below of=vtasoutput, node distance=1.6cm] (betaoutput){};			
	
	\draw [->] (pinput)     -- node[pos=0.2] {\nm{p}}      ($(ADS.west)+(0cm,1.6cm)$);
	\draw [->] (Tinput)     -- node[pos=0.2] {\nm{T}}      ($(ADS.west)+(0cm,0.8cm)$);
	\draw [->] (vtasinput)  -- node[pos=0.2] {\nm{\vtas}}  (ADS.west);
	\draw [->] (alphainput) -- node[pos=0.2] {\nm{\alpha}} ($(ADS.west)-(0cm,0.8cm)$);
	\draw [->] (betainput)  -- node[pos=0.2] {\nm{\beta}}  ($(ADS.west)-(0cm,1.6cm)$);
		
	\draw [->] ($(ADS.east)+(0cm,1.6cm)$) -- node[pos=0.8] {\nm{\widetilde{p}}}            (poutput);	
	\draw [->] ($(ADS.east)+(0cm,0.8cm)$) -- node[pos=0.8] {\nm{\widetilde{T}}}            (Toutput);	
	\draw [->] (ADS.east)                 -- node[pos=0.8] {\nm{\vtastilde}} 				(vtasoutput);	
	\draw [->] ($(ADS.east)-(0cm,0.8cm)$) -- node[pos=0.8] {\nm{\widetilde{\alpha}}}       (alphaoutput);	
	\draw [->] ($(ADS.east)-(0cm,1.6cm)$) -- node[pos=0.8] {\nm{\widetilde{\beta}}}        (betaoutput);	
\end{tikzpicture}
\caption{Sensors flow diagram} \label{fig:Sensors_flow_diagram}
\end{figure}
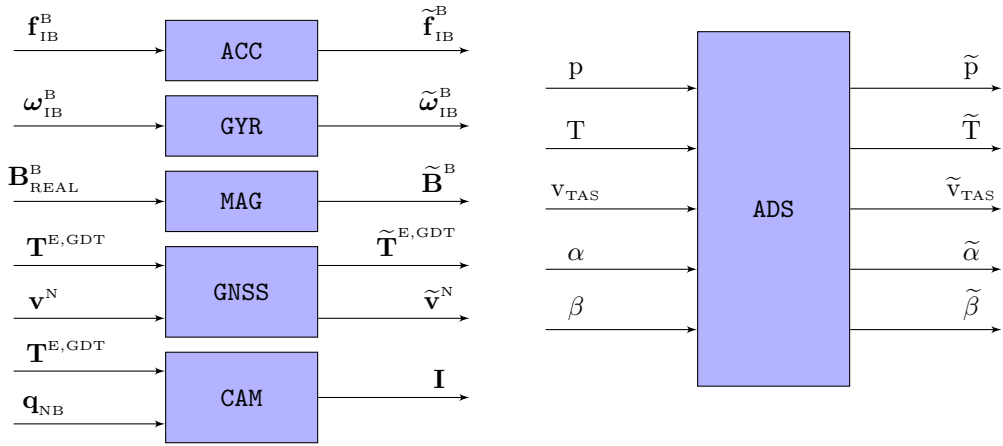

The functionality of the different sensors is graphically shown in figure \ref{fig:Sensors_flow_diagram}. Each sensor measures a different aspect of the true trajectory and returns its estimated value that includes the sensor error. These estimated values make up the sensed trajectory (section \ref{sec:Sensors_ST}), which is the input to the navigation system described in chapter \ref{cha:nav}.

The difficulty of this chapter resides in the development of valid stochastic models for each sensor that are able to capture the different sources of error, and the obtainment of the parameters on which they rely from the sensor data specifications provided by the manufacturers. The validation of the implemented functions is not complex as they all rely on standard normal random variables and random walks, which are validated by comparing their means and variances with their theoretical values, as performed in the different plots of section \ref{subsec:Sensors_Inertial_ErrorModelSingleAxis}.

 \cleardoublepage
\chapter{Pre-Flight Procedures}\label{cha:PreFlight}

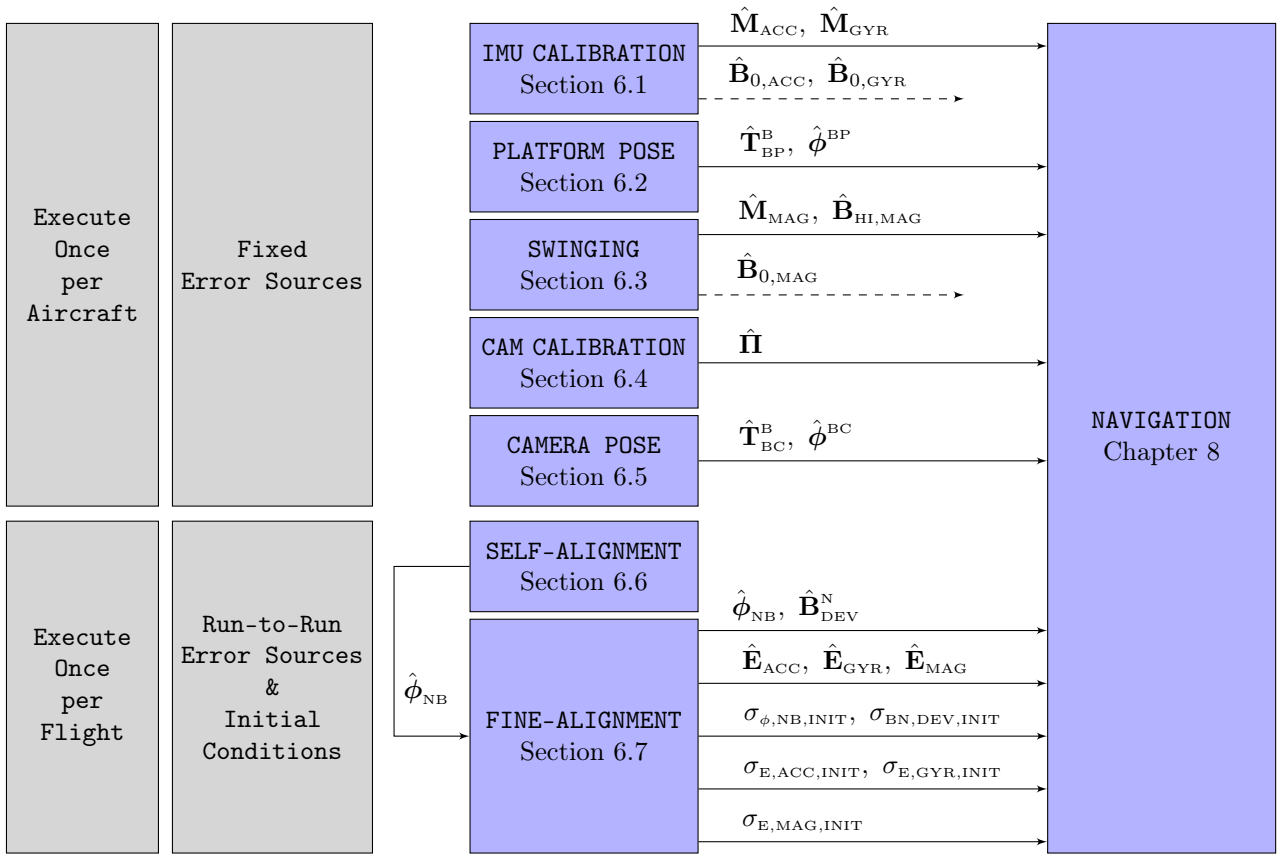
\begin{figure}[h]
\centering
\begin{tikzpicture}[auto, node distance=2cm,>=latex']
	\node [coordinate](origin) {};
	
	\node [blockgrey2, above of=origin, minimum width=2.0cm, node distance=3.3cm, align=center, minimum height=6.4cm] (ONCEPERACFT)   {\texttt{Execute} \\ \texttt{Once} \\ \texttt{per} \\ \texttt{Aircraft}};
	\node [blockgrey2, below of=origin, minimum width=2.0cm, node distance=2.3cm, align=center, minimum height=4.4cm] (ONCEPERFLIGHT) {\texttt{Execute} \\ \texttt{Once} \\ \texttt{per} \\ \texttt{Flight}};
	
	\node [blockgrey2, right of=ONCEPERACFT,   minimum width=2.0cm, node distance=2.5cm, align=center, minimum height=6.4cm] (FIXED) {\texttt{Fixed} \\ \texttt{Error Sources}};
	\node [blockgrey2, right of=ONCEPERFLIGHT, minimum width=2.0cm, node distance=2.5cm, align=center, minimum height=4.4cm] (R2R)   {\texttt{Run-to-Run} \\ \texttt{Error Sources} \\ \texttt{\&} \\ \texttt{Initial} \\ \texttt{Conditions}};
		
	\node [coordinate, right of=FIXED,  node distance=4.1cm] (origin1){};
	\node [coordinate, right of=R2R,    node distance=4.1cm] (origin2){};
	\node [coordinate, above of=origin, node distance=1.0cm] (origin3){};
	\node [coordinate, right of=R2R,    node distance=1.6cm] (pass){};

	\node [block, above of=origin1, minimum width=3.0cm, node distance=2.6cm, align=center, minimum height=1.2cm] (IMUCAL) {\hypertt{IMU} \texttt{CALIBRATION}  \\ Section \ref{sec:PreFlight_Inertial_Calibration}};
	\node [block, above of=origin1, minimum width=3.0cm, node distance=1.3cm, align=center, minimum height=1.2cm] (PLAT)   {\texttt{PLATFORM POSE}    \\ Section \ref{sec:PreFlight_platform_frame}};
	\node [block, right of=FIXED,   minimum width=3.0cm, node distance=4.1cm, align=center, minimum height=1.2cm] (SWING)  {\texttt{SWINGING}         \\ Section \ref{sec:PreFlight_swinging}};
	\node [block, below of=origin1, minimum width=3.0cm, node distance=1.3cm, align=center, minimum height=1.2cm] (CAMCAL) {\hypertt{CAM} \texttt{CALIBRATION}  \\ Section \ref{sec:PreFlight_Camera_Calibration}};
	\node [block, below of=origin1, minimum width=3.0cm, node distance=2.6cm, align=center, minimum height=1.2cm] (CAM)    {\texttt{CAMERA POSE}      \\ Section \ref{sec:PreFlight_camera_frame}};
	\node [block, above of=origin2, minimum width=3.0cm, node distance=1.6cm, align=center, minimum height=1.2cm] (SELF)   {\texttt{SELF-ALIGNMENT}   \\ Section \ref{sec:PreFlight_SelfAlignment}};
	\node [block, below of=origin2, minimum width=3.0cm, node distance=0.65cm,align=center, minimum height=3.1cm] (FINE)   {\texttt{FINE-ALIGNMENT}   \\ Section \ref{sec:PreFlight_FineAlignment}};
	\node [block, right of=origin3, minimum width=3.0cm, node distance=14.2cm,align=center, minimum height=11.0cm] (NAV)   {\texttt{NAVIGATION}       \\ Chapter \ref{cha:nav}};
	
	\node [coordinate, right of=IMUCAL,    node distance=5.0cm] (output1){};
	\node [coordinate, below of=output1,   node distance=0.4cm] (output12){};
	\node [coordinate, right of=SWING,     node distance=5.0cm] (output2){};
	\node [coordinate, below of=output2,   node distance=0.4cm] (output22){};		
		
	\draw [->] ($(IMUCAL.east)+(0cm,0.3cm)$) -- node[pos=0.32] {\nm{\MACCest, \ \MGYRest}} ($(NAV.west)+(0cm,5.2cm)$);
	\draw [dashed,->] ($(IMUCAL.east)-(0cm,0.4cm)$) -- node[pos=0.45] {\nm{\BzeroACCest, \ \BzeroGYRest}} (output12);
	\draw [->] (PLAT.east) -- node[pos=0.28] {\nm{\TBPBest, \ \phiBPest}} ($(NAV.west)+(0cm,3.6cm)$);
	\draw [->] ($(SWING.east)+(0cm,0.4cm)$) -- node[pos=0.38] {\nm{\MMAGest, \ \BhiMAGvecest}} ($(NAV.west)+(0cm,2.7cm)$);
	\draw [dashed,->] ($(SWING.east)-(0cm,0.4cm)$) -- node[pos=0.3] {\nm{\BzeroMAGvecest}} (output22);
	\draw [->] (CAMCAL.east) -- node[pos=0.15] {\nm{\hat{\vec{\Pi}}}} ($(NAV.west)+(0cm,1.0cm)$);
	\draw [->] (CAM.east) -- node[pos=0.28] {\nm{\TBCBest, \ \phiBCest}} ($(NAV.west)-(0cm,0.3cm)$);
	\draw [->] (SELF.west) -| (pass) |- node[pos=0.01] {\nm{\phiNBest}} (FINE.west);
	\draw [->] ($(FINE.east)+(0cm,1.4cm)$) -- node[pos=0.28] {\nm{\phiNBest, \ \BNDEVest}} ($(NAV.west)-(0cm,2.55cm)$);
	\draw [->] ($(FINE.east)+(0cm,0.7cm)$) -- node[pos=0.45] {\nm{\EACCest, \ \EGYRest, \ \EMAGest}} ($(NAV.west)-(0cm,3.25cm)$);
	\draw [->] ($(FINE.east)-(0cm,0.0cm)$) -- node[pos=0.50] {\nm{\sigmaphiNBinit, \ \sigmaBNDEVinit}} ($(NAV.west)-(0cm,3.95cm)$);
	\draw [->] ($(FINE.east)-(0cm,0.7cm)$) -- node[pos=0.50] {\nm{\sigmaEACCinit, \ \sigmaEGYRinit}} ($(NAV.west)-(0cm,4.65cm)$);
	\draw [->] ($(FINE.east)-(0cm,1.4cm)$) -- node[pos=0.30] {\nm{\sigmaEMAGinit}} ($(NAV.west)-(0cm,5.35cm)$);
\end{tikzpicture}
\caption{Summary of pre-flight activities} \label{fig:pre_flight_activities}
\end{figure}

This chapter describes a group of activities or procedures indispensable for navigation that need to be executed before the flight takes place, and are closely related to the determination of the fixed and run-to-run error contributions to the accelerometers, gyroscopes, magnetometers, and onboard camera (chapter \ref{cha:Sensors}). They can be divided into two groups: the \emph{calibration procedures}, which are executed once and do not need to be repeated unless the sensors are replaced or their position inside the aircraft modified\footnote{In addition, the swinging process needs to be performed every time new equipment is installed inside the aircraft, as this may modify the hard and soft iron magnetism and hence the magnetometer readings.}, and the \emph{alignment procedures}, which need to be executed before the flight takes place every time the aircraft systems are switched on. Figure \ref{fig:pre_flight_activities} shows the different procedures described in this chapter together with their results, noting that the dashed lines indicate discarded estimations, as explained below. 

The calibration procedures include the laboratory calibration of the accelerometers and gyroscopes described in section \ref{sec:PreFlight_Inertial_Calibration}, the determination of the relative pose between \nm{\FP} and \nm{\FB} explained in section \ref{sec:PreFlight_platform_frame}, the magnetometer calibration or swinging described in section \ref{sec:PreFlight_swinging}, the calibration of the camera discussed in section \ref{sec:PreFlight_Camera_Calibration}, and the determination of the relative pose between \nm{\FC} and \nm{\FB} explained in section \ref{sec:PreFlight_camera_frame}. Their main objective is the obtainment of the fixed contributions to the sensor error models\footnote{Refer to section \ref{subsec:Sensors_Inertial_ErrorSources} for the different types of sensor error contributions, including fixed, run-to-run, and in-run.}, this is, the scale factor and cross coupling errors of both inertial sensors and magnetometers (\nm{\MACCest, \, \MGYRest, \, \MMAGest})\footnote{Note that \nm{\MMAGest} also includes the soft iron magnetism.}, the magnetometers hard iron magnetism \nm{\BhiMAGvecest}, the platform and camera poses (\nm{\TBPBest, \, \phiBPest, \, \TBCBest, \, \phiBCest}), and the camera projection map (\nm{\hat{\vec \Pi}}). These procedures also provide estimations for the run-to-run error contributions (\nm{\BzeroACCest, \ \BzeroGYRest, \ \BzeroMAGvecest}), but these need to be discarded as they change every time the aircraft systems are switched on.

Alignment is a two phase process that starts with the self-alignment process of section \ref{sec:PreFlight_SelfAlignment} that provides a coarse estimation of the aircraft attitude, which is then fed into the fine-alignment process described in section \ref{sec:PreFlight_FineAlignment}. The alignment procedures need to be executed every time the aircraft systems are switched on and before the flight takes place. In addition to the aircraft attitude \nm{\qNBest \equiv \phiNBest} and the difference between the modeled and real magnetic fields \nm{\BNDEVest} (section \ref{sec:EarthModel_accuracy}), fine-alignment provides estimations for the full sensor errors \nm{\EACCest} (\ref{eq:Sensor_Inertial_acc_error_filter}), \nm{\EGYRest} (\ref{eq:Sensor_Inertial_gyr_error_filter}), and \nm{\EMAGest} (\ref{eq:Sensor_NonInertial_mag_error_filter}). These full sensor error estimations, although including minor contributions from the fixed errors and in-run errors, are mostly made up by the sensors run-to-run error contributions, this is, the bias offsets \nm{\BzeroACCest}, \nm{\BzeroGYRest}, and \nm{\BzeroMAGvecest}.

Self-alignment needs to be executed before flight because inertial navigation systems (chapter \ref{cha:nav}) require sufficiently accurate estimations of the initial value of the state vector for initialization. While in some cases, such as the air data variables, the position, and the ground velocity, it is sufficient to employ the initial sensor measurements as initial conditions, there are no sensors capable of directly measuring other components, such as the aircraft attitude, the gyroscopes, accelerometers, and magnetometers full errors, and the magnetic field model deviation. 

It is worth noting that with the exception of the simple self-alignment algorithms described in section \ref{sec:PreFlight_SelfAlignment}, the author has not simulated any of the processes described in this chapter. However, the author believes that this chapter is necessary to justify the origin of the sensor specifications included in sections \ref{subsec:Sensors_Inertial_Selected_gyr_acc} and \ref{subsec:Sensors_NonInertial_Magnetometers}, as well as the origin of the inertial navigation filter initial conditions in chapter \ref{cha:nav}. The mechanisms implemented to simulate the pre-flight procedures outputs are described in section \ref{sec:PreFlight_implementation}.


\section{Inertial Sensors Calibration}\label{sec:PreFlight_Inertial_Calibration}

\emph{Calibration} is the process of comparing instrument outputs with known references to determine coefficients that force the outputs to agree with the references over a range of output values \cite{Chatfield1997}. The \hypertt{IMU} inertial sensors need to be calibrated to eliminate the fixed errors originated from manufacturing, and also to determine their temperature sensitivity \cite{Rogers2007}. The calibration process requires significant material and time resources, but it greatly reduces the measurement errors, and hence it is indispensable if the \hypertt{IMU} is to be employed in a navigation system with \hypertt{GNSS}-Denied capabilities. While high grade \hypertt{IMU}s are always factory calibrated, low cost ones generally are not, so it is necessary to calibrate the \hypertt{IMU} at the laboratory before mounting it on the aircraft \cite{Groves2008}.

The calibration shall be performed at a location where the position \nm{\TEgdt} and gravity vector \nm{\gc} (section \ref{subsec:EarthModel_GEOP_acc}) have been previously determined with great precision \cite{Chatfield1997}. It relies on a three axis table, which enables rotating the \hypertt{IMU} with known angular velocities into predetermined precisely controlled orientations \cite{Rogers2007,Tidaldi2014}. Inertial measurements are then compared to reference values (gravity for the accelerometers, torquing rate plus Earth angular velocity for the gyroscopes), and the differences employed to generate corrections \cite{Chatfield1997}.

During calibration, the amount of time that the \hypertt{IMU} is maintained stationary at each attitude, as well as the time required to rotate it between two positions, are trade offs based on two opposing influences. On one side, longer periods of time are preferred as the negative influence of system noise in the measurements tends to even out over time, while on the other, shorter times imply smaller variations of the bias drift over the measurement interval.

It is worth noting that as the calibration is performed before the \hypertt{IMU} is installed on the aircraft, it relies on the \nm{\FP} platform and the models contained in sections \ref{subsec:Sensors_Accelerometer_Triad_ErrorModel} and  \ref{subsec:Sensors_Gyroscope_Triad_ErrorModel}. Although it is possible to use a calibration strategy based on selecting platform orientations that isolate sensor input onto a single axis (for example, gravity is only sensed by the accelerometer that is placed vertically with respect to the Earth surface) to then apply least squares techniques, in real life it is better to employ state estimation \footnote{The estimation filter does not only rely on known gravity and angular velocity, but also the fact that the \hypertt{IMU} is stationary and hence its velocity is zero.} \cite{LIE} to obtain the inertial sensors scale factors, cross coupling errors, and bias offsets \cite{Groves2008,Rogers2007}. The process is repeated at different temperatures so the \hypertt{IMU} processor can later apply the correction based on the \hypertt{IMU} sensor temperature \cite{Groves2008}.
\begin{center}
\begin{tabular}{ccc}
	\hline
	Estimation & \# & Coefficients \\
	\hline
	\rule[1pt]{0pt}{12pt}\nm{\MACCest}		& 6 & \nm{\lrb{\sACCXiest, \ \mACCXijest}}	\\
	\nm{\MGYRest}		& 9 & \nm{\lrb{\sGYRXiest, \ \mGYRXijest}}	\\ 
	\nm{\BzeroACCest}	& 3 & \nm{\BzeroACC \ \NuzeroACCiest}		\\
	\nm{\BzeroGYRest}	& 3 & \nm{\BzeroGYR \ \NuzeroGYRiest}		\\
	\hline
\end{tabular}
\end{center}
\captionof{table}{Results of calibration process}\label{tab:PreFlight_Inertial_calibration}

The twenty-one coefficients estimated in the calibration process are listed in table \ref{tab:PreFlight_Inertial_calibration}. Once estimated, they can be introduced into the \hypertt{IMU} processor so it automatically performs the following corrections:
\begin{eqnarray}
\nm{\fIPPtildetilde} & = & \nm{{\MACCest}^{-1} \ \fIPPtilde - \BzeroACCest}\label{eq:PreFlight_Inertial_acc_error_calib_correction} \\
\nm{\wIPPtildetilde} & = & \nm{{\MGYRest}^{-1} \ \wIPPtilde - \BzeroGYRest}\label{eq:PreFlight_Inertial_gyr_error_calib_correction}
\end{eqnarray}

The simulation considers that the bias offset is exclusively a run-to-run source of error that varies every time the \hypertt{IMU} is switched on, so the \nm{\BzeroACCest} and \nm{\BzeroGYRest} calibration outputs are discarded as they have no relation to the offsets that occur during flight. In addition, the author has decided not to simulate the \hypertt{IMU} calibration process to better use the available resources, and instead model its results by reducing the scale factor and cross couplings errors found on the inertial sensors specifications by an arbitrary amount of \nm{95\%} [A\ref{as:PREFLIGHT_calibration}]. This reduction is already included in the tables contained in section \ref{subsec:Sensors_Inertial_Selected_gyr_acc}. To summarize, instead of applying (\ref{eq:PreFlight_Inertial_acc_error_calib_correction}, \ref{eq:PreFlight_Inertial_gyr_error_calib_correction}) to the measurements obtained by (\ref{eq:Sensor_Inertial_gyr_error_final}, \ref{eq:Sensor_Inertial_acc_error_final}), the simulation directly employs (\ref{eq:Sensor_Inertial_gyr_error_final}, \ref{eq:Sensor_Inertial_acc_error_final}) with reduced \nm{\MGYR} and \nm{\MACC} taken from the section \ref{subsec:Sensors_Inertial_Selected_gyr_acc} tables.


\section{Determination of Platform Frame Pose}\label{sec:PreFlight_platform_frame}

The true relative pose between the platform and body frames, given by \nm{\TBPB} and \nm{\phiBP}, as well as their estimated values \nm{\TBPBest} and \nm{\phiBPest}, play a key role in the readings of the inertial sensors, as explained in section \ref{subsec:Sensors_Inertial_ErrorModel}. This section explains how they can be estimated.

The position responds to \nm{\TBPB = \TRPB - \TRBB}, where \nm{\TRPB = \TRPR} is the position of the platform origin \nm{\OP} with respect to the aircraft structure represented by the reference frame \nm{\FR}, and can be determined with near exactitude based on the \hypertt{IMU} attachment point to the aircraft, while \nm{\TRBB = \TRBR} is also known with near exactitude from the mass analysis (\ref{eq:AircraftModel_Trbb}). This results in the very small error \nm{\sigmaTBPBest} assigned in table \ref{tab:Sensor_Inertial_mounting} for the estimation of \nm{\TBPBest} in (\ref{eq:Sensors_Inertial_Mounting_TBPBest}).

The attitude \nm{\phiBP} requires a short discussion. Note that in the appendix \ref{cha:RefSystems} definitions, the \nm{\FB} and \nm{\FR} axes are parallel to each other with their first and third axes being placed within the aircraft plane of symmetry, while those of \nm{\FP} are loosely aligned with them. This constitutes their formal definitions, and in the simulation the distinction is required because of the need to obtain the actual trajectory \nm{\xTRUTH} as described in chapter \ref{cha:FlightPhysics}. In a real aircraft, however, the actual trajectory is never known, so it is actually possible to proceed as follows:
\begin{enumerate}
\item Mount the \hypertt{IMU} platform so two of its axes are approximately aligned with the forward and down directions of an approximate aircraft plane of symmetry, with no particular need for accuracy.

\item Manually measure the angular deviation \nm{\phiBPest}, rely on the self-alignment procedures of section \ref{sec:PreFlight_SelfAlignment}, or even consider that the platform frame axes are fully aligned with the body ones.
\end{enumerate}

It is important to remark that from the point of view of the \hypertt{GNC} system, the \nm{\FB} frame is just a representation of the aircraft fuselage, and its selection has no influence on its behavior as long as it is centered in the aircraft center of mass, and does not rotate with respect to the structure. The estimated aircraft attitude logically shows a bias with respect to the real (and unknown ones), but this has no influence of the resulting trajectory unless one of the aircraft Euler angles is employed as a guidance target (section \ref{sec:GNC_Guidance}), in which case the attitude bias would also be present in the results. Note also that although no particular accuracy is required, the angles need to be sufficiently small to ensure that the control system works as intended and remains stable. 

Although the previous paragraph would justify the use of \nm{\sigmaphiBPest = 0} in table \ref{tab:Sensor_Inertial_mounting}, the author has preferred not to eliminate any error source and instead employ the small error \nm{\sigmaphiBPest} assigned in table \ref{tab:Sensor_Inertial_mounting} for the estimation of \nm{\phiBPest} in (\ref{eq:Sensors_Inertial_Mounting_eulerBPest}).


\section{Swinging or Magnetometer Calibration}\label{sec:PreFlight_swinging}

Magnetometer calibration is inherently more complex than that of the inertial sensors as it must be performed with the sensors already mounted on the aircraft, as otherwise it would not capture the fixed contributions of the hard iron and soft iron magnetisms (section \ref{subsec:Sensors_NonInertial_Magnetometers}). Known as \emph{swinging}, it relies on obtaining magnetometer readings while the aircraft is positioned at different attitudes that encompass a wide array of heading, pitch, and roll values \cite{Groves2008}, and is executed at a location where the magnetic field is precisely known. 

The accuracy of the results is very dependent of the precision with which the different aircraft attitudes can be determined during swinging. This can be done with the alignment methods described in section \ref{sec:PreFlight_SelfAlignment} or with the use of expensive static instruments. In any case, attitude accuracy is always lower than that obtained with a three axis table during inertial sensor calibration. Once the magnetic field readings are obtained, they are compared to the real magnetic field values, and expression (\ref{eq:Sensor_NonInertial_mag_error_final}) employed with least squares techniques to obtain estimations of the bias (sum of hard iron magnetism \nm{\BhiMAGvec} and offset \nm{\BzeroMAGvec}), and the scale factor and cross coupling matrix \nm{\MMAG}, which also includes the soft iron magnetism. The process can be repeated several times to isolate the influence of hard iron magnetism (a fixed effect that does not change) from the offset, which is a run-to-run error source that changes every time the magnetometers are turned on.
\begin{center}
\begin{tabular}{lcc}
	\hline
	Estimation & \# & Coefficients \\
	\hline
	\rule[1pt]{0pt}{12pt}\nm{\MMAGest}			& 9 & \nm{\lrb{\sMAGXiest, \ \mMAGXijest}}	\\ 
	\nm{\BhiMAGvecest}		& 3 & \nm{\BhiMAGXiest} \\
	\nm{\BzeroMAGvecest}	& 3 & \nm{\BzeroMAG \ \NuzeroMAGiest} \\
	\hline
\end{tabular}
\end{center}
\captionof{table}{Results of swinging process}\label{tab:PreFlight_NonInertial_swinging}

The fifteen coefficients estimated with swinging are listed in table \ref{tab:PreFlight_NonInertial_swinging}. Once estimated, they can be introduced into the processor so it automatically performs the following correction:
\neweq{\BBtildetilde = {\MMAGest}^{-1} \ \BBtilde - \BhiMAGvecest - \BzeroMAGvecest}{eq:PreFlight_NonInertial_mag_error_calib_correction}

The simulation assumes that the bias offset \nm{\BzeroMAGvec} is exclusively a run-to-run source of error that varies every time the magnetometer is switched on, so bias offset coefficients obtained by swinging are discarded as they have no relation to the offsets that occur during flight. In addition, the author has decided not to simulate the swinging process to better use the available resources, and instead model its results by reducing the hard iron bias \nm{\BhiMAGvec} and scale factor and cross couplings errors \nm{\MMAG} found on the sensors specifications by an arbitrary amount of \nm{90\%} [A\ref{as:PREFLIGHT_swinging}]. This reduction is already included in the specifications table of section \ref{subsec:Sensors_NonInertial_Magnetometers}. To summarize, instead of applying (\ref{eq:PreFlight_NonInertial_mag_error_calib_correction}) to the measurements obtained by (\ref{eq:Sensor_NonInertial_mag_error_final}), the simulation directly employs (\ref{eq:Sensor_NonInertial_mag_error_final}) with reduced \nm{\BhiMAGvec} and \nm{\MMAG} taken from the table of section \ref{subsec:Sensors_NonInertial_Magnetometers} .


\section{Camera Calibration}\label{sec:PreFlight_Camera_Calibration}

The \emph{camera calibration} procedure is an extremely important step that needs to be performed with as much care as possible if the camera is to be mounted onboard an aircraft and employed for navigation. It is an established procedure that relies on taking several images of a chessboard from different perspectives, and using the location of each chess box corner viewed in the different images to run a minimization problem, which can be linear or nonlinear depending on the type of calibration model applied to the camera projection map. Such a process is described in detail in \cite{Hartley2003, Soatto2001, Kaehler2016}, and provides excellent results if properly executed.

In this document the author has taken special care to identify and model all the different factors that influence the accuracy of the sensors mounted onboard the aircraft, treating them within the simulation as stochastic variables whose value is given by either the sensor specifications or the pre-flight procedures described in this chapter. Adopting the same methodology for the camera perspective map \nm{\vec{\Pi}} defined by (\ref{eq:Vis_camera_pIMGi}, \ref{eq:Vis_camera_pIMGii}) would result in two different maps: the true one \nm{\vec{\Pi}} employed to generate the image, and its estimation \nm{\hat{\vec{\Pi}}} used by visual navigation algorithms. 

The author however has decided to assume perfect calibration [A\ref{as:CAMERA_calibration}] and hence use the same projection \nm{\vec{\Pi}} in the camera and the navigation system. Although slightly optimistic, this assumption reduces the computing costs and is justified given the low distortion present in current lenses, the realism of the distortion models whose parameters are identified in calibration, and the robustness and accuracy of the calibration process.


\section{Determination of Camera Frame Pose}\label{sec:PreFlight_camera_frame}

The images generated by the onboard camera, and simulated by means of the \texttt{Earth Viewer} application introduced in section \ref{subsec:Sensors_camera_earth_viewer}, do not only depend on the relative pose \nm{\zetaEB} between \nm{\FE} and \nm{\FB}, but also on that of the camera with respect to the aircraft structure (\nm{\zetaBC}), represented by the position \nm{\TBCB} and attitude \nm{\qBC \equiv \phiBC} generated when mounting the camera as described in section \ref{subsec:Sensors_Camera_Mounting}.

Visual navigation algorithms however rely on the navigation system best estimate of this pose, this is, \nm{\TBCBest} and \nm{\phiBCest}, which need to be estimated once the already calibrated camera has been mounted on the aircraft. The two-phase process requires a chess board such as that employed in calibration. 

The first phase uses an optimization procedure quite similar to that used in calibration to determine the relative pose between the camera frame \nm{\FC} and one rigidly attached to the chessboard. Instead of using the location of each chess box corner in different images, this process relies on a single photo and imposes that all chess boxes are square and have the same size, which is enough to obtain a solution up to an unknown scale. The size of the chess boxes provides the scale required to unambiguously solve the identification problem with high precision.

The second step is to obtain the pose between the chessboard frame and the \nm{\FR} frame linked to the aircraft structure, and then use the \nm{\TRBR = \TRBB} displacement provided by (\ref{eq:AircraftModel_Trbb}) to obtain the desired pose with the body frame\footnote{The \nm{\FR} and \nm{\FB} frames are aligned.}. This is a straightforward geometric optimization problem that relies on distance measurements between chessboard points and aircraft structure points whose coordinates in the \nm{\FR} frame are known. The resulting accuracy depends on that of the measured distances, so special equipment may be required given the importance of the final estimations for the success of the visual navigation algorithms.

Overall this is a very robust and accurate process if properly executed, which enables the author to employ the very small errors \nm{\sigmaTBCBest} and \nm{\sigmaphiBCest} assigned in table \ref{tab:Sensor_Camera_mounting} for the stochastic estimation of \nm{\TBCBest} and \nm{\phiBCest} in each simulation run by means of (\ref{eq:Sensors_Camera_Mounting_TBCBest}) and (\ref{eq:Sensors_Camera_Mounting_eulerBCest}).
 

\section{Self-Alignment}\label{sec:PreFlight_SelfAlignment}

The objective of the \emph{alignment} algorithms, also known as attitude initialization, is to obtain a coarse estimation of the aircraft attitude \nm{\phiNBest = \lrsb{\psiest \ \ \thetaest \ \ \xiest}^T} required by the fine-alignment process described in section \ref{sec:PreFlight_FineAlignment}. Although it can also be performed while in motion or onboard another vehicle, the results are better if performed with the aircraft stationary, in which case it is known as \emph{self-alignment}.


\subsection{Leveling}\label{subsec:PreFlight_leveling}

The first phase of self-alignment is known as \emph{leveling}, and relies on averaging the accelerometer outputs for a period of time while the aircraft is stationary to provide coarse estimates of its body pitch and bank angles (\nm{\thetaest, \ \xiest}), so it is sensitive to disturbances caused by fueling and wind vibration \cite{Groves2008}. Leveling does not require any previous knowledge about the gravity field \nm{\gc} at the aircraft position, relying exclusively on the fact that gravity is approximately orthogonal to the ellipsoid. As such, the accuracy of the results depends on the accelerometer error sources (section \ref{subsec:Sensors_Inertial_ErrorModel}) as well as the deviations of the local gravity field from vertical (section \ref{sec:EarthModel_accuracy}).

Leveling is performed while the aircraft is stationary, so the specific force given by (\ref{eq:EquationsMotion_speceficforce2}) coincides with the real gravity provided by (\ref{eq:EarthModel_accuracy_gravity}), as shown in (\ref{eq:PreFlight_leveling_fIBN}). Neglecting the influence of the Earth angular velocity \nm{\wIE} (section \ref{sec:EquationsMotion_velocity}) and considering the stationary aircraft, the lever arm between the \hypertt{IMU} reference point and the center of mass does not play any role in the (\ref{eq:Sensor_Inertial_acc_error_final}) accelerometer measurements, which can be simplified into (\ref{eq:PreFlight_leveling_fIBBtilde}).
\begin{eqnarray}
\nm{\fIBN} & = & \nm{- \gcNREAL}\label{eq:PreFlight_leveling_fIBN} \\
\nm{\fIBBtilde} & = & \nm{\vec g_{\ds{\hat{\vec \zeta}_{{\sss BP}*}}} \Bigg(\MACC \bigg(\vec g_{\ds{\vec \zeta_{{\sss BP}*}}}^{-1} \Big(\vec g_{\ds{\vec \zeta_{{\sss NB}*}}}^{-1} \big(\fIBN\big)\Big)\bigg)\Bigg) + \vec e_{\sss BW,ACC}^{\sss B}} \label{eq:PreFlight_leveling_fIBBtilde}\\
\nm{\fIBBtildebar} & = & \nm{E\lrsb{\fIBBtilde}}\label{eq:PreFlight_leveling_fIBBtildebar}
\end{eqnarray}

The only input required for the leveling process is the accelerometer outputs \nm{\fIBBtildebar} averaged over a period of time (\ref{eq:PreFlight_leveling_fIBBtildebar}) that on one side should be as long as possible to diminish the influence of the system noise, and on the other short enough as to lower the risk of contamination by external vibrations. The accelerometer measurements are only recorded after a warm up period intended to stabilize the bias drift. Once the averages are available, the body pitch and bank angles are quickly estimated by imposing a vertical gravity field \cite{Groves2008}:
\neweq{\thetaest = \arctan \dfrac{\fIBBtildebari}{\sqrt{\fIBBtildebarsquareii + \fIBBtildebarsquareiii}} \hspace{80pt} \xiest = \arctan \dfrac{- \fIBBtildebarii}{- \fIBBtildebariii}}{eq:PreFlight_leveling_theta_xi}


\subsection{Gyrocompassing}\label{subsec:PreFlight_gyrocompassing}

The second phase of self-alignment is known as \emph{gyrocompassing}, although its execution generally overlaps with that of leveling. It employs the averages of the gyroscopes outputs, together with the body pitch and roll estimations provided by leveling, to obtain a coarse estimate of the aircraft heading \nm{\psiest}. Gyrocompassing follows a process similar to that of leveling, but where the latter relies on the vertical nature of the gravity field, gyrocompassing is based on the Earth angular velocity \nm{\wIE} (section \ref{sec:EquationsMotion_velocity}) having no component along \nm{\iEii}, this is, no East component. As \nm{\wIE} is very small, gyrocompassing does not work with low quality gyroscopes in which the bias can easily mask the Earth rotation, and hence alternative means to estimate the aircraft heading are necessary, as described in section \ref{subsec:PreFlight_magnetic_alignment} below. The gyrocompassing results do not only depend on the gyroscopes error sources (section \ref{subsec:Sensors_Inertial_ErrorModel}), but also on those of leveling.

Gyrocompassing is performed while the aircraft is stationary, so the aircraft and motion angular velocities (\nm{\wEN, \ \wNB}) (section \ref{sec:EquationsMotion_velocity}) are zero and the inertial angular velocity coincides with that of the Earth, as shown in (\ref{eq:PreFlight_gyrocompassing_wIBN}). The gyroscope measuring equation (\ref{eq:Sensor_Inertial_gyr_error_final}) is repeated here for clarity, resulting in (\ref{eq:PreFlight_gyrocompassing_wIBBtilde}).
\begin{eqnarray}
\nm{\wIBN} & = & \nm{\wIEN = \vec g_{\ds{\vec \zeta_{{\sss EN}*}}}^{-1} \big(\wIEE\big) = \lrsb{\omegaE \; \cos \varphi \ \ \ \ 0 \ \ \ \ - \; \omegaE \; \sin \varphi }^T}\label{eq:PreFlight_gyrocompassing_wIBN} \\
\nm{\wIBBtilde} & = & \nm{\vec {Ad}_{\ds{\hat{\vec q}_{\sss BP}}} \ \MGYR \ \vec {Ad}_{\ds{\vec q_{\sss BP}}}^{-1} \ \wIBB + \vec e_{\sss BW,GYR}^{\sss B}} \label{eq:PreFlight_gyrocompassing_wIBBtilde} \\
\nm{\wIBBtildebar} & = & \nm{E\lrsb{\wIBBtilde}}\label{eq:PreFlight_gyrocompassing_wIBBtildebar}
\end{eqnarray}

The algorithm not only requires the gyroscope outputs \nm{\wIBBtildebar} averaged over a period of time (\ref{eq:PreFlight_gyrocompassing_wIBBtildebar}), but also the body pitch and roll estimates (\nm{\thetaest, \ \xiest}) provided by leveling. The estimation process relies in the Earth angular velocity having no East component, which is equivalent to imposing that the inner product between the second row of \nm{\RNBest} and \nm{\wIBBtildebar} is zero. This results in:
\neweq{\psiest = \arctan \dfrac{- \wIBBtildebarii \ \cos \xiest + \wIBBtildebariii \ \sin \xiest}{\wIBBtildebari \ \cos \thetaest + \wIBBtildebarii \ \sin \xiest \ \sin \thetaest + \wIBBtildebariii \ \cos \xiest \ \sin \thetaest}}{eq:PreFlight_gyrocompassing_psi}


\subsection{Magnetic Alignment}\label{subsec:PreFlight_magnetic_alignment}

The body heading \nm{\psiest} can also be estimated by means of the magnetometers instead of the gyroscopes, although it requires these to have been previously calibrated by swinging (section \ref{sec:PreFlight_swinging}). The process is analogous to gyrocompassing, except that it relies on averaging the measurements of the Earth magnetic field instead of those of the Earth angular velocity. Once corrected with the magnetic deviation obtained from a magnetic model evaluated at the current position (section \ref{sec:EarthModel_WMM}), the measured magnetic field should not have any East component. The accuracy of the results hence not only depends on the magnetometer error sources (section \ref{subsec:Sensors_NonInertial_Magnetometers}) and those of the leveling process, but also on the differences between the real magnetic field at the aircraft location and that provided by the magnetic model (section \ref{sec:EarthModel_accuracy}).

The process starts with the estimation of the magnetic declination \nm{\chiMAGest} at the aircraft location (\ref{eq:PreFlight_magnetic_alignment_chiMAGest}), which relies on the aircraft systems model of the magnetic field described in section \ref{sec:EarthModel_WMM}:
\neweq{\chiMAGest = f_\chi\lrp{\TEgdt} = \arctan \dfrac{\BNMODELii}{\BNMODELi}}{eq:PreFlight_magnetic_alignment_chiMAGest}

As in gyrocompassing, magnetometer readings \nm{\BBtilde} are averaged of a period of time (\ref{eq:PreFlight_magnetic_alignment_BBtildebar}) to lower the influence of the system noise present in the measurements. The magnetometer measuring equation (\ref{eq:Sensor_NonInertial_mag_error}) is repeated here for clarity, resulting in (\ref{eq:PreFlight_magnetic_alignment_BBtilde}):
\begin{eqnarray}
\nm{\BBtilde} & = & \nm{\BzeroMAGvec + \BhiMAGvec + \MMAG \, \Big(\vec g_{\ds{\vec \zeta_{{\sss NB}*}}}^{-1} \big(\BNREAL\big)\Big) + \vec e_{\sss W,MAG}^{\sss B}} \label{eq:PreFlight_magnetic_alignment_BBtilde}\\ 
\nm{\BBtildebar} & = & \nm{E\lrsb{\BBtilde}}\label{eq:PreFlight_magnetic_alignment_BBtildebar}
\end{eqnarray}

Magnetic alignment requires the body pitch and roll estimations (\nm{\thetaest, \ \xiest}) provided by leveling. It relies on the magnetic field belonging to a plane composed by the vertical direction (\nm{\iNiii}) and a horizontal vector (belonging to the plane composed by \nm{\iNi} and \nm{\iNii}) separated from the North direction by an angle \nm{\chiMAG}. This is analogous to stating that there should be no East component once corrected with magnetic declination:
\neweq{\psiest = \chiMAGest +  \arctan \dfrac{- \BBtildebarii \ \cos \xiest + \BBtildebariii \ \sin \xiest}{\BBtildebari \ \cos \thetaest + \BBtildebarii \ \sin \xiest \ \sin \thetaest + \BBtildebariii \ \cos \xiest \ \sin \thetaest}}{eq:PreFlight_magneto_alignment_psi}


\subsection{Self-Alignment Results}\label{subsec:PreFlight_alignment_results}

The main objective of the self-alignment methods is to initialize the aircraft attitude for the fine-alignment process of section \ref{sec:PreFlight_FineAlignment}. As such, they only provide coarse estimations whose accuracy is very dependent on the error sources of the different sensors, specially those of the accelerometers, and also on the accuracy of the gravity and magnetic models at the aircraft location. The author has executed a test with two different grades of accelerometers, gyroscopes, and magnetometers, and the results are shown in table \ref{tab:PreFlight_NonInertial_PreFlight_alignment_results}. The \say{baseline} configuration encompasses the sensors described in tables \ref{tab:Sensors_gyr}, \ref{tab:Sensors_acc} and \ref{tab:Sensors_mag}, while \say{better} relies on the inertial sensors and magnetometers of superior grade listed in \cite{INSE}. 

Based on a random location, the test involves running the different alignment algorithms with the aircraft positioned at 36 different headings equispaced by \nm{10^{\circ}} with no pitch nor bank angles, and computing the mean and standard deviation of the resulting differences between the estimated angles and the real ones. Although not shown in the table, gyrocompassing is also included in the test, but requires sensors of higher grade than those of the baseline configuration to provide acceptable results.
\begin{center}
\begin{tabular}{lccccp{0.1cm}cc}
	\hline
	Sensor Grade & \multicolumn{4}{c}{Leveling} & & \multicolumn{2}{c}{Magnetic Align} \\
	 & \nm{\mu_{\Delta\theta}} & \nm{\sigma_{\Delta\theta}} & \nm{\mu_{\Delta\xi}} & \nm{\sigma_{\Delta\xi}} & & \nm{\mu_{\Delta\psi}} & \nm{\sigma_{\Delta\psi}} \\
	\hline
	Baseline & \nm{-0.515} & \nm{+0.003} & \nm{+0.699} & \nm{+0.003} & & \nm{-0.309} & \nm{+2.976} \\ 
	Better   & \nm{-0.103} & \nm{+0.003} & \nm{+0.083} & \nm{+0.003} & & \nm{-0.306} & \nm{+1.004} \\
	\hline
\end{tabular}
\end{center}
\captionof{table}{Metrics of alignment errors \nm{\lrsb{^{\circ}}}}\label{tab:PreFlight_NonInertial_PreFlight_alignment_results}

The results show that the most complicated task within self-alignment is to estimate the aircraft heading, as gyrocompassing only provides acceptable results with very high quality gyroscopes, while magnetic alignment is intrinsically limited by the differences between the magnetic model and the real magnetic field, as well as the significant error sources present in the magnetometers.


\section{Fine Alignment}\label{sec:PreFlight_FineAlignment}

\emph{Fine alignment} is an attitude and inertial sensor calibration process performed immediately after the navigation system initialization, and has two main objectives \cite{Farrell2008,Groves2008,Chatfield1997}:
\begin{itemize}
\item Reduce the attitude initialization errors remaining from the self-alignment process.
\item Reduce the run-to-run contribution to the errors of the accelerometers, gyroscopes, and magnetometers.
\end{itemize}

Although fine-alignment can also be performed while in motion or onboard another vehicle, the process described here considers that the aircraft is stationary with respect to the Earth while on the ground, and that it has just completed the self-alignment process described in section \ref{sec:PreFlight_SelfAlignment}, which results in good approximations to the aircraft attitude \nm{\phiNBest = \lrsb{\psiest \ \ \thetaest \ \ \xiest}^T}.

With the aircraft stationary on the ground, the fine-alignment process relies on the same inertial navigation algorithms employed during flight (chapter \ref{cha:nav}), but imposing that the velocity is zero and the position and attitude are constant. Note that in these conditions the specific force (\ref{eq:EquationsMotion_speceficforce2}) measured by the accelerometers is the opposite of the gravity acceleration \nm{\gc} (section \ref{subsec:EarthModel_GEOP_acc}), while the aircraft inertial angular velocity \nm{\wIB} coincides with the Earth angular velocity \nm{\wIE} (section \ref{sec:EquationsMotion_velocity}):
\begin{eqnarray}
\nm{\fIBB} & = & \nm{\dfrac{\vec F_{\sss GROUND}^{\sss B}}{m} =  - \ \vec g_{\ds{\vec \zeta_{{\sss NB}*}}}^{-1} \big(\gcNREAL\big)} \label{eq:PreFlight_SpecificForce_static} \\
\nm{\wIBB} & = & \nm{\vec g_{\ds{\vec \zeta_{{\sss NB}*}}}^{-1} \big(\wIEN\big)}\label{eq:PreFlight_AngularVelocity_static}
\end{eqnarray}

Based on (\ref{eq:PreFlight_SpecificForce_static}, \ref{eq:PreFlight_AngularVelocity_static}), false resolving of gravity or angular velocity may occur because of both sensor errors and residual attitude errors remaining from self-alignment. In these conditions the navigation algorithm estimates false velocity and attitude changes that contradict the stationary references, which can be employed to further improve the previous self-alignment estimation of the aircraft attitude \nm{\phiNBest} as well as the sensor errors.

A fully static calibration such as this is unable to completely separate between attitude and sensor errors, as both have the same effects, so maneuvers must be performed to separate them. For this reason, a quasi stationary alignment also includes heading changes. It is also worth noting that fine-alignment is very sensitive to outside disturbances caused by wind or human activity, in which case the best solution is to maintain the process for several minutes to average out the disturbances \cite{Groves2008}.

In a real aircraft there exists no real distinction between the estimation of the observed trajectory performed on the ground by means of the fine-alignment process and that executed on the air by the inertial navigation system (chapter \ref{cha:nav}), as these are just two different names for the same algorithm. When the aircraft is switched on, and after a short period for sensor warm-up required to reduce the influence of the bias drift present in the inertial sensors, a coarse estimation of the aircraft attitude is obtained by means of the self-alignment process described in section \ref{sec:PreFlight_SelfAlignment}. Maintaining the aircraft stationary, this is followed by the initialization of the navigation algorithms, employing the following as initial estimations for the state vector: zero for the linear and angular velocities, an average of the \hypertt{GNSS} receiver outputs for the position, the self-alignment results for the aircraft attitude, and zero for the errors of the different sensors. This is the fine-alignment phase, and when its solution stabilizes, the aircraft proceeds to taxi, take-off, and execution of its flight plan, but the process of estimating the aircraft state is then called navigation, as described in chapter \ref{cha:nav}.

The simulation introduced in section \ref{sec:Intro_traj} can not be used to model the above flight initialization activities. Doing so would require extensive modifications to the aircraft forces (chapter \ref{cha:AircraftModel}) and equations of motion (chapter \ref{cha:FlightPhysics}) to properly model the aircraft and sensors behavior while the aircraft is stationary on the tarmac or accelerating on the runway. The author has preferred to use his limited resources into obtaining a more realistic simulation of the aircraft flight instead of modeling the pre-flight procedures described in this section. The whole purpose of this section, however, is to justify that the inertial navigation algorithms of chapter \ref{cha:nav} can be initialized with attitude and sensor error values that are very close to their real values because they have previously been estimated in stationary conditions by means of the fine-alignment process. This is equivalent to stating that when the aircraft initiates the flight, the navigation system already possesses good estimations of the attitude and full sensor errors, and hence these estimations can be employed to initialize the simulation of the navigation filter algorithms. Table \ref{tab:PreFlight_fine_alignment} shows the fine-alignment accuracy results arbitrarily selected by the author as the baseline:
\begin{center}
\begin{tabular}{lccc}
	\hline
	Initialization Baseline	 & Variable & Value & Unit \\
	\hline
	Attitude error           & \nm{\sigmaqNBinit}   & \nm{0.1} & \nm{^{\circ}} \\
	Full accelerometer error & \nm{\sigmaEACCinit}  & \nm{1}   & \nm{\%}  \\
	Full gyroscope error     & \nm{\sigmaEGYRinit}  & \nm{1}   & \nm{\%}  \\
	Full magnetometer error  & \nm{\sigmaEMAGinit}  & \nm{1}   & \nm{\%}  \\
	Magnetic field deviation & \nm{\sigmaBNDEVinit} & \nm{10}  & \nm{\%}  \\
	\hline
\end{tabular}
\end{center}
\captionof{table}{Fine-alignment accuracy values}\label{tab:PreFlight_fine_alignment}

Based on the above values, the initial conditions for the body attitude, full sensor errors, and deviation between the modeled magnetic field and the real one are obtained based on the following equations, which basically indicate that the difference between the real initial conditions (\nm{\qNBzero, \ \EACCzero, \ \EGYRzero, \ \EMAGzero, \ \BNDEVzero}) and those estimated by the fine-alignment process (\nm{\qNBinit, \ \EACCinit, \ \EGYRinit, \ \EMAGinit, \ \BNDEVinit}) are on average the values indicated in table \ref{tab:PreFlight_fine_alignment}. Note that the symbol \nm{\circ} represents the Hadamart or element-wise matrix product, and that (\ref{eq:PreFlight_fine_alignment_qNB}) makes use of the \nm{\mathbb{SO}\lrp{3}} rotation vector. Also, \nm{N_{\sss qNB,INIT}} is a realization of a standard normal random variable \nm{N\lrp{0, \, 1}}, and \nm{\vec N_{E \sss ACC,INIT}}, \nm{\vec N_{E \sss GYR,INIT}}, \nm{\vec N_{E \sss MAG,INIT}}, and \nm{\vec N_{B\sss N,DEV,INIT}} are realizations of uncorrelated vectors of size three each composed of three uncorrelated standard normal random variables \nm{N\lrp{0, \, 1}}. The angles \nm{\phi_1} and \nm{\phi_2} are obtained from single realizations of the uniform distributions \nm{U\lrp{-179, 180}} and \nm{U\lrp{-90, 90}}, respectively.
\begin{eqnarray}
\nm{\vec n_{\sss qNB}} & = & \nm{\lrsb{\cos \phi_1 \ \cos \phi_2 \ \ \ \ \sin \phi_1 \ \cos \phi_2 \ \ \ \ \sin \phi_2}^T}\label{eq:PreFlight_fine_alignment_qNB1} \\
\nm{\gamma_{\sss qNB}} & = & \nm{\sigmaqNBinit \ N_{\sss qNB,INIT}} \label{eq:PreFlight_fine_alignment_qNB2} \\
\nm{\qNBinit}      & = & \nm{Exp\lrp{\vec n_{\sss qNB} \, \gamma_{\sss qNB}} \otimes \qNBzero} \label{eq:PreFlight_fine_alignment_qNB} \\
\nm{\EACCinit}     & = & \nm{\EACCzeroest - \EACCzero = \sigmaEACCinit \ \EACCzero \circ \vec N_{E \sss ACC,INIT}} \label{eq:PreFlight_fine_alignment_EACC} \\
\nm{\EGYRinit}     & = & \nm{\EGYRzeroest - \EGYRzero = \sigmaEGYRinit \ \EGYRzero \circ \vec N_{E \sss GYR,INIT}} \label{eq:PreFlight_fine_alignment_EGYR} \\
\nm{\EMAGinit}     & = & \nm{\EMAGzeroest - \EMAGzero = \sigmaEMAGinit \ \EMAGzero \circ \vec N_{E \sss MAG,INIT}} \label{eq:PreFlight_fine_alignment_EMAG} \\
\nm{\BNDEVinit}    & = & \nm{\BNDEVzeroest - \BNDEVzero = \sigmaBNDEVinit \ \BNDEVzero \circ \vec N_{B\sss N,DEV,INIT}} \label{eq:PreFlight_fine_alignment_BNDEV}
\end{eqnarray}

As noted above, the estimated values constitute the initial conditions for the inertial navigation algorithms of chapter \ref{cha:nav}, and are required by (\ref{eq:nav_vis_iner_filter_qNBinit_so3_local} , \ref{eq:nav_vis_iner_filter_stestinit_so3_local}). 


\section{Implementation, Validation, and Execution} \label{sec:PreFlight_implementation}

As mentioned throughout this chapter, the different pre-flight procedures have not been simulated, and hence cannot be validated. However, this section describes the implementation of the simple algorithm described at the bottom of section \ref{sec:PreFlight_FineAlignment} to simulate the results of the fine-alignment algorithm that constitute the initial conditions of the inertial navigation filter described in chapter \ref{cha:nav}. This section can be considered and is indeed better understood as the continuation of section \ref{sec:Sensors_implementation}, which describes the implementation of the stochastic algorithms that describe the sensors behavior while maintaining the capability to repeat the results if so desired. As a matter of fact, as the initialization conditions change from run to run, they can be considered as additional sensors that are only executed once at the beginning of the trajectory.

Considering this, the discrete uniform distribution initialized with the flight seed \nm{\seedR} is not only employed to obtain the run sensor seeds \nm{\upsilon_{j,\sss F}} listed in section \ref{sec:Sensors_implementation}, the Earth model seeds described in section \ref{sec:EarthModel_implementation}, and the guidance parameters specified in section \ref{sec:gc_implementation}, but is realized five additional times to generate the \emph{initialization seeds} \nm{\upsilon_{j,\sss F,INIT}} (\nm{\seedRphiNB}, \nm{\seedREACC}, \nm{\seedREGYR}, \nm{\seedREMAG}, \nm{\seedRBNDEV}) for the initial conditions.
\begin{center}
\begin{tabular}{lccc}
	\hline
	Type & Group & Error Sources & Seeds \\
	\hline
	Flight		& \nm{\seedR}	& in-run & \rule[-6pt]{0pt}{12pt}\nm{\seedRphiNB, \, \seedREACC, \, \seedREGYR, \, \seedREMAG, \, \seedRBNDEV} \\
	\hline
\end{tabular}
\end{center}
\captionof{table}{Initialization seeds} \label{tab:PreFlight_seeds}

Each initial condition relies on a single standard normal distribution \nm{N\lrp{0, \, 1}}, which is initialized with the corresponding seed \nm{\upsilon_{j,\sss F,INIT}} for that condition. Then each is realized three times to obtain the stochastic values required by expressions (\ref{eq:PreFlight_fine_alignment_qNB1}) through (\ref{eq:PreFlight_fine_alignment_BNDEV}) above.

 \cleardoublepage
\chapter{Guidance and Control}\label{cha:gc}

The \emph{guidance, navigation, and control} (\hypertt{GNC}) system contains the algorithms required to fly autonomously, this is, without interacting with any human operator. It continuously adjusts the position of the throttle lever and the aerodynamic control surfaces (elevator, ailerons, rudder) with the only input of the sensor measurements and the previously loaded mission objectives. Its simulation considers that the execution of all its algorithms is instantaneous [A\ref{as:GNC_instantaneous}] and that the three systems are perfectly time synchronized not only among themselves but also with the sensors. The navigation system processes all measurements as soon as they become available at a frequency of \nm{100 \ Hz} [A\ref{as:GNC_n_synchronize}], while the guidance and control systems only operate at half that frequency, this is, \nm{50 \ Hz} [A\ref{as:GNC_gc_synchronize}]. Another important assumption is that the position of the throttle lever, as well as that of the aerodynamic control surfaces, respond instantaneously to the commands provided by the control system [A\ref{as:GNC_control_surfaces}], and hence so do the aerodynamic and propulsive actions described in chapter \ref{cha:AircraftModel}.
\begin{figure}[h]
\centering
\begin{tikzpicture}[auto, node distance=2cm,>=latex']
	\node [coordinate](midinput) {};
	\node [coordinate, above of=midinput, node distance=0.4cm] (xSENSEDinput){};
	\node [coordinate, below of=midinput, node distance=0.4cm] (xESTzeroinput){};
	\node [block, right of=midinput, minimum width=3.0cm, node distance=5.3cm, align=center, minimum height=1.5cm] (NAVIGATION) {\texttt{NAVIGATION}};
	\node [coordinate, right of=NAVIGATION, node distance=4.5cm] (midpoint){};
	\filldraw [black] (midpoint) circle [radius=1pt];
	\node [coordinate, above of=midpoint, node distance=1.5cm] (pointup){};
	\node [coordinate, below of=midpoint, node distance=1.5cm] (pointdown){};
	\node [block, right of=pointup, minimum width=3.0cm, node distance=2.8cm, align=center, minimum height=1.5cm] (GUIDANCE) {\texttt{GUIDANCE}};
	\node [block, right of=pointdown, minimum width=3.0cm, node distance=2.8cm, align=center, minimum height=1.5cm] (CONTROL) {\texttt{CONTROL}};
	\node [coordinate, right of=CONTROL, node distance=3.5cm] (deltaCNTRoutput){};
	\node [coordinate, above of=midpoint, node distance=1.9cm] (xREFinputBIS){};
		
	\draw [->] (xSENSEDinput) -- node[pos=0.45] {\nm{\xvectilde\lrp{t_s} = \xSENSED\lrp{t_s}}} ($(NAVIGATION.west)+(0cm,0.4cm)$);
	\draw [->] (xESTzeroinput) -- node[pos=0.1] {\nm{\xvecestzero}} ($(NAVIGATION.west)-(0cm,0.4cm)$);
	\draw [->] (GUIDANCE.south) -- node[pos=0.5] {\nm{\deltaTARGET\lrp{t_c}}} (CONTROL.north);
	\draw [->] (xREFinputBIS) -- node[pos=0.4] {\nm{\xREF}} ($(GUIDANCE.west)+(0cm,0.4cm)$);	
	\draw [->] (NAVIGATION.east) -- node[pos=0.5] {\nm{\xvecest\lrp{t_n} = \xEST\lrp{t_n}}} (midpoint) |- ($(GUIDANCE.west)-(0cm,0.4cm)$);
	\draw [->] (midpoint) |- (CONTROL.west);
	\draw [->] (CONTROL.east) -- node[pos=0.5] {\nm{\deltaCNTR\lrp{t_c}}} (deltaCNTRoutput);
\end{tikzpicture}
\caption{Guidance, navigation, and control flow diagram}
\label{fig:GNC_flow_diagram}
\end{figure}
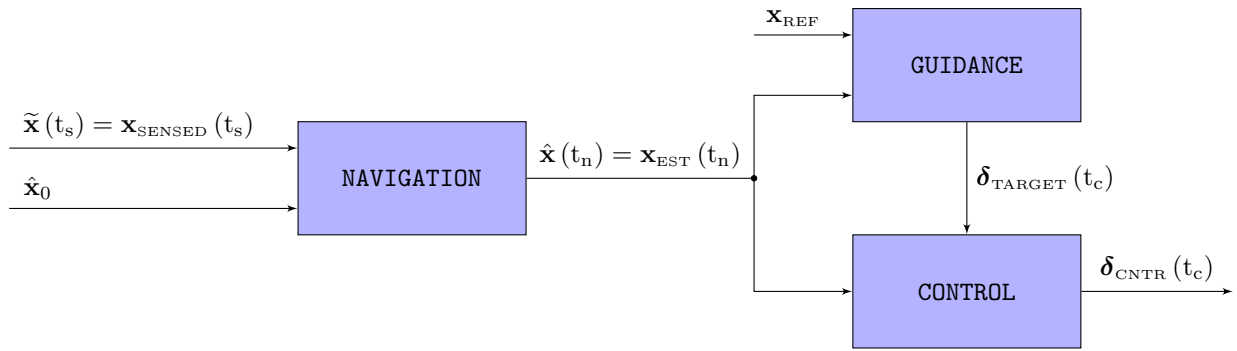

The \hypertt{GNC} system is organized as shown in figure \ref{fig:GNC_flow_diagram}. Considering an ordered evolution from sensor data to control targets, the first task corresponds to the navigation system (chapter \ref{cha:nav}), which takes the sensed state \nm{\xvectilde\lrp{t_s} = \xSENSED\lrp{t_s}} described in section \ref{sec:Sensors_ST}, where \nm{t_s = s \cdot \DeltatSENSED}, and processes it to obtain the observed or estimated state \nm{\xvecest\lrp{t_n} = \xEST\lrp{t_n}} described in section \ref{sec:GNC_OT}, where \nm{t_n = n \cdot \DeltatEST = t_s} as both operate at the same frequency according to table \ref{tab:Intro_trajectory_frequencies}, and which presents two main advantages over the sensed measurements. On one side, it is more complete as it contains variables such as the aircraft attitude that are not measured by the sensors, while in the other, its errors, or the difference with the actual aircraft state, are smaller thanks to the navigation system.

Once the observed state \nm{\xvecest\lrp{t_n} = \xEST\lrp{t_n}} is available, it is taken by the guidance system (section \ref{sec:GNC_Guidance}) to determine how far the aircraft is located along the flight when compared with the mission objectives or reference trajectory \nm{\xREF} (section \ref{subsec:GNC_RT}) previously loaded into the system by the human operator, and then extracts the proper guidance targets \nm{\deltaTARGET\lrp{t_c}} to be tracked by the control system, where \nm{t_c = c \cdot \DeltatCNTR}. Note that both the guidance and control systems work at half the frequency of the navigation filter (table \ref{tab:Intro_trajectory_frequencies}). Last is the turn of the control system (section \ref{sec:GNC_Control}), which compares the guidance targets \nm{\deltaTARGET\lrp{t_c}} with the observed state \nm{\xvecest\lrp{t_n} = \xEST\lrp{t_n}} and generates updates to the position of the control parameters \nm{\deltaCNTR\lrp{t_c}}.

This chapter focuses on the guidance and control systems (sections \ref{sec:GNC_Guidance} and \ref{sec:GNC_Control}), while the navigation system is described in the next chapter. Their implementation in the simulation is discussed in section \ref{sec:gc_implementation}. 


\section{Guidance System}\label{sec:GNC_Guidance}

In an autonomous \hypertt{UAV}, the human operator assigns the \emph{mission objectives} and uploads them by means of an active communication channel. On the aircraft side, the guidance system processes those objectives, and from that point on provides the control system with the appropriate control targets \nm{\deltaTARGET\lrp{t_c}} at every instant, in such a way that the mission objectives will be achieved if the aircraft is capable of successfully adhering to those targets\footnote{Note the difference with remotely piloted aircraft, in which the human operator has the capability, if desired, of remotely moving the throttle and aerodynamic control surfaces to fulfill the mission objectives.}. As long as the data link is open, the operator can continuously monitor the mission progress by means of the estimated trajectory provided by the navigation system, and may decide to modify the mission objectives at any time, as for example if operational reasons mandate a mission update.
\begin{figure}[h]
\centering
\begin{tikzpicture}[auto, node distance=2cm,>=latex']
	\node [block, minimum width=2.6cm, align=center, minimum height=2.0cm] (GUIDANCE) {\texttt{GUIDANCE}};
	\node [coordinate, left of=GUIDANCE, node distance=4.5cm] (inputmiddle){};
	\node [coordinate, above of=inputmiddle, node distance=0.50cm] (xREFinput){};
	\node [coordinate, below of=inputmiddle, node distance=0.50cm] (xESTinput){};
	\node [coordinate, right of=GUIDANCE, node distance=4.0cm] (outputdeltaTRGT){};
	\draw [->] (xREFinput) -- node[pos=0.2] {\nm{\xREF}} ($(GUIDANCE.west)+(0cm,0.5cm)$);
	\draw [->] (xESTinput) -- node[pos=0.5] {\nm{\xvecest\lrp{t_n} = \xEST\lrp{t_n}}} ($(GUIDANCE.west)-(0cm,0.5cm)$);
	\draw [->] (GUIDANCE.east) -- node[pos=0.6] {\nm{\deltaTARGET\lrp{t_c}}} (outputdeltaTRGT);
\end{tikzpicture}
\caption{Guidance system flow diagram}
\label{fig:Guidance_flow_diagram}
\end{figure}
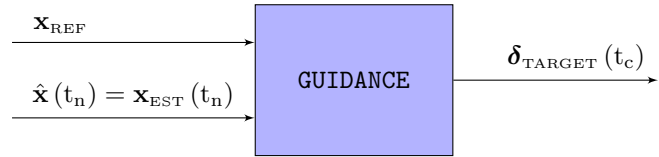

If there is no communications link, which may occur because of technical failure, excessive distance, signal blocking by mountains or buildings, or even malicious actions by a third party, the aircraft is in a completely autonomous mode, and continues using the last received mission objectives until it lands or the communications are restored. This is precisely the case treated in this document, which assumes that the guidance system has received the mission objectives but can not expect to receive any corrections to it.

Figure \ref{fig:Guidance_flow_diagram} represents a partial view of figure \ref{fig:GNC_flow_diagram}, and shows how the guidance system, provided with the reference trajectory, trajectory script, or mission objectives \nm{\xREF} described in section \ref{subsec:GNC_RT}, is capable of providing the control system with a time series of control targets \nm{\deltaTARGET\lrp{t_c}} for the throttle lever and aerodynamic control surfaces, while employing the aircraft estimated state \nm{\xvecest\lrp{t_n} = \xEST\lrp{t_n}} provided by the navigation system to determine its position along the trajectory script and hence generate the proper targets. 


\subsection{Reference Trajectory}\label{subsec:GNC_RT}

The \emph{reference trajectory} or \emph{trajectory script} (\hypertt{RT}) represents the implementation of the mission objectives, and contains a series of targets used by the \emph{guidance system} to unambiguously specify how the aircraft is to be operated at all instants, from the current time (or engine turn on, if the script is provided before the flight has started) until the aircraft has landed (or has reached a location where it can be recovered by a remote pilot). Being unambiguous does not mean that the trajectory script needs to be definite; as a matter of fact, autonomous vehicles require a conditional script that covers what-to-do instructions in case certain contingencies occur. For example, in case the \hypertt{GNSS} signals are lost and hence the aircraft is unable to fulfill the primary mission, it is wise to schedule a back up plan in which the primary mission is aborted and the aircraft instead tries to reach a predetermined recovery location from where it can be remotely landed.

In the case of the simulation, the objective is not to recreate a flexible and robust guidance system and associated trajectory script capable of fulfilling complex guidance objectives\footnote{Some examples of complex but common guidance objectives are the tracking of a minimum distance line or great circle over the \hypertt{WGS84} ellipsoid, the maximization of the aircraft range, the minimization of the fuel consumption, or reaching a given destination while maintaining a minimum safety distance from any populated area for noise or safety reasons.}, but to implement basic guidance modes such as maintaining a constant airspeed or constant altitude, climbing or descending at a given path angle, following a given bearing or turning at a given bank angle, etc., that nevertheless are sufficient to fulfill basic missions. Additional modes can be added at little cost if necessary.

\begin{figure}[h]
\centering
\begin{tikzpicture}[auto, node distance=2cm,>=latex']
	\node [blockgrey1, minimum width=2.0cm, align=center, minimum height=0.5cm] (T1) {\nm{\deltaTARGETTone}};
	\node [blockgrey2, minimum width=2.0cm, align=center, minimum height=0.5cm, below of=T1, node distance=0.8cm] (E1) {\nm{\deltaTARGETEone}};
	\node [blockgrey3, minimum width=2.0cm, align=center, minimum height=0.5cm, below of=E1, node distance=0.8cm] (A1) {\nm{\deltaTARGETAone}};
	\node [blockgrey4, minimum width=2.0cm, align=center, minimum height=0.5cm, below of=A1, node distance=0.8cm] (R1) {\nm{\deltaTARGETRone}};
	\node [blockgrey5, minimum width=2.0cm, align=center, minimum height=0.5cm, below of=R1, node distance=0.8cm] (TR1) {\nm{\deltaTRGone}};
	\node [blockgrey1, minimum width=2.0cm, align=center, minimum height=0.5cm, right of=T1, node distance=2.5cm] (T2) {\nm{\deltaTARGETTtwo}};
	\node [blockgrey2, minimum width=2.0cm, align=center, minimum height=0.5cm, below of=T2, node distance=0.8cm] (E2) {\nm{\deltaTARGETEtwo}};
	\node [blockgrey3, minimum width=2.0cm, align=center, minimum height=0.5cm, below of=E2, node distance=0.8cm] (A2) {\nm{\deltaTARGETAtwo}};
	\node [blockgrey4, minimum width=2.0cm, align=center, minimum height=0.5cm, below of=A2, node distance=0.8cm] (R2) {\nm{\deltaTARGETRtwo}};
	\node [blockgrey5, minimum width=2.0cm, align=center, minimum height=0.5cm, below of=R2, node distance=0.8cm] (TR2) {\nm{\deltaTRGtwo}};
	\node [right of=T2, node distance=2.0cm] (Tdots) {\textbf{...}};
	\node [below of=Tdots, node distance=0.8cm] (Edots) {\textbf{...}};
	\node [below of=Edots, node distance=0.8cm] (Adots) {\textbf{...}};
	\node [below of=Adots, node distance=0.8cm] (Rdots) {\textbf{...}};
	\node [below of=Rdots, node distance=0.8cm] (TRdots) {\textbf{...}};
	\node [blockgrey1, minimum width=2.0cm, align=center, minimum height=0.5cm, right of=Tdots, node distance=2.0cm] (Tn) {\nm{\deltaTARGETTn}};
	\node [blockgrey2, minimum width=2.0cm, align=center, minimum height=0.5cm, below of=Tn, node distance=0.8cm] (En) {\nm{\deltaTARGETEn}};
	\node [blockgrey3, minimum width=2.0cm, align=center, minimum height=0.5cm, below of=En, node distance=0.8cm] (An) {\nm{\deltaTARGETAn}};
	\node [blockgrey4, minimum width=2.0cm, align=center, minimum height=0.5cm, below of=An, node distance=0.8cm] (Rn) {\nm{\deltaTARGETRn}};
	\node [blockgrey5, minimum width=2.0cm, align=center, minimum height=0.5cm, below of=Rn, node distance=0.8cm] (TRn) {\nm{\deltaTRGn}};		
\end{tikzpicture}
\caption{Reference trajectory divided into operations}
\label{fig:Guidance_structure}
\end{figure}
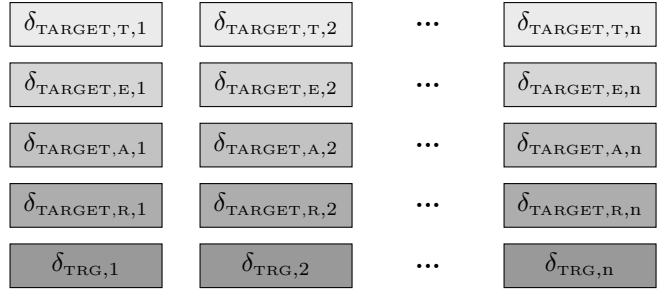

The reference trajectory \nm{\xREF} is divided into \emph{operations} \nm{\deltaTARGET} (\ref{eq:GC_guidance_RT}), and each operation is composed by four instructions and a trigger (\ref{eq:GC_guidance_operation}), as depicted in figure \ref{fig:Guidance_structure}. One and only one operation is active at any time; when it concludes, the script activates the next operation, and so on until the aircraft lands or is recovered by an operator. 
\begin{eqnarray}
\nm{\xREF} & = & \nm{\lrb{\deltaTARGETone, \, \deltaTARGETtwo, \, \ldots , \deltaTARGETn}}\label{eq:GC_guidance_RT} \\
\nm{\deltaTARGET} & = & \nm{\lrsb{\deltaTARGETT \ \ \deltaTARGETE \ \  \deltaTARGETA \ \ \deltaTARGETR \ \ \deltaTRG}^T}\label{eq:GC_guidance_operation}
\end{eqnarray}

The operation \emph{trigger} is a Boolean expression based on the observed state vector \nm{\xvecest\lrp{t_n} = \xEST\lrp{t_n}} provided by the navigation system. It is continuously evaluated when the corresponding operation is active, and its sign monitored. As soon as the sign switches, the operation concludes and is replaced by the next one in the pipeline. Table \ref{tab:GC_guidance_trigger_targets} shows the available trigger targets for each operation: time \emph{t}, operation time \nm{\DeltatOP}, geodetic altitude \emph{h}, pressure altitude \nm{\Hp}, path angle \nm{\gamma}, bearing \nm{\chi}, or heading \nm{\psi}. For example, a trigger \nm{\deltaTRG = \lrb{\Hp = 2500 \ m}} means that the operation concludes when the aircraft first reaches that pressure altitude (\nm{\Hpest = 2500 \ m}), no matter if it is descending or climbing, while a trigger  \nm{\deltaTRG = \lrb{\DeltatOP = 50 \ s}} means that the operation time duration is exactly fifty seconds. Note that the trigger evaluation is based on the observed or estimated trajectory \nm{\xEST}, not on the unknown actual trajectory \nm{\xTRUTH}, so the switch from one operation to the next does not occur exactly as prescribed by the reference trajectory.
\begin{center}
\begin{tabular}{ll}
	\hline
	\multicolumn{2}{c}{Trigger Options} \\
	\hline
	\nm{\deltaTRG} & \nm{t, \ \DeltatOP, \ h, \ \Hp, \ \gamma, \ \chi, \ \psi} \\
	\hline
\end{tabular}
\end{center}
\captionof{table}{Available trigger targets} \label{tab:GC_guidance_trigger_targets}

The operation \emph{instructions} represent the targets employed by the aircraft control system while the operation is active, and are also evaluated based on the observed state vector \nm{\xvecest\lrp{t_n} = \xEST\lrp{t_n}} provided by the navigation system. Most fixed wing aircraft possess four mechanisms to control the flight (throttle lever, elevator, ailerons, and rudder); this number hence coincides with the number of degrees of freedom present in the equations of motion, which must be closed at all times so the system of differential equations of section \ref{sec:EquationsMotion_integration} is well posed and can be integrated to provide an unambiguous trajectory (for given wind and environmental conditions). Each instruction is associated to a given degree of freedom, and they must be compatible. For example, an instruction specifying a given path angle is a valid target for the elevator, as its motion creates a longitudinal moment along \nm{\iBii} that effectively modifies the aircraft path angle; however, the same path angle target is not suitable for the ailerons as their influence on the path angle is extremely small and far from linear.

Table \ref{tab:GC_guidance_instruction_targets} shows the available targets for each instruction type. For example, an elevator instruction \nm{\deltaTARGETE} \nm{= \lrb{\Hp = 2500 \ m}} means that the control system shall employ the elevator to maintain the pressure altitude as close as possible to that target (\nm{\Hpest = 2500 \ m}). The throttle lever \nm{\deltaT} is usually employed to control the airspeed \nm{\vtas}, although it can also be directly set up, for example to execute a maximum power climb or an idle descent. The remaining longitudinal motion targets are controlled by the elevator \nm{\deltaE}, which can use as targets the body pitch angle \nm{\theta}, the pressure altitude \nm{\Hp}, the geodetic altitude \emph{h}, or the aerodynamic path angle \nm{\gammaTAS}. With respect to the lateral motion, the rudder \nm{\deltaR} is always employed to control the sideslip angle \nm{\beta}, while the ailerons \nm{\deltaA} can be used in two different ways: for straight segments it is possible to specify the desired aircraft heading \nm{\psi}, aerodynamic heading \nm{\chiTAS}, or aircraft bearing \nm{\chi}, while turns are indicated by specifying the desired body bank angle \nm{\xi} or airspeed bank angle \nm{\muTAS}.
\begin{center}
\begin{tabular}{lcc}
	\hline
	\multicolumn{3}{c}{Instructions Options} \\
	\hline
	Throttle & \nm{\deltaTARGETT} & \nm{\vtas, \ \deltaT} \\
	Elevator & \nm{\deltaTARGETE} & \nm{\theta, \ \Hp, \ h, \ \gammaTAS} \\ 
	Ailerons & \nm{\deltaTARGETA} & \nm{\xi, \ \chi, \ \psi, \ \muTAS, \ \chiTAS} \\ 
	Rudder   & \nm{\deltaTARGETR} & \nm{\beta} \\ 
	\hline
\end{tabular}
\end{center}
\captionof{table}{Available instruction targets} \label{tab:GC_guidance_instruction_targets}

Note that unlike the actual \nm{\xvec\lrp{t_t} = \xTRUTH\lrp{t_t}} and observed \nm{\xvecest\lrp{t_n} = \xEST\lrp{t_n}} trajectories, which represent complete representations of the aircraft trajectory in the sense that all variables can be obtained from the state vector by means of algebraic equations, the reference trajectory \nm{\xREF} resembles the sensed trajectory \nm{\xvectilde\lrp{t_s} = \xSENSED\lrp{t_s}} in being incomplete, this is, both contain a limited amount of variables from which other relevant aircraft states can not be directly obtained.
	

\section{Control System}\label{sec:GNC_Control}

\begin{figure}[h]
\centering
\begin{tikzpicture}[auto, node distance=2cm,>=latex']
	\node [block, minimum width=2.6cm, align=center, minimum height=2.0cm] (CONTROL) {\texttt{CONTROL}};
	\node [coordinate, left of=CONTROL, node distance=4.5cm] (inputmiddle){};
	\node [coordinate, above of=inputmiddle, node distance=0.50cm] (deltaTRGTinput){};
	\node [coordinate, below of=inputmiddle, node distance=0.50cm] (xESTinput){};
	\node [coordinate, right of=CONTROL, node distance=4.0cm] (outputdeltaCNTR){};
	\draw [->] (deltaTRGTinput) -- node[pos=0.4] {\nm{\deltaTARGET\lrp{t_c}}} ($(CONTROL.west)+(0cm,0.5cm)$);
	\draw [->] (xESTinput) -- node[pos=0.5] {\nm{\xvecest\lrp{t_n} = \xEST\lrp{t_n}}} ($(CONTROL.west)-(0cm,0.5cm)$);
	\draw [->] (CONTROL.east) -- node[pos=0.6] {\nm{\deltaCNTR\lrp{t_c}}} (outputdeltaCNTR);
\end{tikzpicture}
\caption{Control system flow diagram}
\label{fig:Control_flow_diagram}
\end{figure}
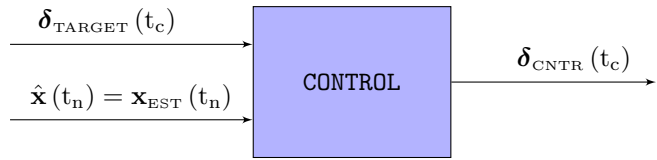

The mission of the aircraft control system is to move the throttle lever \nm{\deltaT} and aerodynamic control surfaces (elevator \nm{\deltaE}, ailerons \nm{\deltaA}, and rudder \nm{\deltaR}) in such a way that the estimated trajectory \nm{\xvecest\lrp{t_n} = \xEST\lrp{t_n}} provided by the navigation system (chapter \ref{cha:nav}) follows the control targets \nm{\deltaTARGET\lrp{t_c}} provided by the guidance system (section \ref{sec:GNC_Guidance}) with as little error as possible \cite{Stevens2003, Franklin1998}. Note that the actual or real trajectory \nm{\xvec\lrp{t_t} = \xTRUTH\lrp{t_t}} is unknown to the aircraft systems, and hence they must operate based on the observed trajectory \nm{\xvecest\lrp{t_n} = \xEST\lrp{t_n}}, which represents the best knowledge the aircraft systems posses about the actual trajectory. This is depicted in figure \ref{fig:Control_flow_diagram}, which represents a partial view of figure \ref{fig:GNC_flow_diagram}.

Figure \ref{fig:Control_flow_diagram} shows the three parts required for a successful aircraft flight that achieves the mission objectives: a well implemented control system, well estimated observed states, and well defined target objectives. If the second objective is achieved, any well designed control system is capable of following the required targets. In fact, not even the best existing control system can execute its job if the observed trajectory is way off from the actual trajectory, but even a simple well implemented control system can be successful if provided with an observed trajectory that closely resembles the real trajectory.

Section \ref{subsec:GNC_Control_PID} describes the \hypertt{PID} controller, which forms the basis for the implementation of the aircraft control system. The specific implementation however depends on which target variables (table \ref{tab:GC_guidance_instruction_targets}) apply to each operation, and is described in section \ref{subsec:GNC_Control_loops}.


\subsection{PID Controller}\label{subsec:GNC_Control_PID}
 
A \emph{proportional integral derivative} (\hypertt{PID}) controller is a control feedback mechanism that provided with a \emph{target} or \emph{setpoint} \nm{r\lrp{t}} and an estimated \emph{process variable} \nm{y\lrp{t}} that is the output of a physical process or plant, computes the position of the \emph{control variable} \nm{u\lrp{t}} to the plant as the sum of three terms, each composed as the product of a given coefficient or gain by either the error value, its integral with time, or its time derivative \cite{Ogata2002}. The \emph{error value} \nm{e\lrp{t}} is defined as the difference between the setpoint and the process variable:
\begin{eqnarray}
\nm{e\lrp{t}} & = & \nm{r\lrp{t} - y\lrp{t}}\label{eq:GNC_Control_PID_error} \\
\nm{u\lrp{t}} & = & \nm{K_P \cdot e\lrp{t} + K_I \int_0^t e\lrp{\tau} \mathrm{d}\tau + K_D \derpar{e\lrp{t}}{t}}\label{eq:GNC_Control_PID}
\end{eqnarray}
\begin{figure}[h]
\centering
\begin{tikzpicture}[auto, node distance=2cm,>=latex']
	\node [coordinate](rinput) {};
	\node [shape=circle, fill=white, draw, right of=rinput, node distance=2cm] (SUM) {\nm{\sss{\sum}}};
    \node [blockyellow, right of=SUM, text width=2cm, node distance=3.0cm] (PID) {\hypertt{PID}};	
	\node [blockgreen, right of=PID, text width=2cm, node distance=4.0cm] (PLANT) {\texttt{PLANT}};	
	\node [coordinate, right of=PLANT, node distance=2cm] (point1){};
	\node [coordinate, right of=point1, node distance=1cm] (youtput){};
	\node [coordinate, below of=point1, node distance=1.2cm] (point2){};
	
	\filldraw [black] (point1) circle [radius=1pt];
	\draw [->] (rinput) -- node[pos=0.5] {\nm{r\lrp{t}}} (SUM.west);
	\draw [->] (SUM.east) -- node[pos=0.5] {\nm{e\lrp{t}}} (PID.west);
	\draw [->] (PID.east) -- node[pos=0.5] {\nm{u\lrp{t}}} (PLANT.west);
	\draw [->] (PLANT.east) -- node[pos=0.5] {\nm{y\lrp{t}}} (youtput);
	\draw [->] (point1) -- (point2) -| (SUM.south);
	
	\node [coordinate, left of=SUM, node distance=0.5cm] (plus){};
	\node [below of=plus, node distance=0.2cm] (plussign) {\nm{+}};	
	\node [coordinate, below of=SUM, node distance=0.5cm] (minus){};
	\node [left of=minus, node distance=0.2cm] (minussign) {\nm{-}};	
\end{tikzpicture}
\caption{\texttt{PID} controller flow diagram}
\label{fig:GNC_PID_flow_diagram}
\end{figure}
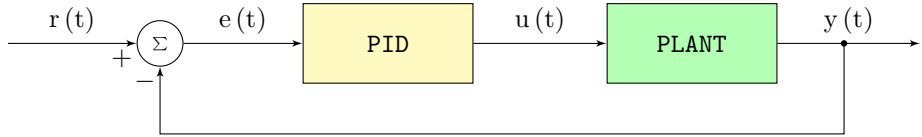

The \emph{proportional} (\hypertt{P}) control issues a response that is proportional to the error value by a factor called the proportional gain \nm{K_P}, and should be the main contributor to the control variable. If operating by itself, a pure \hypertt{P} controller always results in a residual error as it requires it to generate a response. The gain should be adjusted with care, as the control loop can become unstable if it is too high, but be less responsive or sensitive if too low.

The \emph{integral} (\hypertt{I}) term aims to eliminate the residual error caused by the proportional term, enabling the system to reach its target value. Based on the integral gain \nm{K_I}, it can however cause overshooting if the integral gain is too high. A pure \hypertt{I} controller is capable of eliminating the error value, but would be slow to react to deviations, and then brutal in its response, being prone to overshooting and oscillations.

The \emph{derivative} (\hypertt{D}) control aims to flatten the error curve to zero, damping the responsiveness of the system and hence reducing the overshoot. It relies on the derivative gain \nm{K_D} and in theory improves the system settling time and its stability. In practice, however, it is quite sensitive to measurement noise and is generally used with both a low gain to limit its influence and a low pass filter applied to the derivative input to reduce noise \cite{Haugen2008}.
\begin{center}
\begin{tabular}{lcccccc}
	\hline
	Control		 & Gain		& Rise Time		  & Overshoot		& Settling Time	  & Steady State Error & Stability \\
	\hline
	Proportional & \nm{K_P} & \nm{\downarrow} & \nm{\uparrow}   & \nm{\sim}       & \nm{\downarrow} & \nm{\downarrow} \\
	Integral     & \nm{K_I} & \nm{\downarrow} & \nm{\uparrow}   & \nm{\uparrow}   & 0               & \nm{\downarrow} \\ 
	Derivative   & \nm{K_D} & \nm{\sim}       & \nm{\downarrow} & \nm{\downarrow} & \nm{\sim}       & \nm{\uparrow} \\ 
	\hline
\end{tabular}
\end{center}
\captionof{table}{Effects of increasing the gains of a \texttt{PID} controller} \label{tab:GNC_control_PID_effects}

Table \ref{tab:GNC_control_PID_effects} shows the theoretical effect of increasing the value of each of the three gains on the controller rise time, its tendency to overshoot, its settling time, the steady state error, and its stability. In general, \nm{K_P} and \nm{K_I} are a trade off between reducing the overshoot and increasing the controller settling time. A small \nm{K_D} makes it more responsive to real and relatively fast error changes, but also to noise.

The main advantage of \hypertt{PID} controllers and the reason for their popularity in all kinds of applications is their simplicity. However, this also implies some negative consequences \cite{Hagglund2012}:
\begin{itemize}
\item As the gains are constant and do not contain any direct knowledge of the plant physical process, its performance is always reactive and a compromise.
\item Derivative noise can result in instability of the \hypertt{D} control, and should be smoothed with a low pass filter.
\item Setpoint step changes imply significant windup of the \hypertt{I} term and may result in overshooting the target. Possible solutions include disabling \hypertt{I} control until the process variable is closer to the new target or setpoint, and preventing the \hypertt{I} term from accumulating beyond certain boundaries.
\item Setpoint step changes can also result in an excessive response by the \hypertt{P} and \hypertt{D} controllers. This can be addressed by ramping up the setpoint in a linear way instead of steps, or by weighting the setpoints with two adjustable factors (usually between 0 and 1) in these two controllers (but not in \hypertt{I} to avoid steady state errors). As shown in (\ref{eq:GNC_Control_PID2}), where \nm{\lrb{b, \, c}} are the setpoint weights, the response to load disturbances and measurement noise does not depend on the weights, but that to setpoint changes does. It is common to use \nm{c = 0} and \nm{0 \leq b \leq 1} to reduce overshoots in setpoint response.
\end{itemize}
\neweq{u\lrp{t} = K_P \lrsb{b \cdot r\lrp{t} - y\lrp{t} + \dfrac{K_I}{K_P} \, \int_0^t e\lrp{\tau} \mathrm{d}\tau + \dfrac{K_D}{K_P} \, \derpar{\lrp{c \cdot r\lrp{t} - y\lrp{t}}}{t}}}{eq:GNC_Control_PID2}


\subsection{Primary and Secondary Loops}\label{subsec:GNC_Control_loops}

The aircraft control system has been implemented as four independent control mechanisms, where the control variables are each of the control parameters first introduced in section \ref{sec:AircraftModel_Control_Parameters} and repeated here for clarity:
\neweq{\deltaCNTR = \lrsb{\deltaT \ \ \deltaE \ \ \deltaA \ \ \deltaR}^T} {eq:GNC_Control_deltaCNTR}

In each control cycle, the system provides the updated position of the throttle lever \nm{\deltaT}, elevator \nm{\deltaE}, ailerons \nm{\deltaA}, and rudder \nm{\deltaR}, resulting in updated forces and moments that generate changes in the process variables. 
\begin{center}
\begin{tabular}{lcccc}
	\hline
	\multicolumn{2}{c}{Guidance Targets} & Direct Input & Primary Loop & Secondary Loop \\
	\hline
	Throttle & \nm{\deltaTARGETT} & \nm{\deltaT} & \nm{\vtas}  & - \\
	Elevator & \nm{\deltaTARGETE} & -            & \nm{\theta} & \nm{\Hp, \, h, \gammaTAS} \\ 
	Ailerons & \nm{\deltaTARGETA} & -            & \nm{\xi}    & \nm{\chi, \, \psi, \, \muTAS, \, \chiTAS} \\ 
	Rudder   & \nm{\deltaTARGETR} & -            & \nm{\beta}  & - \\ 
	\hline 
\end{tabular}
\end{center}
\captionof{table}{Guidance targets and control mechanisms} \label{tab:GC_control_choices}

Noting that all four control mechanisms (one per control parameter) are fully independent from each other, there exist three specific implementations that are applied or not to a given mechanism depending on the specific target variable provided by the guidance system during a given operation.


\subsubsection{Direct Input of Control Variable}\label{subsubsec:GNC_Control_direct}

This case only applies when the target for the throttle control \nm{\deltaTARGETT} is the throttle parameter \nm{\deltaT} itself, as indicated in table \ref{tab:GC_control_choices}, and may be employed to indicate maximum power climbs \nm{\lrp{\deltaT = 1}} or idle descents \nm{\lrp{\deltaT = 0}}. In this case there is no control whatsoever as the throttle \nm{\deltaT} takes the value indicated by the target without any feedback mechanism. 


\subsubsection{Primary Loops}\label{subsubsec:GNC_Control_primary}

\begin{figure}[h]
\centering
\begin{tikzpicture}[auto, node distance=2cm,>=latex']
	\node [coordinate](vtasinput) {};
	\node [coordinate, below of=vtasinput, node distance=1.2cm]  (thetainput) {};
	\node [coordinate, below of=thetainput, node distance=1.2cm] (xiinput) {};
	\node [coordinate, below of=xiinput, node distance=1.2cm]    (betainput) {};
	\node [coordinate, below of=vtasinput, node distance=1.8cm]    (middleinput) {};

	\node [shape=circle, fill=white, draw, right of=vtasinput, node distance=3.4cm]  (SUMvtas)  {\nm{\sss{\sum}}};
    \node [shape=circle, fill=white, draw, right of=thetainput, node distance=4.2cm] (SUMtheta) {\nm{\sss{\sum}}};
	\node [shape=circle, fill=white, draw, right of=xiinput, node distance=5.0cm]    (SUMxi)    {\nm{\sss{\sum}}};
	\node [shape=circle, fill=white, draw, right of=betainput, node distance=5.8cm]  (SUMbeta)  {\nm{\sss{\sum}}};
	
	\node [blockyellow, right of=SUMvtas, text width=2cm, node distance=4.5cm]  (PIDPRIMT) {\hypertt{PID}\nm{_{\sss{PRIM,T}}}};
	\node [blockyellow, below of=PIDPRIMT, text width=2cm, node distance=1.2cm] (PIDPRIME) {\hypertt{PID}\nm{_{\sss{PRIM,E}}}};	
	\node [blockyellow, below of=PIDPRIME, text width=2cm, node distance=1.2cm] (PIDPRIMA) {\hypertt{PID}\nm{_{\sss{PRIM,A}}}};	
	\node [blockyellow, below of=PIDPRIMA, text width=2cm, node distance=1.2cm] (PIDPRIMR) {\hypertt{PID}\nm{_{\sss{PRIM,R}}}};	
	
	\node [blockgreen, right of=middleinput, text width=2cm, node distance=12.0cm, minimum height=4.5cm] (PLANT) {\texttt{PLANT}};	
	
	\node [coordinate, right of=PLANT, node distance=2cm]      (point1){};
	\node [coordinate, right of=point1, node distance=1cm]     (youtput){};
	\node [coordinate, below of=SUMvtas, node distance=4.5cm]  (pointvtas){};
	\node [coordinate, below of=SUMtheta, node distance=3.3cm] (pointtheta){};
	\node [coordinate, below of=SUMxi, node distance=2.1cm]    (pointxi){};
	\node [coordinate, below of=SUMbeta, node distance=0.9cm]  (pointbeta){};
	
	\filldraw [black] (point1) circle [radius=1pt];
	\filldraw [black] (pointtheta) circle [radius=1pt];
	\filldraw [black] (pointxi) circle [radius=1pt];
	\filldraw [black] (pointbeta) circle [radius=1pt];
		
	\draw [->] (vtasinput)  -- node[pos=0.45]  {\nm{\deltaTARGETT = \vtas}}  (SUMvtas.west);
	\draw [->] (thetainput) -- node[pos=0.30]  {\nm{\deltaTARGETE = \theta}} (SUMtheta.west);
	\draw [->] (xiinput)    -- node[pos=0.25]   {\nm{\deltaTARGETA = \xi}}    (SUMxi.west);
	\draw [->] (betainput)  -- node[pos=0.23]   {\nm{\deltaTARGETR = \beta}}  (SUMbeta.west);
		
	\draw [->] (SUMvtas.east)  -- (PIDPRIMT.west);
	\draw [->] (SUMtheta.east) -- (PIDPRIME.west);
	\draw [->] (SUMxi.east)    -- (PIDPRIMA.west);
	\draw [->] (SUMbeta.east)  -- (PIDPRIMR.west);
		
	\draw [->] (PIDPRIMT.east) -- node[pos=0.5] {\nm{\deltaT}} ($(PLANT.west) + (0cm,1.8cm)$);
	\draw [->] (PIDPRIME.east) -- node[pos=0.5] {\nm{\deltaE}} ($(PLANT.west) + (0cm,0.6cm)$);
	\draw [->] (PIDPRIMA.east) -- node[pos=0.5] {\nm{\deltaA}} ($(PLANT.west) - (0cm,0.6cm)$);
	\draw [->] (PIDPRIMR.east) -- node[pos=0.5] {\nm{\deltaR}} ($(PLANT.west) - (0cm,1.8cm)$);
		
	\draw [->] (PLANT.east) -- node[pos=0.5] {\nm{\xvecest = \xEST}} (youtput);
	
	\draw [->] (point1) |- (pointvtas) -- (SUMvtas.south);
	\draw [-] (SUMvtas.south) -- node[pos=0.07] {\nm{\vtasest}} (pointvtas);
	\node [coordinate, left of=SUMvtas, node distance=0.5cm] (plusvtas){};
	\node [below of=plusvtas, node distance=0.2cm] (plussignvtas) {\nm{+}};	
	\node [coordinate, below of=SUMvtas, node distance=0.5cm] (minusvtas){};
	\node [left of=minusvtas, node distance=0.2cm] (minussignvtas) {\nm{-}};	
	
	\draw [->] (pointtheta) -| (SUMtheta.south);
	\draw [-] (SUMtheta.south) -- node[pos=0.1] {\nm{\thetaest}} (pointtheta);
	\node [coordinate, left of=SUMtheta, node distance=0.5cm] (plustheta){};
	\node [below of=plustheta, node distance=0.2cm] (plussigntheta) {\nm{+}};	
	\node [coordinate, below of=SUMtheta, node distance=0.5cm] (minustheta){};
	\node [left of=minustheta, node distance=0.2cm] (minussigntheta) {\nm{-}};	
	
	\draw [->] (pointxi) -| (SUMxi.south);
	\draw [-] (SUMxi.south) -- node[pos=0.25] {\nm{\xiest}} (pointxi);
	\node [coordinate, left of=SUMxi, node distance=0.5cm] (plusxi){};
	\node [below of=plusxi, node distance=0.2cm] (plussignxi) {\nm{+}};	
	\node [coordinate, below of=SUMxi, node distance=0.5cm] (minusxi){};
	\node [left of=minusxi, node distance=0.2cm] (minussignxi) {\nm{-}};	
	
	\draw [->] (pointbeta) -| (SUMbeta.south);
	\draw [-] (SUMbeta.south) -- node[pos=0.6] {\nm{\betaest}} (pointbeta);
	\node [coordinate, left of=SUMbeta, node distance=0.5cm] (plusbeta){};
	\node [below of=plusbeta, node distance=0.2cm] (plussignbeta) {\nm{+}};	
	\node [coordinate, below of=SUMbeta, node distance=0.5cm] (minusbeta){};
	\node [left of=minusbeta, node distance=0.2cm] (minussignbeta) {\nm{-}};	
\end{tikzpicture}
\caption{Primary control loops}
\label{fig:GNC_PID_flow_diagram_primary}
\end{figure}
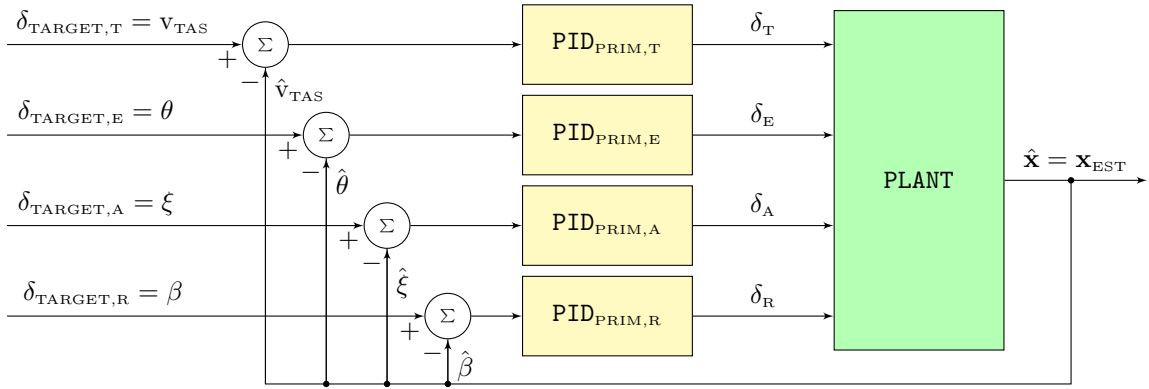

This case applies when the targets provided by the guidance system correspond to those indicated in table \ref{tab:GC_control_choices} for each of the four control mechanisms. If that is the case, then each control mechanism is implemented as represented in figure \ref{fig:GNC_PID_flow_diagram_primary} with a \hypertt{PID} controller similar to those described in section \ref{subsec:GNC_Control_PID}. The outputs of these primary \hypertt{PID} control loops are the control parameters themselves. Note that figure \ref{fig:GNC_PID_flow_diagram_primary} is just a representation of the case in which all four primary loops are employed simultaneously, but each control mechanism is independent from the others and only depends on its own target variable.

In order to maintain the benefits of \hypertt{D} control but reduce its possible instability effects in the presence of noisy measurements, all primary \hypertt{PID} loops include a low pass filter to smooth the derivative inputs. In addition, all primary \hypertt{PID} loops also make use of setpoint ramping to reduce overshooting when the setpoint changes from operation to operation. The ailerons \hypertt{PID} loop differs from the others in that it is implemented as the sum of two \hypertt{PID} loops, with the main one based on the difference between the target and estimated body bank angle \nm{\xi} complemented by an auxiliary one that relies on the target and estimated airspeed \nm{\vtas}. This enables reducing the control errors when the aircraft is simultaneously turning and varying its airspeed.


\subsubsection{Secondary Loops}\label{subsubsec:GNC_Control_secondary}

Secondary control loops have only been implemented in the case of the elevator \nm{\deltaE} and the ailerons \nm{\deltaA}, and are employed when the corresponding target variables provided by the guidance system coincide with those indicated in table \ref{tab:GC_control_choices}. 

\begin{figure}[h]
\centering
\begin{tikzpicture}[auto, node distance=2cm,>=latex']
	\node [coordinate](thetainput) {};
	\node [coordinate, below of=thetainput, node distance=1.4cm] (xiinput) {};
	\node [coordinate, below of=thetainput, node distance=0.7cm] (middleinput) {};

    \node [shape=circle, fill=white, draw, right of=thetainput, node distance=2.3cm] (SUMtheta) {\nm{\sss{\sum}}};
	\node [shape=circle, fill=white, draw, right of=xiinput,    node distance=2.3cm] (SUMxi)    {\nm{\sss{\sum}}};

	\node [blockyellow, right of=SUMtheta, text width=2cm, node distance=2.0cm] (PIDSECE) {\hypertt{PID}\nm{_{\sss{SEC,E}}}};	
	\node [blockyellow, right of=SUMxi,    text width=2cm, node distance=2.0cm] (PIDSECA) {\hypertt{PID}\nm{_{\sss{SEC,A}}}};	
	
	\node [shape=circle, fill=white, draw, right of=PIDSECE, node distance=4.2cm] (SUM2theta) {\nm{\sss{\sum}}};
	\node [shape=circle, fill=white, draw, right of=PIDSECA, node distance=4.2cm] (SUM2xi)    {\nm{\sss{\sum}}};
	
	\node [blockyellow, right of=SUM2theta, text width=2cm, node distance=2.0cm] (PIDPRIME) {\hypertt{PID}\nm{_{\sss{PRIM,E}}}};	
	\node [blockyellow, right of=SUM2xi,    text width=2cm, node distance=2.0cm] (PIDPRIMA) {\hypertt{PID}\nm{_{\sss{PRIM,A}}}};	
	
	\node [blockgreen, right of=middleinput, text width=2cm, node distance=13.5cm, minimum height=2.5cm] (PLANT) {\texttt{PLANT}};	
	
	\node [coordinate, right of=PLANT,    node distance=1.8cm] (point1){};
	\node [coordinate, right of=point1,   node distance=1.7cm] (youtput){};
	\node [coordinate, above of=SUMtheta, node distance=1.0cm] (pointtheta){};
	\node [coordinate, below of=SUMxi,    node distance=1.0cm] (pointxi){};
	\node [coordinate, above of=SUM2theta, node distance=1.0cm] (point2theta){};
	\node [coordinate, below of=SUM2xi,    node distance=1.0cm] (point2xi){};
	
	\filldraw [black] (point1) circle [radius=1pt];
	\filldraw [black] (point2theta) circle [radius=1pt];
	\filldraw [black] (point2xi) circle [radius=1pt];
	
	\draw [->] (thetainput) -- node[pos=0.4] {\nm{\deltaTARGETE}} (SUMtheta.west);
	\draw [->] (xiinput)    -- node[pos=0.4] {\nm{\deltaTARGETA}} (SUMxi.west);

	\draw [->] (SUMtheta.east) -- (PIDSECE.west);
	\draw [->] (SUMxi.east)    -- (PIDSECA.west);
	
	\draw [->] (PIDSECE.east) -- node[pos=0.5] {\nm{\deltaTARGETE = \theta}} (SUM2theta.west);
	\draw [->] (PIDSECA.east) -- node[pos=0.5] {\nm{\deltaTARGETA = \xi}}    (SUM2xi.west);

	\draw [->] (SUM2theta.east) -- (PIDPRIME.west);
	\draw [->] (SUM2xi.east)    -- (PIDPRIMA.west);
		
	\draw [->] (PIDPRIME.east) -- node[pos=0.5] {\nm{\deltaE}} ($(PLANT.west) + (0cm,0.7cm)$);
	\draw [->] (PIDPRIMA.east) -- node[pos=0.5] {\nm{\deltaA}} ($(PLANT.west) - (0cm,0.7cm)$);
		
	\draw [->] (PLANT.east) -- node[pos=0.7] {\nm{\xvecest = \xEST}} (youtput);
	
	\draw [->] (point1) |- (pointtheta) -- (SUMtheta.north);
	\node [coordinate, left of=SUMtheta, node distance=0.5cm] (plustheta){};
	\node [above of=plustheta, node distance=0.2cm] (plussigntheta) {\nm{+}};	
	\node [coordinate, above of=SUMtheta, node distance=0.5cm] (minustheta){};
	\node [left of=minustheta, node distance=0.2cm] (minussigntheta) {\nm{-}};	
	
	\draw [->] (point1) |- (pointxi) -- (SUMxi.south);
	\node [coordinate, left of=SUMxi, node distance=0.5cm] (plusxi){};
	\node [below of=plusxi, node distance=0.2cm] (plussignxi) {\nm{+}};	
	\node [coordinate, below of=SUMxi, node distance=0.5cm] (minusxi){};
	\node [left of=minusxi, node distance=0.2cm] (minussignxi) {\nm{-}};	
	
	\draw [->] (point2theta) -- node[pos=0.5] {\nm{\thetaest}} (SUM2theta.north);
	\node [coordinate, left of=SUM2theta, node distance=0.5cm] (plus2theta){};
	\node [above of=plus2theta, node distance=0.2cm] (plussign2theta) {\nm{+}};	
	\node [coordinate, above of=SUM2theta, node distance=0.5cm] (minus2theta){};
	\node [left of=minus2theta, node distance=0.2cm] (minussign2theta) {\nm{-}};	
	
	\draw [->] (point2xi) -- (SUM2xi.south);
	\draw [-]  (SUM2xi.south) -- node[pos=0.5] {\nm{\xiest}} (point2xi); 
	\node [coordinate, left of=SUM2xi, node distance=0.5cm] (plus2xi){};
	\node [below of=plus2xi, node distance=0.2cm] (plussign2xi) {\nm{+}};	
	\node [coordinate, below of=SUM2xi, node distance=0.5cm] (minus2xi){};
	\node [left of=minus2xi, node distance=0.2cm] (minussign2xi) {\nm{-}};	
\end{tikzpicture}
\caption{Secondary control loops}
\label{fig:GNC_PID_flow_diagram_secondary}
\end{figure}
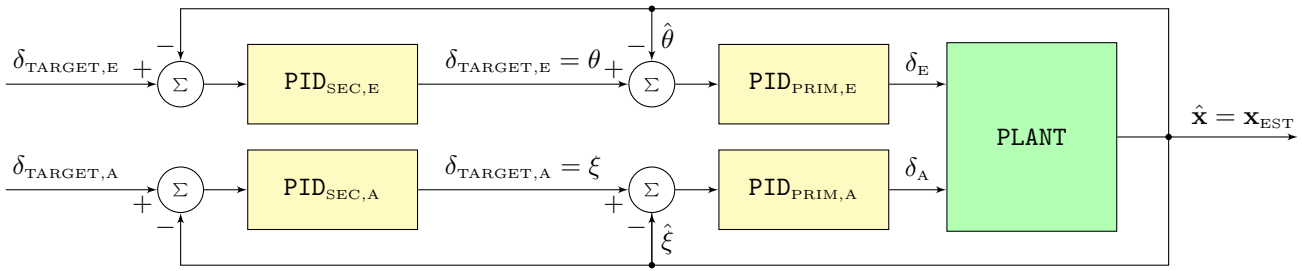

Secondary loops are also \hypertt{PID} loops, but in which the output is the target for the primary loop instead of the control variable \cite{Skogestad2005}, and hence the primary loop acts on a moving target provided by the secondary loop and not on a guidance target (figure \ref{fig:GNC_PID_flow_diagram_secondary}). For example, if in a given operation the guidance system indicates \nm{\deltaTARGETE = \lrb{\Hp = 2500 \ m}}, in each control cycle the secondary loop compares that value with the estimated pressure altitude \nm{\Hpest} and generates as output a given body pitch angle target \nm{\theta_{\sss TARGET}}. This new target is introduced into the primary loop and compared with the estimated body pitch angle \nm{\thetaest}, resulting in the position of the elevator \nm{\deltaE}. The secondary loops also implement a low pass filter applied exclusively to the derivative inputs.


\subsection{Control System Objective}\label{subsec:GNC_Control_Design}

The objective of the control system is to limit the flight technical error \hypertt{FTE} (section \ref{sec:Intro_traj}) or difference between the observed and reference states (sections \ref{sec:GNC_OT} and \ref{subsec:GNC_RT}) for each of the four active control variables. A properly working control loop requires that the \hypertt{FTE} or difference between the setpoint and the process variable (section \ref{subsec:GNC_Control_PID}) has an equal chance of being positive or negative, this is, it is unbiased or zero mean, so the standard deviation and maximum errors become the main metrics to evaluate its performance. A second necessary condition is for the performances of all control loops to be time independent or driftless, as if this is not the case, at some point the errors become big enough as to influence the flight stability, resulting in catastrophic platform loss. With regards to the maximum errors, they generally appear after a setpoint change and are caused by a system too slow to react or by overshooting the target. 


\section{Implementation, Validation, and Execution} \label{sec:gc_implementation}

The guidance and control systems described in this chapter have been implemented as an object oriented \texttt{C++} library. In the case of the guidance system, its implementation is relatively simple as it is based on a container of guidance targets and a method to evaluate the triggers that decide when an operation concludes and the following one is initiated. 

Modeling the control system has not only required the implementation of the primary and secondary loops described in section \ref{sec:GNC_Control}, but also the adjustment of the different gains to obtain an adequate performance. This has involved a continuous refinement process based on executing a series of trajectories as varied as possible, starting with straight constant speed level flights and slowly adding maneuvers (turns, speed changes, and altitude changes), first individually and then simultaneously. This process has been performed with various degrees of difficulty, from no turbulence, perfect sensors, and perfect navigation in open loop configuration, to the much more realistic conditions described in this document. Minor design changes and adjustments to the gains and low pass filters were executed at each step.

As depicted in figure \ref{fig:Trajectory_flow_diagram}, the reference trajectory \nm{\xREF} defined in section \ref{subsec:GNC_RT} is a necessary input to both the guidance system as well as the simulation. Its instructions and triggers can be defined in a deterministic way, so that the same \nm{\xREF} is employed in each of the Monte Carlo executions, or stochastically, this is, based on certain parameters generated pseudo randomly according to predefined rules\footnote{A simple example of a reference trajectory consisting of a single operation (\nm{\xREF = \lrb{\deltaTARGETone}}) would be the following (refer to section \ref{subsec:GNC_RT} for the notation): \begin{itemize}\item Stochastic \nm{\deltaTARGETT} in which \nm{\vTAS = N\lrp{29, \, 1.5^2} m/s}, this is, the constant air speed employed in each Monte Carlo run as the target of the throttle parameter \nm{\deltaT} is the realization of a normal distribution of mean \nm{29 \, m/s} and standard deviation \nm{1.5 \, m/s}. \item Stochastic \nm{\deltaTARGETE} in which \nm{\Hp = N\lrp{2700, \, 200^2} m}, this is, the constant pressure altitude employed in each Monte Carlo run as the target of the elevator \nm{\deltaE} is the realization of a normal distribution of mean \nm{2700 \, m} and standard deviation \nm{200 \, m}. \item Stochastic \nm{\deltaTARGETA} in which \nm{\chi = U\lrp{-179, \, 180} ^\circ}, this is, the constant bearing employed in each Monte Carlo execution as the target of the ailerons \nm{\deltaA} is the realization of a uniform distribution between \nm{-179 ^\circ} and \nm{180 ^\circ}. \item Deterministic \nm{\deltaTARGETR} in which \nm{\beta = 0 ^\circ}, this is, the target of the rudder \nm{\deltaR} is always to maintain symmetric flight. \item Stochastic \nm{\deltaTRG} in which \nm{\DeltatOP = N\lrp{300, \, 50^2} s}, this is, the duration of the operation varies according to a normal distribution of mean \nm{300 \, s} and standard deviation \nm{60 \, s}. \end{itemize}}. If this is the case, and similarly to the case of the Earth model (section \ref{sec:EarthModel_implementation}), sensor errors (section \ref{sec:Sensors_implementation}), and fine alignment results (section \ref{sec:PreFlight_implementation}), the guidance targets vary for each individual simulation run or execution \emph{j}, and hence depend on the flight seed \nm{\seedR} defined in section \ref{sec:Sensors_implementation}.

The flight seed \nm{\seedR}, which is the only input to all the stochastic variables and hence ensures that the results are repeatable if the same seed is employed again, is employed to initialize a discrete uniform distribution, which in turn is realized one extra time (in addition to the realizations required in sections \ref{sec:EarthModel_implementation}, \ref{sec:Sensors_implementation}, and \ref{sec:PreFlight_implementation}) to obtain the guidance seed \nm{\seedRGUID}. The guidance seed becomes the initialization seed for each of the stochastic parameters included in the definition of the reference trajectory.

 \cleardoublepage
\chapter{Navigation}\label{cha:nav}

The navigation system is first introduced in chapter \ref{cha:gc} in the context of the broader \hypertt{GNC} system. It takes the sensor outputs and processes them, providing the results to the guidance and control systems described in sections \ref{sec:GNC_Guidance} and \ref{sec:GNC_Control}. These outputs, called the observed trajectory and defined in section \ref{sec:GNC_OT}, can be considered as an improved version of the sensed trajectory (section \ref{sec:Sensors_ST}), as its differences or errors with respect to the actual aircraft trajectory (section \ref{sec:EquationsMotion_AT}) are smaller, and it contains variables such as aircraft attitude and position that are not present in the sensed trajectory.

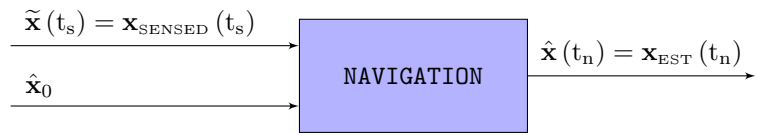
\begin{figure}[h]
\centering
\begin{tikzpicture}[auto, node distance=2cm,>=latex']
	\node [coordinate](midinput) {};
	\node [coordinate, above of=midinput, node distance=0.4cm] (xSENSEDinput){};
	\node [coordinate, below of=midinput, node distance=0.4cm] (xESTzeroinput){};
	\node [block, right of=midinput, minimum width=3.0cm, node distance=5.3cm, align=center, minimum height=1.5cm] (NAVIGATION) {\texttt{NAVIGATION}};
	\node [coordinate, right of=NAVIGATION, node distance=4.5cm] (output){};
		
	\draw [->] (xSENSEDinput) -- node[pos=0.45] {\nm{\xvectilde\lrp{t_s} = \xSENSED\lrp{t_s}}} ($(NAVIGATION.west)+(0cm,0.4cm)$);
	\draw [->] (xESTzeroinput) -- node[pos=0.1] {\nm{\xvecestzero}} ($(NAVIGATION.west)-(0cm,0.4cm)$);
	\draw [->] (NAVIGATION.east) -- node[pos=0.5] {\nm{\xvecest\lrp{t_n} = \xEST\lrp{t_n}}} (output);
\end{tikzpicture}
\caption{Navigation system flow diagram}
\label{fig:Navigation_flow_diagram}
\end{figure}

Figure \ref{fig:Navigation_flow_diagram} represents a partial view of figure \ref{fig:GNC_flow_diagram}, and shows how the navigation system, provided with the sensed state \nm{\xvectilde\lrp{t_s} = \xSENSED\lrp{t_s}} and an initial estimation of the observed state \nm{\xvecestzero}, provides the guidance and control systems with the observed or estimated state \nm{\xvecest\lrp{t_n} = \xEST\lrp{t_n}} defined in section \ref{sec:GNC_OT}. 

This chapter describes a standard \emph{inertial navigation system} (\hypertt{INS}) \cite{Farrell2008, Groves2008, Chatfield1997, Rogers2007, Kayton1997, Stevens2003, Titterton2004}, which makes use of the various sensors listed in chapter \ref{cha:Sensors} with the exception of the camera\footnote{As noted in chapter \ref{cha:Intro}, the simulation is designed so it can accept visual and visual inertial navigation algorithms that make use of the images obtained by the onboard camera, although these are not implemented.}. Section \ref{sec:GNC_Navigation_Complimentary} contains a detailed description of the two filters that constitute the \hypertt{INS}, followed by section \ref{sec:nav_iner_implementation} with some brief notes about their implementation in the simulation.

The objective of the navigation system is to diminish the \emph{navigation system error} \hypertt{NSE} (section \ref{sec:Intro_traj}) or difference between the observed and actual trajectories for all those variables required by the guidance and control systems (tables \ref{tab:GC_guidance_trigger_targets} and \ref{tab:GC_guidance_instruction_targets}), as well as the \nm{\vTASB} air velocity components (section \ref{sec:EquationsMotion_velocity}) required for flight envelope protection. The estimation of the aircraft attitude \nm{\qNBest} and position \nm{\TEgdtest} are of particular importance.


\section{Observed Trajectory}\label{sec:GNC_OT}

The \emph{observed} or \emph{estimated trajectory} (\hypertt{OT}) is the output of the \hypertt{INS} and represents the best knowledge that the \hypertt{GNC} system possesses about the actual trajectory described in section \ref{sec:EquationsMotion_AT}. First introduced in section \ref{sec:Intro_traj} and represented by \nm{\xvecest\lrp{t_n} = \xEST\lrp{t_n}}, where \nm{t_n = n \cdot \DeltatEST}, the observed trajectory contains the variation with time of the \emph{observed state}. It can be considered as an improved version of the sensed state described in section \ref{sec:Sensors_ST}, in the sense that it is both more complete as it includes variables that are not measured by the sensors, and more accurate as the filtering process reduces its difference with respect to the actual trajectory. As the output of the \hypertt{INS}, the observed state comprises the main input to the control system described in section \ref{sec:GNC_Control}. The differences between the actual and observed trajectories have a negative influence on the performance of the \hypertt{GNC} system, as the control system may react to observed deviations from the guidance trajectory that are not real (the observed trajectory deviates from the guidance targets but in fact the actual trajectory in on target), or may not react to real deviations that are not observed.
\begin{center}
\begin{tabular}{lcc}
	\hline
	Primary Components	& Variable		  & Section \\
	\hline
	Attitude		    & \nm{\qNBest}    & \ref{subsec:RefSystems_B}                  \\
	Geodetic position	& \nm{\TEgdtest}  & \ref{subsubsec:RefSystems_E_GeodeticCoord} \\	
	Pose				& \nm{\zetaEBest} & \ref{subsec:RefSystems_B}				   \\	
	\hline
\end{tabular}
\captionof{table}{Primary components of observed trajectory} \label{tab:GNC_Navigation_OT_primary}
\end{center}

The components of the \hypertt{OT} can be classified into primary, secondary, and auxiliary, according to their relevance to the \hypertt{INS}. The estimation of the aircraft pose constitutes the \hypertt{INS} primary objective, as listed in table \ref{tab:GNC_Navigation_OT_primary}. It can be represented in two equivalent ways, either as the geodetic coordinates plus an \nm{\mathbb{SO}(3)} representation of the body attitude, or as a complete \nm{\mathbb{SE}(3)} pose representation. If \nm{\zetaEBest} is employed, \cite{LIE} describes how to obtain the Cartesian coordinates \nm{\TEBEest = \TEcarest} from \nm{\zetaEBest}, leading to the geodetic coordinates \nm{\TEgdtest} and \nm{\qENest} determination of the \hypertt{NED} frame. The attitude can then be obtained per \nm{\qNBest = \qENest^{\ast} \otimes \qEBest}, where \nm{\qEBest} is also obtained from \nm{\zetaEBest} \cite{LIE}. If \nm{\qNBest} and \nm{\TEgdtest} are preferred, just reverse the process to obtain \nm{\zetaEBest}.
\begin{center}
\begin{tabular}{lcc}
	\hline
	Secondary Components	& Variable		  & Section \\
	\hline
	Angular velocity	& \nm{\wNBBest}   & \ref{sec:EquationsMotion_velocity} \\
	Ground velocity		& \nm{\vNest}     & \ref{sec:EquationsMotion_velocity} \\
	Twist				& \nm{\xiEBBest}  & \cite{LIE} \\
	\hline
\end{tabular}
\captionof{table}{Secondary components of observed trajectory} \label{tab:GNC_Navigation_OT_secondary}
\end{center}

The time derivatives of the primary variables constitute the secondary components (table \ref{tab:GNC_Navigation_OT_secondary}). There also exist two equivalent formats, either the Euclidean ground velocity combined with an \nm{\mathfrak{so}(3)} representation of the angular velocity, or as a complete \nm{\mathfrak{se}(3)} twist. It \nm{\xiEBBest} is employed, its second half constitutes the body ground velocity \nm{\nuEBBest = \vBest}, and \nm{\vNest} can be obtained by applying the \nm{\qNBest} rotation action. The twist first half directly provides \nm{\wEBBest}, which leads to \nm{\wNBBest} by subtracting the motion angular velocity \nm{\wENBest} (section \ref{sec:EquationsMotion_velocity}), which is known based on the other components. If \nm{\wNBBest} and \nm{\vNest} are preferred, just reverse the process to obtain \nm{\xiEBBest}.
\begin{center}
\begin{tabular}{lccllcc}
	\hline
	Auxiliary Components & Variable & Section & & Auxiliary Components & Variable & Section \\
	\hline
	\rule[1pt]{0pt}{12pt}Airspeed		     		& \nm{\vtasest}  	& \ref{sec:EquationsMotion_velocity}	& & Gyroscopes full error		& \nm{\EGYRest}		& \ref{subsec:Sensors_Inertial_ErrorModel} \\	
	Angle of attack          	& \nm{\alphaest}   	& \ref{subsec:RefSystems_W}             & & Accelerometers full error	& \nm{\EACCest}		& \ref{subsec:Sensors_Inertial_ErrorModel} \\
	Angle of sideslip			& \nm{\betaest}    	& \ref{subsec:RefSystems_W}				& & Magnetometers full error	& \nm{\EMAGest}		& \ref{subsec:Sensors_NonInertial_Magnetometers} \\
	Air temperature           	& \nm{\Test}       	& \ref{sec:EarthModel_ISA}              & & Magnetic field deviation	& \nm{\BNDEVest}	& \ref{sec:EarthModel_accuracy} \\
	Pressure altitude         	& \nm{\Hpest}      	& \ref{sec:EarthModel_ISA}              & & Specific force      		& \nm{\fIBBest}		& \ref{subsubsec:EquationsMotion_specific_force} \\
	Temperature offset        	& \nm{\DeltaTest}  	& \ref{sec:EarthModel_ISA}              & & Wind field				    & \nm{\vWINDNest}	& \ref{sec:EarthModel_WIND} \\
	Pressure offset             & \nm{\Deltapest}	& \ref{sec:EarthModel_ISA}              & & Mass                        & \nm{\mest}        & \ref{sec:AircraftModel_MassInertia} \\
	\hline
\end{tabular}
\captionof{table}{Auxiliary components of observed trajectory} \label{tab:GNC_Navigation_OT_auxiliary}
\end{center}

The remaining estimations of the \hypertt{INS}, listed in table \ref{tab:GNC_Navigation_OT_auxiliary}, constitute the auxiliary \hypertt{OT} variables.


\section{Inertial Navigation Filter}\label{sec:GNC_Navigation_Complimentary}

The proposed \hypertt{INS} is composed by two filters that together estimate all components of the observed or estimated state (section \ref{sec:GNC_OT}). These two filters, which run one after the other, are the air data filter (\nm{\first} to execute, section \ref{subsec:GNC_Navigation_AirDataFilter}), and the navigation filter (\nm{\second}, section \ref{subsec:nav_vis_iner_filter_so3_local}). 

Two filters are employed instead of a single one because those observed state components that describe the aircraft position and motion with respect to the air instead of the ground, such as the airspeed, the angles of attack and sideslip, the air temperature, and the pressure altitude, are independent from those components related to the Earth surface, such as the aircraft attitude, velocity, and position, in the sense that any equation containing variables in both groups also contains variables unknown to the \hypertt{INS}, such as the wind velocity or the pressure offset. The lack of relationships between these two groups of variables makes it more computationally efficient to create an independent smaller filter to estimate the air data variables.

This is best explained based on (\ref{eq:EquationsMotion_auxiliary_ground_speed1}), which states that the ground velocity \nm{\vvec} is the sum of the airspeed \nm{\vTAS} plus the wind field \nm{\vWIND} plus the turbulence \nm{\vTURB}. As there are no sensors capable of directly measuring neither the wind field nor the turbulence, (\ref{eq:EquationsMotion_auxiliary_ground_speed1}) can not be part of any filter, and can only be employed afterwards to obtain the merged wind plus turbulence speed (\nm{\vWIND + \vTURB}) once the ground and air velocities have been estimated.

The following sections include a detailed description of both filters, which have been implemented as \emph{extended Kalman filters} or \hypertt{EKF}s. Refer to \cite{LIE} for a detailed description of the \hypertt{EKF}, both when all state vector variables are Euclidean and when they also include Lie group elements. Given a system whose state can be described by a set of variables (state variables or state vector variables), first order continuous time nonlinear dynamics that model the variation with time of those variables, a second group of variables (observations or measurements) whose values are provided at discrete times, and a discrete time nonlinear observations system that links the observations with the state variables, the \hypertt{EKF} provides estimations of the mean and covariance of the state vector variables at every time instant at which measurements are provided. In this document, the measurements are available at a fixed frequency (\nm{t_n = n \cdot \DeltatEST}), and hence so are the state vector estimations.
\begin{center}
\begin{tabular}{cl}
	\hline
	Section & Objective \\
	\hline
	   				& Introduction of state and observation vectors. \\
	\nm{\first}		& Description of state dynamics and its linearization. \\
	\nm{\second}	& Explanation of observation system and its linearization. \\
	\nm{\third}		& Lie group element reset (only when not Euclidean). \\
	\nm{\fourth}	& Justification of employed process and observation noises. \\
	\nm{\fifth}		& Initialization of state variables and its covariances. \\
	\nm{\sixth}		& Extra steps to estimate other variables. \\
	\hline
\end{tabular}
\end{center}
\captionof{table}{Scheme employed for the description of each \texttt{EKF}} \label{tab:GNC_Navigation_Design_scheme}

The description of both filters follows the same script depicted in table \ref{tab:GNC_Navigation_Design_scheme}, with minor modifications based on whether the state vector is fully Euclidean or not. It starts by presenting the state variables (those to be estimated) and the observations or measurements on which the filter relies. The sampled data system describes the state vector dynamics, and includes the definition of the continuous time nonlinear state system and its linearization based on the system matrix \nm{\vec A\lrp{t}} and the process noise transform matrix \nm{\vec L\lrp{t}}. Note that although part of the linearized state system, the modified input vector \nm{\utilde \lrp{t}} is not part of the \hypertt{EKF} solution and hence is not computed. The sampled observations system describes the relationships between the measurement and state vectors, and includes the initial nonlinear system as well as its linearization based on the output matrix \nm{\vec H_n} and the measurement noise transform matrix \nm{\Mvec_n}. As above, the \nm{\vec z_n} input vector does not play any role in the solution and hence is not shown. 

The linearized state system relies on a zero mean continuous time white noise random process \nm{\wtilde\lrp{t}} that is discretized into \nm{\wtilde_n}. This discrete white noise is characterized by its covariance \nm{\Qtilde_{dn}}, which can be taken from the state system equations in the rare cases when these are exact, but that in general are based on a trade off between higher confidence in the noisy measurements (high covariance values, noisier state estimations, quick detection of variable changes) or more reliance on the state system equations (low covariance values, smoother results, slow detection of variable changes). Note that the non-diagonal terms of \nm{\Qtilde_{dn}} are zero as the noise components are considered independent. On the other hand, while the linearized observation system also relies on a zero mean discrete time white noise random process \nm{\vtilde_n}, its covariance \nm{\Rtilde_n} can generally be obtained from the sensor characteristics described in chapter \ref{cha:Sensors}. As in the case of the system noise, the non-diagonal terms of \nm{\Rtilde_n} are also zero.

The description of each \hypertt{EKF} concludes with its initialization based on the initial estimations of both the state variables \nm{\xvecest_0^+} and their covariances \nm{\Pvec_0^+}, as well as the obtainment of any other variables present in the observed trajectory that are not part of the filter state vector.


\subsection{Air Data Filter in Euclidean Space}\label{subsec:GNC_Navigation_AirDataFilter}

The objective of the air data filter is to provide an estimation of the variables contained in the state vector (\ref{eq:GNC_Navigation_AirDataFilter_st}) based on the periodic measurements present in the observations vector (\ref{eq:GNC_Navigation_AirDataFilter_y}).

\begin{figure}[h]
\centering
\begin{tikzpicture}[auto, node distance=2cm,>=latex']

	\node [coordinate](midinput) {};
	\pgfdeclarelayer{background};
	\node [rectangle, draw=black!50, fill=cyan!25, dashed, rounded corners, right of=midinput, node distance=4.0cm, minimum width=8.2cm, minimum height=2.4cm] (EUCLIDEAN) {};
	\pgfdeclarelayer{foreground};

	\coordinate (center) at ($(midinput) + (+0.0cm,-0.3cm)$);	
	\node [rectangle, draw=black!50, fill=pink!80, right of=center, minimum width=2.0cm, node distance=3.5cm, align=center, minimum height=1.5cm] (FILTER) {\texttt{Air Data} \\ \texttt{Filter}};
	\node [rectangle, draw=black!50, fill=pink!80, left of=center, dashed, minimum width=2.0cm, node distance=1.5cm, align=center, minimum height=1.5cm] (SENSORS) {\hypertt{TAS}-\hypertt{AOA}-\hypertt{AOS} \\ \hypertt{OAT}-\hypertt{OSP}};
	\draw [->] (SENSORS.east)  -- node[pos=0.55] {\nm{\xvectilde = \xSENSED}} (FILTER.west);

	\draw [->] ($(FILTER.east)+(+0.0cm,+0.4cm)$) -- node[pos=0.50] {\nm{\vtasest, \, \alphaest, \, \betaest, \, \Test, \, \Hpest}} ($(FILTER.east)+(+4.0cm,+0.4cm)$);
	\coordinate (ref1) at ($(FILTER.east) + (0.6cm,0.4cm)$);	
	\filldraw [black] (ref1) circle [radius=1pt];
	\coordinate (ref2) at ($(FILTER.east) + (1.8cm,0.4cm)$);	
	\node [rectangle, draw=black!50, fill=pink!80, below of=ref2, minimum width=0.8cm, node distance=0.7cm, align=center, minimum height=0.7cm] (EXTRA) {(\ref{eq:GNC_Navigation_AirDataFilter_DeltaT})};
	\draw [->] (ref1) |- (EXTRA.west);
	\draw [->] (EXTRA.east) -- node[pos=0.40] {\nm{\DeltaTest}} ($(EXTRA.east)+(+1.6cm,+0.0cm)$);
	
	\node at ($(EUCLIDEAN.east) +(-1.3cm,0.95cm)$) {\texttt{Euclidean} \hypertt{EKF}};
	
\end{tikzpicture}
\caption{Air data filter flow diagram}
\label{fig:filter_air_data_flow_diagram}
\end{figure}
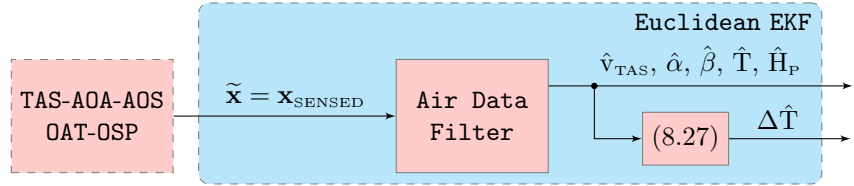

\begin{eqnarray}
\nm{\xvec\lrp{t}} & = & \nm{\lrsb{\vtas \ \ \alpha \ \ \beta \ \ T \ \ \Hp \ \ ROC \ \ \vtasdot \ \ \alphadot \ \ \betadot}^T \hspace{0.2cm} \in \mathbb{R}^9\lrp{t}}\label{eq:GNC_Navigation_AirDataFilter_st} \\
\nm{\yvec_n} & = & \nm{\lrsb{\vtastilde \ \ \alphatilde \ \ \betatilde \ \ \Ttilde \ \ \ptilde}^T \hspace{0.2cm} \in {\mathbb{R}^5}_n}\label{eq:GNC_Navigation_AirDataFilter_y}
\end{eqnarray}

All measurements are provided by the air data system (section \ref{subsec:Sensors_NonInertial_ADS}), and \emph{rate of climb} (\hypertt{ROC}) is the time derivative of the pressure altitude. The measurements \nm{\yvec_n = \yvec\lrp{t_n}} are equispaced per \nm{t_n = n \cdot \DeltatEST}, as described in table \ref{tab:Intro_trajectory_frequencies}. All (\ref{eq:GNC_Navigation_AirDataFilter_st}) components are Euclidean, so the classical \hypertt{EKF} scheme applies \cite{LIE}. Note that the filter does not attempt to estimate air data system biases \nm{\lrb{\BzeroTAS, \, \BzeroAOA, \, \BzeroAOS, \, \BzeroOAT, \ \BzeroOSP}}, as there is not enough information (observations) to do so; these. The biases errors hence accumulate into the estimated variables.


\subsubsection{Air Data State System}\label{subsubsec:GNC_Navigation_AirDataFilter_statesystem}

The air data filter state equations can be divided into two groups: those in which the derivative with time of a certain state variable coincides with a different variable, and those in which the \hypertt{INS} does not posses any knowledge about the dynamics of the state variable, so its derivative with time is set to zero\footnote{Note that although present, system noise \nm{\wvec\lrp{t}} is not shown in equations (\ref{eq:GNC_Navigation_AirDataFilter_vtasdot}) through (\ref{eq:GNC_Navigation_AirDataFilter_otherdot}).}. The system noise described in section \ref{subsubsec:GNC_Navigation_AirDataFilter_whitenoises_initialization} specifies how much the estimated values of the state variables should adhere to the noisy measurements. The following linear continuous time state system can hence be established:
\begin{eqnarray}
\nm{\xvecdot\lrp{t}} & = & \nm{\vec f_{AIR}\big(\xvec\lrp{t}, \uvec\lrp{t}, \wvec\lrp{t}, t\big)} \label{eq:GNC_Navigation_AirDataFilter_xdot} \\
\nm{\dot {\vec x}_1} & = & \nm{\vtasdot  = \xvec_7}           \label{eq:GNC_Navigation_AirDataFilter_vtasdot} \\
\nm{\dot {\vec x}_2} & = & \nm{\alphadot = \xvec_8}           \label{eq:GNC_Navigation_AirDataFilter_alphadot} \\
\nm{\dot {\vec x}_3} & = & \nm{\betadot  = \xvec_9}           \label{eq:GNC_Navigation_AirDataFilter_betadot} \\
\nm{\dot {\vec x}_5} & = & \nm{\Hpdot    = \xvec_6 = ROC}     \label{eq:GNC_Navigation_AirDataFilter_Hpdot} \\
\nm{\dot {\vec x}_{OTHER}} & = & \nm{\vec 0_5}                  \label{eq:GNC_Navigation_AirDataFilter_otherdot} \\
\nm{\uvec\lrp{t}}             & = & \nm{\vec 0}               \label{eq:GNC_Navigation_AirDataFilter_u}
\end{eqnarray}

The linear state system above can also be written in the following form:
\begin{eqnarray}
\nm{\xvecdot\lrp{t}} & = & \nm{\Avec\lrp{t} \, \xvec\lrp{t} + \Bvec\lrp{t} \, \utilde\lrp{t} + \wtilde\lrp{t}} \label{eq:GNC_Navigation_AirDataFilter_stdot} \\
\nm{\vec A \lrp{t}}  & = & \nm{\begin{bmatrix} \nm{\vec O_{3x5}} & \nm{\vec O_{3x1}} & \nm{\vec I_{3x3}} \\
                                               \nm{\vec O_{1x5}} & \nm{0}            & \nm{\vec O_{1x3}} \\
											   \nm{\vec O_{1x5}} & \nm{1}            & \nm{\vec O_{1x3}} \\
											   \nm{\vec O_{4x5}} & \nm{\vec O_{4x1}} & \nm{\vec O_{4x3}} \end{bmatrix} \hspace{0.2cm} \in \mathbb{R}^{9x9}\lrp{t}} \label{eq:GNC_Navigation_AirDataFilter_A} \\
\nm{\utilde\lrp{t}} & = & \nm{\vec 0} \label{eq:GNC_Navigation_AirDataFilter_utilde}
\end{eqnarray}

Because of the lack of knowledge about the system dynamics, the only purpose of the air data filter is to smooth the estimations by reducing the noise present in the measurements. Although it could have been implemented as a low pass filter \cite{Haugen2008}, the author has preferred to employ an \hypertt{EKF} for commonality with the navigation filter (section \ref{subsec:nav_vis_iner_filter_so3_local}).


\subsubsection{Air Data Observations}\label{subsubsec:GNC_Navigation_AirDataFilter_observations}

The relationship between the observation and state vectors is not linear because of the ratio between pressure and pressure altitude (\ref{eq:GNC_Navigation_AirDataFilter_ptilde}), which coincides with (\ref{eq:EarthModel_ISA_p_Hp}), but is nevertheless quite simple as the other equations are straightforward\footnote{Note that although present,  measurement noise \nm{\vvec_n} is not shown in equations (\ref{eq:GNC_Navigation_AirDataFilter_vtastilde}) through (\ref{eq:GNC_Navigation_AirDataFilter_ptilde}).}:
\begin{eqnarray}
\nm{\yvec_n}     & = & \nm{\vec h_{AIR}\lrp{\xvec_n, \, \vvec_n, \, t_n}}\label{eq:GNC_Navigation_AirDataFilter_ynonlinear} \\
\nm{\vtastilde}  & = & \nm{\vtas}\label{eq:GNC_Navigation_AirDataFilter_vtastilde} \\
\nm{\alphatilde} & = & \nm{\alpha}\label{eq:GNC_Navigation_AirDataFilter_alphatilde} \\
\nm{\betatilde}  & = & \nm{\beta}\label{eq:GNC_Navigation_AirDataFilter_betatilde} \\
\nm{\Ttilde}     & = & \nm{T}\label{eq:GNC_Navigation_AirDataFilter_Ttilde} \\
\nm{\ptilde}     & = & \nm{\pzero \ \lrp{1 + \frac{\betaT}{\Tzero} \ \Hp}^{\sss \gBR}}\label{eq:GNC_Navigation_AirDataFilter_ptilde}
\end{eqnarray}

The above system is linearized resulting in a discrete time linear observation system:
\begin{eqnarray}
\nm{\yvec_n} & \nm{\approx} & \nm{\Hvec_n \, \xvec_n + \zvec_n + \vtilde_n}\label{eq:GNC_Navigation_AirDataFilter_obser} \\
\nm{\Hvec_n} & = & \nm{\begin{bmatrix} \nm{\vec I_{4x4}} & \nm{\vec O_{4x1}} & \nm{\vec O_{4x4}} \\ 
                                       \nm{\vec O_{1x4}} & \nm{\frac{- \gzero \, p}{R \ \lrp{\Tzero + \betaT \, \Hp}}} & \nm{\vec O_{1x4}} \end{bmatrix} \hspace{0.2cm} \in {\mathbb{R}^{5x9}}_n} \label{eq:GNC_Navigation_AirDataFilter_H}
\end{eqnarray}

where the output matrix \nm{\Hvec_n} relies on the ratio between \nm{\Hp} and \emph{p} provided by (\ref{eq:EarthModel_ISA_static_equilibrium4}), and it is not necessary to evaluate the observations input vector \nm{\zvec_n} as it plays no role in the solution.


\subsubsection{Air Data White Noises and Initialization}\label{subsubsec:GNC_Navigation_AirDataFilter_whitenoises_initialization}

The covariances of the process noise \nm{\wtilde_n} are assigned by the author with the objective of determining the weight of the dynamic equations versus the observations when estimating the value of the state vector variables. Smaller values are assigned to those state equations that are more exact, such as (\ref{eq:GNC_Navigation_AirDataFilter_Hpdot}), while higher values are applied where there are more inaccuracies, as is the case for the remaining state variables in section \ref{subsubsec:GNC_Navigation_AirDataFilter_statesystem}, in which the time derivative is set to zero as the \hypertt{INS} has no knowledge about the dynamics:
\begin{eqnarray}
\nm{\Lvec} & = & \nm{\vec I_{9x9}} \label{eq:GNC_Navigation_AirDataFilter_L} \\
\nm{Diag \ \Qtilde_d} & = & \nm{\lrsb{10^{-15} \ \ 10^{-15} \ \ 10^{-15} \ \ 10^{-4} \ \ 10^{-15} \ \ 10^{-3} \ \ 10^{-4} \ \ 10^{-6} \ \ 10^{-6}}^T \hspace{0.2cm} \in \mathbb{R}^{9}} \label{eq:GNC_Navigation_AirDataFilter_Qtilde}
\end{eqnarray}

where \nm{Diag\lrp{}} stands for diagonal of a matrix. In the case of the measurement noise \nm{\vvec_n}, its covariances coincide with the system noise introduced by the different air data sensors (section \ref{subsec:Sensors_NonInertial_ADS}). Note that as both the system and measurement noises are applied to the same variables contained in the (\ref{eq:GNC_Navigation_AirDataFilter_st}) and (\ref{eq:GNC_Navigation_AirDataFilter_y}) vectors, the noise transform matrices \nm{\Lvec} and \nm{\Mvec} are unitary.
\begin{eqnarray}
\nm{\Mvec} & = & \nm{\vec I_{5x5}} \label{eq:GNC_Navigation_AirDataFilter_M} \\
\nm{Diag \ \Rtilde} & = & \nm{\lrsb{\sigma_{\sss TAS}^2 \ \ \sigma_{\sss AOA}^2 \ \ \sigma_{\sss AOS}^2 \ \ \sigma_{\sss OAT}^2 \ \ \sigma_{\sss OSP}^2}^T \hspace{0.2cm} \in \mathbb{R}^{5}}\label {eq:GNC_Navigation_AirDataFilter_Rtilde}
\end{eqnarray}

For initialization purposes, the air data filter employs the initial measurements as initial estimations \nm{\xvecest_0^+}, and uses the system noise and bias offset of the air data sensors (section \ref{subsec:Sensors_NonInertial_ADS}) for the covariances of the initial estimations \nm{\Pvec_0^+}. Equations (\ref{eq:EarthModel_ISA_p_Hp2}) and (\ref{eq:EarthModel_ISA_Hp_p}) are used to obtain the covariance of the pressure altitude, and zero is employed when there is no data available, as is the case with the rate of climb and other time derivatives.
\begin{eqnarray}
\nm{\xvecest_0^+} & = & \nm{\lrsb{{\widetilde{v}_{{\sss TAS},0}} \ \ {\widetilde{\alpha}_0} \ \ {\widetilde{\beta}_0} \ \ {\widetilde{T}_0} \ \ \Hp\lrp{{\widetilde{p}_0}} \ \ 0 \ \ 0 \ \ 0 \ \ 0}^T \hspace{0.2cm} \in \mathbb{R}^9}\label{eq:GNC_Navigation_AirDataFilter_stestinit} \\
\nm{Diag \ \Pvec_0^+} & = & \nm{\begin{bmatrix} \nm{\sigma_{\sss TAS}^2 + B_{0, \sss TAS}^2} \\\nm{\sigma_{\sss AOA}^2 + B_{0,\sss AOA}^2} \\ \nm{\sigma_{\sss AOS}^2 + B_{0,\sss AOS}^2} \\ \nm{\sigma_{\sss OAT}^2 + B_{0, \sss OAT}^2} \\ \nm{   \lrsb{\derpar{\Hp}{p}\big(p, \Hp\lrp{p}\big)}^2 \, \lrp{\sigma_{\sss OSP}^2 + B_{0,\sss OSP}^2}    } \\ \nm{\vec O_{4x1}} \end{bmatrix} \hspace{0.2cm} \in \mathbb{R}^{9}}\label{eq:GNC_Navigation_AirDataFilter_Pestinit} 
\end{eqnarray}


\subsubsection{Air Data Extra Steps}\label{subsubsec:GNC_Navigation_AirDataFilter_extra}

Once the air data filter step has concluded, the temperature offset \nm{\DeltaT} is estimated based on (\ref{eq:EarthModel_ISA_T_Hp}):
\neweq{\DeltaTest = \Test - \Tzero - \betaT \ \Hpest}{eq:GNC_Navigation_AirDataFilter_DeltaT}


\subsection{Navigation Filter in Local Manifold of Rigid Body Rotations}\label{subsec:nav_vis_iner_filter_so3_local}

The objective of the navigation filter is to provide an estimation of \nm{\qNB} together with the variables contained in the (\ref{eq:nav_vis_iner_filter_st_so3_local}) state vector based on the periodic measurements provided by the (\ref{eq:nav_vis_iner_filter_y_so3_local}) observations vector. As the body attitude \nm{\qNB} representing \nm{\mathbb{SO}(3)} is not Euclidean, the classical \hypertt{EKF} scheme is modified ensuring that \nm{\qNB} is updated by means of local (body) perturbations represented by \nm{\DeltarNBB} and hence never deviates from its \nm{\mathbb{SO}(3)} manifold \cite{LIE}.
\begin{eqnarray}
\nm{\mathcal R\lrp{t}} & = & \nm{\qNB \hspace{0.2cm} \in \mathbb{SO}(3)\lrp{t}} \label{eq:nav_vis_iner_filter_X_so3_local} \\
\nm{\xvec\lrp{t}}      & = & \nm{\lrsb{\DeltarNBB \ \ \wNBB \ \ \zvec}^T} \nonumber \\
                       & = & \nm{\lrsb{\DeltarNBB \ \ \wNBB \ \ \TEgdt \ \ \vN \ \ \fIBB \ \ \EGYR \ \ \EACC \ \ \EMAG \ \ \BNDEV}^T \hspace{0.2cm} \in \mathbb{R}^{27}\lrp{t}}\label{eq:nav_vis_iner_filter_st_so3_local} \\
\nm{\yvec_n}           & = & \nm{\lrsb{\wIBBtilde \ \ \fIBBtilde \ \ \BBtilde \ \ \TEgdttilde \ \ \vNtilde}^T \hspace{0.2cm} \in {\mathbb{R}^{15}}_n}\label{eq:nav_vis_iner_filter_y_so3_local}
\end{eqnarray}

Note that \nm{\DeltarNBB} contains the body perturbations to \nm{\qNB}, and the measurements are provided at different frequencies: the inertial angular velocity, specific force, and magnetic field readings provided by the gyroscopes (section \ref{subsec:Sensors_Inertial_ErrorModel}), accelerometers (section \ref{subsec:Sensors_Inertial_ErrorModel}), and magnetometers (section \ref{subsec:Sensors_NonInertial_Magnetometers}), are available every \nm{\DeltatSENSED = 0.01 \ s}, while the ground velocity and position measurements provided by the \hypertt{GNSS} receiver are generated every \nm{\DeltatGNSS = 1 \ s} (section \ref{subsec:Sensors_NonInertial_GNSS}).


\subsubsection{Navigation Local \nm{\mathbb{SO}(3)} State System}\label{subsubsec:nav_vis_iner_filter_statesystem_so3_local}

The variation with time of the state variables comprises a continuous time nonlinear system\footnote{Note that although present, system noise \nm{\wvec\lrp{t}} is not shown in equations (\ref{eq:nav_vis_iner_filter_DeltarNBBdot_so3_local}) through (\ref{eq:nav_vis_iner_filter_other_so3_local}).} composed by three exact equations (with no simplifications) containing the rotational motion time derivative \cite{LIE}, the kinematics of the geodetic coordinates (\ref{eq:EquationsMotion_equations_kinematic}), as well as the aircraft velocity dynamics (\ref{eq:EquationsMotion_equations_force8_N}), together with six other differential equations in which the \hypertt{INS} does not posses any knowledge about the time derivatives of the state variables, which are set to zero: 
\begin{eqnarray}
\nm{\xvecdot\lrp{t}} & = & \nm{\lrsb{\DeltarNBBdot \ \ \wNBBdot \ \ \vec {\dot z}}^T} \nonumber \\
                     & = & \nm{\vec f_{LSO3}\big(\qNB\lrp{t} \oplus \DeltarNBB\lrp{t}, \, \wNBB\lrp{t}, \, \vec z\lrp{t}, \, \uvec\lrp{t}, \ \wvec\lrp{t}, \, t\big)} \label{eq:nav_vis_iner_cont_time_system_so3_local} \\
\nm{\vec {\dot x}_{1-3}}   & = & \nm{\DeltarNBBdot = \wNBB = {\vec x}_{4-6}} \label{eq:nav_vis_iner_filter_DeltarNBBdot_so3_local} \\  
\nm{\vec {\dot x}_{7-9}}   & = & \nm{\TEgdtdot = \lrsb{\dfrac{\vNii}{\lrsb{N\lrp{\varphi} + h} \, \cos\varphi} \ \ \ \ \dfrac{\vNi}{M\lrp{\varphi} + h} \ \ \ \ - \vNiii}^T} \label{eq:nav_vis_iner_filter_TEgdtdot_so3_local} \\
\nm{\vec {\dot x}_{10-12}} & = & \nm{\vNdot = \vec g_{{\ds{\qNB \oplus \DeltarNBB}}*}\lrp{\fIBB} - \wENNskew \; \vN + \gcNMODEL - \acorN} \label{eq:nav_vis_iner_filter_vNdot_so3_local} \\  
\nm{\vec {\dot x}_{OTHER}} & = & \nm{\lrsb{\wNBBdot \ \ \fIBBdot \ \ \EGYRdot \ \ \EACCdot \ \ \EMAGdot \ \ \BNDEVdot}^T = \vec{0}_{18}} \label{eq:nav_vis_iner_filter_other_so3_local} \\
\nm{\uvec\lrp{t}}	       & = & \nm{\vec 0} \label{eq:nav_vis_iner_filter_u_so3_local}
\end{eqnarray}

\begin{figure}[h]
\centering
\begin{tikzpicture}[auto, node distance=2cm,>=latex']

	\node [coordinate](midinput) {};
	\filldraw [black] (midinput) circle [radius=1pt];
	\pgfdeclarelayer{background};
	\node [rectangle, draw=black!50, fill=cyan!30, dashed, rounded corners, right of=midinput, node distance=5.3cm, minimum width=10.3cm, minimum height=2.8cm] (SO3) {};
	\pgfdeclarelayer{foreground};

	\coordinate (centerright) at ($(midinput) + (+0.4cm,-0.0cm)$);	
	\coordinate (centerleft) at ($(midinput)  + (-1.8cm,+0.0cm)$);
		
	\node [rectangle, draw=black!50, fill=pink!80, right of=centerright, minimum width=2.0cm, node distance=2.8cm, align=center, minimum height=1.5cm] (FILTER) {\texttt{Navigation} \\ \texttt{Filter}};
	\node [rectangle, draw=black!50, fill=pink!80, above of=centerleft, dashed, minimum width=2.3cm, node distance=0.5cm, align=center, minimum height=0.8cm] (SENSORS1) {\hypertt{GYR}-\hypertt{ACC}-\hypertt{MAG}};
	\node [rectangle, draw=black!50, fill=pink!80, below of=centerleft, dashed, minimum width=2.3cm, node distance=0.5cm, align=center, minimum height=0.8cm] (SENSORS2) {\hypertt{GNSS}};
			
	\draw [->] (SENSORS1.east) -| (midinput) --  node[pos=0.55] {\nm{\xvectilde = \xSENSED}}  (FILTER.west);
	\draw (SENSORS2.east) -| (midinput);

	\coordinate (pright) at ($(FILTER.east) + (+0.5cm,0.0cm)$);	
	\filldraw [black] (pright) circle [radius=1pt];
	\coordinate (topright) at ($(pright) + (+6.0cm,+0.7cm)$);	
	\coordinate (botright) at ($(pright) + (+6.0cm,-0.7cm)$);	
		
	\draw [->] (FILTER.east) -- (pright) |- node[pos=0.77] {\nm{\qNBest, \, \wNBBest, \, \TEgdtest, \, \vNest, \, \fIBBest}} (topright);
	\node at ($(topright) + (-2.8cm,-0.4cm)$) {\nm{\EGYRest, \, \EACCest, \, \EMAGest, \, \BNDEVest}};
	
	\node [rectangle, draw=black!50, fill=pink!80, left of=botright, minimum width=2.0cm, node distance=4.5cm, align=center, minimum height=0.7cm] (AUX) {\texttt{Sec} \ref{subsubsec:nav_vis_iner_filter_extra_so3_local}};
	
	\draw (pright) |- (AUX.west);
	
	\draw [->] (AUX.east) --  node[pos=0.45] {\nm{\mest, \, \Deltapest, \, \vWINDNest}}  (botright);
	\node at ($(botright) +(-1.5cm,-0.4cm)$) {\nm{\zetaEBest, \, \xiEBBest}};
	
	\node at ($(FILTER.west) +(-0.50cm,1.10cm)$) {\texttt{Local} \nm{\mathbb{SO}(3)} \hypertt{EKF}};
		
\end{tikzpicture}
\caption{Local \nm{\mathbb{SO}(3)} navigation filter flow diagram}
\label{fig:filter_navigation_local_so3_flow_diagram}
\end{figure}

Its linearization results in the following system matrix \nm{\vec A\lrp{t} \in \mathbb{R}^{27x27}}, which makes use of the motion angular velocity, geodetic coordinates, and Coriolis acceleration Jacobians with respect to the \hypertt{NED} velocity (\ref{eq:EquationsMotion_jac_wENN_vN}, \ref{eq:EquationsMotion_jac_TEgdt_vN}, \ref{eq:EquationsMotion_jac_acorN_vN}), as well as the Jacobians of the \nm{\mathbb{SO}(3)} vector rotation action \nm{\vec g_{\mathcal R*}\lrp{\vec v}} listed in \cite{LIE}. Note that the linearization neglects the influence on the geodetic coordinates \nm{\TEgdt} of four different terms (\nm{\TEgdtdot}, \nm{\wENN}, \nm{\gcNMODEL}, and \nm{\acorN}) present in the (\ref{eq:nav_vis_iner_cont_time_system_so3_local}) differential equations, but includes the influence on the aircraft velocity \nm{\vN} of three of them (gravity does not depend on velocity). This is because the \nm{\vN} variation amount that can be experienced in a single state estimation step can have a significant influence on the values of the variables considered, while that of the geodetic coordinates \nm{\TEgdt} does not and hence can be neglected.
\begin{eqnarray}
\nm{\xvecdot\lrp{t}} & = & \nm{\Avec\lrp{t} \, \xvec\lrp{t} + \Bvec\lrp{t} \, \utilde\lrp{t} + \wtilde\lrp{t}} \label{eq:nav_vis_iner_filter_stdot_so3_local} \\
\nm{\vec A_{1-3,4-6}}     & = & \nm{\derpar{\DeltarNBBdot}{\wNBB} = \vec I_{3x3}} \label{eq:nav_vis_iner_filter_A_DeltarNBBdot_wNBB_so3_local} \\
\nm{\vec A_{7-9,10-12}}   & = & \nm{\derpar{\TEgdtdot}{\vN} = \vec J_{\ds{\; \vN}}^{\ds{\; \vec {\dot{T}}^{\sss E,GDT}}}} \label{eq:nav_vis_iner_filter_A_TEgdt_vN_so3_local} \\ 
\nm{\vec A_{10-12,1-3}}   & = & \nm{\derpar{\vNdot}{\DeltarNBB} = \vec J_{\ds{\oplus \; \mathcal R}}^{\ds{- \; \vec g_{\qNB*}(\fIBB)}}} \label{eq:nav_vis_iner_filter_A_vNdot_DeltarNBB_so3_local} \\
\nm{\vec A_{10-12,10-12}} & = & \nm{\derpar{\vNdot}{\vN} = - \wENNskew + \vNskew \, \vec J_{\ds{\; \vN}}^{\ds{\; \wENN}} - \vec J_{\ds{\; \vN}}^{\ds{\; \acorN}}} \label{eq:nav_vis_iner_filter_A_vNdot_vN_so3_local} \\
\nm{\vec A_{10-12,13-15}} & = & \nm{\derpar{\vNdot}{\fIBB} = \vec J_{\ds{+ \; \vec v}}^{\ds{- \; \vec g_{\qNB \oplus \DeltarNBB*}(\fIBB)}}} \label{eq:nav_vis_iner_filter_A_vNdot_fIBB_so3_local} \\
\nm{A_{ij,OTHER}}         & = & \nm{0} \label{eq:nav_vis_iner_filter_A_other_so3_local} \\
\nm{\utilde\lrp{t}}       & = & \nm{\vec 0} \label{eq:nav_vis_iner_filter_utilde_so3_local}
\end{eqnarray}


\subsubsection{Navigation Local \nm{\mathbb{SO}(3)} Observations}\label{subsubsec:nav_vis_iner_filter_observations_so3_local}

The (\ref{eq:nav_vis_iner_filter_yn_so3_local}) discrete time nonlinear observations system\footnote{Note that although present, measurement noise \nm{\vvec_n} is not shown in equations (\ref{eq:nav_vis_iner_filter_wIBBtilde_so3_local}) through (\ref{eq:nav_vis_iner_filter_vNtilde_so3_local}).} is based on the measurements provided by the gyroscopes, accelerometers, magnetometers, and \hypertt{GNSS} receiver, respectively, and contains no simplifications.
\begin{eqnarray}
\nm{\yvec_n}     & = & \nm{\vec h_{LSO3}\lrp{\vec q_{{\sss NB}n} \oplus \Delta \vec r_{{\sss NB}n}^{\sss B}, \, \vec \omega_{{\sss NB}n}^{\sss B}, \, \vec z_n, \, \, \vvec_n, \, t_n}}\label{eq:nav_vis_iner_filter_yn_so3_local} \\
\nm{\wIBBtilde}  & = & \nm{\wNBB + \vec {Ad}_{{\ds{\vec q_{\sss NB} \oplus \DeltarNBB}}}^{-1} \lrp{\wIEN + \wENN} + \EGYR}\label{eq:nav_vis_iner_filter_wIBBtilde_so3_local} \\
\nm{\fIBBtilde}  & = & \nm{\fIBB + \EACC} \label{eq:nav_vis_iner_filter_fIBBtilde_so3_local} \\
\nm{\BBtilde}    & = & \nm{\vec g_{{\ds{\vec q_{\sss NB} \oplus \DeltarNBB}}*}^{-1} \lrp{\BNMODEL - \BNDEV} + \EMAG} \label{eq:nav_vis_iner_filter_BBtilde_so3_local} \\
\nm{\TEgdttilde} & = & \nm{\TEgdt} \label{eq:nav_vis_iner_filter_TEgdttilde_so3_local} \\
\nm{\vNtilde}    & = & \nm{\vN} \label{eq:nav_vis_iner_filter_vNtilde_so3_local}
\end{eqnarray}

Its linearization results in the following output matrix \nm{\vec H_n \in \mathbb{R}^{15x27}}, which makes use of the motion angular velocity Jacobian with respect to the \hypertt{NED} velocity (\ref{eq:EquationsMotion_jac_wENN_vN}), as well as the Jacobians of the \nm{\mathbb{SO}(3)} inverse vector rotation action \nm{\vec g_{\mathcal R*}^{-1}\lrp{\vec v}} and inverse adjoint action \nm{\vec {Ad}_{\mathcal R}^{-1}\lrp{\vec w}} listed in \cite{LIE}. As in the state system case, the linearization also neglects the influence on the geodetic coordinates \nm{\TEgdt} of three different terms (\nm{\wIEN}, \nm{\wENN}, and \nm{\BNMODEL}) present in the (\ref{eq:nav_vis_iner_filter_yn_so3_local}) observation equations. Note that it is not necessary to evaluate the observations input vector \nm{\zvec_n} as it plays no role in the solution.
\begin{eqnarray}
\nm{\yvec_n} & \nm{\approx} & \nm{\Hvec_n \, \xvec_n + \zvec_n + \vtilde_n}\label{eq:nav_vis_iner_filter_obser_so3_local} \\
\nm{\vec H_{1-3,1-3}}	  & = & \nm{\derpar{\wIBBtilde}{\DeltarNBB} = \vec J_{\ds{\oplus \; \mathcal R}}^{\ds{- \; \vec {Ad}_{\qNB}^{-1}(\wIEN + \wENN)}}} \label{eq:nav_vis_iner_filter_H_wIBBtilde_DeltarNBB_so3_local} \\
\nm{\vec H_{1-3,4-6}}	  & = & \nm{\derpar{\wIBBtilde}{\wNBB} = \vec{I}_{3x3}} \label{eq:nav_vis_iner_filter_H_wIBBtilde_wNBB_so3_local} \\
\nm{\vec H_{1-3,10-12}}	  & = & \nm{\derpar{\wIBBtilde}{\vN} = \vec J_{\ds{+ \; \vec \omega}}^{\ds{- \; \vec {Ad}_{\qNB \oplus \DeltarNBB}^{-1}(\wIEN + \wENN)}} \, \vec J_{\ds{\; \vN}}^{\ds{\; \wENN}}} \label{eq:nav_vis_iner_filter_H_wIBBtilde_vN_so3_local} \\
\nm{\vec H_{1-3,16-18}}	  & = & \nm{\derpar{\wIBBtilde}{\EGYR} = \vec{I}_{3x3}} \label{eq:nav_vis_iner_filter_H_wIBBtilde_EGYR_so3_local} \\
\nm{\vec H_{4-6,13-15}}   & = & \nm{\derpar{\fIBBtilde}{\fIBB} = \vec{I}_{3x3}} \label{eq:nav_vis_iner_filter_H_fIBBtilde_fIBB_so3_local} \\
\nm{\vec H_{4-6,19-21}}   & = & \nm{\derpar{\fIBBtilde}{\EACC} = \vec{I}_{3x3}} \label{eq:nav_vis_iner_filter_H_fIBBtilde_EACC_so3_local} \\
\nm{\vec H_{7-9,1-3}}     & = & \nm{\derpar{\BBtilde}{\DeltarNBB} = \vec J_{\ds{\oplus \; \mathcal R}}^{\ds{- \; \vec g_{\qNB*}^{-1}(\BNMODEL - \BNDEV)}}} \label{eq:nav_vis_iner_filter_H_BBtilde_DeltarNBB_so3_local} \\
\nm{\vec H_{7-9,22-24}}   & = & \nm{\derpar{\BBtilde}{\EMAG} = \vec{I}_{3x3}} \label{eq:nav_vis_iner_filter_H_BBtilde_EMAG_so3_local} \\
\nm{\vec H_{7-9,25-27}}   & = & \nm{\derpar{\BBtilde}{\BNDEV} = - \vec J_{\ds{+ \; \vec v}}^{\ds{- \; \vec g_{\qNB \oplus \DeltarNBB*}^{-1}(\BNMODEL - \BNDEV)}}} \label{eq:nav_vis_iner_filter_H_BBtilde_BNDEV_so3_local} \\
\nm{\vec H_{10-12,7-9}}   & = & \nm{\derpar{\TEgdttilde}{\TEgdt} = \vec{I}_{3x3}} \label{eq:nav_vis_iner_filter_H_TEgdttilde_TEgdt_so3_local} \\
\nm{\vec H_{13-15,10-12}} & = & \nm{\derpar{\vNtilde}{\vN} = \vec{I}_{3x3}} \label{eq:nav_vis_iner_filter_H_vNtilde_vN_so3_local} \\
\nm{H_{ij,OTHER}}         & = & \nm{0} \label{eq:nav_vis_iner_filter_H_other_so3_local} 
\end{eqnarray}


\subsubsection{Navigation Local \nm{\mathbb{SO}(3)} Reset}\label{subsubsec:nav_vis_iner_filter_reset_so3_local}

As described in \cite{LIE}, the modified \hypertt{EKF} scheme capable of accepting state vectors containing non-Euclidean (Lie group) components contains an additional step in every state estimation cycle. In the navigation filter, this step incorporates the \nm{\Delta \hat{\vec r}_{{\sss NB}n}^{{\sss B}+}} perturbation into the \nm{\hat {\vec q}_{{\sss NB}n}^+} \nm{\mathbb{SO}(3)} parameterization, which also implies updating the error covariance estimation \nm{\vec P_n^+}. The latter is achieved by means of the \nm{\vec J_{\ds{\oplus \; \mathcal R}}^{\ds{\ominus \; \hat{\vec q}_{{\sss NB}n}^+ \oplus \Delta \hat{\vec r}_{{\sss NB}n}^{{\sss B}+}}}} Jacobian \cite{LIE}, and the former by means of:
\begin{eqnarray}
\nm{\hat{\vec q}_{{\sss NB}n}^+} & \nm{\longleftarrow} & \nm{\hat{\vec q}_{{\sss NB}n}^+ \oplus \Delta \hat{\vec r}_{{\sss NB}n}^{{\sss B}+}} \label{eq:nav_vis_iner_filter_qNB_reset_so3_local} \\
\nm{\Delta \hat{\vec r}_{{\sss NB}n}^{{\sss B}+}} & \nm{\longleftarrow} & \nm{\vec{0}_3} \label{eq:nav_vis_iner_filter_DeltarNB_reset_so3_local} 
\end{eqnarray}


\subsubsection{Navigation Local \nm{\mathbb{SO}(3)} White Noises and Initialization}\label{subsubsec:nav_vis_iner_filter_whitenoises_initialization_so3_local}

The process noise \nm{\wtilde_n} should be adjusted employing smaller values for those equations within (\ref{eq:nav_vis_iner_cont_time_system_so3_local}) or (\ref{eq:nav_vis_iner_filter_stdot_so3_local}) that are more exact, and higher values otherwise. Note that the geodetic coordinates covariances employ linear units, and hence need to be converted to geodetic ones per (\ref{eq:nav_vis_iner_filter_L1_so3_local}), resulting in a non-unitary transform matrix \nm{\Lvec}.
\begin{eqnarray}
\nm{Diag \ \Lvec_{1-6,1-6}} & = & \nm{\vec I_{6}} \label{eq:nav_vis_iner_filter_L0_so3_local} \\
\nm{\Lvec_{7-9,7-9}} & = & \nm{\vec J_{\ds{\; \vN}}^{\ds{\; \vec {\dot{T}}^{\sss E,GDT}}}} \label{eq:nav_vis_iner_filter_L1_so3_local} \\
\nm{Diag \ \Lvec_{10-27,10-27}} & = & \nm{\vec I_{18}} \label{eq:nav_vis_iner_filter_L2_so3_local} 
\end{eqnarray}
\begin{eqnarray}
\nm{Diag \ \Qtilde_d} & = & \nm{\begin{bmatrix} \nm{5 \cdot 10^{-15} \, \vec I_3 \ rad^2/s^2 \ \ \ 1 \cdot 10^{-6} \ \vec I_3 \, rad^2/s^4 \ \ \ 1 \cdot 10^{-3} \, \vec I_2 \ m^2/s^2} \\ \nm{5 \cdot 10^{-6} \ m^2/s^2 \ \ \ 5 \cdot 10^{-13} \, \vec I_3 \ m^2/s^4 \ \ \ 1 \cdot 10^{-4} \, \vec I_3 \ m^2/s^6} \\ \nm{3 \cdot 10^{-14} \, \vec I_3 \ rad^2/s^4 \ \ \ 1 \cdot 10^{-3} \, \sigmauACC^2 \, \DeltatEST \, \vec I_3 \ m^2/s^6} \\ \nm{1 \cdot 10^{-14} \, \vec I_3 \ nT^2/s^2 \ \ \ 1 \cdot 10^{-12} \, \vec I_3 \ nT^2/s^2} \end{bmatrix}^T \hspace{0.2cm} \in \mathbb{R}^{27}} \label{eq:nav_vis_iner_filter_Qtilde_so3_local} 
\end{eqnarray}

As there are no simplifications within the (\ref{eq:nav_vis_iner_filter_yn_so3_local}) or (\ref{eq:nav_vis_iner_filter_obser_so3_local}) observation equations, the navigation filter can rely on the system noises of the specific sensors. Note that the \nm{\Mvec} transform matrix is non-unitary because the linear position error needs to be converted to geodetic coordinates (\ref{eq:nav_vis_inter_filter_M1_so3_local}).
\begin{eqnarray}
\nm{Diag \ \Mvec_{1-9,1-9}} & = & \nm{\vec I_9} \label{eq:nav_vis_iner_filter_M0_so3_local} \\
\nm{\Mvec_{10-12,10-12}} & = & \nm{\vec J_{\ds{\; \vN}}^{\ds{\; \vec {\dot{T}}^{\sss E,GDT}}}} \label{eq:nav_vis_inter_filter_M1_so3_local} \\
\nm{Diag \ \Mvec_{13-15,13-15}} & = & \nm{\vec I_3} \label{eq:nav_vis_iner_filter_M2_so3_local} 
\end{eqnarray}
\begin{eqnarray}
\nm{Diag \ \Rtilde} & = & \nm{\begin{bmatrix} \nm{\sigmavGYR^2 \cdot \DeltatEST^{-1} \, \vec I_3 \ {^{\circ}}{^2}/s^2 \ \ \ \sigmavACC^2 \cdot \DeltatEST^{-1} \, \vec I_3 \ m^2/s^4 } \\ \nm{ \sigmavMAG^2 \cdot \DeltatEST^{-1} \, \vec I_3 \ nT^2/s^2 \ \ \ \sigmaPOSvec^2 \ m^2 \ \ \ \sigmaVELvec^2 \ m^2/s^2} \end{bmatrix}^T \hspace{0.2cm} \in \mathbb{R}^{15}} \label{eq:nav_vis_iner_filter_R_so3_local}
\end{eqnarray}

The initialization of the navigation filter relies on a combination of the results of the fine alignment process described in section \ref{sec:PreFlight_FineAlignment} and the initial measurements provided by the onboard sensors. The initial estimations of the \nm{\mathbb{SO}(3)} representation \nm{\hat{\vec q}_{{\sss NB}0}^+} and the state variables \nm{\xvecest_0^+} are based on the initial gyroscope, accelerometer, and \hypertt{GNSS} receiver measurements (\nm{{\widetilde{\vec \omega}}_{{\sss IB},0}^{\sss B}, \, \widetilde{\vec f}_{{\sss {IB}},0}^{\sss B}, \, \widetilde{\vec T}_0^{\sss E,GDT}, \, \widetilde{\vec v}_0^{\sss N}}) combined with fine alignment estimations of the aircraft attitude \nm{\qNBinit}, gyroscope error \nm{\EGYRinit}, accelerometer error \nm{\EACCinit}, magnetometer error \nm{\EMAGinit}, and magnetic field deviation \nm{\BNDEVinit}. 

In the case of the \nm{\Pvec_0^+} covariances for these initial estimations, they rely on a hand adjusted value for the attitude\footnote{Directly setting the attitude covariance to the results of fine alignment resulted in erratic results, so this value was carefully adjusted by the author.} coupled with covariances obtained by combining the fine alignment results of table \ref{tab:PreFlight_fine_alignment} with various fixed parameters from chapters \ref{cha:EarthModel} and \ref{cha:Sensors}. As in previous cases, note that the initial covariance of the horizontal position is converted from distance to longitude and latitude when introduced into (\ref{eq:nav_vis_iner_filter_Pestinit_so3_local}):
\begin{eqnarray}
\nm{Diag \ \Pvec_0^+} & = & \nm{\begin{bmatrix} 
\nm{1 \cdot 10^{-8} \, \vec I_3 \ rad^2} \\ 
\nm{\Big(\sigmavGYR^2 \, {\DeltatEST}^{-1} + \lrp{B_{0\sss{GYR}}^2 + \sigma_{u\sss{GYR}}^2 \, \DeltatEST} \, \sigmaEGYRinit^2\Big) \, \vec I_3} \\ 
\nm{\vec J_{\ds{\; \vN}}^{\ds{\; \vec {\dot{T}}^{\sss E,GDT}}} \, \lrsb{\sigma_{\sss GNSS,POS,HOR}^2 \ \ \sigma_{\sss GNSS,POS,HOR}^2 \ \ \sigma_{\sss GNSS,POS,VER}^2}^T \, \vec J_{\ds{\; \vN}}^{\ds{\; \vec {\dot{T}}^{\sss E,GDT}} \ ^T}} \\
\nm{\sigma_{\sss GNSS,VEL}^2 \, \vec I_3} \\
\nm{\Big(\sigmavACC^2 \, {\DeltatEST}^{-1} + \lrp{B_{0\sss{ACC}}^2 + \sigma_{u\sss{ACC}}^2 \, \DeltatEST} \, \sigmaEACCinit^2\Big) \, \vec I_3} \\
\nm{\lrp{B_{0\sss{GYR}}^2 + \sigma_{u\sss{GYR}}^2 \, \DeltatEST} \, \sigmaEGYRinit^2 \, \vec I_3} \\
\nm{\lrp{B_{0\sss{ACC}}^2 + \sigma_{u\sss{ACC}}^2 \, \DeltatEST} \, \sigmaEACCinit^2 \, \vec I_3} 
\nm{\lrp{B_{0,\sss MAG}^2 + B_{\sss{HI,MAG}}^2} \, \sigmaEMAGinit^2 \, \vec I_3} \\ 
\nm{\lrsb{\sigma_{B \sss{DEV,1}}^2 \ \ \sigma_{B \sss{DEV,2}}^2 \ \ \sigma_{B \sss{DEV,1}}^3}^T \, \sigmaBNDEVinit^2}
\end{bmatrix} \hspace{0.2cm} \in \mathbb{R}^{27}}\label{eq:nav_vis_iner_filter_Pestinit_so3_local} \\
\nm{\hat{\vec q}_{{\sss NB}0}^+} & = & \nm{\qNBinit \hspace{0.2cm} \in \mathbb{SO}(3)}\label{eq:nav_vis_iner_filter_qNBinit_so3_local} \\
\nm{\xvecest_0^+} & = & \nm{\begin{bmatrix} \nm{\vec 0_3 \ \ \ \ {\widetilde{\vec \omega}}_{{\sss{IB}},0}^{\sss B} - \EGYRinit \ \ \ \ {\widetilde{\vec T}}_0^{\sss E,GDT} \ \ \ \ {\widetilde{\vec v}}_0^{\sss N}} \\ \nm{{\widetilde {\vec f}}_{{\sss {IB}},0}^{\sss B} - \EACCinit \ \ \ \ \EGYRinit \ \ \ \ \EACCinit \ \ \ \ \EMAGinit \ \ \ \ \BNDEVinit} \end{bmatrix}^T \hspace{0.2cm} \in \mathbb{R}^{27}}\label{eq:nav_vis_iner_filter_stestinit_so3_local} 
\end{eqnarray}


\subsubsection{Navigation Local \nm{\mathbb{SO}(3)} Extra Steps}\label{subsubsec:nav_vis_iner_filter_extra_so3_local}

Each navigation step is not complete until estimating those variables contained in the observed state \nm{\xvecest = \xEST} (section \ref{sec:GNC_OT}) that have not been estimated by the filter; these are the aircraft pose \nm{\zetaEBest} and local twist \nm{\xiEBBest} obtained per the section \ref{sec:GNC_OT} instructions, the atmospheric pressure offset \nm{\Deltapest}, the wind field \nm{\vWINDNest}, and the aircraft mass \nm{\mest}. 

The pressure offset \nm{\Deltap} and wind field \nm{\vWINDN} are both computed after the filter execution step based on the remaining components of the observed trajectory. After estimating the geopotential altitude \nm{\hat H} according to (\ref{eq:EarthModel_GEOP_h2H}) based on the geometric altitude (part of \nm{\TEgdtest}), it is possible to estimate \nm{\Deltapest} by iterating around (\ref{eq:EarthModel_ISA_H_Hp}) based on \nm{\Hpest} and the temperature offset \nm{\DeltaTest} estimated by the air data filter in section \ref{subsec:GNC_Navigation_AirDataFilter}. Finally, the wind velocity is estimated neglecting the turbulence by subtracting the airspeed obtained by the air data filter from the ground velocity. Both \nm{\Deltapest} and \nm{\vWINDNest} should be smoothed by means of a low pass filter \cite{Haugen2008}.
\begin{eqnarray}
\nm{\hat H} & = & \nm{\dfrac{\RE \cdot \hat h}{\RE + \hat h}}\label{eq:GNC_Navigation_AirDataFilterGNSS_H} \\
\nm{\Deltapest} & = & \nm{f_{\Deltap}\lrp{\Hpest, \, \hat H, \, \DeltaTest}}\label{eq:GNC_Navigation_AirDataFilterGNSS_Deltap} \\ 
\nm{\vWINDNest} & = & \nm{\vNest - \vec g_{\ds{\hat{\vec q}_{{\sss NB}*}}} \big(\vTASB(\vtasest, \, \alphaest, \, \betaest) \big)}\label{eq:GNC_Navigation_AirDataFilterGNSS_vN}  
\end{eqnarray}

The aircraft mass \nm{\mest} is also estimated independently of the filters\footnote{As in only depends on variables estimated during the previous cycle (\ref{eq:GNC_Navigation_mass1}), the mass can be estimated before executing any of the two filters.}, and enables the estimation of the relative position between the body \nm{\FB} and platform \nm{\FP} frames (\nm{\TBPBest}), by means of (\ref{eq:Sensors_Inertial_Mounting_Tbpb}, \ref{eq:Sensors_Inertial_Mounting_TBPBest}), required to generate the measurements of the accelerometers per (\ref{eq:Sensor_Inertial_acc_error_final}). If required, the mass is also necessary to estimate the relative position between the body \nm{\FB} and camera \nm{\FC} frames (\nm{\TBCBest}), by means of (\ref{eq:Sensors_Camera_Mounting_Tbcb}, \ref{eq:Sensors_Camera_Mounting_TBCBest}).

The aircraft mass at a given time \nm{t_n} can be estimated by integrating the fuel consumption provided by (\ref{eq:AircraftModel_power}, \ref{eq:AircraftModel_fuel_consumption}), where the atmospheric expressions (\ref{eq:EarthModel_ISA_delta}, \ref{eq:EarthModel_ISA_theta}, \ref{eq:EarthModel_ISA_Hp_p}, \ref{eq:EarthModel_ISA_T_Hp}) are employed to convert the pressure and temperature ratios (\nm{\delta, \, \theta}) into \nm{\Hp} and \nm{\DeltaT}, which have both been estimated by the air data filter. The throttle parameter \nm{\deltaT} can be obtained from the \nm{\deltaCNTR} output of the control system\footnote{Note that the dependency of the \hypertt{INS} on the throttle parameter is not shown in figures \ref{fig:Trajectory_flow_diagram}, \ref{fig:GNC_flow_diagram}, and \ref{fig:Navigation_flow_diagram}.}, by means of (\ref{eq:GNC_Control_deltaCNTR}).
\neweq{\mest = \hat{m}_{n-1} - \DeltatEST \cdot F\lrp{\delta_{{\sss T},n-1}, \, \hat{\delta}_{n-1}, \, \hat{\theta}_{n-1}} = \hat{m}_{n-1} - \DeltatEST \cdot F\lrp{\delta_{{\sss T},n-1}, \, \hat{H}_{{\sss P},n-1}, \, \Delta \hat{T}_{n-1}}} {eq:GNC_Navigation_mass1}  

Note that an initial mass estimation \nm{\hat{m}_0} is required for initialization, which is obtained by weighting the aircraft before departure or by adding the aircraft empty mass \nm{m_{empty}} (section \ref{sec:AircraftModel_MassInertia}) with that of the fuel load.

This modus operandi assumes that the \hypertt{INS} possess an exact knowledge of the real variation of the center of mass location with the aircraft mass \nm{\TRBR} (\ref{eq:AircraftModel_Trbb}) [A\ref{as:GNC_TBPB_mass}], as well as the real fuel consumption model [A\ref{as:GNC_F}]. The author could have opted to include some random differences between the real dependencies and those employed by the \hypertt{INS} to simulate that the onboard models are never completely equal to the real dependencies, but the maximum displacement of the center of mass between the full and empty tanks configurations is so small (section \ref{sec:AircraftModel_MassInertia}) that the final effects of these assumptions on the realism of the simulation are negligible.


\section{Implementation, Validation, and Execution} \label{sec:nav_iner_implementation}

The \hypertt{INS} described in this chapter have been implemented as an object oriented \texttt{C++} library following a process similar to that of the control system (section \ref{sec:gc_implementation}), but significantly more challenging. Two reasons explain this: on one side, the stochastic nature of the filters makes their implementation intrinsically more difficult, but also special care is required to obtain an efficient implementation that makes the best possible use of the available computing resources, and hence minimizes the time required to run repeated simulation executions. The process has also involved testing progressively more complex missions, and slowly making modifications to the filter algorithms to improve the results.

 \cleardoublepage

\addcontentsline{toc}{chapter}{Bibliography} 
\bibliographystyle{ieeetr}   
\bibliography{bib/part_intro,bib/part_01,bib/part_02,bib/part_04,bib/books,bib/a3_algebra,bib/a7_estim,bib/references}
\cleardoublepage

\appendix
\pagenumbering{gobble} 
\chapter{Assumptions} \label{cha:Assumptions}	

\renewcommand{\thepage}{\thechapter-\arabic{page}} 
\setcounter{page}{1}    			

\hspace{20pt} \hyperbf{WGS84} \textbf{and Gravity}

\begin{list}{[A\arabic{level_1}]}{\usecounter{level_1}}	


\item The surface of the Earth is a revolution ellipsoid. \label{as:WGS84_ellipsoid}

\item The Earth ellipsoid is centered at the Earth's center of mass. \label{as:WGS84_center_mass}
	
\item The Earth rotates at constant speed around the ellipsoid symmetry axis. \label{as:WGS84_rotation}
	
\item The gravitational field varies only with the geocentric distance and colatitude. \label{as:EGM96_gravity}

\item The geopotential expression is an explicit approximation obtained with spherical Earth surface, spherical gravitation, and average centrifugal effect. \label{as:GEOP_geopotential}

\item The difference between the real and modeled gravity vectors does not change during the duration of the flight. \label{as:MODELS_gc}

\textbf{Atmosphere}

\item The atmosphere is composed by pure air, which is a perfect gas. \label{as:ISA_perfect_gas}

\item The atmosphere is static in relation to the Earth, so it is in vertical fluid static equilibrium.\label{as:ISA_static_equilibrium}

\item The temperature decreases with altitude according to a given gradient, which is constant when expressed in terms of pressure altitude.\label{as:ISA_T_gradient}

\item The tropopause altitude is constant when expressed in terms of pressure altitude.\label{as:ISA_tropopause}

\item The air humidity does not have any influence on the value of any other atmospheric property.\label{as:ISA_humidity}

\textbf{Wind and Turbulence}

\item The wind experienced by the aircraft is composed of low and high frequency components. \label{as:WIND_composition}

\item The air turbulence is frozen in time and space, implying that the turbulence experienced by the aircraft results exclusively from its motion relative to the frozen turbulence, not from the variations or motion of the turbulence itself. \label{as:WIND_turbulence}

\textbf{Magnetism}

\item The Earth magnetic field can be locally approximated by a bilinear function on longitude and latitude while neglecting the influence of altitude and time. \label{as:WMM_linear}

\item The difference between the real and modeled magnetic field vectors does not change during the duration of the flight. \label{as:MODELS_B}

\vspace{20pt}

\textbf{Aircraft}

\item The aircraft is a rigid body, discarding the aeroelastic effects. \label{as:APM_rigid_body}

\item The aircraft is symmetric from a geometric and mass distribution point of view. \label{as:APM_symmetric}

\item The aircraft center of mass position varies linearly with the aircraft mass between its empty (no fuel) and full locations. \label{as:APM_center_mass}

\item The aircraft inertia matrix varies linearly with the aircraft mass between its empty (no fuel) and full values.\label{as:APM_inertia}

\item The \hypertt{GNSS} receiver estimates the position of the aircraft center of mass, not the receiver position. \label{as:APM_GNSS}

\item The air stream is quasi stationary in its motion through the propeller and around the aircraft, implying that an air particle does not experience any variation in its own properties or in the aircraft configuration while interacting with it. This is the result of the air particles residence time being much shorter than the characteristic time of the aircraft or propeller to modify their operating conditions. \label{as:APM_quasi_stationary}

\item The Reynold's number (proportional to the airspeed) of the interaction of the air stream with both the propeller and the aircraft structure is sufficiently elevated to discard the influence of the air viscosity.\label{as:APM_viscosity}

\item The air stream perturbed by both the propeller and the aircraft structure is incompressible. \label{as:APM_incompressible}

\item The propeller rotation axis is located in the aircraft plane of symmetry, aligned with the fuselage, and passes through its center of mass.\label{as:APM_propeller_axis}

\item The thrust force generated by the propeller is aligned with its axis, independently of the direction of the air stream.\label{as:APM_propeller_thrust}

\item The piston engine thermodynamic efficiency, depicting the percentage of the fuel internal energy that gets transformed into power, is constant.\label{as:APM_power_efficiency}

\item The direction of the air flow with respect to the aircraft does not influence the power generated by the piston engine as the air intake is capable of recovering all its kinetic energy.\label{as:APM_flow_piston}

\item The performance of the propeller depends exclusively on its axial air flow, discarding the influence of the air stream lateral and vertical components.\label{as:APM_flow_propeller}

\item The energy loses incurred by the power plant shaft are negligible.\label{as:APM_propulsion_shaft}

\item The propulsive forces and moments do not depend on the position of the aircraft center of mass. \label{as:APM_propulsion_center_mass}

\textbf{Physics of Flight}

\item A reference frame centered at the Sun center of mass and with axes fixed with respect to other stars is inertial.\label{as:MOTION_inertial_system}

\item The accelerations caused by the translation motion of the Earth center of mass around the Sun can be discarded.\label{as:MOTION_earth_translation}

\item The variation with time of the aircraft mass is slow enough as to be considered quasi stationary. \label{as:MOTION_mass_stationary}

\item The variation with time of the aircraft inertia matrix when expressed in the body frame is slow enough as to be considered quasi stationary.\label{as:MOTION_inertia_matrix_stationary}

\textbf{Sensors}

\item The accelerometers, gyroscopes, and magnetometers are physically attached to the aircraft structure in a strapdown configuration.\label{as:SENSOR_strapdown}

\item The measurement process is instantaneous for all sensors.\label{as:SENSOR_instantaneous}

\item All sensors, with the exception of the camera and the \hypertt{GNSS} receiver, work at a frequency of \nm{100 \ Hz}, and are synchronized so they generate their outputs at exactly the same discrete times. The digital camera and \hypertt{GNSS} receiver are also synchronized with all other sensors, but their frequencies are \nm{10} and \nm{1 \ Hz}, respectively.\label{as:SENSOR_synchronize}

\item The accelerometers, gyroscopes, magnetometers, and \hypertt{GNSS} receiver are infinitesimally small and located at the \hypertt{IMU} reference point.\label{as:SENSOR_small}

\item The measurement error introduced by the triad of accelerometers includes contributions from the bias offset, bias drift, system noise, scale factors, and cross coupling errors.\label{as:SENSOR_ACC}

\item The measurement error introduced by the triad of gyroscopes includes contributions from the bias offset, bias drift, system noise, scale factors, and cross coupling errors.\label{as:SENSOR_GYR}

\item The measurement error introduced by the triad of magnetometers includes contributions from hard iron magnetism, bias offset, system noise, soft iron magnetism, scale factors, and cross coupling errors.\label{as:SENSOR_MAG}

\item The measurement error introduced by the barometer includes contributions from bias offset and system noise.\label{as:SENSOR_OSP}

\item The measurement error introduced by the thermometer includes contributions from bias offset and system noise.\label{as:SENSOR_OAT}

\item The measurement error introduced by the Pitot tube measurement of airspeed includes contributions from bias offset and system noise.\label{as:SENSOR_TAS}

\item The measurement error introduced by either the Pitot tube or the air vanes when measuring the angles of attack and sideslip includes contributions from bias offset and system noise.\label{as:SENSOR_AOA_AOS}

\item The position error of a \hypertt{GNSS} receiver is the sum of a system or white noise plus slow varying ionospheric effects, while the velocity error contains system noise exclusively.\label{as:SENSOR_GNSS}

\textbf{Camera}

\item The camera is physically attached to the aircraft structure in a strapdown configuration.\label{as:CAMERA_strapdown}

\item The camera works with a sufficiently high shutter speed so that all images are equally sharp. \label{as:CAMERA_sharp}

\item The camera image generation process is instantaneous.\label{as:CAMERA_instantaneous}

\item The camera works with a constant \hypertt{ISO} setting, so the luminosity with which a given light ray is recorded does not change from image to image. \label{as:CAMERA_iso}

\item All images generated by the camera are noise free. \label{as:CAMERA_noise}

\item The Earth surface luminosity does not vary with time.  \label{as:CAMERA_luminosity}

\item All objects viewed by the camera have a Lambertian surface, this is, the viewing direction does not change their appearance. \label{as:CAMERA_lambertian}

\item The image generation process relies on a perspective projection or pinhole camera model. \label{as:CAMERA_pinhole}

\item The camera is perfectly calibrated, so there is no distortion in the projection. \label{as:CAMERA_calibration}

\textbf{Pre-Flight Procedures}

\item The calibration of the inertial sensors results in a \nm{95\%} reduction in the accelerometers and gyroscopes scale factor and cross coupling errors, instead of corrections to the sensor readings. \label{as:PREFLIGHT_calibration}

\item The calibration of the magnetometers or swinging results in a \nm{90\%} reduction in their scale factor and cross coupling errors, which also include the soft iron errors, and in the hard iron magnetism effects, instead of corrections to the magnetometers readings. \label{as:PREFLIGHT_swinging}

\item The fine alignment process results in an error of \nm{0.1^{\circ}} in the estimation of each of the three body Euler angles, a \nm{1 \, \%} error in the estimation of the accelerometers, gyroscopes, and magnetometers full errors, and a \nm{10 \, \%} error in the estimation of the difference between the real and modeled Earth magnetic fields. \label{as:PREFLIHGT_fine_alignment}

\textbf{Guidance, Navigation, and Control}

\item The execution of the algorithms in the guidance, navigation, and control systems is instantaneous.\label{as:GNC_instantaneous}

\item The navigation system works at a frequency of \nm{100 \ Hz}, and is fully synchronized with the sensors, operating at the same discrete times.\label{as:GNC_n_synchronize}

\item The guidance and control systems work at a frequency of \nm{50 \ Hz}, operate at the same discrete times, and are fully time synchronized with the sensors and navigation system.\label{as:GNC_gc_synchronize}

\item The throttle, elevator, ailerons, and rudder react instantaneously to the adjustments commanded by the control system.\label{as:GNC_control_surfaces}

\item The dependency between the aircraft mass (fuel load) and the position of the aircraft center of mass with respect to the aircraft structure is known with exactitude by the navigation system.\label{as:GNC_TBPB_mass}

\item The dependency of the power plant fuel consumption with the throttle position and atmospheric properties is known with exactitude by the navigation system.\label{as:GNC_F}

\item The visual navigation system works at a frequency of \nm{10 \ Hz}, and is fully synchronized with the camera, operating at the same discrete times.\label{as:GNC_vis_synchronize}

\end{list}

 \cleardoublepage
\chapter{Physical Constants} \label{cha:PhysicalConstants}  

\renewcommand{\thepage}{\thechapter-\arabic{page}} 
\setcounter{page}{1}


\renewcommand{\arraystretch}{1.25}
\begin{table}[ht]
\begin{tabular}{cp{6.5cm}rcc}
	\rule[-8pt]{0pt}{12pt}\textbf{Constant} & \textbf{Definition} & \textbf{Value} & \textbf{Units} & \textbf{Source} \\
	\nm{a{\sss{WGS84}}} & \hypertt{WGS84} ellipsoid semi-major axis	& 6378137.0			& m & \cite{WGS84} \\
	\nm{f{\sss{WGS84}}}	& \hypertt{WGS84} ellipsoid flattening & 1/298.257223563	& - & \cite {WGS84} \\ 
	\nm{GM} & Earth gravitational constant (atmosphere included) &	\nm{3986004.418 \cdot 10^8} &	\nm{m^3/\,s^2} & \cite{WGS84} \\ 
	\nm{\bar{C}_{20}} & \hypertt{EGM96} \nm{\second} degree zonal harmonical & \nm{- 0.484166774985 \cdot 10^{-3}} & - & \cite{EGM96} \\
	\nm{\omega_{\sss E}}	& \hypertt{WGS84} ellipsoid angular velocity &	\nm{7.292115 \cdot 10^{-5}}	& rad/s & \cite{WGS84} \\ 
	\nm{\gcMSLE} & Theoretical \hypertt{WGS84} gravity at Equator & \nm{9.7803253359} & \nm{m/\,s^2} & \cite{WGS84} \\
	\nm{\gcMSLP} & Theoretical \hypertt{WGS84} gravity at poles   & \nm{9.8321849378} & \nm{m/\,s^2} & \cite{WGS84} \\	
	\nm{\gzero} & Standard acceleration of free fall &	9.80665 & \nm{m/\,s^2} & \cite{WGS84} \\
	\nm{\RE}	& Earth nominal radius & 6356766.0 & m & \cite{SMITHSONIAN} \\ 
	\nm{\Tzero} & Standard temperature at mean sea level	& 288.15 & K & \cite{ISA} \\ 
	\nm{\pzero} & Standard pressure at mean sea level & 101325 &	\nm{kg/\,m\,s^2} & \cite{ISA} \\ 
	\nm{\rhozero} & Standard density at mean sea level &	1.225 & \nm{kg/\,m^3} & \cite{ISA} \\ 	
	\nm{R} & Specific air constant &	287.05287 &	\nm{m^2/\,K\,s^2} & \cite {ISA} \\ 
	\nm{\betaT}	& \raggedright Temperature gradient below tropopause & \nm{-6.5 \cdot 10^{-3}} & \nm{K/m} & \cite{ISA} \\ 
	\nm{\Hptrop} & Tropopause pressure altitude & 11000 & m & \cite{ISA} \\
	\nm{R_{\sss E-S}} & Distance between Earth and Sun & \nm{\sim 1.5 \cdot 10^{11}} & m & - \\
	\nm{\omega_{\sss E-S}} & Earth angular velocity with respect to Sun & \nm{\sim 2 \pi / \left(365 \cdot 24 \cdot 3600\right)} & \nm{^{\circ}/\,s} & - \\
	\nm{\kappa} & Air adiabatic index & 1.4 & - & \cite{ISA} \\ 	
\end{tabular}
\end{table}
\renewcommand{\arraystretch}{1.25}

 \cleardoublepage
\chapter{Coordinate Frames}\label{cha:RefSystems}

\renewcommand{\thepage}{\thechapter-\arabic{page}} 
\setcounter{page}{1}   

A \emph{coordinate frame} provides an origin and a set of axes in terms of which the motion and pose (position plus attitude) of rigid bodies may be described \cite{Groves2008}. Most coordinate frames listed in table \ref{tab:RefSystems_RefSystems} below are \emph{Cartesian}, this is, their basis vectors are orthonormal and right handed \cite{LIE}\footnote{The accelerometer and gyroscope frames of sections \ref{subsec:RefSystems_A} and \ref{subsec:RefSystems_Y} are non-orthogonal.}.
\begin{center}
\begin{tabular}{lccccl}
	\hline
	\multicolumn{2}{c}{Frame} & Section & Origin & Parameters & Main Purpose \\
	\hline
	Inertial 		& \nm{\FI}	& \ref{subsec:RefSystems_I}	& Earth\footnotemark	& -   							& Equations of motion 			\\
	Earth \hypertt{ECEF} & \nm{\FE}	& \ref{subsec:RefSystems_E}	& Earth				& -								& Earth representation			\\
	Spherical		& \nm{\FS}	& \ref{subsec:RefSystems_S}	& Aircraft\footnotemark	& \nm{\theta, \ \lambda, \ r}	& Gravitation					\\
	\hypertt{NED}	& \nm{\FN}	& \ref{subsec:RefSystems_N}	& Aircraft				& \nm{\lambda, \ \varphi, \ h} 	& Aircraft position				\\
	Body			& \nm{\FB}	& \ref{subsec:RefSystems_B}	& Aircraft				& \nm{\psi, \ \theta, \ \xi}	& Aircraft structure attitude	\\
	Wind			& \nm{\FW}	& \ref{subsec:RefSystems_W}	& Aircraft				& \nm{\alpha, \ \beta}			& Aircraft air velocity attitude	\\
	Ground			& \nm{\FG}  & \ref{subsec:RefSystems_G} & Aircraft				& \nm{\chi, \ \gamma, \ \mu}	& Aircraft ground velocity attitude \\
	Turbulence		& \nm{\FT}	& \ref{subsec:RefSystems_T} & Aircraft				& \nm{\chiWIND}					& Wind turbulence				\\
	Reference		& \nm{\FR}	& \ref{subsec:RefSystems_R} & Arbitrary				& \nm{\TRBR}					& Aerodynamic forces			\\
	Platform		& \nm{\FP}	& \ref{subsec:RefSystems_P} & \hypertt{IMU}\footnotemark		& \nm{\TBPB, \ \psiP, \ \thetaP, \ \xiP}	& \hypertt{IMU} calibrated output			\\
	Accelerometer	& \nm{\FA}	& \ref{subsec:RefSystems_A} & \hypertt{IMU}					& \nm{\alphaACCi, \ \alphaACCii, \ \alphaACCiii}		& Accelerometers sensing axes 	\\
	Gyroscope             & \nm{\FY}	& \ref{subsec:RefSystems_Y} & \hypertt{IMU}               & \nm{\alphaGYRiXii, \ \alphaGYRiiXi, \ \alphaGYRiXiii} & Gyroscopes sensing axes \\	
		                    &	          &	                          &	                            & \nm{\alphaGYRiiiXi, \ \alphaGYRiiXiii, \ \alphaGYRiiiXii} & 	\\	
	Camera			& \nm{\FC}	& \ref{subsec:RefSystems_C} & Camera\footnotemark	& \nm{\TBCB, \ \psiC, \ \thetaC, \ \xiC}				& Vision 	\\
	\hline
\end{tabular}
\end{center}
\captionof{table}{Summary of coordinate frames}\label{tab:RefSystems_RefSystems}	
\addtocounter{footnote}{-3}
\footnotetext{Refers to the Earth center of mass.} 
\addtocounter{footnote}{1}
\footnotetext{Refers to the aircraft center of mass.} 
\addtocounter{footnote}{1}
\footnotetext{Refers to the \hypertt{IMU} reference point (section \ref{sec:Sensors_Inertial}).}
\addtocounter{footnote}{1}
\footnotetext{Refers to the camera optical center (section  \ref{subsec:Vis_Camera}).}

This chapter begins with the description of those coordinate frames employed to establish positions and attitudes with respect to the Earth (section \ref{sec:RefSystems_Earth}), then focuses on those that do the same with respect to the aircraft (section \ref{sec:RefSystems_Acft}), and finally describes the frames required to express the sensor measurements (section \ref{sec:RefSystems_Sensor}). Table \ref{tab:RefSystems_RefSystems} summarizes the different coordinate frames employed in this document.


\section{Earth Focused Frames}\label{sec:RefSystems_Earth}

The coordinate frames defined in this section are employed to describe the position of any point as well as the attitude or orientation of any vector with respect to the Earth. These frames are employed by many flight dynamics and navigation textbooks \cite{Farrell2008,Groves2008,Chatfield1997,Rogers2007,Etkin1972,Ashley1974,Miele1962,Titterton2004}.


\subsection{Inertial Frame}\label{subsec:RefSystems_I}

An inertial (\texttt{I}) frame is that in which bodies whose net force acting upon them is zero do not experience any acceleration and remain either at rest or moving at constant velocity in a straight line. The inertial frame \nm{\FI} obtained in section \ref{sec:EquationsMotion_inertial_selection} is one centered at the Earth center of mass and whose axes do not rotate with respect to any stars other than the Sun. Such a frame is necessary to establish the equations of motion as in it there are no inertial accelerations. 


\subsection{Earth Frame}\label{subsec:RefSystems_E}

The \hypertt{WGS84} model introduced in section \ref{sec:EarthModel_WGS84} is based on the \emph{Earth Centered Earth Fixed} (\hypertt{ECEF}) frame, also known as the Earth (\texttt{E}) frame, a Cartesian frame \nm{\FE = \{\OECEF,\,\iEi,\,\iEii,\,\iEiii\}}, shown in figure \ref{fig:RefSystems_E}. It is defined as having \nm{\OECEF} located at the Earth center of mass, \nm{\iEiii} pointing towards the geodetic North pole along the Earth rotation axis (poles axis), \nm{\iEi} contained in both the Equator and Greenwich meridian plane, pointing towards zero longitude, and \nm{\iEii} orthogonal to \nm{\iEi} and \nm{\iEiii} forming a right handed system.
\begin{figure}[h]
\centering
\begin{tikzpicture}[auto,>=latex',scale=1.0]
	\coordinate (origin) at (+0.0,+0.0);
	\coordinate (origin1) at (+0.0,+1.6);
	\coordinate (origin2) at (+0.0,+2.2);
		
	\filldraw [black] (origin) circle [radius=2pt] node [left=1pt] {\nm{\OECEF}};
	\draw [name path = ellhor] (origin) ellipse [x radius=4.0, y radius=1.65];
	\draw [name path = ellhor2, dashed, ultra thin] (origin1) ellipse [x radius=(4.0/1.19), y radius=(1.25/1.19)];
	\draw [name path = ellver] (origin) ellipse [x radius=1.3, y radius=3.0];
	\draw [name path = ellmain, very thick] (origin) ellipse [x radius=4.0, y radius=3.0];
	
	\draw [name path = iECEFi, name intersections={of=ellhor and ellver, by={int1,int2,int3,int4}}, very thick] [->] (origin) -- ($(origin) + 1.25*(int3)$) node [below=3pt] {\nm{\iEi}};
	\draw [name path = iECEFii, very thick] [->] (origin) -- (+4.5,+0.0) node [right=3pt]  {\nm{\iEii}};
	\draw [name path = iECEFiii, very thick] [->] (origin) -- (+0.0,+3.5) node [right=4pt]  {\nm{\iEiii}};
	
	\path [red, name path = path1] [-] (origin1) -- (+3.4,-0.6);
	\path [name intersections={of=path1 and ellhor2, by={pointQ}}] [-] (origin) -- (pointQ);
	\draw [dashed, ultra thin, name path = path11] [-] (origin1) -- (pointQ);
	\filldraw [black] (pointQ) circle [radius=1pt];
	
	\path [red, name path = path2] [-] (origin2) -- (+3.4,-0.0);
	\path [red, name path = path3] [-] (origin) -- ($(origin) + 6/3*(pointQ)$);
	\draw [dashed, ultra thin, name intersections={of=path2 and path3, by={pointP}}] [-] (origin) -- (pointP) node[above=2pt] {\nm{\vec P}};
	\draw [dashed, ultra thin, name path = path21] [-] (origin2) -- (pointP);
	\filldraw [black] (pointP) circle [radius=1pt];
	
	\path [red, name path = path4] [-] (origin) -- (+3.4,-2.2);
	\path [red, name path = path5] [-] (pointQ) ++(0,-2) -- (pointQ);
	\draw [dashed, ultra thin, name intersections={of=path4 and path5, by={pointq}}] [-] (pointQ) -- (pointq);
	\path [red, name path = path6] [-] (pointP) ++(0,-2.5) -- (pointP);
	\draw [dashed, ultra thin, name intersections={of=path4 and path6, by={pointp}}] [-] (origin) -- (pointp);	
	
	\path [red, name path = path7] [-] (pointp)++(-4.0,0.0) -- (pointp);
	\path [red, name intersections={of=path7 and iECEFi, by={pointp1}}] [-] (pointp) -- (pointp1);	
	\path [red, name path = path8] [-] (pointp)++(+1.5,+1.5) -- (pointp); 
	\path [red, name intersections={of=path8 and iECEFii, by={pointp2}}] [-] (pointp) -- (pointp2);	
		
	\draw [blue, ultra thick] [->] (pointp1) -- (pointp) node [pos=0.5, above=0pt]  {\nm{\TEii}};		
	\draw [blue, ultra thick] [->] (pointp2) -- (pointp) node [pos=0.4, right=4pt]  {\nm{\TEi}};		
	\draw [blue, ultra thick] [->] (pointp) -- (pointP) node [pos=0.75, right=0pt]  {\nm{\TEiii}};
	
	\node [left of=origin, node distance=2.4cm] (Greenwich) {\st{Greenwich Mer.}};	
	\coordinate (origin3) at (+0.0,+1.2);
	\node [left of=origin3, node distance=4.6cm] (Ellipsoid) {\st{\hypertt{WGS84} Ell.}};	
	\coordinate (origin4) at (+0.0,-1.0);
	\node [left of=origin4, node distance=2.4cm] (Equator) {\st{Equator}};	
		
\end{tikzpicture}
\caption{Earth frame \nm{\FE} and cartesian coordinates \nm{\TEcar}}
\label{fig:RefSystems_E}
\end{figure}

Being fixed to the Earth, the \nm{\FE} frame is well suited to express the position of any object with respect to the Earth surface. Three different sets of coordinates can be employed for this purpose: Cartesian, geocentric, and geodetic. Either set can be unambiguously obtained from any of the other two.

        
\subsubsection{Cartesian Coordinates}\label{subsubsec:RefSystems_E_CartesianCoord}
   
The \emph{Cartesian coordinates} \nm{\TEBE = \TENE = \TEcar = \lrsb{\TEi \ \ \TEii \ \ \TEiii}^T} result from the direct projection of the vector connecting \nm{\OECEF} with the point \nm{\vec P} over the \nm{\FE} axes, as depicted in figure \ref{fig:RefSystems_E}. Although easy to obtain, they are not very practical as their values do not provide any intuitive indication about the point \nm{\vec P} position with respect to the Earth surface. The Cartesian coordinates can be obtained from either the geocentric or geodetic ones defined below\footnote{Refer to section \ref{sec:EarthModel_WGS84} and figure \ref{fig:EarthModel_ellipsoid} for more information on the terms present in (\ref{eq:RefSystems_E_geoc2car}) and (\ref{eq:RefSystems_E_geod2car}).}:
\begin{eqnarray}
\nm{\TEcar} & = & \nm{\Big[r \, \sin\theta \, \cos\lambda \ \ \ \ r \, \sin\theta \, \sin\lambda \ \ \ \ r \, \cos\theta \Big]^T} \label{eq:RefSystems_E_geoc2car} \\
\nm{\TEcar} & = & \nm{\Big[ (N+h) \, \cos\varphi \, \cos\lambda \ \ \ \ (N+h) \, \cos\varphi \, \sin\lambda \ \ \ \ \big(N(1-e^2)+h\big) \, \sin\varphi \Big]^T} \label{eq:RefSystems_E_geod2car}
\end{eqnarray}

        
\subsubsection{Geocentric Coordinates}\label{subsubsec:RefSystems_E_GeocentricCoord}

The \emph{geocentric coordinates} \nm{\TEgct = \lrsb{\theta \ \ \lambda \ \ r}^T} represent a more intuitive way to position a certain point \nm{\vec P} with respect to \nm{\FE}. Shown in figure \ref{fig:RefSystems_S}, they are defined as follows:
\begin{itemize}
\item \emph{Geocentric colatitude} \nm{\theta} \nm{\lrp{0 \leq \theta \leq \pi }} is the angle formed between the Earth rotation axis (\nm{\iEiii}) and the line passing through the point \nm{\vec P} and the Earth center of mass.
\item \emph{Longitude} \nm{\lambda} \nm{\lrp{0 \leq \lambda < 2\pi }} is the angle formed between the Greenwich meridian plane (formed by \nm{\iEi} and \nm{\iEiii}) and the point meridian plane\footnote{It coincides with the geodetic longitude defined in section \ref{subsubsec:RefSystems_E_GeodeticCoord}.}.
\item \emph{Geocentric distance} \emph{r} \nm{\lrp{r \geq 0 }} is the distance between the point and the Earth center of mass.
\end{itemize}
        	
The geocentric coordinates can be obtained from either the Cartesian or the geodetic coordinates defined below. Note that the longitude \nm{\lambda} belongs to both the geocentric and geodetic coordinates:
\begin{eqnarray}
\nm{\TEgct} & = & \nm{\lrsb{\arctan\frac{\sqrt{{\TEi}^2 + {\TEii}^2}}{\TEiii} \ \ \ \ \arctan\frac{\TEii}{\TEi} \ \ \ \ \sqrt{{\TEi}^2 + {\TEii}^2 + {\TEiii}^2} }^T} \label{eq:RefSystems_E_car2geoc} \\
\nm{\TEgct} & = & \nm{\lrsb{\arctan \frac{(N+h) \ \cos\varphi}{\lrsb{N(1-e^2)+h} \, \sin\varphi} \ \ \ \ \lambda \ \ \ \ \sqrt{(N+h)^2 \, \cos^2 \varphi + \lrsb{N(1-e^2)+h}^2 \, \sin^2 \varphi} }^T} \label{eq:RefSystems_E_geod2geoc}
\end{eqnarray}

        
\subsubsection{Geodetic Coordinates}\label{subsubsec:RefSystems_E_GeodeticCoord}

The \emph{geodetic coordinates} \nm{\TEgdt = \lrsb{\lambda \ \ \varphi \ \ h}^T} represent the most common means of designating position with respect to \nm{\FE}. Depicted in figure \ref{fig:RefSystems_N}, they are defined as follows:
\begin{itemize}
\item \emph{Longitude} \nm{\lambda} \nm{\lrp{0 \leq \lambda < 2\pi }} is the angle formed between the Greenwich meridian plane (formed by \nm{\iEi} and \nm{\iEiii}) and the point meridian plane\footnote{It coincides with the geocentric longitude defined in section \ref{subsubsec:RefSystems_E_GeocentricCoord}.}.
\item \emph{Latitude} \nm{\varphi} \nm{\lrp{-\pi/2 \leq \varphi \leq \pi/2 }} is the angle formed between the Equator plane and the line passing through the point that is orthogonal to the ellipsoid surface at the point where it intersects it.
\item \emph{Geodetic altitude} \emph{h} is the distance between the ellipsoid surface and the point measured along a line that is orthogonal to the ellipsoid surface at the point where it intersects it.
\end{itemize}
 
The geodetic coordinates are more intuitive than the Cartesian or geocentric ones for geopositioning. By using one distance (h) and two angles (\nm{\lambda} and \nm{\varphi}), the geodetic coordinates take advantage of the ellipsoidal Earth surface described in section \ref{sec:EarthModel_WGS84} (which represents the surface of zero geodetic altitude h) to define the position with respect to the Earth via three coordinates that posses a clear physical meaning. The use of geodetic coordinates is very convenient to express the position of any point placed near the Earth surface. 
                	 
As the ellipsoid radius of curvature of the prime vertical N depends on the latitude per (\ref{eq:EarthModel_WGS84_ELL_N}), it is not possible to reverse (\ref{eq:RefSystems_E_geod2car}) or (\ref{eq:RefSystems_E_geod2geoc}) to obtain explicit expressions providing the geodetic coordinates based on the Cartesian or geocentric ones. They can however be unambiguously obtained through iteration.


\subsection{Spherical Frame}\label{subsec:RefSystems_S}

The spherical (\texttt{S}) frame is a Cartesian frame \nm{\FS = \{\OS,\,\iStheta,\,\iSlambda,\,\iSr\}} (figure \ref{fig:RefSystems_S}) whose main purpose is to express the gravitational acceleration (section \ref{subsec:EarthModel_EGM96}). It is defined by having \nm{\OS} located at the aircraft center of mass, \nm{\iSr} parallel to the line connecting the origin with the Earth center of mass pointing away from the Earth, \nm{\iSlambda} orthogonal to \nm{\iSr} and parallel to the Equator plain, pointing toward geodetic East, and \nm{\iStheta} orthogonal to \nm{\iSr} and \nm{\iSlambda} forming a right handed system.
\begin{figure}[h]
\centering
\begin{tikzpicture}[auto,>=latex',scale=1.0]
	\coordinate (origin) at (+0.0,+0.0);
	\coordinate (origin1) at (+0.0,+1.6);
	\coordinate (origin2) at (+0.0,+2.2);
		
	\filldraw [black] (origin) circle [radius=2pt] node [left=1pt] {\nm{\OECEF}};
	\draw [name path = ellhor] (origin) ellipse [x radius=4.0, y radius=1.65];
	\draw [dashed, ultra thin, name path = ellhor2, dashed] (origin1) ellipse [x radius=(4.0/1.19), y radius=(1.25/1.19)];
	\draw [name path = ellver] (origin) ellipse [x radius=1.3, y radius=3.0];
	\draw [name path = ellmain, very thick] (origin) ellipse [x radius=4.0, y radius=3.0];
	
	\draw [name path = iECEFi, name intersections={of=ellhor and ellver, by={int1,int2,int3,int4}}, very thick] [->] (origin) -- ($(origin) + 1.25*(int3)$) node [below=3pt] {\nm{\iEi}};
	\draw [name path = iECEFii, very thick] [->] (origin) -- (+4.5,+0.0) node [right=3pt]  {\nm{\iEii}};
	\draw [name path = iECEFiii, very thick] [->] (origin) -- (+0.0,+3.5) node [right=4pt]  {\nm{\iEiii}};
	
	\path [red, name path = path1] [-] (origin1) -- (+3.4,-0.6);
	\path [name intersections={of=path1 and ellhor2, by={pointQ}}] [-] (origin) -- (pointQ);
	\draw [dashed, ultra thin, name path = path11] [-] (origin1) -- (pointQ);
	\filldraw [black] (pointQ) circle [radius=1pt];
	
	\path [red, name path = path2] [-] (origin2) -- (+3.4,-0.0);
	\path [red, name path = path3] [-] (origin) -- ($(origin) + 6/3*(pointQ)$);
	\draw [blue, name path = pathP, name intersections={of=path2 and path3, by={pointP}}, ultra thick] [->] (origin) -- (pointP) node [pos=0.5, below=0pt] {\nm{r}} node[black, above=4pt] {\nm{\OS}};
	\draw [dashed, ultra thin, name path = path21] [-] (origin2) -- (pointP);
	\filldraw [black] (pointP) circle [radius=1pt];
		
	\path [red, name path = path4] [-] (origin) -- (+3.4,-2.2);
	\path [red, name path = path5] [-] (pointQ) ++(0,-2) -- (pointQ);
	\draw [dashed, ultra thin, name intersections={of=path4 and path5, by={pointq}}] [-] (pointQ) -- (pointq);
	\path [red, name path = path6] [-] (pointP) ++(0,-2.5) -- (pointP);
	\draw [dashed, ultra thin, name intersections={of=path4 and path6, by={pointp}}] [-] (origin) -- (pointp);	
	\draw [dashed, ultra thin, name path = path22] [-] (pointp) -- (pointP);
	
	\path [red, name path = path7] [-] (-1.2,-0.7) -- (+1.0,-0.3);
	\draw [blue, name intersections={of=iECEFi and path7, by={lambda1}}, name intersections={of=path4 and path7, by={lambda2}}, very thick] [->] (lambda1) .. controls (-0.2,-0.65) and (0.15,-0.55) .. (lambda2) node [pos=0.5, below=0pt] {\nm{\lambda}};
			
	\path [red, name path = path8] [-] (-0.5,+1.4) -- (+0.8,+0.2);
	\draw [blue, name intersections={of=iECEFiii and path8, by={theta1}}, name intersections={of=pathP and path8, by={theta2}}, very thick] [->] (theta1) .. controls (0.3,0.9) and (0.45,0.8) .. (theta2) node [pos=0.5, above=0pt] {\nm{\theta}};
	
	\draw [ultra thick] [->] (pointP) -- ($(pointQ)!3.0!(pointP)$) node [above left=0pt]  {\nm{\iSr}};
	\draw [ultra thick] [<-] (pointP) ++(0.9,0.2) node [below=2pt]  {\nm{\iSlambda}} -- (pointP);
	\draw [ultra thick] [<-] (pointP) ++(0.4,-0.6) node [right=0pt]  {\nm{\iStheta}} -- (pointP);
		
	\node [left of=origin, node distance=2.4cm] (Greenwich) {\st{Greenwich Mer.}};	
	\coordinate (origin3) at (+0.0,+1.2);
	\node [left of=origin3, node distance=4.6cm] (Ellipsoid) {\st{\hypertt{WGS84} Ell.}};	
	\coordinate (origin4) at (+0.0,-1.0);
	\node [left of=origin4, node distance=2.4cm] (Equator) {\st{Equator}};		
	
\end{tikzpicture}
\caption{Spherical frame \nm{\FS} and geocentric coordinates \nm{\TEgct}}
\label{fig:RefSystems_S}
\end{figure}

If \nm{\TESE = \TEcar} contains the Cartesian coordinates of \nm{\OS} with respect to the \hypertt{ECEF} frame, the \nm{\FS} frame can be obtained from \nm{\FE} by first executing two consecutive rotations followed by a \nm{\TESE} translation. The two rotations are the following:
\begin{enumerate}
\item a rotation angle \nm{\lambda} (longitude) about \nm{\iEiii}. 
\item a rotation angle \nm{\theta} (colatitude) about \nm{\iSlambda} (the orientation of \nm{\iEii} after the previous rotation).
\end{enumerate}
\begin{eqnarray}
\nm{\vec q_{\sss ES}} & \nm{\equiv} & \nm{\vec R_{\sss ES} = \vec R_3(\lambda) \ \vec R_2(\theta)} \label{eq:RefSystems_qES} \\
\nm{\vec \zeta_{\sss ES}} & = & \nm{\vec q_{\sss ES} + \dfrac{\epsilon}{2} \, \TESE \otimes \vec q_{\sss ES}} \label{eq:RefSystems_zetaES}
\end{eqnarray}


\subsection{North East Down Frame}\label{subsec:RefSystems_N}

The North East Down (\hypertt{NED} or \texttt{N}) frame is a Cartesian frame \nm{\FN = \{\ON,\,\iNi,\,\iNii,\,\iNiii\}}  (figure \ref{fig:RefSystems_N}) very useful for navigation because its three axes are oriented along easily understood directions. It is defined by \nm{\ON} being located at the aircraft center of mass, \nm{\iNiii} orthogonal to the ellipsoid surface pointing toward nadir, \nm{\iNii} orthogonal to \nm{\iNiii} and parallel to the Equator plain, pointing towards geodetic East, and \nm{\iNi} orthogonal to the former ones and pointing towards geodetic North in such a way that they form a right handed system.
\begin{figure}[h]
\centering
\begin{tikzpicture}[auto,>=latex',scale=1.0]
	\coordinate (origin) at (+0.0,+0.0);
	\coordinate (origin1) at (+0.0,+1.6);
	\coordinate (origin2) at (+0.0,+2.2);
	\coordinate (origin3) at (+0.0,-0.5);
			
	\filldraw [black] (origin) circle [radius=2pt] node [left=1pt] {\nm{\OECEF}};
	\draw [name path = ellhor] (origin) ellipse [x radius=4.0, y radius=1.65];
	\draw [name path = ellhor2, dashed] (origin1) ellipse [x radius=(4.0/1.19), y radius=(1.25/1.19)];
	\draw [name path = ellver] (origin) ellipse [x radius=1.3, y radius=3.0];
	\draw [name path = ellmain, very thick] (origin) ellipse [x radius=4.0, y radius=3.0];
	
	\draw [name path = iECEFi, name intersections={of=ellhor and ellver, by={int1,int2,int3,int4}}, very thick] [->] (origin) -- ($(origin) + 1.25*(int3)$) node [below=3pt] {\nm{\iEi}};
	\draw [name path = iECEFii, very thick] [->] (origin) -- (+4.5,+0.0) node [right=3pt]  {\nm{\iEii}};
	\draw [name path = iECEFiii, very thick] [->] (origin) -- (+0.0,+3.5) node [right=4pt]  {\nm{\iEiii}};
	\draw [dashed, ultra thin] (origin) -- (origin3);
	\filldraw [black] (origin3) circle [radius=1pt];
	
	\path [red, name path = path1] [-] (origin1) -- (+3.4,-0.6);
	\draw [dashed, ultra thin, name path = path12, name intersections={of=path1 and ellhor2, by={pointQ}}] [-] (origin3) -- (pointQ);
	\draw [dashed, ultra thin, name path = path11] [-] (origin1) -- (pointQ);
	\filldraw [black] (pointQ) circle [radius=1pt];
	
	\path [red, name path = path2] [-] (origin2) -- (+3.4,-0.0);
	\path [red, name path = path3] [-] (origin3) -- ($(origin3)!1.4!(pointQ)$);
	\path [name path = pathP, name intersections={of=path2 and path3, by={pointP}}] [->] (pointQ) -- (pointP);
	\path node at ($(pointP) + (0.2,0.3)$) {\nm{\ON}};
	
	\draw [dashed, ultra thin, name path = path21] [-] (origin2) -- (pointP);
	\filldraw [black] (pointP) circle [radius=1pt];
		
	\path [red, name path = path4] [-] (origin) -- (+3.4,-2.2);
	\path [red, name path = path5] [-] (pointQ) ++(0,-2) -- (pointQ);
	\draw [dashed, ultra thin, name intersections={of=path4 and path5, by={pointq}}] [-] (pointQ) -- (pointq);
	\path [red, name path = path6] [-] (pointP) ++(0,-2.5) -- (pointP);
	\draw [dashed, ultra thin, name path = path66, name intersections={of=path4 and path6, by={pointp}}] [-] (origin) -- (pointp);	
	\draw [dashed, ultra thin, name path = path22] [-] (pointp) -- (pointP);
	
	\path [red, name path = path7] [-] (-1.2,-0.7) -- (+1.0,-0.3);
	\draw [blue, name intersections={of=iECEFi and path7, by={lambda1}}, name intersections={of=path4 and path7, by={lambda2}}, very thick] [->] (lambda1) .. controls (-0.2,-0.65) and (0.15,-0.55) .. (lambda2) node [pos=0.7, below=0pt] {\nm{\lambda}};
			
	\path [red, name path = path8] [-] (+0.9,-0.8) -- (+0.85,+0.8);
	\draw [blue, name intersections={of=path66 and path8, by={phi1}}, name intersections={of=path12 and path8, by={phi2}}, very thick] [->] (phi1) .. controls (1.0,-0.4) and (1.0,-0.2) .. (phi2) node [pos=0.5, right=-2pt] {\nm{\varphi}};
	
	\draw [ultra thick] [->] (pointP) -- ($(pointP)!2.0!(pointQ)$) node [left =0pt]  {\nm{\iNiii}};
	\draw [ultra thick] [<-] (pointP) ++(0.9,0.2) node [above=0pt]  {\nm{\iNii}} -- (pointP);
	\draw [ultra thick] [<-] (pointP) ++(-0.4,+0.6) node [above=-2pt]  {\nm{\iNi}} -- (pointP);
		
	\draw [blue, ultra thick] [->] (pointQ) -- (pointP) node [pos=0.5, above left] {\nm{h}};
	
	\node [left of=origin, node distance=2.4cm] (Greenwich) {\st{Greenwich Mer.}};	
	\coordinate (origin3) at (+0.0,+1.2);
	\node [left of=origin3, node distance=4.6cm] (Ellipsoid) {\st{\hypertt{WGS84} Ell.}};	
	\coordinate (origin4) at (+0.0,-1.0);
	\node [left of=origin4, node distance=2.4cm] (Equator) {\st{Equator}};		
	
\end{tikzpicture}
\caption{\texttt{NED} frame \nm{\FE} and geodetic coordinates \nm{\TEgdt}}
\label{fig:RefSystems_N}
\end{figure}

If \nm{\TENE = \TEcar} contains the Cartesian coordinates of \nm{\ON} with respect to the \hypertt{ECEF} frame, the \nm{\FN} frame can be obtained from \nm{\FE} by first executing two consecutive rotations followed by a \nm{\TENE} translation. The two rotations are the following:
\begin{enumerate}
	\item a rotation angle \nm{\lambda} about \nm{\iEiii}.
	\item a rotation angle \nm{-\pi/2 - \varphi} about \nm{\iNii} (the orientation of \nm{\iEii} after the previous rotation).  
\end{enumerate}
\begin{eqnarray}
\nm{\qEN} & \nm{\equiv} & \nm{\REN = \vec R_3(\lambda) \ \vec R_2(-\frac{\pi}{2}-\varphi)} \label{eq:RefSystems_qEN} \\
\nm{\zetaEN} & = & \nm{\qEN + \dfrac{\epsilon}{2} \, \TENE \otimes \qEN} \label{eq:RefSystems_zetaEN}
\end{eqnarray}

As they share the same origin (\nm{\vec T_{\sss SN}^{\sss S} = \vec 0_3}), \nm{\FN} can also be obtained from \nm{\FS} by executing a single rotation of \nm{- \theta - \pi/2 - \varphi} about \nm{\iSlambda}:
\begin{eqnarray}
\nm{\vec q_{\sss SN}} & \nm{\equiv} & \nm{\vec R_{\sss SN} = \vec R_2(- \theta -\frac{\pi}{2}-\varphi)} \label{eq:RefSystems_qSN} \\
\nm{\vec \zeta_{\sss SN}} & = & \nm{\vec q_{\sss SN} + \dfrac{\epsilon}{2} \, \vec T_{\sss SN}^{\sss S} \otimes \vec q_{\sss SN} = \vec q_{\sss SN}} \label{eq:RefSystems_zetaSN}
\end{eqnarray}


\section{Aircraft Focused Frames}\label{sec:RefSystems_Acft}

The coordinate frames defined in this section are employed to describe the position of any point as well as the attitude or orientation of any vector with respect to the aircraft. 
\begin{figure}[h]
\centering
\begin{tikzpicture}[auto,>=latex',scale=1.0]
	\coordinate (origin)     at (+0.0,+0.0);
	\coordinate (iNi)      at (+180:4.0);
	\coordinate (iNii)     at (+45:3.0);
	\coordinate (iNiii)    at (-90:4.0);
	\coordinate (iTEMPi)     at (+168:3.3);
	\coordinate (iTEMPii)    at (+28:3.7);
	\coordinate (iTEMPiii)   at (-105:3.4);
	\coordinate (iBi)      at (+153:3.6);
	\coordinate (iBii)     at (+15:3.1);
	\coordinate (iBiii)    at (-117:4.0);
	\coordinate (turn_phi3)  at (-90:2.5);
	\coordinate (turn_phi2)  at (+32:2.2);
	\coordinate (turn_phi1)    at (+153:2.2);
				
	\coordinate (aNi)    at ($(origin)!0.95!(iNi)$);
	\coordinate (aNii)   at ($(origin)!0.95!(iNii)$);
	\coordinate (aNiii)  at ($(origin)!0.95!(iNiii)$);
	\coordinate (aTEMPi)   at ($(origin)!0.95!(iTEMPi)$);
	\coordinate (aTEMPii)  at ($(origin)!0.95!(iTEMPii)$);
	\coordinate (aTEMPiii) at ($(origin)!0.95!(iTEMPiii)$);
	\coordinate (aBi)    at ($(origin)!0.95!(iBi)$);
	\coordinate (aBii)   at ($(origin)!0.95!(iBii)$);
	\coordinate (aBiii)  at ($(origin)!0.95!(iBiii)$);
	
	\fill [color=gray!10] (origin) -- (aNi) .. controls ($(aNi)!0.5!10:(aTEMPi)$) .. (aTEMPi) -- cycle;
	\fill [color=gray!10] (origin) -- (aNii) .. controls ($(aNii)!0.5!10:(aTEMPii)$) .. (aTEMPii) -- cycle;
	\fill [color=gray!35] (origin) -- (aNiii) .. controls ($(aNiii)!0.5!10:(aTEMPiii)$) .. (aTEMPiii) -- cycle;
	\fill [color=gray!35] (origin) -- (aTEMPi) .. controls ($(aTEMPi)!0.5!10:(aBi)$) .. (aBi) -- cycle;
	\fill [color=gray!60] (origin) -- (aTEMPii) .. controls ($(aTEMPii)!0.5!10:(aBii)$) .. (aBii) -- cycle;
	\fill [color=gray!60] (origin) -- (aTEMPiii) .. controls ($(aTEMPiii)!0.5!10:(aBiii)$) .. (aBiii) -- cycle;
			
	\filldraw [black] (origin) circle [radius=2pt] node [below right] {\nm{\ON}};
	\draw [ultra thick] [->] (origin) -- (iNi)   node [below=3pt] {\nm{\iNi}};
	\draw [ultra thick] [->] (origin) -- (iNii)  node [left=5pt] {\nm{\iNii}};
	\draw [ultra thick] [->] (origin) -- (iNiii) node [above right] {\nm{\iNiii}};
	
	\draw [dashed, ultra thick] (origin) -- (iTEMPi);
	\draw [dashed, ultra thick] (origin) -- (iTEMPii);
	\draw [dashed, ultra thick] (origin) -- (iTEMPiii);
	\draw [->] ($(iNi)!0.1!(iTEMPi)$) .. controls ($(iNi)!0.5!10:(iTEMPi)$) .. ($(iNi)!0.95!(iTEMPi)$) node [pos=0.7, left=3pt] {\nm{\phi_3}};
	\draw [->] ($(iNii)!0.1!(iTEMPii)$) .. controls ($(iNii)!0.5!5:(iTEMPii)$) .. ($(iNii)!0.95!(iTEMPii)$) node [pos=0.6, above] {\nm{\phi_3}};
	\draw [->] ($(iNiii)!0.1!(iTEMPiii)$) .. controls ($(iNiii)!0.5!5:(iTEMPiii)$) .. ($(iNiii)!0.95!(iTEMPiii)$) node [pos=0.6, below=2pt] {\nm{\phi_2}};

	\draw [ultra thick] [->] (origin) -- (iBi) node [left] {\nm{\vec i_1}};
	\draw [ultra thick] [->] (origin) -- (iBii) node [below=3pt] {\nm{\vec i_2}};
	\draw [ultra thick] [->] (origin) -- (iBiii) node [left] {\nm{\vec i_3}};
	\draw [->] ($(iTEMPi)!0.1!(iBi)$) .. controls ($(iTEMPi)!0.5!5:(iBi)$) .. ($(iTEMPi)!0.95!(iBi)$) node [pos=0.5, left] {\nm{\phi_2}};
	\draw [->] ($(iTEMPii)!0.1!(iBii)$) .. controls ($(iTEMPii)!0.5!5:(iBii)$) .. ($(iTEMPii)!0.95!(iBii)$) node [pos=0.5, right=1pt] {\nm{\phi_1}};
	\draw [->] ($(iTEMPiii)!0.1!(iBiii)$) .. controls ($(iTEMPiii)!0.5!5:(iBiii)$) .. ($(iTEMPiii)!0.95!(iBiii)$) node [pos=0.4, below] {\nm{\phi_1}};
		
	\draw [blue, thick,->] (turn_phi3) arc [x radius=0.4, y radius=0.2, start angle=90, end angle=15] node [right] {\nm{\phi_3}};
	\draw [blue, thick,->] (turn_phi3) arc [x radius=0.4, y radius=0.2, start angle=90, end angle=-75];
	\draw [blue, thick,->] (turn_phi3) arc [x radius=0.4, y radius=0.2, start angle=90, end angle=-145];
	\draw [blue, thick,->] (turn_phi2) arc [x radius=0.22, y radius=0.3, start angle=160, end angle=75];			
	\draw [blue, thick,->] (turn_phi2) arc [x radius=0.22, y radius=0.3, start angle=160, end angle=-5] node [right] {\nm{\phi_2}};	
	\draw [blue, thick,->] (turn_phi2) arc [x radius=0.22, y radius=0.3, start angle=160, end angle=-90];		
	\draw [blue, thick,<-<] (turn_phi1) node [above right] {\nm{\phi_1}} arc [x radius=0.2, y radius=0.4, start angle=0, end angle=125];
	\draw [blue, thick,-<] (turn_phi1) arc [x radius=0.2, y radius=0.4, start angle=0, end angle=200];
	\draw [blue, thick,-] (turn_phi1) arc [x radius=0.2, y radius=0.4, start angle=0, end angle=240];
	
\end{tikzpicture}
\caption{Euler angles \nm{3-2-1} from \texttt{NED}}
\label{fig:RefSystems_N321}
\end{figure}

With the sole exception of the reference frame (section \ref{subsec:RefSystems_R}), these frames are centered in the aircraft center of mass, and hence their poses with respect to the \hypertt{NED} frame \nm{\FN} can be defined exclusively by a rotational motion \cite{LIE}. By convention, the chosen representation to define the attitude of these frames with respect to \nm{\FN} are the Euler angles \nm{\vec \phi = \lrsb{\phi_3 \ \ \phi_2 \ \ \phi_1}^T}, this is, three successive rotations graphically shown in figure \ref{fig:RefSystems_N321} and which shall always be executed in the following order:
\begin{enumerate}
\item a rotation angle \nm{\phi_3} (yaw) about \nm{\iNiii}.
\item a rotation angle \nm{\phi_2} (pitch) about \nm{\textbf i_2'} (the orientation of \nm{\iNii} after the previous rotation).
\item a rotation angle \nm{\phi_1} (bank or roll) about {\nm{\vec i_1}} (the orientation of \nm{\iNi} after the previous rotations).
\end{enumerate}


\subsection{Body Frame}\label{subsec:RefSystems_B}

The body (\texttt{B}) frame \cite{Farrell2008,Groves2008,Chatfield1997,Rogers2007,Etkin1972,Ashley1974,Miele1962,Titterton2004} is a Cartesian frame \nm{\FB = \{\OB,\,\iBi,\,\iBii,\,\iBiii\}} fixed to the aircraft, defined by having \nm{\OB} located at the aircraft center of mass, \nm{\iBi} contained in the plane of symmetry of the aircraft pointing forward along a fixed direction, \nm{\iBiii} contained in the plane of symmetry of the aircraft, normal to \nm{\iBi}, and pointing downward, and \nm{\iBii} orthogonal to \nm{\iBi} and \nm{\iBiii}, in such a way that they form a right handed system.

The body Euler angles \nm{\phiNB = \lrsb{\psi \ \ \theta \ \ \xi}^T} follow the rotation schedule shown in section \ref{sec:RefSystems_Acft} and describe the \nm{\FB} attitude with respect to \nm{\FN}. Their physical meaning is the following:
\begin{itemize}
\item The aircraft \emph{heading} or \emph{body yaw angle} \nm{\psi} represents the angle that exists between the geodetic North represented by \nm{\iNi} and the projection of \nm{\iBi} (fuselage) onto the horizontal plane defined by \nm{\iNi} and \nm{\iNii}.
\item The \emph{body pitch angle} \nm{\theta} represents the angle existing between \nm{\iBi} (aircraft fuselage) and its projection onto the horizontal plane.
\item The \emph{body bank angle} \nm{\xi} represents the angle existing between \nm{\iBii} (normal to the aircraft plane of symmetry) and the intersection between the plane defined by \nm{\iBii} and \nm{\iBiii}  (normal to fuselage) with the horizontal plane.
\end{itemize}

As \nm{\FB} and \nm{\FN} share the same origin (\nm{\TNBN = \vec 0_3}), \nm{\FB} can be obtained from \nm{\FN} as follows:
\begin{eqnarray}
\nm{\qNB} & \nm{\equiv} & \nm{\RNB = \vec R_3(\psi) \ \vec R_2(\theta) \ \vec R_1(\xi) \equiv \phiNB} \label{eq:RefSystems_qNB} \\
\nm{\zetaNB} & = & \nm{\qNB + \dfrac{\epsilon}{2} \, \TNBN \otimes \qNB = \qNB} \label{eq:RefSystems_zetaNB}
\end{eqnarray}

The relative pose between \nm{\FE} and \nm{\FB} results in:
\neweq{\zetaEB = \zetaEN \otimes \zetaNB} {eq:RefSystems_zetaEB}


\subsection{Wind Frame}\label{subsec:RefSystems_W}

The wind (\texttt{W}) frame \cite{Etkin1972, Ashley1974, Miele1962} is a Cartesian frame \nm{\FW = \{\OW,\,\iWi,\,\iWii,\,\iWiii\}} by having \nm{\OW} located at the aircraft center of mass, \nm{\iWi} directed at every moment along the aircraft air velocity and looking forward, \nm{\iWiii} contained in the plane of symmetry of the aircraft, orthogonal to \nm{\iWi}, and pointing downward, and \nm{\iWii} normal to \nm{\iWi} and \nm{\iWiii} in such a way that they form a right handed system.

The wind frame is necessary to properly express the aerodynamic forces, as the aircraft drag acts in a direction opposite \nm{\iWi} (air velocity), and the aircraft lift acts in a direction opposite \nm{\iWiii}. As noted in its definition above, the air velocity \nm{\vTAS} (section \ref{sec:EquationsMotion_velocity}) does not have any component along the \second\ and \third\ axes. If the body frame can be considered as a geometric representation of the aircraft structure because it is rigidly attached to it, the wind frame represents the aircraft from an aerodynamic point of view, describing the directions in which the air flow reaches the aircraft.
 
The airspeed Euler angles \nm{\phiNW = \lrsb{\chiTAS \ \ \gammaTAS \ \ \muTAS}^T} follow the rotation schedule described in section \ref{sec:RefSystems_Acft}, and their physical meaning is the following:
\begin{itemize}
\item The \emph{aerodynamic heading} or \emph{airspeed yaw angle} \nm{\chiTAS} represents the angle that exists between \nm{\iNi} (geodetic North) and the projection of \nm{\iWi} (airspeed) onto the horizontal plane (defined by \nm{\iNi} and \nm{\iNii}).
\item The \emph{aerodynamic path angle} or \emph{airspeed pitch angle} \nm{\gammaTAS} represents the angle existing between \nm{\iWi} (airspeed) and its projection onto the horizontal plane.
\item The \emph{airspeed bank angle} \nm{\muTAS} represents the angle existing between \nm{\iWii} and the intersection between the plane defined by \nm{\iWii} and \nm{\iWiii} (normal to airspeed) with the horizontal plane.
\end{itemize}

As \nm{\FW} and \nm{\FN} share the same origin (\nm{\vec T_{\sss NW}^{\sss N} = \vec 0_3}), \nm{\FW} can be obtained from \nm{\FN} as follows:
\begin{eqnarray}
\nm{\vec q_{\sss NW}} & \nm{\equiv} & \nm{\RNW = \vec R_3(\chiTAS) \ \vec R_2(\gammaTAS) \ \vec R_1(\muTAS) \equiv \phiNW} \label{eq:RefSystems_qNW} \\
\nm{\vec \zeta_{\sss NW}} & = & \nm{\vec q_{\sss NW} + \dfrac{\epsilon}{2} \, \vec T_{\sss NW}^{\sss N} \otimes \vec q_{\sss NW} = \vec q_{\sss NW}} \label{eq:RefSystems_zetaNW}
\end{eqnarray}

The relative pose between \nm{\FE} and \nm{\FW} results in:
\neweq{\vec \zeta_{\sss EW} = \vec \zeta_{\sss EN} \otimes \vec \zeta_{\sss NW}} {eq:RefSystems_zetaEW}

\begin{figure}[h]
\centering
\begin{tikzpicture}[auto,>=latex',scale=1.0]
	\coordinate (origin)     at (+0.0,+0.0);
	\coordinate (iWi)      at (+180:4.0);
	\coordinate (iWii)     at (+45:3.0);
	\coordinate (iWiii)    at (-90:4.0);
	\coordinate (iTEMPi)     at (+168:3.3);
	\coordinate (iBi)      at (+153:3.6);
	\coordinate (iBii)     at (+28:3.7);
	\coordinate (iBiii)    at (-105:3.4);
	\coordinate (turn_beta)  at (-90:2.5);
	\coordinate (turn_alpha) at (+32:2.2);
				
	\coordinate (aWi)    at ($(origin)!0.95!(iWi)$);
	\coordinate (aWii)   at ($(origin)!0.95!(iWii)$);
	\coordinate (aWiii)  at ($(origin)!0.95!(iWiii)$);
	\coordinate (aTEMPi)   at ($(origin)!0.95!(iTEMPi)$);
	\coordinate (aBi)    at ($(origin)!0.95!(iBi)$);
	\coordinate (aBii)   at ($(origin)!0.95!(iBii)$);
	\coordinate (aBiii)  at ($(origin)!0.95!(iBiii)$);
	
	\fill [color=gray!10] (origin) -- (aWi) .. controls ($(aWi)!0.5!10:(aTEMPi)$) .. (aTEMPi) -- cycle;
	\fill [color=gray!10] (origin) -- (aWii) .. controls ($(aWii)!0.5!10:(aBii)$) .. (aBii) -- cycle;
	\fill [color=gray!35] (origin) -- (aWiii) .. controls ($(aWiii)!0.5!10:(aBiii)$) .. (aBiii) -- cycle;
	\fill [color=gray!35] (origin) -- (aTEMPi) .. controls ($(aTEMPi)!0.5!10:(aBi)$) .. (aBi) -- cycle;
			
	\filldraw [black] (origin) circle [radius=2pt] node [below right] {\nm{\OW = \OB}};
	\draw [ultra thick] [->] (origin) -- (iWi)   node [below=3pt] {\nm{\iWi}};
	\draw [ultra thick] [->] (origin) -- (iWii)  node [left=5pt] {\nm{\iWii}};
	\draw [ultra thick] [->] (origin) -- (iWiii) node [above right] {\nm{\iWiii}};
	
	\draw [dashed, ultra thick] (origin) -- (iTEMPi);
	\draw [->] ($(iWi)!0.1!(iTEMPi)$) .. controls ($(iWi)!0.5!10:(iTEMPi)$) .. ($(iWi)!0.95!(iTEMPi)$) node [pos=0.6, left=2pt] {\nm{- \beta}};

	\draw [ultra thick] [->] (origin) -- (iBi) node [left] {\nm{\iBi}};
	\draw [ultra thick] [->] (origin) -- (iBii) node [below=5pt] {\nm{\iBii}};
	\draw [ultra thick] [->] (origin) -- (iBiii) node [left] {\nm{\iBiii}};
	\draw [->] ($(iTEMPi)!0.1!(iBi)$) .. controls ($(iTEMPi)!0.5!5:(iBi)$) .. ($(iTEMPi)!0.95!(iBi)$) node [pos=0.5, left=2pt] {\nm{\alpha}};
	\draw [->] ($(iWii)!0.1!(iBii)$) .. controls ($(iWii)!0.5!5:(iBii)$) .. ($(iWii)!0.95!(iBii)$) node [pos=0.5, above right] {\nm{- \beta}};
	\draw [->] ($(iWiii)!0.1!(iBiii)$) .. controls ($(iWiii)!0.5!5:(iBiii)$) .. ($(iWiii)!0.95!(iBiii)$) node [pos=0.6, below] {\nm{\alpha}};
		
	\draw [blue, thick,->] (turn_beta) arc [x radius=0.4, y radius=0.2, start angle=90, end angle=15] node [right] {\nm{- \beta}};
	\draw [blue, thick,->] (turn_beta) arc [x radius=0.4, y radius=0.2, start angle=90, end angle=-75];
	\draw [blue, thick,->] (turn_beta) arc [x radius=0.4, y radius=0.2, start angle=90, end angle=-145];
	\draw [blue, thick,->] (turn_alpha) arc [x radius=0.22, y radius=0.3, start angle=160, end angle=75];			
	\draw [blue, thick,->] (turn_alpha) arc [x radius=0.22, y radius=0.3, start angle=160, end angle=-5] node [right] {\nm{\alpha}};	
	\draw [blue, thick,->] (turn_alpha) arc [x radius=0.22, y radius=0.3, start angle=160, end angle=-90];		
	\draw [blue, thick,->] (turn_alpha) arc [x radius=0.22, y radius=0.3, start angle=160, end angle=-150];
	
\end{tikzpicture}
\caption{Wind to body conversion}
\label{fig:RefSystems_W2B}
\end{figure}
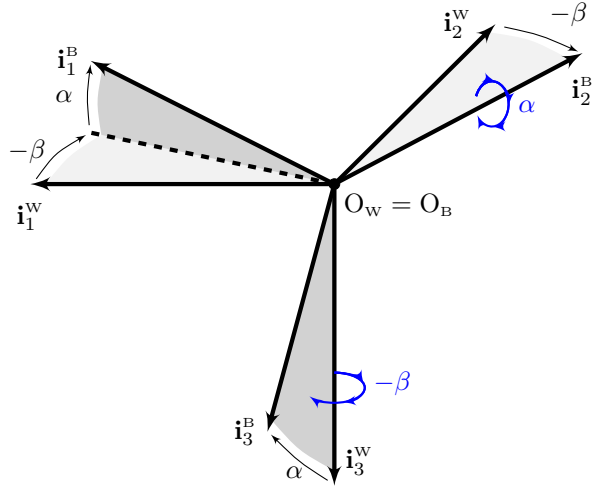

The obtainment of \nm{\FB} from \nm{\FW}, represented in figure \ref{fig:RefSystems_W2B}, requires only two rotations because \nm{\iBi}, \nm{\iBiii} and \nm{\iWiii} are all contained in the aircraft plane of symmetry:
\begin{enumerate}
\item a rotation angle \nm{- \beta} (sideslip) about \nm{\iWiii}.
\item a rotation angle \nm{\alpha} (angle of attack) about \nm{\iBii}.
\end{enumerate}

The angles of attack and sideslip, \nm{\phiWB = \lrsb{- \beta \ \ \alpha \ \ 0}^T},  represent the orientation of the air flow with respect to the aircraft structure, and as such play a significant role in the aerodynamic forces described in section \ref{sec:AircraftModel_aerodynamics}:
\begin{itemize}
\item The \emph{sideslip angle} \nm{\beta} represents the angle existing between \nm{\iWi} (air velocity) and its projection onto the aircraft plane of symmetry (defined by \nm{\iBi}, \nm{\iBiii}, and \nm{\iWiii}).
\item The \emph{angle of attack} \nm{\alpha} represents the angle that exists between the projection of \nm{\iWi} (air velocity) onto the aircraft plane of symmetry and \nm{\iBi} (aircraft fuselage).
\end{itemize}

As \nm{\FB} and \nm{\FW} share the same origin (\nm{\vec T_{\sss WB}^{\sss W} = \vec 0_3}), \nm{\FB} can be obtained from \nm{\FW} as follows:
\begin{eqnarray}
\nm{\vec q_{\sss WB}} & \nm{\equiv} & \nm{\RWB = \vec R_3(-\beta) \ \vec R_2(\alpha) \equiv \phiWB} \label{eq:RefSystems_qWB} \\
\nm{\vec \zeta_{\sss WB}} & = & \nm{\vec q_{\sss WB} + \dfrac{\epsilon}{2} \, \vec T_{\sss WB}^{\sss W} \otimes \vec q_{\sss WB} = \vec q_{\sss WB}} \label{eq:RefSystems_zetaWB}
\end{eqnarray}
 

\subsection{Ground Velocity Frame}\label{subsec:RefSystems_G}

The ground velocity (\texttt{G}) frame is a Cartesian frame \nm{\FG = \{\OG,\,\iGi,\,\iGii,\,\iGiii\}} defined by having \nm{\OG} located at the aircraft center of mass, \nm{\iGi} directed at every moment along the aircraft ground velocity, looking forward, \nm{\iGiii} contained in the plane of symmetry of the aircraft, normal to \nm{\iGi}, and pointing downward, and \nm{\iGii} orthogonal to \nm{\iGi} and \nm{\iGiii} in such a way that they form a right handed system.

Although \nm{\FG} is not generally found in the literature, the author has chosen to include it as it is the best way to define the ground velocity Euler angles below. Another characteristic is that the ground velocity \nm{\vG} (section \ref{sec:EquationsMotion_velocity}) only contains one component when viewed in \nm{\FG}. 

The ground velocity Euler angles \nm{\phiNG = \lrsb{\chi \ \ \gamma \ \ \mu}^T} follow the rotation schedule described in section \ref{sec:RefSystems_Acft}, and their physical meaning is the following:
\begin{itemize}
\item The \emph{bearing} or \emph{ground velocity yaw angle} \nm{\chi} represents the angle that exists between \nm{\iNi} (geodetic North) and the projection of \nm{\iGi} (ground velocity) onto the horizontal plane (defined by \nm{\iNi} and \nm{\iNii}).
\item The \emph{path angle} or \emph{ground velocity pitch angle} \nm{\gamma} represents the angle existing between \nm{\iGi} (ground velocity) and its projection onto the horizontal plane.
\item The \emph{ground velocity bank angle} \nm{\mu} represents the angle existing between \nm{\iGii} and the intersection between the plane defined by \nm{\iGii} and \nm{\iGiii} (normal to ground velocity) with the horizontal plane.
\end{itemize}

As \nm{\FG} and \nm{\FN} share the same origin (\nm{\vec T_{\sss NG}^{\sss N} = \vec 0_3}), \nm{\FG} can be obtained from \nm{\FN} as follows:
\begin{eqnarray}
\nm{\qNG} & \nm{\equiv} & \nm{\RNG = \vec R_3(\chi) \ \vec R_2(\gamma) \ \vec R_1(\mu) \equiv \phiNG} \label{eq:RefSystems_qNG} \\
\nm{\zetaNG} & = & \nm{\qNG + \dfrac{\epsilon}{2} \, \TNGN \otimes \qNG = \qNG} \label{eq:RefSystems_zetaNG}
\end{eqnarray}

The relative pose between \nm{\FE} and \nm{\FG} results in:
\neweq{\zetaEG = \zetaEN \otimes \zetaNG} {eq:RefSystems_zetaEG}


\subsection{Turbulence Frame}\label{subsec:RefSystems_T}

The turbulence (\texttt{T}) frame is a Cartesian frame \nm{\FT = \{\OT ,\, \iTi, \, \iTii, \, \iTiii\}} centered at the aircraft center of mass, with \nm{\iTi} located on the horizontal plane (defined by \nm{\iNi} and \nm{\iNii}) and aligned with the nonturbulent wind field horizontal direction provided by the wind bearing \nm{\chiWIND} (section \ref{sec:EarthModel_WIND}), \nm{\iTiii} parallel to \nm{\iNiii}, and \nm{\iTii} forming a right handed system.

The only purpose of the turbulence frame \nm{\FT} is to facilitate the computation of the turbulent wind as described in section \ref{sec:EarthModel_WIND}. As \nm{\FT} and \nm{\FN} share the same origin (\nm{\vec T_{\sss NT}^{\sss N} = \vec 0_3}), \nm{\FT} can be obtained from \nm{\FN} by a single rotation \nm{\chiWIND} about \nm{\iNiii}:
\begin{eqnarray}
\nm{\vec q_{\sss NT}} & \nm{\equiv} & \nm{\vec R_{\sss NT} = \vec R_3(\chiWIND)} \label{eq:RefSystems_qNT} \\
\nm{\vec \zeta_{\sss NT}} & = & \nm{\vec q_{\sss NT} + \dfrac{\epsilon}{2} \, \vec T_{\sss NT}^{\sss N} \otimes \vec q_{\sss NT} = \vec q_{\sss NT}} \label{eq:RefSystems_zetaNT}
\end{eqnarray}

The relative pose between \nm{\FE} and \nm{\FT} results in:
\neweq{\vec \zeta_{\sss ET} = \vec \zeta_{\sss EN} \otimes \vec \zeta_{\sss NT}} {eq:RefSystems_zetaET}


\subsection{Reference Frame}\label{subsec:RefSystems_R}

The reference (\texttt{R}) frame is a Cartesian frame \nm{\FR = \{\OR ,\, \iRi, \, \iRii, \, \iRiii\}} employed in chapter \ref{cha:AircraftModel} as an intermediate step to obtain the aircraft aerodynamic and propulsive forces and moments. Its axes are parallel to those of the body frame \nm{\lrp{\iRi \parallel \iBi, \ \iRii \parallel \iBii, \ \iRiii \parallel \iBiii}}, but it is centered on a reference point (hence its name) that the author has arbitrarily located at the aircraft plane of symmetry, below its intersection with the wing trailing edge. The translation from \nm{\OR} to the aircraft center of mass \nm{\OB} is given by \nm{\TRBR}, defined in section \ref{sec:AircraftModel_MassInertia}.
\begin{eqnarray}
\nm{\vec q_{\sss RB}} & = & \nm{\vec{q_1} \equiv \vec R_{\sss RB} = {\vec I}_3} \label{eq:RefSystems_qRB} \\
\nm{\vec \zeta_{\sss RB}} & = & \nm{\vec q_{\sss RB} + \dfrac{\epsilon}{2} \, \vec T_{\sss RB}^{\sss R} \otimes \vec q_{\sss RB} = \vec{q_1} + \dfrac{\epsilon}{2} \, \vec T_{\sss RB}^{\sss R} \otimes \vec{q_1} = \vec{q_1} + \dfrac{\epsilon}{2} \, \vec T_{\sss RB}^{\sss R}} \label{eq:RefSystems_zetaRB}
\end{eqnarray}

\nm{\FB} and \nm{\FR} both represent the aircraft structure and describe its attitude. However, the aircraft center of mass \nm{\OB} slowly moves with respect to \nm{\OR} as the diminishing fuel load varies the aircraft mass distribution. The aircraft propeller and aerodynamic surfaces are fixed to the aircraft structure, and hence the forces and moments they generate are best viewed in \nm{\FR}. As the aircraft motion is based on the application of those forces and moments about the aircraft center of mass, they are better converted to \nm{\FB} when employed in the equations of motion of chapter \ref{cha:FlightPhysics}. 


\section{Sensor Focused Frames}\label{sec:RefSystems_Sensor}

These coordinate frames are necessary to interpret the measurements of some sensors described in chapter \ref{cha:Sensors}, such as the accelerometers, gyroscopes, and digital camera. Note that none are centered on the aircraft center of mass, but on other fuselage points fixed with respect to \nm{\FR}, and hence slowly moving with respect to \nm{\FB}.


\subsection{Platform Frame}\label{subsec:RefSystems_P}

The platform (\texttt{P}) frame \cite{Farrell2008,Chatfield1997,Rogers2007} is a Cartesian frame \nm{\FP = \{\OP,\,\iPi,\,\iPii,\,\iPiii\}} defined by having \nm{\OP} located at the \hypertt{IMU} reference point (section \ref{subsec:Sensors_Inertial_Mounting}); its axes \nm{\iPi}, \nm{\iPii}, \nm{\iPiii} form a right handed system that is loosely aligned with the \nm{\FB} body axes, meaning that they point in the general directions of the aircraft fuselage (forward), aircraft wings (rightwards), and downward, respectively.

A proper definition of \nm{\FP} is indispensable for inertial navigation, as the calibrated outputs of the accelerometers and gyroscopes are viewed on it (sections \ref{subsec:Sensors_Accelerometer_Triad_ErrorModel} and \ref{subsec:Sensors_Gyroscope_Triad_ErrorModel}). The \nm{\FP} frame can be obtained from \nm{\FB} by a rotation best described by the Euler angles \nm{\phiBP = \lrsb{\psiP \ \ \thetaP \ \ \xiP}^T} followed by a translation \nm{\TBPB} (section \ref{subsec:Sensors_Inertial_Mounting}) from the aircraft center of mass \nm{\OB} to the \hypertt{IMU} reference point \nm{\OP}:
\begin{itemize}
\item The \emph{platform heading} or \emph{platform yaw angle} \nm{\psiP} represents the angle that exists between \nm{\iBi} (forward fuselage) and the projection of \nm{\iPi} onto the plane defined by \nm{\iBi} and \nm{\iBii}.
\item The \emph{platform pitch angle} \nm{\thetaP} represents the angle existing between \nm{\iPi} and its projection onto the plane defined by \nm{\iBi} and \nm{\iBii}.
\item The \emph{platform bank angle} \nm{\xiP} represents the angle existing between \nm{\iPii} and the intersection between the plane defined by \nm{\iPii} and \nm{\iPiii} with the plane defined by \nm{\iBi} and \nm{\iBii}.
\end{itemize}
\begin{eqnarray}
\nm{\vec q_{\sss BP}} & \nm{\equiv} & \nm{\vec R_{\sss BP} = \vec R_3(\psiP) \ \vec R_2(\thetaP) \ \vec R_1(\xiP) \equiv \phiBP} \label{eq:RefSystems_qBP} \\
\nm{\vec \zeta_{\sss BP}} & = & \nm{\vec q_{\sss BP} + \dfrac{\epsilon}{2} \, \vec T_{\sss BP}^{\sss B} \otimes \vec q_{\sss BP}} \label{eq:RefSystems_zetaBP}
\end{eqnarray}


\subsection{Accelerometers Frame}\label{subsec:RefSystems_A}

The accelerometers (\texttt{A}) frame \cite{Farrell2008,Chatfield1997,Rogers2007} is a non-orthogonal frame \nm{\FA = \{\OA,\,\iAi,\,\iAii,\,\iAiii\}} centered at the \hypertt{IMU} reference point (\nm{{\vec T}_{\sss PA}^{\sss P} = \vec 0_3}). The basis vectors \nm{\lrb{\iAi,\,\iAii,\,\iAiii}} are aligned with each of the three accelerometers sensing axes\footnote{Each accelerometer hence only senses the specific force component parallel to its sensing axis.} (section \ref{subsec:Sensors_Accelerometer_Triad_ErrorModel}), but they are not orthogonal among them due to manufacturing inaccuracies, which implies that the angles between the \nm{\FA} and \nm{\FP} axes are very small.  

It is always possible, with no loss of generality, to impose that \nm{\iPi} coincides with {\nm{\iAi} and that \nm{\iPii} is located in the plane defined by \nm{\iAi} and \nm{\iAii}. If this is the case, \nm{\iAi  \perp \iPii}, \nm{\iAi \perp \iPiii}, and \nm{\iAii \perp \iPiii}, and the relative attitude between the \nm{\FP} and \nm{\FA} frames can be defined by three independent small rotations. 
\begin{itemize}
\item The \nm{\iAii} axis can be obtained from \nm{\iPii} by means of a small rotation \nm{\alphaACCiii} about \nm{\iPiii}.
\item The \nm{\iAiii} axis can be obtained from \nm{\iPiii} by two small rotations: \nm{\alphaACCi} about \nm{\iPi} and \nm{\alphaACCii} about \nm{\iPii}.
\end{itemize}

Although the exact relationships can be obtained \cite{Chatfield1997}, and given that the angles are very small, it is possible to consider \nm{\cos \alphaACCXi = 1, \ \sin \alphaACCXi = \alphaACCXi}, and \nm{\alphaACCXi \cdot \alphaACCXj = 0 \ \forall \ i,j \in \lrb{1,\ 2, \ 3}, i \neq j}, resulting in the following transformation matrices between \nm{\FP} and \nm{\FA}\footnote{As \nm{\FA} is not orthogonal, the transformation matrices are denoted with \nm{\star} to indicate that they are not proper rotation matrices as defined in \cite{LIE}. These matrices hence can not be converted into unit quaternions.}: 
\neweq{\RPA = \begin{bmatrix} 1 & 0 & 0 \\ \nm{\alphaACCiii} & 1 & 0 \\ - \nm{\alphaACCii} & \nm{\alphaACCi} & 1 \end{bmatrix} \hspace{50pt} \RAP = \begin{bmatrix} 1 & 0 & 0 \\ - \nm{\alphaACCiii} & 1 & 0 \\ \nm{\alphaACCii} & - \nm{\alphaACCi} & 1 \end{bmatrix}} {eq:RefSystems_PA_AP}


\subsection{Gyroscopes Frame}\label{subsec:RefSystems_Y}

The gyroscopes (\texttt{Y}) frame \cite{Farrell2008,Chatfield1997,Rogers2007} is similar to \nm{\FA}, but aligned with the gyroscopes sensing axes instead of those of the accelerometers. It is also a non-orthogonal frame \nm{\FY = \{\OY,\,\iYi,\,\iYii,\,\iYiii\}} centered at the \hypertt{IMU} reference point (\nm{{\vec T}_{\sss PY}^{\sss P} = \vec 0_3}), but no simplifications can be made about the relative position of the \nm{\FY} axes with respect to those of \nm{\FP}, so their relative attitude is defined by six small rotations \nm{\alphaGYRXij \ \forall \ i,j \in \lrb{1,\ 2, \ 3}, i \neq j}, where \nm{\alphaGYRXij} is the rotation of \nm{\vec i_{i}^{\sss Y}} about \nm{\vec i_{j}^{\sss P}}.

An approach similar to that employed in section \ref{subsec:RefSystems_A} leads to the following transformations: 
\neweq{\RPY = \begin{bmatrix} 1 & -\nm{\alphaGYRiiXiii} & \nm{\alphaGYRiiiXii} \\ \nm{\alphaGYRiXiii} & 1 & -\nm{\alphaGYRiiiXi} \\ - \nm{\alphaGYRiXii} & \nm{\alphaGYRiiXi} & 1 \end{bmatrix} \hspace{50pt} \RYP = \begin{bmatrix}  1 & \nm{\alphaGYRiiXiii} & -\nm{\alphaGYRiiiXii} \\ -\nm{\alphaGYRiXiii} & 1 & \nm{\alphaGYRiiiXi} \\ \nm{\alphaGYRiXii} & -\nm{\alphaGYRiiXi} & 1 \end{bmatrix}} {eq:RefSystems_PY_YP}


\subsection{Camera Frame}\label{subsec:RefSystems_C}

The camera frame (\texttt{C}) \cite{Soatto2001} is a Cartesian frame \nm{\FC = \{\OC ,\, \iCi, \, \iCii, \, \iCiii\}} that has its origin \nm{\OC} located at the camera optical center (section \ref{sec:Sensors_camera}), \nm{\iCi} located parallel to the camera focal plane pointing towards the right of the image, \nm{\iCiii} located in the camera principal axis pointing forward, and \nm{\iCii} orthogonal to \nm{\iCi} and \nm{\iCiii} in such a way that they form a right handed system.

\nm{\FC} is indispensable for visual navigation. It can be obtained from \nm{\FB} by a rotation best described by the Euler angles \nm{\phiBC = \lrsb{\psiC \ \ \thetaC \ \ \xiC}^T} followed by a translation \nm{\TBCB} (section \ref{sec:Sensors_camera}) from the aircraft center of mass \nm{\OB} to the camera optical center \nm{\OC}:
\begin{itemize}
\item The \emph{camera heading} or \emph{camera yaw angle} \nm{\psiC} represents the angle that exists between \nm{\iBi} (forward fuselage) and the projection of \nm{\iCi} onto the plane defined by \nm{\iBi} and \nm{\iBii}.
\item The \emph{camera pitch angle} \nm{\thetaC} represents the angle existing between \nm{\iCi} and its projection onto the plane defined by \nm{\iBi} and \nm{\iBii}.
\item The \emph{camera bank angle} \nm{\xiC} represents the angle existing between \nm{\iCii} and the intersection between the plane defined by \nm{\iCii} and \nm{\iCiii} with the plane defined by \nm{\iBi} and \nm{\iBii}.
\end{itemize}
\begin{eqnarray}
\nm{\qBC} & \nm{\equiv} & \nm{\RBC = \vec R_3(\psiC) \ \vec R_2(\thetaC) \ \vec R_1(\xiC) \equiv \phiBC} \label{eq:RefSystems_qBC} \\
\nm{\zetaBC} & = & \nm{\qBC + \dfrac{\epsilon}{2} \, \TBCB \otimes \qBC} \label{eq:RefSystems_zetaBC}
\end{eqnarray}

The relative pose between \nm{\FE} and \nm{\FC} results in:
\neweq{\zetaEC = \zetaEN \otimes \zetaNB \otimes \zetaBC} {eq:RefSystems_zetaEC}

The simulation also employs the image frame \nm{\FIMG}, a two dimensional Cartesian frame associated to \nm{\FC} better described within the context of the camera sensor in section \ref{sec:Sensors_camera}.

 \cleardoublepage


\end{document}